\pdfoutput=1

\documentclass[11pt]{article}

\usepackage[final]{acl}

\usepackage{times}
\usepackage{latexsym}

\usepackage[T1]{fontenc}

\usepackage[utf8]{inputenc}

\usepackage{microtype}
\usepackage{subcaption}
\usepackage{inconsolata}

\usepackage{graphicx}

\usepackage{booktabs}       
\usepackage{amsfonts}       
\usepackage{nicefrac}       
\usepackage{microtype}      
\usepackage[table, dvipsnames]{xcolor}         
\usepackage{graphicx}
\usepackage{adjustbox}
\usepackage{multirow}
\usepackage{wrapfig}
\usepackage{amsmath}
\usepackage{enumitem}
\usepackage{makeidx} 
\usepackage{appendix} 
\usepackage{titletoc}
\usepackage{floatflt}
\usepackage{ragged2e}
\usepackage[most]{tcolorbox} 
\usepackage{tabularx}
\usepackage{makecell}
\usepackage{svg}
\usepackage{soul,xcolor}
\soulregister\cite7

\tcbuselibrary{listings, breakable}
\usepackage[dvipsnames]{xcolor}
\usepackage{listings}
\usepackage{fancyvrb}
\usepackage{float}
\usepackage{placeins}
\usepackage{caption} 
\fvset{
    formatcom=\color{py-doc},
}
\lstdefinestyle{pydoc}{
    basicstyle=\ttfamily\color{py-doc},
    breaklines=true,
    breakatwhitespace=false,
    xleftmargin=2em
}

\usepackage{lipsum}

\newcommand{\chartagent}{\texttt{ChartAgent}}
\newcolumntype{C}[1]{>{\centering\arraybackslash}p{#1}} 
\newcolumntype{M}[1]{>{\centering\arraybackslash}m{#1}}

\newcommand{\hlpink}[1]{\sethlcolor{pink!30}\hl{#1}}
\title{\centering\includegraphics[scale = 0.03]{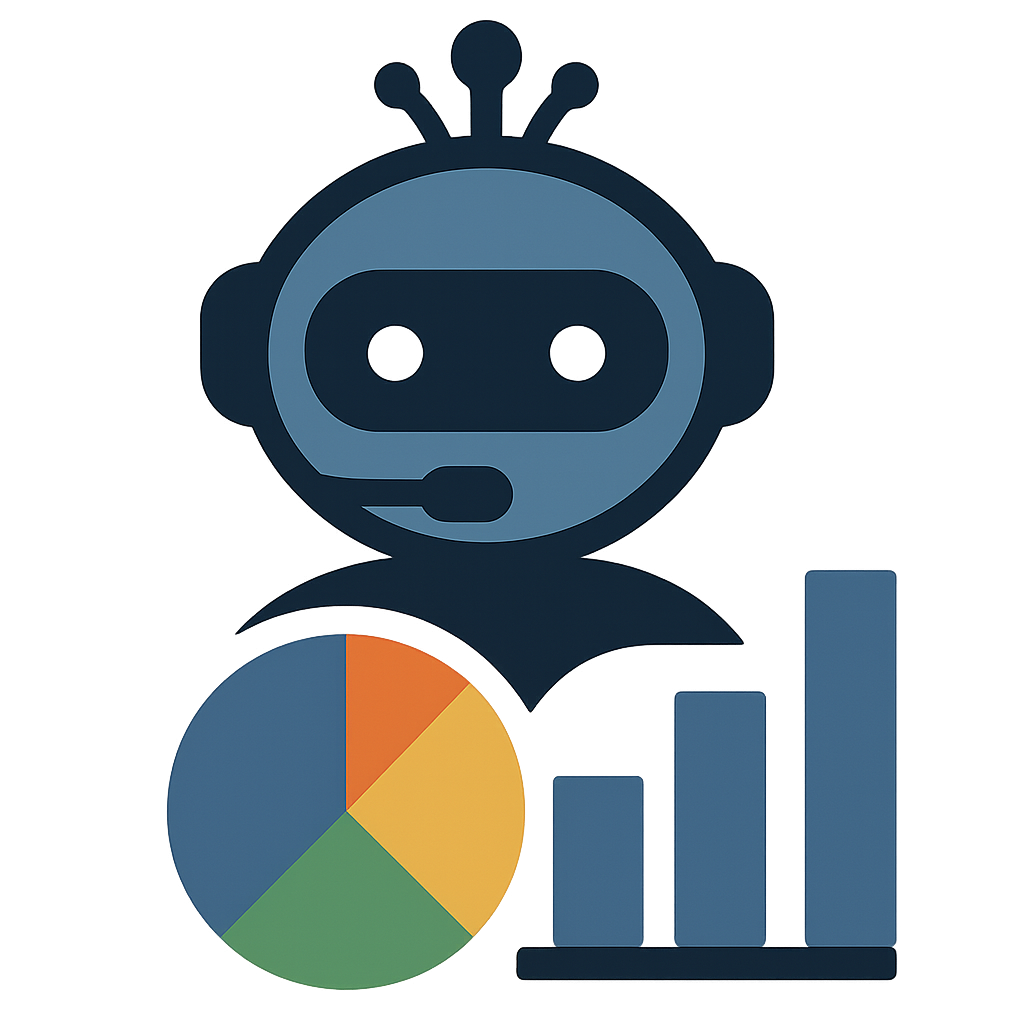}
ChartAgent: A Multimodal Agent for Visually Grounded Reasoning in Complex Chart Question Answering
}

\author{%
Rachneet Kaur \quad \textbf{Nishan Srishankar} 
\quad \textbf{Zhen Zeng} 
\\
\quad \textbf{Sumitra Ganesh} 
\quad \textbf{Manuela Veloso}
\\
J.P. Morgan AI Research\\
\texttt{\{rachneet.kaur, nishan.srishankar, zhen.zeng\}@jpmorgan.com}
\\
\texttt{\{sumitra.ganesh, manuela.veloso\}@jpmorgan.com}\\
}

\colorlet{pale1}{blue!10}
\colorlet{pale2}{green!10}
\colorlet{pale3}{red!10}
\colorlet{pale4}{orange!10}
\colorlet{pale5}{cyan!10}
\colorlet{pale6}{magenta!10}
\colorlet{pale7}{gray!10}
\colorlet{pale8}{teal!10}
\colorlet{pale9}{purple!10}
\newcolumntype{P}[1]{>{\centering\arraybackslash}p{#1}}
\usepackage{pifont}

\begin{document}
\maketitle

\begin{abstract}
Recent multimodal LLMs have shown promise in chart-based visual question answering, 
but their performance declines sharply on unannotated charts---those requiring precise visual interpretation rather than relying on textual shortcuts. To address this, we introduce \textbf{\chartagent{}}, a novel agentic framework that explicitly performs visual reasoning directly within the chart's spatial domain. 
Unlike textual chain-of-thought reasoning, \chartagent{} iteratively decomposes queries into visual subtasks and actively manipulates and interacts with chart images through specialized actions such as drawing annotations, cropping regions (e.g., segmenting pie slices, isolating bars), and localizing axes, using a library of chart-specific vision tools to fulfill each subtask. 
This iterative reasoning process closely mirrors human cognitive strategies for chart comprehension.
\chartagent{} achieves state-of-the-art accuracy on the ChartBench and ChartX benchmarks, surpassing prior methods by up to 16.07\% absolute gain overall and 17.31\% on unannotated, numerically intensive queries.
Furthermore, our analyses show that \chartagent{} is (a) effective across diverse chart types, (b) achieves the highest scores across varying visual and reasoning complexity levels, and (c) serves as a plug-and-play framework that boosts performance across diverse underlying LLMs.
Our work is among the first to demonstrate visually grounded reasoning for chart understanding using tool-augmented multimodal agents.

\end{abstract}

\section{Introduction}\label{sec:intro}
Charts, including bar plots, pie charts, line graphs, and their many variants, are foundational tools for communicating quantitative information across domains such as finance, science, and journalism~\cite{chishtie2022interactive, srivastava2025enhancing}. Enabling computational systems to answer natural-language questions about charts, referred to as \textit{chart visual question answering} (Chart VQA), remains an essential yet challenging problem in multimodal machine learning research~\cite{masry-etal-2022-chartqa, xu2023chartbench, xia2024chartx, wangcharxiv}.
Recent advances in 
multimodal large language models (MLLMs) have driven substantial progress in general visual reasoning tasks~\cite{liu2023visual, hurst2024gpt, llava-onevision}. 
However, their performance degrades significantly on Chart VQA, especially when dealing with charts that lack explicit textual annotations of key values or labels, commonly referred to as \textit{unannotated} charts~\cite{xu2023chartbench, xia2024chartx, islam-etal-2024-large} (see Appendix~\ref{appendix:annotated_vs_unannotated_examples} for examples). These scenarios demand accurate visual grounding and interpretation (e.g., estimating numerical values from graphical elements), a setting where even state-of-the-art (SoTA) MLLMs often struggle.

\begin{figure*}
\centering
    \includegraphics[scale = 0.365]{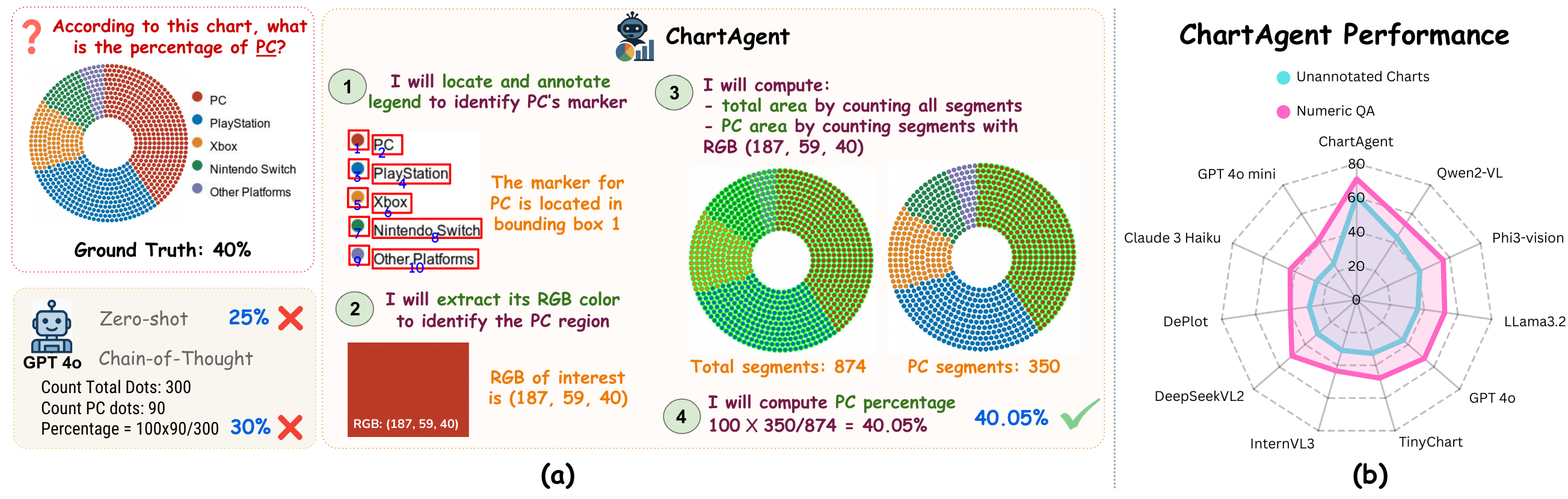}
   \caption{\textbf{Comparison of our work with the existing SoTA.}
    \textbf{(a)} \chartagent{} performs visually grounded reasoning in the chart domain. For this unannotated chart, MLLM fails to produce the correct answer, whereas \chartagent{} succeeds. 
    \textbf{(b)} \chartagent{} performance on unannotated charts and numeric QA compared with the top-10 SoTA.  
}
    \label{fig:teaser_image}
\end{figure*}

To address these shortcomings, we draw inspiration from how humans reason with charts. Humans typically process graphical elements sequentially, interpreting axes, legends, and segments, and often add annotations to support intermediate reasoning, such as tracing bars and lines to compare values, circling or shading pie slices to judge proportions, and highlighting legends or markers to align categories. Building on these cognitive strategies, we propose \textbf{\chartagent{}}, a novel agentic framework explicitly designed for visually grounded reasoning in the chart domain (see Figure~\ref{fig:teaser_image}). 
At the core of \chartagent{} lies a multi-turn interaction loop that progressively decomposes chart queries into subtasks that are primarily visual 
and occasionally numerical, 
while simultaneously manipulating and interacting with chart images through precise, modular perception tools 
tailored to fulfill these subtasks, thereby augmenting MLLM reasoning with chart-specialized visual capabilities.
To the best of our knowledge, and complementary to existing chart VQA approaches that rely on prompting or fine-tuning MLLMs~\cite{masry-etal-2025-chartgemma, masry-etal-2024-chartinstruct, han2023chartllamamultimodalllmchart, liu-etal-2023-matcha}, this work is the \underline{first} to demonstrate visually grounded reasoning for chart understanding through tool-augmented multimodal agents, achieving SoTA performance. 
Importantly, the perception tools are designed to generate interpretable visualizations (see Figures~\ref{fig:universal-tools-viz-self-verify},~\ref{fig:chart-specific-tools-viz-self-verify}) that the agent can inspect. This allows it to dynamically adjust its strategy, such as tuning 
parameters or switching to alternative tools, when the 
outputs are unsatisfactory.
Our \textbf{key contributions} are:
\begin{itemize}[leftmargin=*]
    \item \textbf{\textit{Multimodal Agent for Charts:}} We introduce \chartagent{}, the \underline{first} framework to augment MLLM reasoning with chart-specialized visual capabilities for Chart VQA, systematically demonstrating visually grounded reasoning in charts via a tool-augmented multimodal agent.
    
    \item \textbf{\textit{Modular Vision Tool Library with Self-Verification:}} An agent-compatible library of chart-specialized perception tools covering 40+ chart types, generating interpretable visualizations (see Figures~\ref{fig:universal-tools-viz-self-verify},~\ref{fig:chart-specific-tools-viz-self-verify}) that not only support grounded reasoning in \chartagent{} but also enable a visual self-verification mechanism, allowing the agent to inspect intermediate results and adaptively adjust reasoning and tool use.

    \item \textbf{\textit{State-of-the-Art Performance:}} \chartagent{} achieves new SoTA,
    surpassing 30+ baselines by up to 16.07\% absolute gain overall and 17.31\% on unannotated, numerically intensive queries, evaluated on the well-established ChartBench and ChartX 
    datasets
    spanning 40+ chart types.

    \item \textbf{\textit{In-Depth Analysis:}} We conduct extensive analyses to demonstrate the effectiveness of \chartagent{}. Specifically, we show that (a) it is effective across diverse chart types, (b) it achieves the highest scores across varying visual and reasoning complexity levels of chart–QA pairs, and (c) it serves as a plug-and-play framework that enhances performance across different base MLLMs, thereby validating both effectiveness and generalization. We also present a failure mode analysis highlighting common errors.
\end{itemize}

The remainder of this paper is organized as follows: Section~\ref{sec:related_works} discusses related work, Section~\ref{sec:methods} details the methodology behind \chartagent{}, Section~\ref{sec:experiments} and~\ref{sec:results} presents experiments and results, and Section~\ref{sec:conclusion_future_work} concludes the paper.

\section{Related Work}\label{sec:related_works}
We review related work in three areas: chart VQA (\ref{related-chartvqa}), MLLMs and visual grounding (\ref{related-mllm}), and agentic frameworks (\ref{related-agents}). See Appendix~\ref{appendix:related_works} for an extended review.

\subsection{Chart Visual Question Answering}\label{related-chartvqa}
Chart VQA interprets charts to answer natural-language queries. Early synthetic datasets~\cite{kahou2017figureqa,kafle2018dvqa} emphasized visual reasoning but lacked real-world diversity. Later benchmarks~\cite{methani2020plotqa,masry-etal-2022-chartqa,huang2024evochart,xu2023chartbench,xia2024chartx,wangcharxiv} introduced realistic, diverse, and numerically intensive charts. Chart-specific MLLMs~\cite{zhang2024tinychart,masry-etal-2023-unichart,liu2024mmc,masry-etal-2024-chartinstruct} enhanced instruction tuning and vision–language alignment, while hybrid approaches~\cite{luo2021chartocr} integrated vision tools with rule-based parsing. However, recent studies~\cite{xu2023chartbench,razeghi-etal-2024-plot,islam-etal-2024-large} reveal sharp performance drops on unannotated charts, highlighting poor visual grounding. Our work addresses this gap through chart-specialized, visually grounded reasoning.

\subsection{Multimodal LLMs and Visual Grounding}\label{related-mllm}
General-purpose MLLMs such as GPT-4~\cite{achiam2023gpt}, GPT-4o~\cite{hurst2024gpt}, Gemini~\cite{team2023gemini}, LLaVA~\cite{liu2023visual}, and Visual CoT~\cite{shao2024visual} have advanced visual reasoning. For stronger grounding, models integrate tools or visual prompts: Visual ChatGPT~\cite{wu2023visual}, MM-ReAct~\cite{yang2023mm}, ViperGPT~\cite{suris2023vipergpt}, and VisProg~\cite{gupta2023visual} employ structured tools, while Visual Sketchpad~\cite{huvisualsketchpad} and Set-of-Marks~\cite{yang2023set} iteratively refine and annotate inputs. Inspired by these, our approach unites iterative reasoning, visual prompting, and modular vision tools for chart-grounded understanding.

\subsection{Agentic Frameworks}\label{related-agents}
Agent-based AI systems, defined by perception, cognition, and action, have advanced with LLM integration. The ReAct framework~\cite{yao2023react} structures interactions into iterative reasoning, action, and observation, while platforms such as AutoGen~\cite{wu2024autogen}, CrewAI~\cite{crewai}, LangChain~\cite{LangChain}, LangGraph~\cite{langgraph}, and AutoGPT~\cite{autogpt} support practical implementations. MLLM agents extend this paradigm to robotics~\cite{nasirianypivot, hori2025interactive}, vision-language reasoning~\cite{liu2025llava, yang2023mm}, and GUI navigation~\cite{verma2025adaptagent, he2024webvoyager, xie2024osworld, zhenggpt, koh-etal-2024-visualwebarena}. Similarly, \chartagent{} 
integrates multimodal reasoning with modular, chart-oriented vision tools in an agentic framework.
\begin{figure*}[!t]
    \centering
    \includegraphics[width = 0.99\textwidth, height = 0.51\textheight]{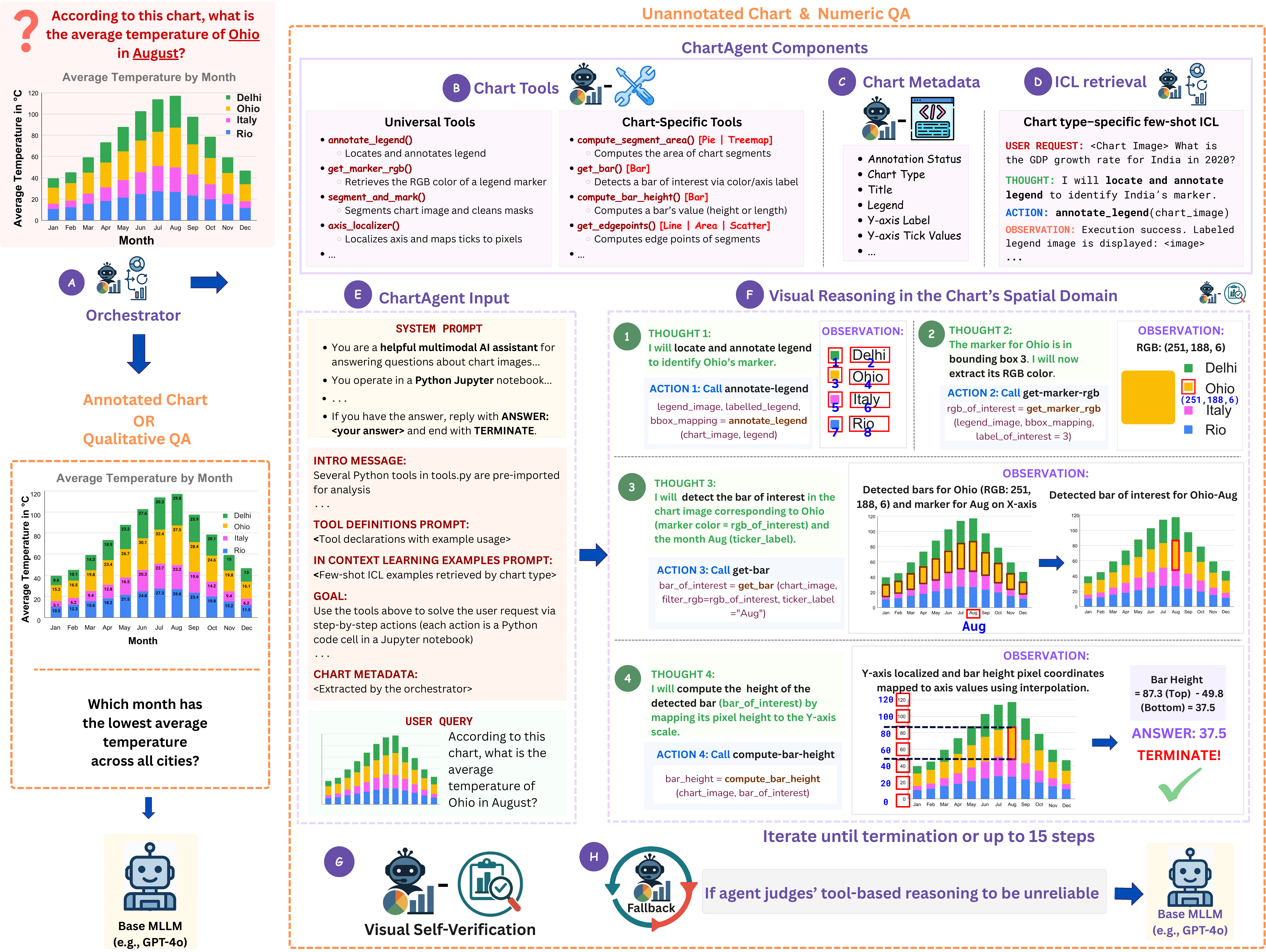}
    \caption{\textbf{\chartagent{}.} The \textbf{(A)} orchestrator extracts chart metadata and routes annotated charts with textual shortcuts and qualitative QA to the base MLLM, while unannotated charts and numeric queries trigger the ReAct-style loop. The system includes \textbf{(B)} a library of universal and chart-specific tools, \textbf{(C)} metadata for parameterizing tool usage and retrieving chart-type-specific ICL examples, and \textbf{(D)} few-shot ICL retrieval. Using these components as the \textbf{(E)} input, \chartagent{} performs \textbf{(F)} iterative visual reasoning, supported by \textbf{(G)} visual self-verification of intermediate outputs. When tool-based reasoning is unreliable, \textbf{(H)} the agent falls back to the base MLLM.}
    \label{fig:chartagent}
\end{figure*}
\section{ChartAgent: A Multimodal Agent for Visually Grounded Reasoning in Charts}\label{sec:methods}
Given a multimodal query consisting of a chart image and a natural-language question about the chart, the goal is to generate an answer that accurately reflects the information conveyed in the chart.
Building on human strategies for chart comprehension, such as highlighting legend entries to clarify category mappings, sketching guide lines across bars or axes to compare values, or shading portions of a pie chart to approximate proportions, we propose \textbf{\chartagent{}}. As illustrated in Figure~\ref{fig:chartagent}, \chartagent{} is a novel agentic framework that equips MLLMs with structured visual reasoning capabilities for charts, by decomposing queries into visual subtasks and directly interacting with chart images in their spatial domain through specialized vision tools to accomplish these subtasks.
These tools are supported by interpretable intermediate visualizations that enable adaptive refinement of reasoning and grounding until a confident answer is reached or the iteration limit is exhausted.

\subsection{Visually Grounded Chart Reasoning}
The foundation of \chartagent{} is a structured, iterative ReAct~\cite{yao2023react}-style multi-turn interaction loop within the chart’s visual environment, which at each time step $t$ generates a sequence of Thought, Action, and Observation phases to guide the agent in interpreting charts and answering user queries.

\textbullet \; \textbf{Thought (Reasoning)}: The MLLM evaluates the current state $s_t$, which includes the multimodal query along with previous thoughts, actions, and observations, to derive the next subtask (goal) $g_t$ that guides the subsequent action toward answering the user’s query. These sub-goals primarily involve visual perception tasks (e.g., segmenting chart elements, detecting and annotating legends, or localizing axes), but may also include numerical operations (e.g., interpolation, arithmetic).

\textbullet \; \textbf{Action (Chart Tool Execution)}: Based on the subtask $g_t$ from the Thought phase, the agent selects and executes an appropriate tool $a^{\text{chart-tool}}_t$ from a modular chart-specialized library (see Appendix Table~\ref{appendix_tab:vision_tool_description}) that directly manipulates the chart image. Examples include pie segmentation, bar isolation, legend detection, axis tick localization, and interpolation. Each tool returns structured outputs (e.g., numeric estimates, labels, detected coordinates) and, when applicable, interpretable intermediate or final visualizations (e.g., segmentation masks with labels, colored overlays for pie slices, bar height markers, annotated legends, or bounding boxes) (see Appendix Figures~\ref{fig:universal-tools-viz-self-verify},~\ref{fig:chart-specific-tools-viz-self-verify}), which support the agent’s subsequent visual self-verification.

\textbullet \; \textbf{Observation (Visual Self-Verification and Adaptive Tool Use)}: Based on the invoked action $a^{\text{chart-tool}}_t$, \chartagent{} receives new perception-friendly visualizations and outputs $o_{t+1}$. The multimodal state is then updated as $s_{t+1} = (s_t, g_t, a^{\text{chart-tool}}_t, o_{t+1})$. \chartagent{} then interprets and verifies these multimodal outputs, particularly for perception-related tools, by visually inspecting the provided visualizations to assess their accuracy. 
If verification reveals unsatisfactory results (e.g., incomplete segmentation, mismatched legend associations, overly small pie slices, incorrect colors, negative bar heights, or outputs inconsistent with axis values), the agent adaptively adjusts its tool use in the next iteration $t+1$, for instance by invoking an alternative tool or tweaking parameters such as detection thresholds. 
This iterative correction loop mimics human-like debugging, enabling \chartagent{} to reason and ground with visualizations it generated on the chart, thereby ensuring 
improved chart VQA capabilities (see Section~\ref{results: additional_analysis}).
Further, if tool outputs remain insufficient after multiple iterations, this design enables \chartagent{} to recognize the limits of its perception capabilities with the available tools, a key feature for trustworthy agent design.

\subsection{Chart Interaction and Manipulation} 
The effectiveness of \chartagent{} hinges on the careful design of a modular library of perception and numeric tools tailored for chart understanding (a detailed taxonomy is provided in Appendix~\ref{appendix:vision-tool-description} and Table~\ref{appendix_tab:vision_tool_description}). Inspired by primitive visual tasks in natural image domains (e.g., object detection, segmentation, relational inference), we define analogous primitive tasks for the chart domain, treating chart elements (e.g., bars, pie slices, lines, legends, tick marks, and axis labels) as fundamental visual ``objects.'' By targeting shared components such as legends, axes, ticks, bar segments, and pie slices, these tools enable broad generalization across diverse chart formats (see Appendix~\ref{appendix:chart_types_figures} for the 40+ chart types supported in \chartagent{}).  
Guided by this perspective, we designed a structured, chart-domain-specific set of \textit{primitive} tools, organized into two categories:  
\begin{enumerate}[leftmargin=*, noitemsep, topsep=1pt, partopsep=1pt]
    \item \textbf{Universal chart tools:} General-purpose perception tools applicable across chart types, such as segmentation, legend detection, axis localization, and numeric interpolation.  
    \item \textbf{Chart-specific tools:} Tools specialized for particular chart types (e.g., pie, bar, line, box), targeting subtasks unique to their visual structures.  
\end{enumerate}
Each tool is deliberately scoped to remain clear and distinct, avoiding overly fine-grained or excessively complex functionalities, thereby ensuring robust implementations with modern vision techniques.
\subsection{Architecture and Components} 
\textbullet \; \textbf{Chart Metadata Extraction and Orchestration}: \chartagent{} begins with an LLM-based orchestrator (e.g., GPT-4o) that extracts comprehensive chart metadata, including chart type, title, legend details, axis labels and tick marks, annotation status (annotated or unannotated), and a concise visual description (see Appendix~\ref{appendix:chart-metadata-prompt}). 
This metadata is critical for orchestrating the smart routing mechanism, which first determines whether perception tools are necessary for the user task.
For annotated charts containing explicit textual shortcuts (e.g., numerical annotations or clear labels) or for queries requiring mainly qualitative reasoning, direct reasoning by the base MLLM is often sufficient. In such cases, the orchestrator routes the query directly to the MLLM 
balancing accuracy and computational efficiency. In contrast, for unannotated charts (see Appendix~\ref{appendix:annotated_vs_unannotated_examples}), where accurate interpretation of graphical elements, such as bar-height/pie-area estimation, 
or legend association, 
is essential, the orchestrator initiates a deeper, iterative routine of visual reasoning to derive the answer.
In the unannotated case, the extracted metadata is also used to retrieve appropriate chart-type-specific few-shot in-context learning (ICL) examples and to parameterize subsequent tool usage.

\textbullet \; \textbf{Chart Tools Implementation}: Chart tools are implemented as Python functions callable by \chartagent{}. Some of these tools internally leverage SoTA computer vision and OCR methods, such as Segment Anything (SAM)~\cite{kirillov2023segment}, Semantic SAM~\cite{li2307semantic}, Tesseract~\cite{tesseract}, and EasyOCR~\cite{easyocr}. They also handle edge cases (e.g., rotated text, fuzzy label matching for legends or axis ticks, and filtering small, background, or overlapping segments) and return structured outputs (e.g., numeric values, bounding boxes, text labels) along with visualizations (e.g., segmentation masks with labels or bounding box annotations; see details and examples in Appendix~\ref{appendix:self-verification} and Figures~\ref{fig:universal-tools-viz-self-verify},~\ref{fig:chart-specific-tools-viz-self-verify}) that are explicitly designed to facilitate \chartagent{}’s visual self-verification. 
See Appendix~\ref{appendix:vision-tool-description} for detailed tool descriptions, Appendix~\ref{appendix-chartagent-tool-definition-prompts} for prompt details, and Section~\ref{results: additional_analysis} for an analysis of their effectiveness.

\textbullet \; \textbf{ICL}: \chartagent{} uses few-shot (1–2) ICL examples that are specifically retrieved based on the chart type identified during metadata extraction (see Appendix~\ref{appendix:chartagent-ICL-prompt}). 
For instance, if a chart is classified as a pie chart, only pie chart ICL examples are appended to the prompt. 
If no ICL examples exist for the detected chart type, then none are added.
Each ICL example consists of a complete ReAct trajectory that successfully answers sample queries (see Appendix~\ref{appendix:chartagent-ICL-prompt}). 

\textbullet \; \textbf{Multimodal Agentic Framework}: \chartagent{} uses GPT-4o (\texttt{gpt-4o-2024-08-06}) as the base MLLM, serving as both reasoning backbone and orchestrator. 
With its plug-and-play design, \chartagent{} benefits from advances in both perception tools and MLLM reasoning, enabling seamless integration and sustained cumulative performance gains. We also experiment with other MLLMs to validate this generalization; see Section~\ref{results: effectiveness_chartagent}.
\chartagent{} is built on AutoGen~\cite{wu2024autogen}, which enables tool orchestration; see Appendix~\ref{appendix:Prompts} for the structured set of prompts.
After each ReAct cycle, \chartagent{} evaluates the updated multimodal state $s_{t+1}$ and decides whether to continue or terminate with a final answer.
If satisfactory results cannot be achieved after multiple iterations, the agent gracefully falls back to direct MLLM reasoning (see Section~\ref{results: additional_analysis} for evaluation). The maximum number of ReAct iterations is set to 15. Qualitative illustrations of agent trajectories are provided in Appendix~\ref{appendix:qualitative_examples}, with further implementation details in Appendix~\ref{appendix:implementation_details}.

\section{Experimental Protocol and Details}\label{sec:experiments}
\subsection{Datasets}\label{experiments:datasets}
We benchmark on two widely used datasets: \textbf{ChartBench}~\cite{xu2023chartbench}, which spans 9 chart categories and 42 subtypes, including standard charts (bar, line, pie) and complex ones (area, radar, box, scatter, node, and combinations), with 3,800 chart--QA pairs (76.2\% unannotated). We evaluate two QA types: (1) \textit{Numeric QA}, requiring precise value extraction, and (2) \textit{Relationship QA}, involving structural reasoning (e.g., connectivity in graphs), with 96.7\% numeric QA.  \textbf{ChartX}~\cite{xia2024chartx}, which covers 18 chart types, ranging from standard to domain-specific formats (e.g., treemaps, heatmaps, candlestick charts), with 1,152 chart--QA pairs (61.7\% unannotated). The questions span (1) \textit{Numeric QA}, and (2) \textit{Value Comparison / Global Perception QA}, which involves reasoning over relative or extremum-based patterns, with 71.9\% numeric QA. Both benchmarks are visually grounded, requiring models to reason about chart logic (e.g., bar heights, pie-slice areas) beyond OCR. Their high proportion of unannotated charts and numeric QA makes them particularly well-suited for evaluating complex visual reasoning. See Appendix~\ref{appendix:datasets_details} and~\ref{appendix:chart_types_figures} for dataset details.
\begin{table*}[htb]
\caption{\textbf{Comparison of accuracy (\%).} \textcolor{red}{\textbf{Red:}} Best,
\textcolor{blue}{\underline{Blue:}} Second best. All values correspond to the highest performance achieved across zero-shot and CoT prompting styles for each MLLM. Ann./Unann. denote Annotated and Unannotated charts. RL QA: Relationship QA; VC/GC QA: Value Comparison \& Global Conception QA.
}
\begin{subtable}[t]{0.53\textwidth}
\centering
\begin{adjustbox}{max width=\linewidth}
\begin{tabular}{l|cc|cc|c}
\toprule
\multirow{2}{*}{\textbf{Model}} & \multicolumn{2}{c|}{\textbf{Chart Types}} & \multicolumn{2}{c|}{\textbf{Question Types}} & \cellcolor{gray!15} \textbf{Overall}\\ 
\cmidrule(lr{1em}){2-3}
\cmidrule(lr{1em}){4-5}
\cmidrule(lr{1em}){6-6}
& Ann. & \cellcolor{green!5} \textbf{Unann.} & \cellcolor{green!5} \textbf{Numeric QA} & RL QA & \cellcolor{gray!15} Avg.  $ \mathbf{\uparrow}$ \\
\midrule
\multicolumn{6}{l}{\cellcolor{orange!15}\textit{Proprietary Multimodal Large Language Models}}  \\
\midrule
GPT 4o~\cite{hurst2024gpt} & \textbf{\textcolor{red}{94.33}} & \cellcolor{green!5} 36.15 & \cellcolor{green!5} 52.50 & \textcolor{red}{\textbf{91.00}} & \cellcolor{gray!15} 54.53\\
GPT 4o-mini~\cite{GPT4omini} & 84.83 & \cellcolor{green!5} 25.19 & \cellcolor{green!5} 41.50 & \textcolor{blue}{\underline{89.50}} & \cellcolor{gray!15} 44.03\\
Claude 3 Haiku~\cite{TheC3} & 84.58 & \cellcolor{green!5} 26.04 & \cellcolor{green!5} 42.94 & 73.00 & \cellcolor{gray!15} 44.53\\
Gemini 1.5~\cite{team2024gemini} & 89.72 & \cellcolor{green!5} 27.27 & \cellcolor{green!5} 46.69 & 53.85 & \cellcolor{gray!15} 47.08\\
\midrule
\multicolumn{6}{l}{\cellcolor{orange!15}\textit{Open-weights Multimodal Large Language Models}}  \\
\midrule
BLIP-2~\cite{li2023blip} & 3.67 & \cellcolor{green!5} 2.92 & \cellcolor{green!5} 3.11 & 4.00 & \cellcolor{gray!15} 3.16\\
CogAgent~\cite{hong2023cogagent} & 69.92 & \cellcolor{green!5} 11.62 & \cellcolor{green!5} 30.28 & 27.00 & \cellcolor{gray!15} 30.03\\
CogVLM~\cite{wang2023cogvlm} & 64.83 & \cellcolor{green!5}  11.62 & \cellcolor{green!5}  29.03 & 21.50 & \cellcolor{gray!15} 28.42\\
DeepSeek-VL2~\cite{wu2024deepseekvl2mixtureofexpertsvisionlanguagemodels} & \textcolor{blue}{\underline{90.75}} & \cellcolor{green!5}  30.31 & \cellcolor{green!5}  50.28 & 33.50 & \cellcolor{gray!15} 49.39\\
DocOwl1.5~\cite{hu2024mplug} & 67.50 & \cellcolor{green!5}  23.58 & \cellcolor{green!5} 37.06 & 44.50 & \cellcolor{gray!15} 37.45\\
InstructBLIP~\cite{instructblip} & 3.92 & \cellcolor{green!5} 5.92 & \cellcolor{green!5} 4.22 & 24.50 & \cellcolor{gray!15} 5.29\\
InternVL3~\cite{zhu2025internvl3exploringadvancedtraining} & 72.67 & \cellcolor{green!5} 30.92 & \cellcolor{green!5} 43.39 & 57.00 & \cellcolor{gray!15} 44.11\\
LLama3.2~\cite{grattafiori2024llama} & 87.58 & \cellcolor{green!5} 36.38 & \cellcolor{green!5} 52.22 & 50.00 & \cellcolor{gray!15} 52.11\\
Llava1.6~\cite{llavanext} & 35.58 & \cellcolor{green!5} 9.92 & \cellcolor{green!5} 16.69 & 42.00 & \cellcolor{gray!15} 18.03\\
Llava1.5~\cite{llava1.5} & 26.75 & \cellcolor{green!5} 7.00 & \cellcolor{green!5} 13.06 & 16.50 & \cellcolor{gray!15} 13.24\\
LlaVA-OneVision~\cite{llava-onevision} & 13.25 & \cellcolor{green!5} 10.50 & \cellcolor{green!5} 9.94 & 37.00 & \cellcolor{gray!15} 11.37\\
mPLUG-Owl3~\cite{mplugowl3} & 31.08 & \cellcolor{green!5} 12.65 & \cellcolor{green!5} 16.92 & 46.50 & \cellcolor{gray!15} 18.47\\
Phi3-vision~\cite{phi3} & 86.92 & \cellcolor{green!5} 40.77 & \cellcolor{green!5} \textcolor{blue}{\underline{55.89}} & 52.00 & \cellcolor{gray!15} \textcolor{blue}{\underline{55.32}}\\
Pixtral~\cite{agrawal2024pixtral} & 66.58 & \cellcolor{green!5} 28.73 & \cellcolor{green!5} 39.53 & 63.50 & \cellcolor{gray!15} 40.50\\
Qwen2-VL~\cite{wang2024qwen2vlenhancingvisionlanguagemodels} & 78.42 & \cellcolor{green!5} \textcolor{blue}{\underline{43.50}} & \cellcolor{green!5} 52.94 & 83.00 & \cellcolor{gray!15} 54.53\\
Qwen-VL-Chat~\cite{Qwen-VL} & 27.17 & \cellcolor{green!5} 6.54 & \cellcolor{green!5} 12.61 & 21.00 & \cellcolor{gray!15} 13.05\\
SmolVLM~\cite{marafioti2025smolvlmredefiningsmallefficient} & 47.75 & \cellcolor{green!5} 14.46 & \cellcolor{green!5} 23.14 & 58.00 & \cellcolor{gray!15} 24.97\\
SPHINX-V~\cite{lin2025drawandunderstand} & 35.91 & \cellcolor{green!5} 12.30 & \cellcolor{green!5} 18.08 & 0.5  & \cellcolor{gray!15} 19.76\\
VisualGLM~\cite{glm2024chatglmfamilylargelanguage} & 4.83 & \cellcolor{green!5} 7.65 & \cellcolor{green!5} 3.92 & 58.00 & \cellcolor{gray!15} 6.76\\
\midrule
\multicolumn{6}{l}{\cellcolor{orange!15}\textit{Chart-related Models}}  \\
\midrule
ChartGemma~\cite{masry-etal-2025-chartgemma} & 75.92 & \cellcolor{green!5} 22.42 & \cellcolor{green!5} 39.56 & 35.00 & \cellcolor{gray!15} 39.32\\
ChartInstruct~\cite{masry-etal-2024-chartinstruct} & 55.17 & \cellcolor{green!5} 20.19 & \cellcolor{green!5} 31.75 & 22.00 & \cellcolor{gray!15} 31.24\\
ChartLlama~\cite{han2023chartllamamultimodalllmchart} & 38.25 & \cellcolor{green!5} 11.42 & \cellcolor{green!5} 18.81 & 39.50 & \cellcolor{gray!15} 19.89\\
ChartVLM~\cite{xia2024chartx} & 61.00 & \cellcolor{green!5} 23.92 & \cellcolor{green!5} 36.97 & 11.50 & \cellcolor{gray!15} 35.63\\
DePlot~\cite{liu-etal-2023-deplot} & 70.08 & \cellcolor{green!5} 28.15 & \cellcolor{green!5} 39.33 & 78.50 & \cellcolor{gray!15} 41.39\\
MatCha~\cite{liu-etal-2023-matcha} & 59.50 & \cellcolor{green!5} 9.69 & \cellcolor{green!5} 25.86 & 17.50 & \cellcolor{gray!15} 25.42\\
OneChart~\cite{chen2024onechart} & 56.78 & \cellcolor{green!5} 26.81 & \cellcolor{green!5} 35.22 & 62.76 & \cellcolor{gray!15} 36.81\\
TinyChart~\cite{zhang2024tinychart} & 77.33 & \cellcolor{green!5} 32.77 & \cellcolor{green!5} 47.86 & 28.50 & \cellcolor{gray!15} 46.84\\
UniChart~\cite{masry-etal-2023-unichart} & 53.50 & \cellcolor{green!5} 15.96 & \cellcolor{green!5} 27.44 & 34.50 & \cellcolor{gray!15} 27.82\\
\midrule
\multicolumn{6}{l}{\cellcolor{orange!15}\textit{Multimodal Agentic Framework (Ours)}}  \\
\midrule
\textbf{\chartagent{}} &  \textcolor{red}{\textbf{94.33}} &  \cellcolor{green!5} \textcolor{red}{\textbf{60.81}} &  \cellcolor{green!5} \textcolor{red}{\textbf{70.91}} &  \textcolor{red}{\textbf{91.00}} &  \cellcolor{gray!15} \textcolor{red}{\textbf{71.39}} \\
\bottomrule
\end{tabular}
\end{adjustbox}
\vspace{0.05mm}
\caption{\textbf{ChartBench} (76.2\% unannotated charts; 96.7\% numeric QA)}
\label{tab:chartbench}
\end{subtable}
\hfill
\begin{subtable}[t]{0.46\textwidth}
\centering
\begin{adjustbox}{max width=\linewidth}
\begin{tabular}{l|cc|cc|c}
\toprule
\multirow{2}{*}{\textbf{Model}} & \multicolumn{2}{c|}{\textbf{Chart Types}} & \multicolumn{2}{c|}{\textbf{Question Types}} & \cellcolor{gray!15} \textbf{Overall}\\ 
\cmidrule(lr{1em}){2-3}
\cmidrule(lr{1em}){4-5}
\cmidrule(lr{1em}){6-6}
& Ann. & \cellcolor{green!5} \textbf{Unann.} & \cellcolor{green!5} \textbf{Numeric QA} & VC/GC QA & \cellcolor{gray!15} Avg.  $ \mathbf{\uparrow}$  \\
\midrule
\multicolumn{6}{l}{\cellcolor{orange!15}\textit{Proprietary Multimodal Large Language Models}}  \\
\midrule
GPT 4o & \textcolor{red}{\textbf{84.84}} & \cellcolor{green!5} 39.44 & \cellcolor{green!5} \textcolor{blue}{\underline{52.05}} & \textcolor{red}{\textbf{69.14}} & \cellcolor{gray!15} \textcolor{blue}{\underline{56.86}} \\
GPT 4o-mini & 71.95 & \cellcolor{green!5} 33.94 & \cellcolor{green!5} 42.51 & 63.89 & \cellcolor{gray!15} 48.52\\
Claude 3 Haiku & 63.57 & \cellcolor{green!5} 25.77 & \cellcolor{green!5} 35.99 & 51.23 & \cellcolor{gray!15} 40.28\\
Gemini 1.5 & 68.09 &  \cellcolor{green!5}  31.41 &  \cellcolor{green!5} 40.22 & 58.95 & \cellcolor{gray!15} 45.48 \\
\midrule
\multicolumn{6}{l}{\cellcolor{orange!15}\textit{Open-weights Multimodal Large Language Models}}  \\
\midrule
BLIP-2 & 1.13 & \cellcolor{green!5} 1.69 & \cellcolor{green!5} 0.72 & 3.40 & \cellcolor{gray!15} 1.48\\
CogAgent & 46.15 & \cellcolor{green!5} 24.93 & \cellcolor{green!5} 27.05 & 48.46 & \cellcolor{gray!15} 33.07 \\
CogVLM & 46.38 & \cellcolor{green!5} 24.23 & \cellcolor{green!5} 24.28 & 54.32 & \cellcolor{gray!15} 32.73\\
DeepSeek-VL2 & 66.74 & \cellcolor{green!5} 35.63 & \cellcolor{green!5} 43.84 & 57.10 & \cellcolor{gray!15} 47.57\\
DocOwl1.5 & 42.53 &\cellcolor{green!5}  24.37 & \cellcolor{green!5} 26.81 & 42.90 & \cellcolor{gray!15} 31.34\\
InstructBLIP & 10.41 & \cellcolor{green!5} 8.87 & \cellcolor{green!5} 7.37 & 14.81 & \cellcolor{gray!15} 9.46\\
InternVL3 & 65.84 & \cellcolor{green!5} 36.62 & \cellcolor{green!5} 44.20 & 57.10 & \cellcolor{gray!15} 47.83\\
LLama3.2 & \textcolor{blue}{\underline{78.51}} & \cellcolor{green!5} 39.86 & \cellcolor{green!5} 50.36 & 65.74 & \cellcolor{gray!15} 54.69\\
Llava1.6 & 26.24 & \cellcolor{green!5} 18.17 & \cellcolor{green!5}  16.55 & 33.33 & \cellcolor{gray!15} 21.27\\
Llava1.5 & 18.55 & \cellcolor{green!5} 14.51 & \cellcolor{green!5} 10.63 & 29.94 & \cellcolor{gray!15} 16.06\\
LlaVA-OneVision & 20.14 & \cellcolor{green!5} 12.82 & \cellcolor{green!5} 13.89 & 20.06 & \cellcolor{gray!15} 15.62\\
mPLUG-Owl3 & 23.98 & \cellcolor{green!5} 18.31 & \cellcolor{green!5} 14.49 & 35.80 & \cellcolor{gray!15} 20.49\\
Phi3-vision & 59.95 & \cellcolor{green!5} 41.69 & \cellcolor{green!5} 41.06 & 68.21 & \cellcolor{gray!15} 48.70\\
Pixtral & 64.93 & \cellcolor{green!5} 38.17 & \cellcolor{green!5} 41.55 & 66.05 & \cellcolor{gray!15} 48.44\\
Qwen2-VL & 76.24 & \cellcolor{green!5} \textcolor{blue}{\underline{42.96}} & \cellcolor{green!5}  51.81 & 65.74 & \cellcolor{gray!15} 55.73\\
Qwen-VL-Chat & 24.66 & \cellcolor{green!5} 20.42 & \cellcolor{green!5} 11.59 & 48.77 & \cellcolor{gray!15} 22.05\\
SmolVLM & 28.51 & \cellcolor{green!5} 22.11 & \cellcolor{green!5} 19.93 & 36.42 & \cellcolor{gray!15} 24.57\\
SPHINX-V & 27.37 & \cellcolor{green!5} 20.70 & \cellcolor{green!5} 14.49 & 45.67 & \cellcolor{gray!15} 23.26\\
VisualGLM & 9.28 & \cellcolor{green!5} 13.10 & \cellcolor{green!5} 4.47 & 29.94 & \cellcolor{gray!15} 11.63\\
\midrule
\multicolumn{6}{l}{\cellcolor{orange!15}\textit{Chart-related Models}}  \\
\midrule
ChartGemma & 45.93 & \cellcolor{green!5} 28.87 & \cellcolor{green!5} 27.54 & 55.56 & \cellcolor{gray!15} 35.42\\
ChartInstruct & 27.38 & \cellcolor{green!5} 17.75 & \cellcolor{green!5} 20.29 & 24.38 & \cellcolor{gray!15} 21.44\\
ChartLlama & 30.54 & \cellcolor{green!5} 21.55 & \cellcolor{green!5} 18.72 & 41.05 & \cellcolor{gray!15} 25.00\\
ChartVLM & 46.83 & \cellcolor{green!5} 29.01 & \cellcolor{green!5} 35.75 & 36.11 & \cellcolor{gray!15} 35.85\\
DePlot & 60.63 & \cellcolor{green!5} 34.51 & \cellcolor{green!5} 41.30 & 52.78 & \cellcolor{gray!15} 44.53 \\
MatCha & 28.28 & \cellcolor{green!5} 17.04 & \cellcolor{green!5} 18.24 & 29.32 & \cellcolor{gray!15} 21.35 \\
OneChart & 54.48 & \cellcolor{green!5} 37.14 & \cellcolor{green!5} 41.61 & 51.50 & \cellcolor{gray!15} 44.33\\
TinyChart & 57.01 & \cellcolor{green!5} 33.38 & \cellcolor{green!5} 36.11 & 58.64 & \cellcolor{gray!15} 42.45\\
UniChart & 24.66 & \cellcolor{green!5} 18.87 & \cellcolor{green!5} 16.06 & 33.95 & \cellcolor{gray!15} 21.09\\
\midrule
\multicolumn{6}{l}{\cellcolor{orange!15}\textit{Multimodal Agentic Framework (Ours)}}  \\
\midrule
\textbf{\chartagent{}} &  \textcolor{red}{\textbf{84.84}} &  \cellcolor{green!5} \textcolor{red}{\textbf{44.16}} &  \cellcolor{green!5} \textcolor{red}{\textbf{55.93}} &  \textcolor{red}{\textbf{69.14}} &  \cellcolor{gray!15} \textcolor{red}{\textbf{59.69}}\\
\bottomrule
\end{tabular}
\end{adjustbox}
\vspace{0.05mm}
\caption{\textbf{ChartX} (61.7\% unannotated; 71.9\% numeric QA)}
\label{tab:chartx}
\end{subtable}
\label{tab:overall_results}
\end{table*}

\subsection{Baselines}\label{experiments:baselines}
We evaluate against 42 baseline models to ensure a comprehensive comparison:  
\textbf{(A) Proprietary MLLMs:} GPT-4o, GPT-4o-mini, Claude 3 Haiku, Gemini 1.5;  
\textbf{(B) Open-Weight General-Purpose MLLMs:} BLIP-2, CogAgent, CogVLM, DeepSeek-VL2, DocOwl1.5, InstructBLIP, InternVL3, LLaMA-3.2, LLaVA-1.6/1.5/OneVision, mPLUG-Owl3, Phi-3 Vision, Pixtral, Qwen2-VL, Qwen-VL-Chat, SmolVLM, SPHINX-V, VisualGLM;  
\textbf{(C) Chart-Specific MLLMs:} ChartGemma, ChartInstruct, ChartLLaMA, ChartVLM, DePlot, MatCha, OneChart, TinyChart, UniChart. \textit{Concurrent Works:} We additionally include recently released models whose knowledge cutoffs are later than the dataset release or whose launch dates are concurrent with ours: GPT-o3/o4-mini/4.1/5/5-mini, Gemini 2.0 Flash, Claude 3.7 Sonnet/3.5 Sonnet/3.5 Haiku, and Mistral. 
We compare zero-shot and Chain-of-Thought (CoT) prompting; see Appendix~\ref{appendix:baseline_prompts} for the corresponding prompts.
Further details in Appendix~\ref{appendix:baselines} and Table~\ref{appendix_tab:baseline_modelarchs}.

\subsection{Evaluation Metrics} \label{experiments:evaluation}
We use accuracy as the primary evaluation metric, computed via a two-step procedure. First, GPT-4o standardizes both the model’s response and the ground truth—stripping units (e.g., “M” for million, “B” for billion), converting scales, removing symbols, and formatting numbers consistently (see Appendix~\ref{appendix:standardization_examples}). 
If responses are numeric, we then apply an arithmetic correctness check with a strict 5\% relative error tolerance, as commonly adopted in the literature~\cite{masry-etal-2022-chartqa, methani2020plotqa, xu2023chartbench} (see Appendix~\ref{appendix:evaluation_tolerance_analysis} for analysis across multiple numerical tolerance settings); for non-numeric responses, we perform an exact string match after standardization. 
Prior work often uses the LLM-as-a-Judge paradigm~\cite{masry-etal-2023-unichart, masry-etal-2022-chartqa, xia2024chartx, xu2023chartbench}, but we find it suboptimal for numerically precise answers under a 5\% tolerance, as LLMs may inconsistently enforce thresholds or miss small deviations (see Appendix~\ref{appendix:acc_vs_llm_as_judge}). See Appendix~\ref{appendix:prompts_evaluation_metrics} for evaluation prompts.

\section{Results and Analysis}\label{sec:results}
\subsection{Performance}\label{results:main_performance}
\paragraph{\textit{Comparison to State-of-the-art}}

\begin{table*}[htb]
\caption{\textbf{Accuracy on unannotated charts (\%) by chart type.}
\textcolor{red}{\textbf{Red:}} Best,
\textcolor{blue}{\underline{Blue:}} Second best. \textbf{Abbreviations:} Over: Overlay $\vert$ Stack: Stacked $\vert$ Mul: Multi $\vert$ Sing: Single $\vert$ Hor: Horizontal $\vert$ Vert: Vertical $\vert$ B-L: Bar-Line $\vert$ L-L: Line-Line $\vert$ Dir: Directed $\vert$ Undir: Undirected $\vert$ Combo: Combination. See App.~\ref{appendix:chart_types_figures} for examples of each chart type.}
\begin{subtable}{1.0\textwidth}
\centering
\small                                  
\setlength{\tabcolsep}{4pt}
\renewcommand{\arraystretch}{1.1}
\begin{adjustbox}{max width=\linewidth}
\begin{tabular}{l|cc|cccccccc|ccc|cc|cc|cc|ccc|ccc|c|c}
\toprule
\multirow[c]{2}{*}{\textbf{Model}} & \multicolumn{2}{c|}{\textbf{Area}} & \multicolumn{3}{c}{\textbf{Horizontal Bar}} & \multicolumn{2}{c}{\textbf{3D Bar}} & \multicolumn{3}{c|}{\textbf{Vertical Bar}} & \multicolumn{3}{c|}{\textbf{Box}} & \multicolumn{2}{c|}{\textbf{Combo}} & \multicolumn{2}{c|}{\textbf{Line}} & \multicolumn{2}{c|}{\textbf{Node}} & \multicolumn{3}{c|}{\textbf{Pie}} & \multicolumn{3}{c|}{\textbf{Radar}} & \textbf{Scatter} & \cellcolor{gray!15} \multirow[c]{2}{*}{\textbf{Avg.} $ \mathbf{\uparrow}$} \\
\cmidrule(lr{1em}){2-3}
\cmidrule(lr{1em}){4-6}
\cmidrule(lr{1em}){7-8}
\cmidrule(lr{1em}){9-11}
\cmidrule(lr{1em}){12-14}
\cmidrule(lr{1em}){15-16}
\cmidrule(lr{1em}){17-18}
\cmidrule(lr{1em}){19-20}
\cmidrule(lr{1em}){21-23}
\cmidrule(lr{1em}){24-26}
\cmidrule(lr{1em}){27-27}
& Over & Stack & Mul & Sing & Stack & Mul & Stack & Mul & Sing & Stack & Hor & Vert & Stock & B-L & L-L & Mul & Sing & Dir & Undir & Mul & Ring & Sector & Mul & Fill & Sing & 3D & \cellcolor{gray!15} \\
\midrule
\multicolumn{28}{l}{\cellcolor{orange!15}\textit{Proprietary Multimodal Large Language Models}}  \\
\midrule
GPT 4o & 21.0 & 18.0 &  24.0 &   59.0 &10.0 &  20.0 & 6.0 &38.0 & 73.0 &  12.0 &   \textcolor{blue}{\underline{20.0}} &   26.0 &   \textcolor{blue}{\underline{63.0}} &  35.0 &   41.0 & 37.0 &  75.0 & \textcolor{red}{\textbf{91.0}} &   \textcolor{red}{\textbf{91.0}} &  3.0 &  \textcolor{blue}{\underline{32.0}} & \textcolor{blue}{\underline{34.0}} &   \textcolor{blue}{\underline{22.0}} & 20.0 & 6.0 & 63.0 & \cellcolor{gray!15} 36.15 \\
Gemini 1.5 &   5.0 &  4.0 &  28.0 &   52.0 & 7.0 &  14.0 & 4.0 & 39.05 & 49.0 & 5.0 &   13.0 & 18.0 & 24.0 &   28.0 &  5.0 &   7.0 & \textcolor{red}{\textbf{91.0}} &  48.0 & 59.26 &   1.0 &  14.0 &   29.52 & 1.0 & 7.0 & 0.0 & 45.0 & \cellcolor{gray!15} 27.27 \\
\midrule
\multicolumn{28}{l}{\cellcolor{orange!15}\textit{Open-weights Multimodal Large Language Models}}  \\
\midrule
DeepSeek-VL2 & 29.0 & 11.0 & 25.0 & 57.0 & 8.0 & 36.0 & 8.0 & 58.0 & 82.0 & 13.0 & 11.0 & 3.0 & 51.0 & 46.0 & 48.0 & 51.0 & 8.0 & 31.0 & 36.0 & 0.0 & 6.0 & 15.0 & 13.0 & 21.0 & 5.0 & 44.0 &\cellcolor{gray!15} 30.31 \\
InternVL3 &   25.0 & 16.0 &  45.0 &   80.0 & \textcolor{blue}{\underline{19.0}} &  \textcolor{blue}{\underline{38.0}} & 1.0 &44.0 & 80.0 &  16.0 &   16.0 &   23.0 &   60.0 &  27.0 &   24.0 & 30.0 &  56.0 & 62.0 &   52.0 &  0.0 &   2.0 & 9.0 &   \textcolor{red}{\textbf{24.0}} & \textcolor{blue}{\underline{24.0}} & 6.0 &25.0 & \cellcolor{gray!15}  30.92 \\
LLama3.2 & \textcolor{blue}{\underline{46.0}} & 21.0 & 58.0 & \textcolor{blue}{\underline{91.0}} & 11.0 & 31.0 & 4.0 & 71.0 & 89.0 & 10.0 & 6.0 & 6.0 & 49.0 & 42.0 & 46.0 & 63.0 & \textcolor{blue}{\underline{87.0}} & 42.0 & 58.0 & 5.0 & 4.0 & 25.0 & 8.0 & 17.0 & 10.0 & 46.0 &\cellcolor{gray!15} 36.38 \\
Phi3-vision & 27.0 & \textcolor{blue}{\underline{37.0}} & 43.0 & 78.0 & 8.0 & \textcolor{red}{\textbf{40.0}} & 7.0 & 86.0 & \textcolor{blue}{\underline{92.0}}  & 30.0 & 9.0 & 15.0 & 48.0 & 31.0 & 55.0 & \textcolor{blue}{\underline{66.0}} & 84.0 & 39.0 & 51.0 & 2.0 & 14.0 & 21.0 & 11.0 & \textcolor{red}{\textbf{26.0}} & \textcolor{red}{\textbf{66.0}} & \textcolor{red}{\textbf{73.0}} &\cellcolor{gray!15} 40.77 \\
Pixtral & 26.0 & 10.0 & 25.0 & 51.0 & 6.0 & 30.0 & 5.0 & 39.0 & 89.0 & 10.0 & 16.0 & 29.0 & 39.0 & 19.0 & 24.0 & 17.0 & 32.0 & 68.0 & 59.0 & 2.0 & 21.0 & 28.0 & 13.0 & 9.0 & 8.0 & 72.0 & \cellcolor{gray!15} 28.73 \\
Qwen2VL &   \textcolor{red}{\textbf{57.0}} & 18.0 &  \textcolor{red}{\textbf{87.0}} &   \textcolor{red}{\textbf{97.0}} & 17.0 &  \textcolor{red}{\textbf{40.0}} & 7.0 & \textcolor{red}{\textbf{94.0}} & \textcolor{red}{\textbf{97.0}} &  \textcolor{blue}{\underline{24.0}} &   13.0 & 4.0 &  \textcolor{red}{\textbf{64.0}} &  37.0 &   46.0 & \textcolor{red}{\textbf{80.0}} &  85.0 & 80.0 &   \textcolor{blue}{\underline{86.0}} &  1.0 &  12.0 & 9.0 &9.0 &11.0 & 9.0 &47.0 & \cellcolor{gray!15}   \textcolor{blue}{\underline{43.50}} \\
\midrule
\multicolumn{28}{l}{\cellcolor{orange!15}\textit{Chart-related Models}}  \\
\midrule
DePlot &   18.0 &  2.0 &  43.0 &   74.0 &13.0 &  34.0 & \textcolor{blue}{\underline{9.0}} &66.0 & 78.0 &   7.0 &   \textcolor{blue}{\underline{20.0}} &   20.0 &0.0 &  \textcolor{blue}{\underline{48.0}} &   45.0 & 14.0 &  63.0 & \textcolor{blue}{\underline{84.0}}&   73.0 &  4.0 &   3.0 & 5.0 &2.0 & 2.0 & 3.0 & 2.0 & \cellcolor{gray!15}  28.15 \\
TinyChart &   32.0 & 22.0 &  71.0 &   88.0 &13.0 &  37.0 & \textcolor{red}{\textbf{15.0}} &76.0 & 82.0 &  21.0 &2.0 &3.0 &4.0 &  46.0 &   50.0 & 51.0 &  \textcolor{red}{\textbf{91.0}} & 22.0 &   35.0 &  1.0 &  20.0 &21.0 &   10.0 & 8.0 & 4.0 &27.0 & \cellcolor{gray!15}  32.77 \\
\midrule
\multicolumn{28}{l}{\cellcolor{orange!15}\textit{Multimodal Agentic Framework (Ours)}} \\
\midrule
\textbf{\chartagent{}} &  30.0 &  \textcolor{red}{\textbf{38.0}} &  \textcolor{blue}{\underline{79.0}} &  76.0 &  \textcolor{red}{\textbf{82.0}} &  20.0 &  6.0 &  \textcolor{blue}{\underline{88.0}} &  88.0 &  \textcolor{red}{\textbf{76.0}} &  \textcolor{red}{\textbf{89.0}} &  \textcolor{red}{\textbf{83.0}} & \textcolor{red}{\textbf{64.0}} & \textcolor{red}{\textbf{67.0}} & \textcolor{red}{\textbf{65.0}} &  63.0 &  81.0 &  \textcolor{red}{\textbf{91.0}} & \textcolor{red}{\textbf{91.0}} &  \textcolor{red}{\textbf{18.0}} &  \textcolor{red}{\textbf{94.0}} &  \textcolor{red}{\textbf{80.0}} & \textcolor{blue}{\underline{22.0}} &  20.0 &  6.0 &  \textcolor{blue}{\underline{64.0}} & \cellcolor{gray!15}  \textcolor{red}{\textbf{60.81}} \\
\bottomrule
\end{tabular}%
\end{adjustbox}
\label{chartbench_unannotated_results}
\end{subtable}
\label{tab:unannotated-results-mainpaper}
\end{table*}

\begin{figure}[t!]
    \centering
    \includegraphics[width=\linewidth]{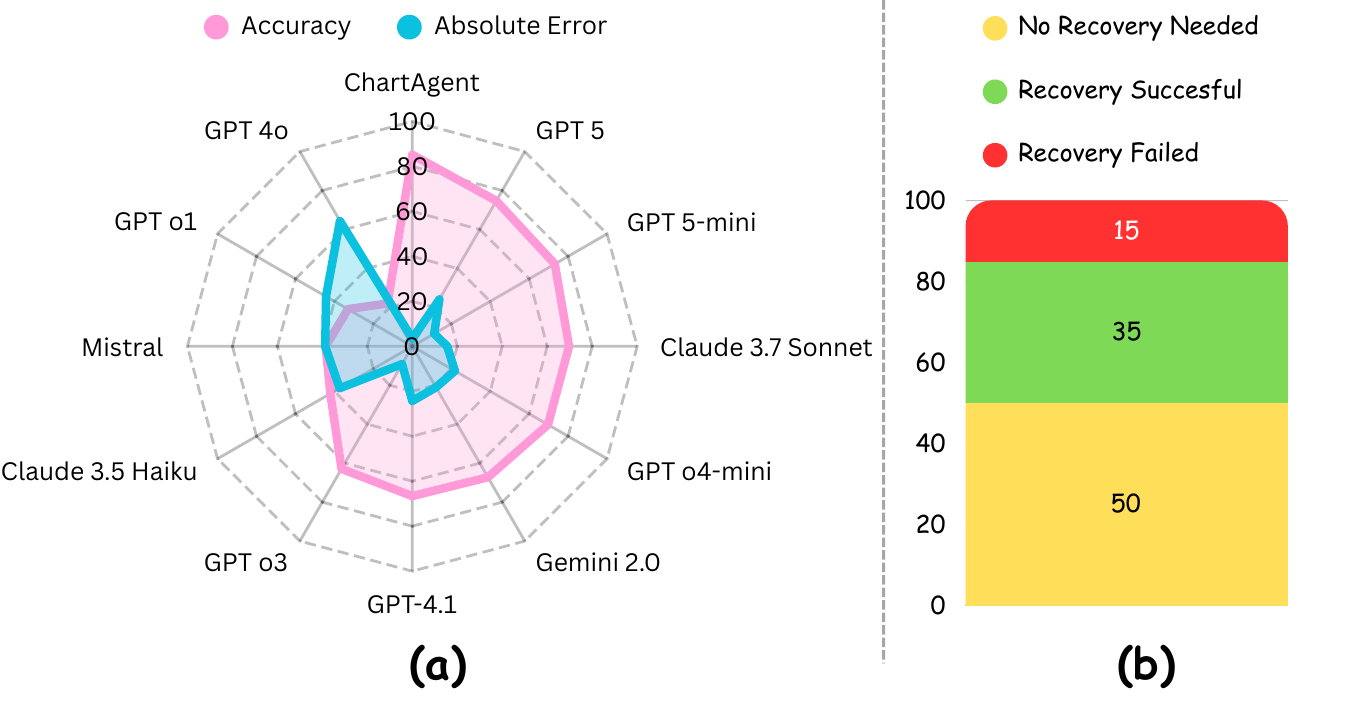}
    \caption{\textbf{(a) Left: \chartagent{} vs. concurrent works:} overall accuracy ($\uparrow$) and average absolute error ($\downarrow$). \textbf{(b) Right: Effectiveness of visual self-verification:} enabled 70\% successful recoveries when invoked.}
    \label{fig:concurrent-self-verify-results-figure}
\end{figure}

Table~\ref{tab:overall_results} presents a comparative analysis of \chartagent{} against 32 baselines on the ChartBench and ChartX benchmarks, stratified by annotation status and QA type.  
\textit{\chartagent{} consistently outperforms all competing methods}, showing particularly strong gains on unannotated charts and numeric QA—the dominant categories across both datasets. On ChartBench, \chartagent{} achieves 71.39\% overall accuracy, a +16.07\% absolute gain over the second-best model (Phi-3 Vision), including 60.81\% on unannotated charts (+17.31\% over Qwen2-VL) and 70.91\% on numeric QA (+15.02\% over Phi-3 Vision). 
As expected, performance on annotated charts remains comparable to GPT-4o, owing to the routing mechanism that preserves both accuracy and computational efficiency. 
A similar trend is observed on ChartX, where \chartagent{} attains 59.69\% overall accuracy (+2.83\% absolute gain over GPT-4o), with top scores on unannotated (44.16\%) and numeric QA (55.93\%). 
Furthermore, Figure~\ref{fig:concurrent-self-verify-results-figure}(a) and Appendix Table~\ref{tab:appendix-concurrent-internal-dataset} present results comparing \chartagent{} with 10 additional concurrent works on a newly curated dataset designed to ensure fair comparison and mitigate potential data leakage (see Appendix~\ref{appendix:concurrent_works}). 
\textit{\chartagent{} outperforms all concurrent models} by a significant margin, achieving a +10.48\% absolute accuracy gain over the second-best model (GPT-5) and a 5.72-point reduction in average absolute error relative to GPT-o3. 
Overall, these results establish \hlpink{\textbf{\chartagent{} as the new SoTA in Chart VQA}}, with major gains in numeric QA on unannotated charts, highlighting the value of visually grounded agentic reasoning for charts.


\paragraph{\textit{Performance by Chart Type}}
Table~\ref{tab:unannotated-results-mainpaper} compares \chartagent{} with the top-10 baselines on unannotated charts, stratified by chart type on ChartBench (see Appendix~\ref{appendix:comparision_chart_type_full_table} for the full table and ChartX results). 
On ChartBench, \chartagent{} achieves the largest gains on Bar (particularly horizontal and stacked variants, up to +65\%), Box (up to +69\%), Combination (Bar-Line, Multi-Line, up to +23\%), and Pie (Ring, Sector, up to +62\%) charts.
On ChartX, the most substantial improvements occur on Bubble, Ring, and Treemap charts.
On ChartX, major gains are observed for Bubble, Ring, and Treemap charts.
See Appendix~\ref{appendix:qualitative_examples} for qualitative examples and trajectories across chart types. 
Overall, these results underscore \hlpink{\textbf{\chartagent{}’s robustness across a wide range of chart types}}.

\subsection{Effectiveness of \chartagent{}}\label{results: effectiveness_chartagent}
\begin{figure*}
\centering
    \includegraphics[scale = 0.335]{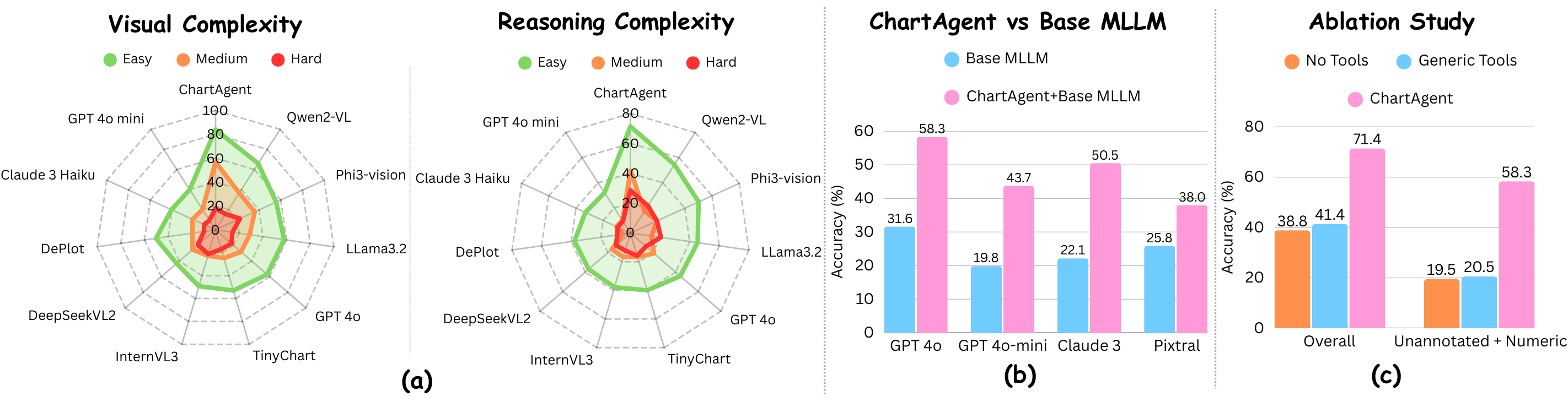}
    \caption{\textbf{Analysis of \chartagent{} Performance.}
    \textbf{(a) Left:} Stratified by visual complexity of charts and reasoning complexity of chart–QA pairs on unannotated charts, compared with top-10 SoTA. 
    \textbf{(b) Middle:} \chartagent{} performance on unannotated+numeric chartQA when instantiated with different base MLLMs. 
    \textbf{(c) Right:} Ablation study comparing \chartagent{} with ReAct using no tools and ReAct with natural image–based generic tools.}
    \label{fig:additional_results_figure}
\end{figure*}
\paragraph{\textit{Performance Across Visual and Reasoning Complexity Levels}}
We analyze \chartagent{}’s performance across difficulty levels, stratified by (1) the \textit{visual complexity of charts} and (2) the \textit{reasoning complexity of chart–QA pairs}, each categorized into three levels: Easy, Medium, and Hard. 
Visual complexity reflects the perceptual effort required to interpret a chart, while reasoning complexity measures the depth of reasoning needed to answer a question.
See Appendix~\ref{appendix:complexity_analysis} for details and statistics, and Appendix~\ref{appendix:prompts_complexity_analysis} for corresponding prompts.
Figure~\ref{fig:additional_results_figure}(a) compares \chartagent{} with the top-10 baselines on unannotated charts, stratified by these complexity levels on ChartBench (see Appendix~\ref{appendix:complexity_results} for full results). All models show a consistent decline from Easy to Hard across both dimensions, confirming that visual clutter and multi-step reasoning increase Chart VQA difficulty. \textit{\chartagent{} achieves the best performance at all levels except visually Hard}, with notable gains on visually Easy (+18\%) and Medium (+20.1\%) charts, and reasoning Easy (+21.2\%) and Medium (+20.8\%) tasks. Visually Hard charts (17.9\%) remain challenging due to 3D, radar, and overlapping structures that obscure segment boundaries and axis references. However, on reasoning Hard tasks involving multi-step numerical reasoning, \chartagent{} still delivers a +6.9\% gain.
A similar pattern is observed on ChartX, where it consistently ranks first or second across both complexity dimensions. 
These results demonstrate \hlpink{\textbf{\chartagent{}'s strong generalization across varying visual and reasoning complexities}} in chart–QA pairs.

\paragraph{\textit{Plug-and-Play Generalization Across MLLMs}}
\chartagent{} follows a plug-and-play design, enabling seamless integration with any MLLM to provide chart-specialized, visually grounded reasoning. To assess generalization beyond GPT-4o as the base MLLM, we evaluate \chartagent{} with three additional models: GPT-4o-mini, Claude 3 Haiku, and Pixtral, covering both closed- and open-source variants. 
Figure~\ref{fig:additional_results_figure}(b) compares the performance of \chartagent{}+Base MLLM versus the Base MLLM alone on unannotated and numeric Chart VQA. \textit{\chartagent{} consistently outperforms its corresponding base models}, yielding absolute accuracy gains of +26.7\% on GPT-4o, +23.9\% on GPT-4o-mini, +28.4\% on Claude 3 Haiku, and +12.2\% on Pixtral. 
Thus, \hlpink{\textbf{\chartagent{} serves as an effective plug-and-play framework that enhances performance across diverse MLLMs}}, demonstrating both robustness and generalization.

\subsection{Additional Analysis}\label{results: additional_analysis}
\paragraph{\textit{Effectiveness of Visual Self-Verification and Recovery}}  
We evaluated \chartagent{}’s ability to detect unsatisfactory tool outputs and recover using its visual self-verification mechanism. Figure~\ref{fig:concurrent-self-verify-results-figure}(b) and Appendix Table~\ref{tab:self_verification_stats} summarize these results. Across 30 randomly sampled trajectories from ChartBench, tool outputs were correct and required no recovery in 50\% of cases. In the remaining 50\%, \chartagent{} correctly flagged unsatisfactory outputs and triggered its self-verification mechanism, \textit{recovering successfully 70\% of the time} and failing 30\%, with the latter contributing to a 15\% overall error rate attributable to unresolved tool-level failures. Thus, \hlpink{\textbf{\chartagent{}’s visual self-verification mechanism is both frequently invoked and often effective}}, enhancing its robustness in the presence of imperfect tool outputs.

\paragraph{\textit{Ablation Study}}
Prior frameworks for visually grounding MLLMs primarily focus on natural images and rely on generic tools such as cropping and zooming~\cite{zheng2025deepeyes, su2025thinking, jegham2025visual, huvisualsketchpad, gupta2023visual, suris2023vipergpt}. While effective for object localization or text spotting, these tools lack the fine-grained capabilities required for structured, quantitative reasoning in charts.
We compare three ReAct-style agents, all using GPT-4o as the base MLLM with visual self-verification: (i) ReAct (No Tools), (ii) ReAct + Natural Image Tools, with generic natural-image operations, and (iii) \chartagent{}. 
All variants use the same 15-step iteration limit.
Figure~\ref{fig:additional_results_figure}(c) shows that \textit{\chartagent{} outperforms both variants} by +32.6\% over ReAct (No Tools) and +30.0\% over ReAct + Image Tools overall, and by +38.8\% and +37.8\% respectively on the unannotated + numeric subset. These findings highlight the \hlpink{\textbf{limitations of generic tools and the necessity of chart-specialized visual grounding}}. See Appendix~\ref{sec:appendix-ablation} for further details.

\paragraph{\textit{Fallback Behavior and Common Triggers}}
We conducted a manual analysis of 30 randomly selected ChartBench trajectories (unannotated, numeric QA) to understand when and why \chartagent{} reverts to the base MLLM. 
The \textit{fallback rate was relatively low (below 10\%)} and was typically triggered by: (1) bar charts with negative or axis-inconsistent bar-height estimates; (2) OCR tools returning \texttt{None} for legends or axis labels; and (3) edge-point detection or interpolation tools producing empty or axis-inconsistent outputs. In such cases, the agent identified tool-based reasoning as unreliable and reverted to the base MLLM, \hlpink{\textbf{a rare but effective fail-safe mechanism}} that helps maintain robustness. See Appendix~\ref{appendix:fallback_analysis} for further details on fallback behavior.

See Appendix~\ref{appendix:extended_results} for extended discussion and analysis on tool usage, inference time, and monetary costs.

\subsection{Failure Mode Analysis}\label{failure_analysis}
We conducted a failure mode analysis to identify common errors in \chartagent{}, which fall into two main categories:  
\textbf{(1) Perception-based failures.}  
These stem from visual misinterpretations such as:  (1.1)~\textit{OCR obstruction} from overlays or dense elements;  
(1.2)~\textit{Poor color contrast} (e.g., white text on yellow background);  
(1.3)~\textit{Legend occlusion} over key regions;  
(1.4)~\textit{Element invisibility} where lines or markers blend with background;  
(1.5)~\textit{Segmentation errors} caused by axis lines overlapping chart elements;  
(1.6)~\textit{Overlapping series} obscuring category distinctions; and  
(1.7)~\textit{Axis interpretation issues} in 3D or multi-axis charts with distorted or inconsistent scales across multiple axis.  
\textbf{(2) Reasoning-based failures.}  
(2.1)~\textit{Incorrect tool choice} (e.g., using area instead of height);  
(2.2)~\textit{Ambiguous queries} (e.g., missing denominators in multi-ring pies);  and
(2.3)~\textit{Label duplication} across hierarchy levels (e.g., “Netflix” as both parent and child).
See Appendix~\ref{appendix:failure_modes} and Figures~\ref{fig:failure_mode_illustrations},~\ref{fig:failure_mode_qual_examples} for details. 
Most failures are perception-driven, originating from tool-level errors rather than high-level reasoning or planning.

\section{Conclusion}\label{sec:conclusion_future_work}
We introduced \chartagent{}, a novel multimodal agentic framework for visually grounded reasoning in charts. 
Inspired by human cognitive strategies of iterative reasoning and annotation-based chart comprehension, \chartagent{} employs a multi-turn, tool-augmented interaction loop to achieve SoTA performance on well-established benchmarks spanning 40+ chart types, surpassing 40+ baselines with particularly strong gains on unannotated charts and numeric QA. 
Comprehensive analyses demonstrate its robustness across varying visual and reasoning complexity levels, its plug-and-play generalization across MLLMs, and the effectiveness of each agent component, supported by a failure mode analysis.
\section{Limitations and Broader Perspective} \label{sec:limitations} 

\textbf{\textit{Limitations and future work:}}
We highlight several remaining challenges and areas for future improvement in \chartagent{}.

\begin{itemize}
    \item \textbf{Task Coverage and Context.} 
    The current approach focuses on question answering, which functions as a core building-block task and can naturally extend to data extraction, summarization, description, and fact-checking. Reliable QA requires accurate perception and reasoning, and once these components are established, downstream tasks can be derived more systematically. Evaluation so far is restricted to single charts; future work will explore multi-chart and slide-level scenarios. Our ICL examples are textual rather than multimodal; integrating visual ICL may improve accuracy but introduces a trade-off between richer supervision and context length. Future work should systematically examine this balance.  

    \item \textbf{Computation and Latency.}
    Inference with large proprietary models (OpenAI, Claude, etc.) adds latency and cost due to the agentic design involving iterative reasoning, tool executions, and verification loops (details in Appendices~\ref{appendix:inference_time_analysis} and~\ref{appendix_sec:monetary_costs}). Despite this overhead, the accuracy gains, particularly on unannotated charts and numeric QA, remain valuable for precision-critical settings. We also outline directions for reducing latency, including parallelization, smart routing, and caching strategies, in Appendix~\ref{appendix:inference_time_analysis}.

    \item \textbf{Vision Tools and Query Handling.}
    While manually designed, our vision tools generalize across 40+ chart types by operating at the component level. Future work includes on-the-fly tool construction and enabling the agent to detect ambiguous queries and request clarification. Finally, since \chartagent{} is designed to be modular and plug-and-play, it can directly benefit from future advances in vision tools (e.g., stronger OCR or segmentation models).

    \item \textbf{Evaluation of Tool-Level Behavior.}
    There is currently no standard method for quantitatively assessing tool-level accuracy because intermediate visual outputs, such as which segment, region, or axis tick should be considered ``correct", do not come with ground-truth annotations. In line with earlier agentic frameworks (e.g., Visual Sketchpad~\cite{huvisualsketchpad}, ViperGPT~\cite{suris2023vipergpt}, VideoAgent~\cite{wang2024videoagent}, VideoAgent2~\cite{zhi2025videoagent2}), we report end-task performance rather than supervising each intermediate tool step. To increase transparency, we provide tool usage statistics (Appendix~\ref{sec:tool_selection}) and analyze error propagation and recovery (Section~\ref{results: additional_analysis}, Figure~\ref{fig:concurrent-self-verify-results-figure} (right)), and include several qualitative agent trajectories illustrating how \chartagent{} interprets and verifies tool outputs (Appendix~\ref{appendix:qual_agent_trajectories}). \chartagent{}’s agent-driven visual self-verification mechanism further mitigates this challenge by allowing the model to internally evaluate tool sufficiency without manual heuristics (details in Appendices~\ref{appendix:self-verification} and~\ref{appendix:adaptive-heuristic-free-self-verify}).

    \item \textbf{Enhancing Coverage for More Chart Types.} While ChartAgent performs strongly on the chart types most common in real-world analytics, future work can further improve performance on harder formats such as 3D and radar plots, which are affected by depth distortion and radial coordinate structures. We plan to explore dedicated processing modules, such as 2D projection correction and angle-to-numerical conversion, to better support these formats.
\end{itemize}

\textbf{\textit{Broader perspective:}} 
Prior work has highlighted the new and unpredictable risks associated with using automated agents in sensitive contexts \cite{aiagents-safety}. We advise against using this framework or MLLM agents to automate critical chart- or image-related tasks without human oversight. Additionally, the resources accompanying this study will be responsibly released for research purposes only. 

\textbf{\textit{Datasets:}} The benchmarks used in this study are publicly available and were curated by previous research. Specifically, we include the following datasets: ChartBench \cite{xu2023chartbench}, ChartX \cite{xia2024chartx}, and ChartQA-unannotated \cite{islam-etal-2024-large}. We abide by their terms of use.

\section*{Acknowledgements}
The authors would like to thank David Westera of J.P. Morgan AI Research for his valuable discussions and feedback on this work.

\section*{Disclaimer}
This paper was prepared for informational purposes by the Artificial Intelligence Research group of JPMorgan Chase \& Co and its affiliates ("J.P. Morgan") and is not a product of the Research Department of J.P. Morgan.  J.P. Morgan makes no representation and warranty whatsoever and disclaims all liability, for the completeness, accuracy or reliability of the information contained herein.  This document is not intended as investment research or investment advice, or a recommendation, offer or solicitation for the purchase or sale of any security, financial instrument, financial product or service, or to be used in any way for evaluating the merits of participating in any transaction, and shall not constitute a solicitation under any jurisdiction or to any person, if such solicitation under such jurisdiction or to such person would be unlawful. 


\bibliography{main}

\newcommand{\thought}[2]{\textbf{\textcolor{blue}{\texttt{THOUGHT #1:}}} #2}
\newcommand{\action}[2]{\textbf{\textcolor{teal}{\texttt{ACTION #1:}}} #2}
\newcommand{\observation}[1]{\textbf{\textcolor{purple}{\texttt{OBSERVATION:}}} #1}
\newcommand{\answer}[1]{\textbf{\textcolor{orange!70!black}{\texttt{ANSWER:}}} #1}
\newcommand{\terminate}[1]{\textbf{\textcolor{red!70!white}{\texttt{TERMINATE.}}} #1}

\definecolor{py-keyword}{RGB}{0,0,0}       
\definecolor{py-func}{RGB}{163,21,21}        
\definecolor{py-arg}{RGB}{0,16,128}          
\definecolor{py-type}{RGB}{38,127,153}       
\definecolor{py-num}{RGB}{9,134,88}          
\definecolor{py-doc}{RGB}{0,128,0}       

\lstdefinestyle{pydoc}{
  basicstyle=\ttfamily\small\color{py-doc},
  breaklines=true,
  breakatwhitespace=false,
  columns=fullflexible,
  keepspaces=false,
  showstringspaces=false,
  xleftmargin=2em,
}

\newpage
\clearpage
\appendix
\section*{Appendix}
\section*{Table of Contents} 
\setcounter{tocdepth}{1}  
\startcontents[sections] 
\printcontents[sections]{}{1}{}

\section{Annotated vs. Unannotated Charts}\label{appendix:annotated_vs_unannotated_examples}
An \textbf{\textit{annotated chart}} contains explicit textual annotations or shortcuts. For instance, in bar charts, exact values may be printed above or inside the bars; in pie charts, percentage labels may appear alongside slices. In some cases, answers to questions may even be embedded in the title or legend. 
\begin{figure*}[htbp]
    \centering
    \includegraphics[scale = 0.27]{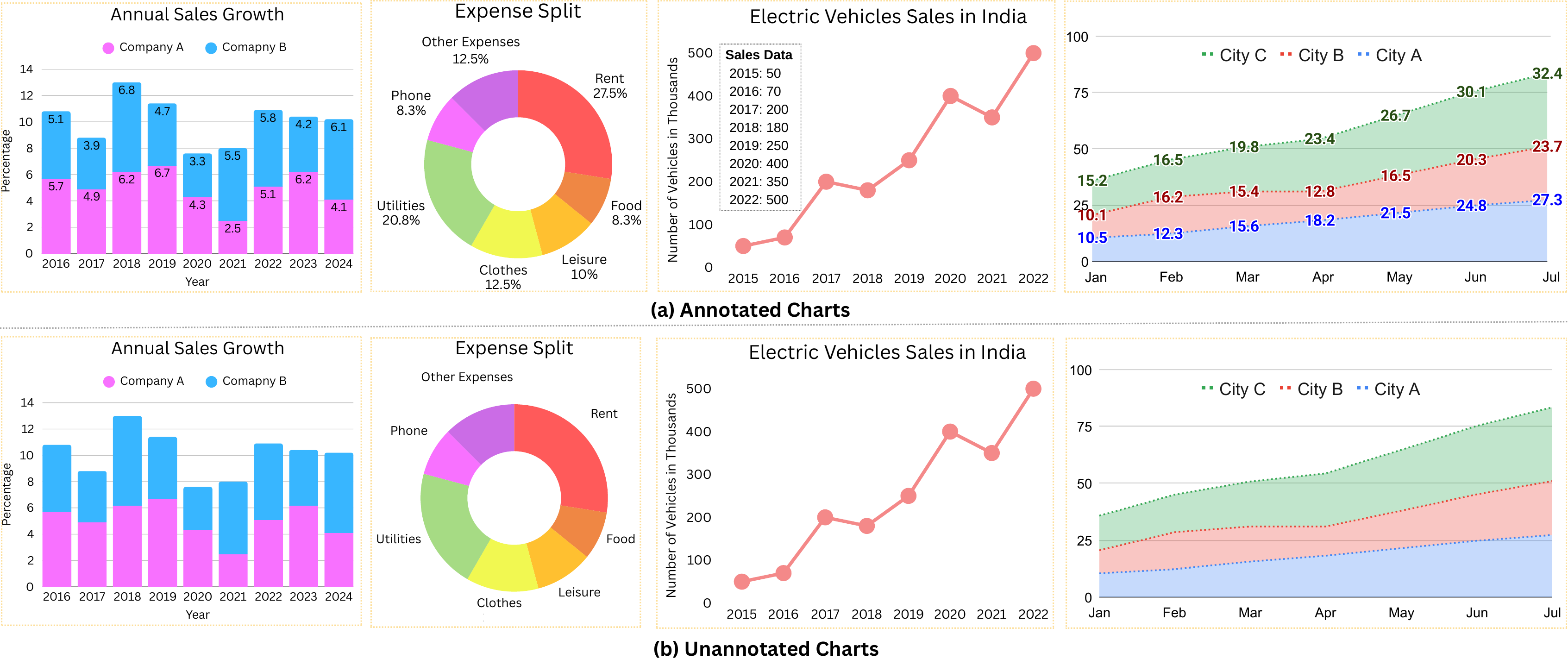}
    \caption{\textbf{Examples of annotated (top) vs. unannotated (bottom) charts.} An \textbf{\textit{annotated chart}} contains explicit textual annotations or shortcuts, whereas an \textbf{\textit{unannotated chart}} lacks such explicit value indicators. For instance, in the first column (top), the bar chart includes printed bar values, while in the corresponding bottom chart, the values must be inferred through visual interpretation.}
    \label{fig:Charttypes_annotation}
    \vspace{-0.3cm}
\end{figure*}
Generally, an annotated chart includes values visibly placed near the relevant graphical elements, though the information may also appear elsewhere within the chart image, such as in captions or legends. These textual cues allow models like GPT-4o to directly extract information from the image, often producing correct answers without requiring complex visual reasoning.

In contrast, \textbf{\textit{unannotated charts}} lack such explicit value indicators. Consequently, the model must infer values by interpreting graphical features—such as bar heights, pie slice angles, or positions along axes. These tasks demand fine-grained visual perception and structured reasoning, often exceeding the capabilities of general-purpose LLMs or MLLMs alone. 
\subsection{Examples}
Figure~\ref{fig:Charttypes_annotation} illustrates this distinction, showing representative examples of both annotated and unannotated charts from the datasets.

\subsection{How \chartagent{} Handles Annotated vs. Unannotated Charts} 
Given a chart, \chartagent{} first classifies it as annotated or unannotated using an LLM-based orchestrator (e.g., GPT-4o). On a uniformly sampled subset, this classification step achieves 100\% accuracy. \chartagent{} dynamically adapts its execution pathway based on the annotation type. For \textit{annotated charts}—where text extraction alone is sufficient—the agent directly forwards the query to the base model (e.g., GPT-4o), which already achieves over 90\% accuracy (see Table \ref{tab:overall_results}). This approach ensures both high performance and computational efficiency. For \textit{unannotated charts}, however, \chartagent{} triggers its full ReAct-style loop. Here, the agent’s iterative reasoning and specialized visual tool use become essential to accurately extract values and answer queries, as detailed in Section~\ref{sec:methods}.

\section{Related Work} \label{appendix:related_works}
We review related work in three areas: chart VQA (\ref{related-work-chartVQA}), MLLMs and visual grounding (\ref{related-work-VisualGrounding}), and agentic frameworks (\ref{related-work-Agents}). 
\subsection{Chart Visual Question Answering}\label{related-work-chartVQA}
Chart visual question answering (Chart VQA) aims to automatically interpret visual charts to answer natural-language queries. Early datasets such as FigureQA~\cite{kahou2017figureqa} and DVQA~\cite{kafle2018dvqa} introduced synthetic charts designed to evaluate specific reasoning skills but lacked real-world diversity. This gap was subsequently addressed by more comprehensive datasets like PlotQA~\cite{methani2020plotqa}, ChartQA~\cite{masry-etal-2022-chartqa}, and EvoChart~\cite{huang2024evochart}, which incorporated complex, real-world charts coupled with natural-language queries. Recent benchmarks such as ChartBench~\cite{xu2023chartbench}, ChartX~\cite{xia2024chartx}, and CharXiv~\cite{wangcharxiv} have further expanded the complexity and diversity of tasks, covering a wide range of chart types and numeric-intensive queries. These benchmarks reflect a growing trend toward datasets that demand sophisticated visual comprehension combined with nuanced quantitative reasoning.

Advancements in chart-focused multimodal large language models (MLLMs)~\cite{zhang2024tinychart, masry-etal-2023-unichart, han2023chartllamamultimodalllmchart, wu2024chartinsights, mukhopadhyay2024unraveling, liu2024mmc, masry-etal-2024-chartinstruct} have demonstrated notable progress by leveraging instruction-tuned datasets and vision-language alignment methods. Alternatively, ChartOCR~\cite{luo2021chartocr} combines computer vision tools and rule-based techniques, such as keypoint detection and chart-specific rules, for enhanced chart understanding. However, recent studies~\cite{xu2023chartbench, razeghi-etal-2024-plot, islam-etal-2024-large} reveal persistent limitations, particularly in precise numerical interpretation tasks involving unannotated charts—visualizations lacking textual shortcuts such as numeric annotations or labels. In particular, ~\cite{xu2023chartbench} showed a significant performance drop when transitioning from annotated charts (containing textual cues) to unannotated charts, highlighting models' dependence on optical character recognition (OCR) rather than genuine visual reasoning. Addressing this limitation requires enhanced visual grounding capabilities that enable accurate interpretation and numerical reasoning directly from graphical elements (e.g., bar heights, segment areas).

Our approach specifically targets this challenge by enhancing MLLMs with modular, specialized vision tools tailored explicitly to the chart domain, thereby significantly improving visual reasoning and grounding in Chart VQA.

\subsection{General-Purpose Multimodal LLMs and Visual Grounding}\label{related-work-VisualGrounding}
While recent chart-specific multimodal models have made notable progress, broader developments in general-purpose multimodal large language models (MLLMs)—such as GPT-4~\cite{achiam2023gpt}, GPT-4o~\cite{hurst2024gpt}, Gemini~\cite{team2023gemini}, and LLaVA~\cite{liu2023visual}, Visual CoT~\cite{shao2024visual}—have significantly advanced general visual reasoning and understanding across various tasks and domains. However, these general-purpose MLLMs also face challenges when tasks demand precise visual grounding and fine-grained interpretation of visual information.

To address these limitations, recent approaches have explored augmenting language and multimodal models with external tools or visual prompting. For instance, ToolFormer~\cite{schick2023toolformer} integrates text-based language models with external APIs, demonstrating improved reasoning through external knowledge retrieval. Simiarly, Visual ChatGPT~\cite{wu2023visual} and MM-ReAct~\cite{yang2023mm} enhance text-only ChatGPT with vision expert tools for multimodal tasks. For MLLMs, ViperGPT~\cite{suris2023vipergpt} and VisProg~\cite{gupta2023visual} generate executable code via LLMs to perform sequences of tool invocations, though their execution follows a fixed plan without flexibility for dynamic adaptation based on intermediate tool outcomes. In contrast, methods like Visual Sketchpad~\cite{huvisualsketchpad} explicitly incorporate intermediate visual results into iterative reasoning, enabling dynamic refinement of action plans based on observed outcomes.

Additionally, visual prompting methods such as Set-of-Marks (SoM)~\cite{yang2023set} augment input images with visual annotations (e.g., bounding boxes or segmentation masks), providing richer context to LLMs for informed reasoning. Inspired by SoM, our approach similarly presents the multimodal agent with explicit visualizations of intermediate tool outputs, enabling visual inspection and informed decision-making at each reasoning step.

Motivated by these advancements, our work extends multimodal LLM capabilities specifically into the chart domain, combining iterative reasoning, dynamic visual prompting, and modular external tools. Unlike fixed-sequence approaches, our framework enables adaptive replanning and precise visual grounding, effectively addressing complex chart interpretation tasks.

\subsection{Agentic Frameworks}\label{related-work-Agents}
The concept of agents—entities capable of perception, cognition, and action—has long been foundational in artificial intelligence research. Traditional agents perceive their environment, reason about possible actions, and execute these actions to achieve specific goals. Recent advances in large language models (LLMs) have inspired a new generation of LLM-based agents, leveraging powerful reasoning capabilities and dynamic interactions with external tools. A notable example of aligning LLM reasoning explicitly with the agent paradigm is the ReAct framework~\cite{yao2023react}, which organizes model interactions into iterative cycles of reasoning (cognition), action execution (action), and observing results (perception). This structured loop allows LLM-based agents to refine their decisions dynamically, closely mirroring traditional agent definitions.

Several software frameworks and platforms now support the practical implementation of LLM-based agents, enabling seamless integration of external tool usage within iterative reasoning loops. Examples include AutoGen~\cite{wu2024autogen}, CrewAI~\cite{crewai}, LangChain~\cite{LangChain}, LangGraph~\cite{langgraph}, and AutoGPT~\cite{autogpt}, each providing flexible infrastructures to orchestrate sophisticated LLM-driven workflows.

Extending this agentic paradigm into multimodal settings has further expanded agent capabilities across diverse applications. Multimodal agents effectively handle tasks in software engineering~\cite{jimenezswe, hongmetagpt}, robotics~\cite{nasirianypivot}, general vision-language reasoning~\cite{liu2025llava, yang2023mm}, and GUI navigation~\cite{xie2024osworld, koh-etal-2024-visualwebarena, zhenggpt, verma2025adaptagent}. These frameworks dynamically combine visual perception with iterative LLM reasoning, adjusting action plans based on multimodal feedback. Chart VQA introduces unique challenges that specifically require chart-oriented perception and numeric reasoning capabilities.

Our proposed \chartagent{} explicitly adopts the ReAct agentic framework~\cite{yao2023react}, integrating iterative multimodal reasoning with carefully designed modular perception tools specifically tailored for chart understanding tasks. The practical implementation of our agent leverages AutoGen~\cite{wu2024autogen}, providing a flexible infrastructure for orchestrating dynamic interactions between the multimodal LLM and external tools, enabling effective iterative refinement and visual grounding.

\section{Datasets}\label{appendix:datasets_details}
To evaluate our agent’s ability to understand charts, we design experiments that require complex visual reasoning, specifically focusing on question answering over \emph{unannotated} charts, where accurate numerical interpretation and output precision are critical. We evaluate \chartagent{} on two well-established and widely used chart QA benchmarks: ChartBench \cite{xu2023chartbench} and ChartX \cite{xia2024chartx}. These benchmarks are visually grounded—models must interpret the visual logic of the chart to answer questions, without relying solely on OCR. They are designed to assess chart comprehension and data reliability through complex reasoning, and the majority of their charts are unannotated (see Appendix~\ref{appendix:annotated_vs_unannotated_examples}), making them ideal for testing visual understanding. 

\subsection{ChartBench}
\textbf{\textit{ChartBench}}~\cite{xu2023chartbench} comprises charts from 9 major categories and 42 subcategories, with unannotated charts present across all 9 categories and over 75\% of images being unannotated. 
It includes both regular chart types (line, bar, pie) and diverse, complex types such as area, box, radar, scatter, node, and combination charts (e.g., bar+line, bar+pie). The test set originally contained 2,100 images (50 per subcategory), but we discarded 4 subcategories with corrupted or incorrect ground-truth labels, yielding a final set of 1,900 images. We use two subsets of the ChartBench QA data: Numeric Question Answering (NQA) and Value Extraction (VE), resulting in 3,800 image-QA pairs. ChartBench includes two primary types of questions:
1) \textit{Numeric QA} — questions requiring precise numerical extraction (e.g., ``What is the value of India in 2021?'' or ``How much more is A than B?'');
2) \textit{Relationship QA} — questions involving relational understanding (e.g., ``Is node A connected to node B?'' or ``Is node A directed toward node B?'').

\subsection{ChartX}
\textbf{\textit{ChartX}}~\cite{xia2024chartx} comprises charts from 18 categories, including regular types such as line, bar, and pie charts, as well as fine-grained and domain-specific charts such as ring charts, radar charts, box plots, 3D-bar charts, histograms, treemaps, rose charts, bubble charts, multi-axes charts, area charts, heatmaps, funnels, and candlestick charts. The dataset includes 1,152 image-question pairs in the test set, with more than 60\% of the images being unannotated. 
ChartX includes two primary types of questions: 1) Numeric QA — questions that require precise numerical extraction; 2) Value Comparison and Global Perception QA — questions that require relative or extremum-based reasoning (e.g., identifying the highest, lowest, or most relevant entity), where exact values are not necessary. Examples of global perception questions include:
``Which country has the highest GDP?'', ``Which region planted the most trees?'', ``Are there more trees planted in 2021 in region A or region B?''

It is important to note that \textbf{ChartX is a much harder dataset, both in terms of questions and chart samples}. The questions are more varied and open-ended; for example, ``How many countries have CO$_2$ emissions greater than or equal to 350 million metric tons?'' and ``How many nonprofits received donations in the range of 50K to 100K?'' require computing all entries and then applying careful numeric filtering, which increases error susceptibility. The chart samples themselves are also more challenging: a significant fraction are occluded charts, where legends often overlap bars or chart elements of interest; many multi-axis plots involve three or more Y-axes; and in some cases grid lines are the same color as the bar or the box in box plots, making it difficult to distinguish regions of interest even after segmentation. Overall, ChartX presents a substantially more challenging testbed.

\subsection{Dataset Statistics}
Table~\ref{table:datasets_stats_anno_chart_types} presents the chart type, annotation, and QA type distribution across the two evaluation datasets, ChartBench and ChartX.
\begin{table*}[htbp]
\caption{\textbf{Dataset Statistics.} Chart type, annotation, and QA type distribution in the evaluation datasets.}
\label{table:datasets_stats_anno_chart_types}
\centering
\subfloat[\textbf{ChartBench (3800 Image-QA pairs):} Over 75\% unannotated charts; approximately 95\% numeric QA.]{
\begin{minipage}{\linewidth}
\centering
\resizebox{\linewidth}{!}{%
\begin{tabular}{@{}l|cc|ccc|ccc|ccc@{}}
\toprule
\textbf{Chart Type} & \textbf{\% Annotated} & \textbf{\% Unannotated} & 
\multicolumn{3}{c|}{\textbf{Regular Types}} & 
\multicolumn{6}{c}{\textbf{Extra (Diverse/Complex) Types}} \\
\cmidrule(r){4-6} \cmidrule(l){7-12}
& & & Line & Bar & Pie &
Area & Box & Radar & Scatter & Node & Combination \\
\midrule
ChartBench & 23.80\% & \textcolor{red}{\textbf{76.20\%}} &
11.90\% & 31.00\% & 11.90\% &
7.10\% & 7.10\% & 9.50\% &
7.10\% & 4.80\% & 11.90\% \\
\bottomrule
\end{tabular}
}
\medskip

\resizebox{0.45\linewidth}{!}{%
\begin{tabular}{@{}l|cc@{}}
\toprule
\textbf{QA Type} & \textbf{\% Numeric QA} & \textbf{\% Non-Numeric QA} \\
\midrule
ChartBench & \textcolor{red}{\textbf{94.74\%}} & 5.26\% \\
\bottomrule
\end{tabular}
}
\end{minipage}
}

\bigskip

\subfloat[\textbf{ChartX (1152 Image-QA pairs):} Over 60\% unannotated charts; over 70\% numeric QA.]{
\begin{minipage}{\linewidth}
\centering
\resizebox{\linewidth}{!}{%
\begin{tabular}{@{}l|cc|ccc|ccccccccccccc@{}}
\toprule
\textbf{Chart Type} & \textbf{\% Annotated} & \textbf{\% Unannotated} & 
\multicolumn{3}{c|}{\textbf{Regular Types}} & 
\multicolumn{13}{c}{\textbf{Extra (Diverse/Complex) Types}} \\
\cmidrule(r){4-6} \cmidrule(l){7-19}
& & & Line & Bar & Pie &
Area & Box & Radar & Ring & 3D-Bar & Histogram & Treemap & Rose & Bubble & Multi-axes & Heatmap & Funnel & Candlestick \\
\midrule
ChartX & 38.28\% & \textcolor{red}{\textbf{61.72\%}} &
17.36\% & 17.36\% & 8.68\% &
4.34\% & 4.34\% & 4.34\% &
4.34\% & 4.34\% & 4.34\% &
4.34\% & 4.34\% & 4.34\% & 4.34\% & 4.51\% & 4.34\% & 4.34\% \\
\bottomrule
\end{tabular}
}
\medskip

\resizebox{0.45\linewidth}{!}{%
\begin{tabular}{@{}l|cc@{}}
\toprule
\textbf{QA Type} & \textbf{\% Numeric QA} & \textbf{\% Non-Numeric QA} \\
\midrule
ChartX & \textcolor{red}{\textbf{71.88\%}} & 28.12\% \\
\bottomrule
\end{tabular}
}
\end{minipage}
}
\end{table*}
A key observation is the dominance of unannotated charts, which constitute over 76\% of ChartBench and over 61\% of ChartX. As discussed in Appendix~\ref{appendix:annotated_vs_unannotated_examples}, such unannotated samples require visual extraction of values from chart elements rather than relying on textual annotations or shortcuts, thereby posing greater difficulty. Another important characteristic is the prevalence of numeric QA, comprising more than 94\% in ChartBench and nearly 72\% in ChartX. Taken together, these properties underscore that both datasets serve as rigorous testbeds for evaluating chart reasoning systems under visually demanding and numerically intensive conditions.

Note that we did not use the popular ChartQA~\cite{masry-etal-2022-chartqa} dataset, as all charts are annotated and MLLM performance on it already exceeds 85\% due to strong OCR capabilities.
We also excluded the CharXiv~\cite{wangcharxiv} dataset, as it lacks numerically precise questions—only approximately 20\% of its data involves numeric QA on unannotated charts. In contrast, a key strength and focus of our framework is unannotated numeric ChartQA, where most current SOTA models struggle. CharXiv primarily emphasizes descriptive and reasoning-based queries rather than precise numeric extraction.
Thus, ChartBench and ChartX were selected for evaluation as they emphasize unannotated charts and require models to demonstrate true visual understanding and numerical reasoning beyond text extraction. 
See Appendix~\ref{appendix:chart_types_figures} for visualizations of the diverse chart types included in our benchmark datasets.

\section{Chart Types Supported in \chartagent{}} \label{appendix:chart_types_figures}
\begin{figure*}[htb]
    \centering
    \includegraphics[width=0.94\textwidth, height = 0.9\textheight]{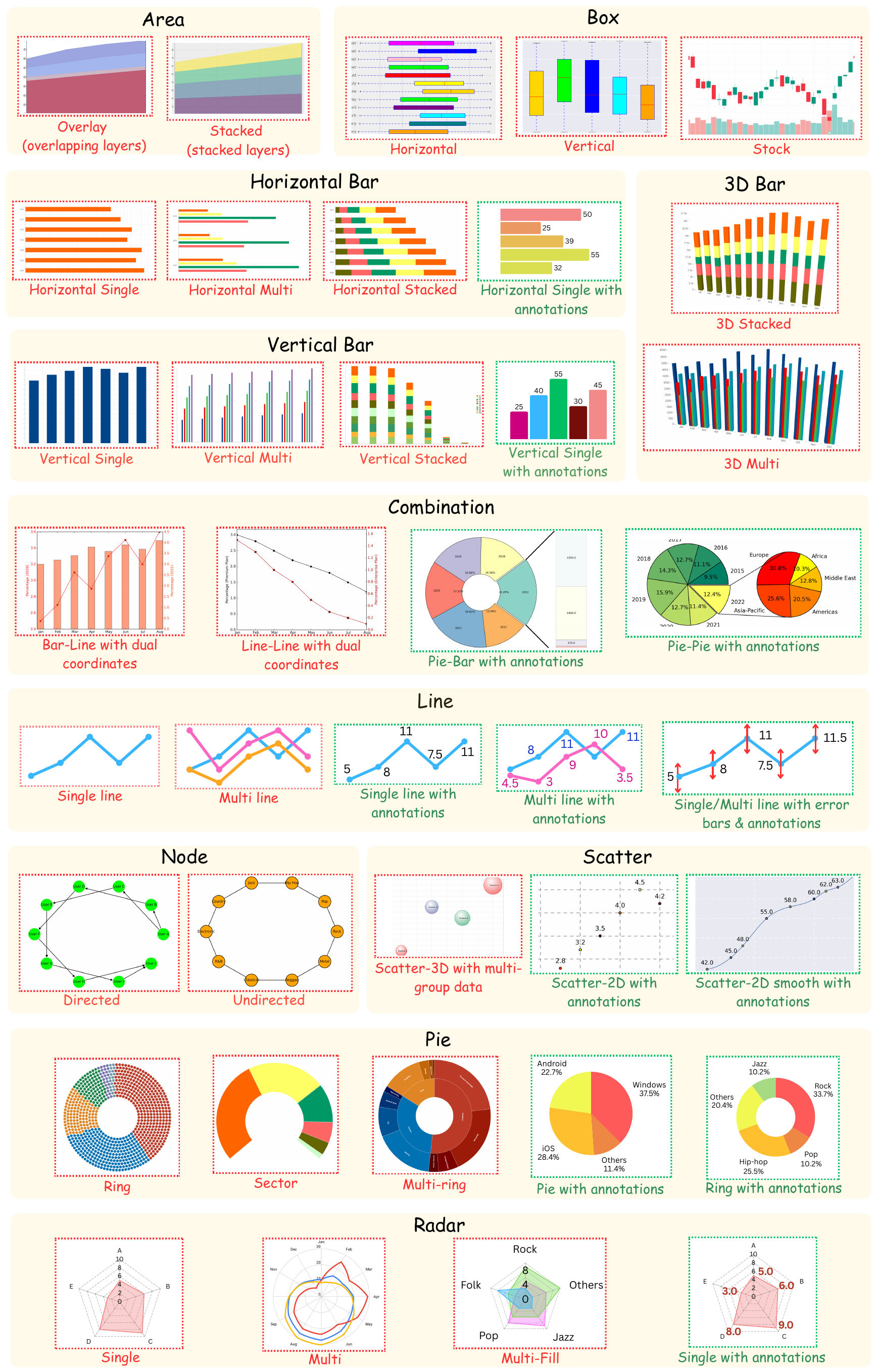}
    \caption{\textbf{Chart types in the ChartBench dataset}: 9 major types with 38 subtypes (excluding 4 subtypes with corrupted or incorrect ground-truth labels). Annotated subtypes are marked in \textcolor{green!50!black}{green}, and unannotated subtypes are marked in \textcolor{red}{red}. Over 75\% of the data is unannotated, making ChartBench a robust testbed for visual reasoning in charts.}
    \label{fig:ChartBench-charttypes}
\end{figure*}

\begin{figure*}[htb]
    \centering
    \includegraphics[width=\textwidth, height = 0.5\textheight]{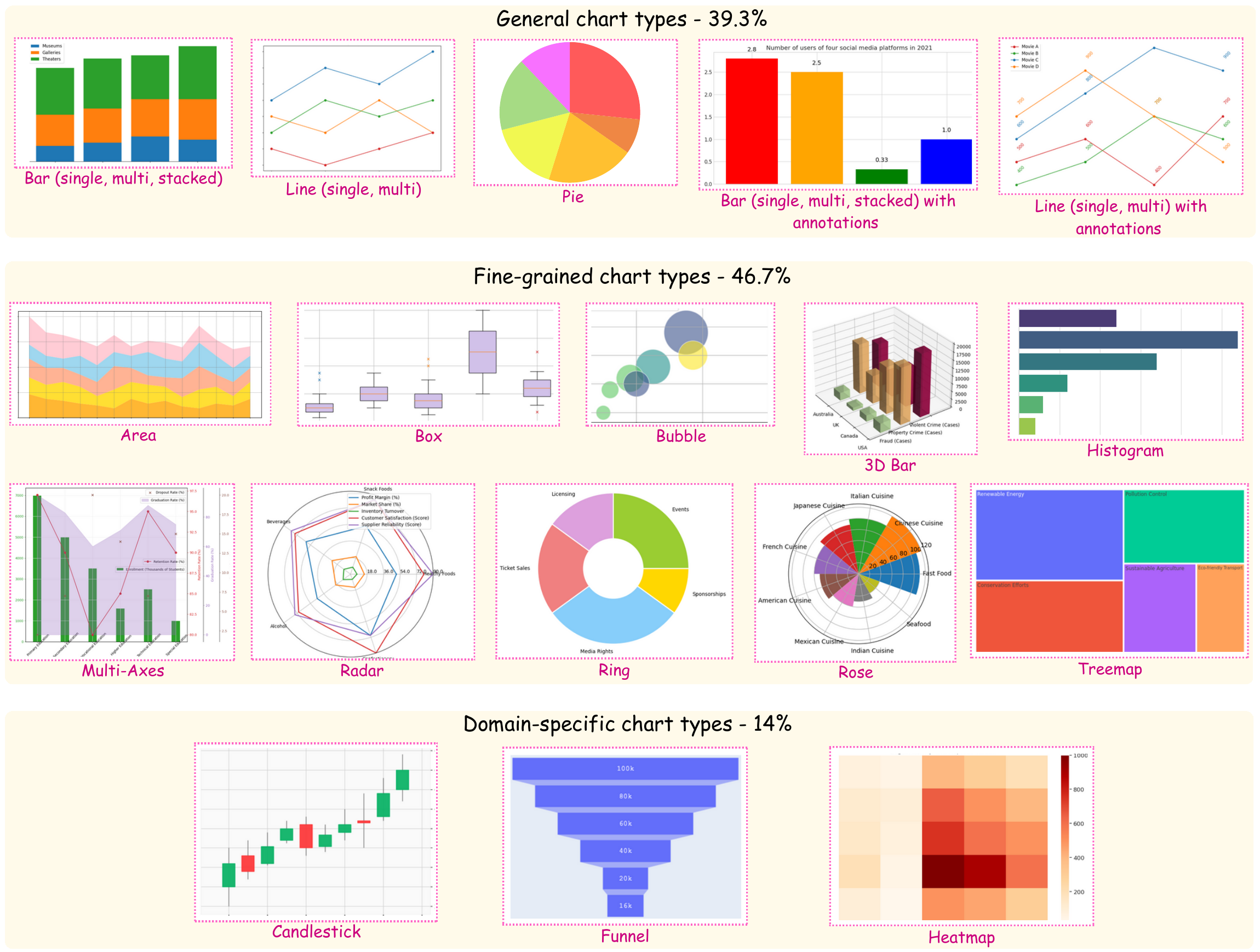}
    \caption{\textbf{Chart types in the ChartX dataset}: 18 types organized into three subcategories—general, fine-grained, and domain-specific chart types, with the percentage of data in each subcategory indicated. Over 60\% of the data is unannotated, making ChartX a robust testbed for visual reasoning in charts.}
    \label{fig:ChartX-charttypes}
\end{figure*}

\chartagent{} supports a wide range of chart types across both the ChartBench and ChartX datasets. Specifically, ChartBench contains 9 major categories and 38 subcategories of charts (excluding 4 with corrupted or incorrect ground-truth labels), while ChartX comprises 18 types organized into three subcategories—general, fine-grained, and domain-specific. 
The majority of charts in both datasets are unannotated, making them an ideal testbed for evaluating visual reasoning in charts. 
Figure~\ref{fig:ChartBench-charttypes} illustrates examples of each ChartBench chart type, and Figure~\ref{fig:ChartX-charttypes} presents the corresponding examples from the ChartX dataset.

\clearpage
\onecolumn
\section{Baselines}\label{appendix:baselines}
Table~\ref{appendix_tab:baseline_modelarchs} summarizes the model architecture details of all baseline MLLMs compared in our experiments, including both proprietary and open-weight models—covering general-purpose as well as chart-specific open-weight MLLMs. See Appendix~\ref{appendix:implementation_details_baselines} for implementation details and Appendix~\ref{appendix:baseline_prompts} for prompts.
\begin{table*}[!htbp]
\centering
\captionof{table}{\textbf{Model architectures of baseline MLLMs} considered in our experiments, including both proprietary and open-weight models—covering general-purpose and chart-based open-weight MLLMs. We report the model version (for proprietary models) or the underlying component architectures (for open-weight models), along with the name and parameter sizes of the vision encoder and language model (where applicable), and official access links. 
Concurrent works with knowledge cutoff dates after the release of our benchmark datasets (ChartBench, ChartX) are highlighted in \colorbox{orange!15}{orange}.
}
\label{appendix_tab:baseline_modelarchs}
\begin{subtable}{\linewidth}
\centering
\begin{adjustbox}{max width=0.7\linewidth}
\begin{tabular}{lll}
\toprule
\makecell{\textbf{Model}} & \makecell{\textbf{Version}} & \makecell{\textbf{Link}}\\
\midrule
GPT 4o~\cite{hurst2024gpt} & \texttt{gpt-4o-2024-08-06} & \href{https://platform.openai.com/docs/models/gpt-4o}{OpenAI\slash gpt4o}\\
GPT 4o-mini~\cite{GPT4omini} & \texttt{gpt-4o-mini-2024-07-18} & \href{https://platform.openai.com/docs/models/gpt-4o-mini}{OpenAI\slash gpt4o-mini}\\
\cellcolor{orange!15} GPT 5~\cite{GPT5system} & \texttt{gpt-5-2025-08-07} & \href{https://platform.openai.com/docs/models/gpt-5}{OpenAI\slash gpt5}\\
\cellcolor{orange!15} GPT 5-mini~\cite{GPT5system} & \texttt{gpt-5-mini-2025-08-07} & \href{https://platform.openai.com/docs/models/gpt-5-mini}{OpenAI\slash gpt5-mini}\\
Gemini 1.5 flash~\cite{geminiteam2024gemini15unlockingmultimodal} & \texttt{gemini-1.5-flash-002} & \href{https://cloud.google.com/vertex-ai/generative-ai/docs/learn/model-versions}{Google\slash gemini-1.5-flash}\\
Claude 3 Haiku~\cite{TheC3} & \texttt{claude-3-haiku-20240307} & \href{https://docs.anthropic.com/en/docs/about-claude/models/overview}{Anthropic\slash claude-3-haiku}\\
GPT o1~\cite{openai2024o1systemcard} & \texttt{o1-2024-12-17} &  \href{https://platform.openai.com/docs/models/o1}{OpenAI\slash o1}\\
\cellcolor{orange!15} GPT o4-mini~\cite{openai2023o3o4} & \texttt{o4-mini-2025-04-16} & \href{https://platform.openai.com/docs/models/o4-mini}{OpenAI\slash o4-mini}\\
\cellcolor{orange!15} GPT o3~\cite{openai2023o3o4} & \texttt{o3-2025-04-16} &  \href{https://platform.openai.com/docs/models/o3}{OpenAI\slash o3}\\
\cellcolor{orange!15} GPT o4-mini~\cite{openai2023o3o4} & \texttt{o4-mini-2025-04-16} & \href{https://platform.openai.com/docs/models/o4-mini}{OpenAI\slash o4-mini}\\
\cellcolor{orange!15} GPT 4.1~\cite{openai_gpt41} & \texttt{gpt-4.1-2025-04-14} & \href{https://platform.openai.com/docs/models/gpt-4.1}{OpenAI\slash gpt4.1}\\

\cellcolor{orange!15} Gemini 2.0 flash~\cite{gemini2_flash} &  \texttt{gemini-2.0-flash-001} & \href{https://cloud.google.com/vertex-ai/generative-ai/docs/models/gemini/2-0-flash}{Google\slash gemini-2.0-flash}\\
\cellcolor{orange!15} Claude 3.7 Sonnet~\cite{claude37_sonnet} & \texttt{claude-3-7-sonnet-20250219} & \href{https://docs.anthropic.com/en/docs/about-claude/models/overview}{Anthropic\slash claude-3.7-sonnet}\\
\cellcolor{orange!15} Claude 3.5 Sonnet~\cite{claude35_haiku_sonnet} & \texttt{claude-3-5-sonnet-20240620} & \href{https://docs.anthropic.com/en/docs/about-claude/models/overview}{Anthropic\slash claude-3.5-sonnet}\\
\cellcolor{orange!15} Claude 3.5 Haiku~\cite{claude35_haiku_sonnet} & \texttt{claude-3-5-haiku-20241022} & \href{https://docs.anthropic.com/en/docs/about-claude/models/overview}{Anthropic\slash claude-3.5-haiku}\\
\bottomrule
\end{tabular}
\end{adjustbox}
\vspace{0.2cm}
\caption{Proprietary Multimodal Large Language Models}
\label{appendix_tab:baseline_modelarchs_closedweight}
\end{subtable}

\vspace{0.3em}

\begin{subtable}{\linewidth}
\centering
\begin{adjustbox}{max width=0.9\linewidth}
\begin{tabular}{llllll}
\toprule
\makecell{\textbf{Model}} & \multicolumn{2}{c}{\makecell{\textbf{Vision} \\ \textbf{Encoder}}} & \multicolumn{2}{c}{\makecell{\textbf{Language} \\ \textbf{Model}}} & \makecell{\textbf{Link}} \\
\cmidrule(lr){2-3} \cmidrule(lr){4-5}
 & \textbf{Name} & \textbf{Size} & \textbf{Name} & \textbf{Size} & \\

\midrule
\multicolumn{6}{c}{\textbf{General$-$Purpose MLLMs}} \\ 
\midrule
BLIP-2~\cite{li2023blip} & EVA-CLIP ViT & 1.1B & OPT & 2.7B & \href{https://huggingface.co/Salesforce/blip2-opt-2.7b}{Salesforce\slash blip2-opt-2.7b} \\
CogAgent~\cite{hong2023cogagent} & EVA2-CLP-E & 11B & Vicuna1.5 & 7B & \href{https://huggingface.co/THUDM/cogagent-vqa-hf}{THUDM\slash cogagent-vqa-hf} \\
CogVLM~\cite{wang2023cogvlm} & EVA2-CLP-E & 11B & Vicuna1.5 & 7B & \href{https://huggingface.co/THUDM/cogvlm-chat-hf}{THUDM\slash cogvlm-vqa-hf} \\
DeepSeek-VL2~\cite{wu2024deepseekvl2mixtureofexpertsvisionlanguagemodels} & SigLIP-SO400M & 878M & DeepSeek MoE & 16.4B & \href{https://huggingface.co/deepseek-ai/deepseek-vl2-small}{deepseek-ai\slash deepseek-vl2-small} \\
DocOwl1.5-Chat~\cite{hu2024mplug} & ViT/L & 304M & Llama2 & 7B & \href{https://github.com/X-PLUG/mPLUG-DocOwl/tree/main/DocOwl1.5}{mPLUG-DocOwl\slash DocOwl1.5} \\
InstructBLIP~\cite{instructblip} & BLIP2-QFormer & 188M & Vicuna & 7B & \href{https://github.com/salesforce/LAVIS/tree/main/projects/instructblip}{LAVIS\slash instrucblip} \\
InternVL3~\cite{zhu2025internvl3exploringadvancedtraining} & InternViT & 300M & Qwen2.5 & 1.5B & \href{https://huggingface.co/OpenGVLab/InternVL3-2B}{OpenGVLab\slash InternVL3-2B}\\
LLama3.2~\cite{grattafiori2024llama} & ViT & 630M & Llama3.1 & 8B & \href{https://huggingface.co/meta-llama/Llama-3.2-11B-Vision}{meta-llama\slash Llama-3.2-11B-Vision} \\
Llava1.6~\cite{llavanext} & CLIP ViT & 304M & Mistral & 6.74B & \href{https://huggingface.co/llava-hf/llava-v1.6-mistral-7b-hf}{llava-hf\slash llava-v1.6-mistral-7b-hf} \\
Llava1.5~\cite{llava1.5} & CLIP ViT & 304M & Vicuna1.5 & 7B & \href{https://huggingface.co/liuhaotian/llava-v1.5-13b}{liuhaotian\slash llava-v1.5-7b}\\
LlaVA-OneVision~\cite{llava-onevision} & SigLIP & 894M & Qwen2 & 494M & \href{https://huggingface.co/lmms-lab/llava-onevision-qwen2-7b-ov}{lmms-lab\slash llava-onevision-quen2-0.5b-ov}\\
mPLUG-Owl3~\cite{mplugowl3} & SigLIP & 400M & Qwen2 & 7B & \href{https://huggingface.co/mPLUG/mPLUG-Owl3-7B-240728}{mPLUG\slash mPLUG-Owl3-7B-240728} \\
Phi3-vision~\cite{phi3} & CLIP ViT & 428M & Phi-3 Mini & 3.8B & \href{https://huggingface.co/microsoft/Phi-3-vision-128k-instruct}{microsoft\slash Phi-3-vision-128k-instruct} \\
Pixtral~\cite{agrawal2024pixtral} & Pixtral-ViT & 400M & Mistral-Nemo & 12B & \href{https://huggingface.co/mistralai/Pixtral-12B-2409}{mistralai\slash Pixtral-12B-2409} \\
Qwen2-VL~\cite{wang2024qwen2vlenhancingvisionlanguagemodels} & ViT & 675M & QwenLM & 7.6B & \href{https://huggingface.co/Qwen/Qwen2-VL-7B-Instruct}{Qwen\slash Qwen2-VL-7B-Instruct} \\
Qwen-VL-Chat~\cite{Qwen-VL} & ViT-bigG & 1.9B & QwenLM & 7.7B &  \href{https://huggingface.co/Qwen/Qwen-VL-Chat}{Qwen\slash Qwen-VL-Chat} \\
SmolVLM~\cite{marafioti2025smolvlmredefiningsmallefficient} & SigLIP & 428M & SmolLM2 & 1.7B & \href{https://huggingface.co/HuggingFaceTB/SmolVLM-Instruct}{HuggingFaceTB\slash SmolVLM-Instruct} \\
SPHINX-V~\cite{lin2025drawandunderstand} & ViT-H SAM  & 636M & Llama2 & 13B & \href{https://github.com/AFeng-x/Draw-and-Understand.git}{AFeng-x\slash Draw-and-Understand} \\
VisualGLM~\cite{glm2024chatglmfamilylargelanguage} & BLIP2-QFormer & 188M & ChatGLM & 6.2B & \href{https://huggingface.co/THUDM/visualglm-6b}{THUDM\slash visualglm-6b} \\
\cellcolor{orange!15} Mistral~\cite{mistralsmall3} & \multicolumn{4}{c}{Mistral-small 22.2B} &  \href{https://huggingface.co/mistralai/Mistral-Small-Instruct-2409}{mistralai\slash Mistral-Small-Instruct-2409} \\

\midrule
\multicolumn{6}{c}{\textbf{Chart$-$related MLLMs}} \\ 
\midrule
ChartGemma~\cite{masry-etal-2025-chartgemma} & \multicolumn{4}{c}{PaliGemma 3B} & \href{https://huggingface.co/ahmed-masry/chartgemma}{ahmed-masry\slash chartgemma} \\
ChartInstruct~\cite{masry-etal-2024-chartinstruct} & UniChart & 201M & LLama2 & 7B & \href{https://huggingface.co/ahmed-masry/ChartInstruct-LLama2}{ahmed-masry\slash ChartInstruct-LLama2} \\
ChartLlama~\cite{han2023chartllamamultimodalllmchart} & \multicolumn{4}{c}{LLaVA1.5 13B} & \href{https://github.com/tingxueronghua/ChartLlama-code.git}{tingxueronghua\slash ChartLlama-code} \\
ChartVLM~\cite{xia2024chartx} & Pix2Struct & 282M & Vicuna1.5 & 7B & \href{https://huggingface.co/U4R/ChartVLM-base}{U4R\slash ChartVLM-base} \\
DePlot~\cite{liu-etal-2023-deplot} & \multicolumn{4}{c}{Pix2Struct 282M} & \href{https://huggingface.co/google/deplot}{google\slash deplot} \\
MatCha~\cite{liu-etal-2023-matcha} & \multicolumn{4}{c}{Pix2Struct 282M} & \href{https://huggingface.co/google/matcha-chartqa}{google\slash matcha-chartqa}\\
OneChart~\cite{chen2024onechart} & SAM-base & 86M & OPT & 125M & \href{https://huggingface.co/kppkkp/OneChart/tree/main}{kppkkp\slash OneChart} \\
TinyChart~\cite{zhang2024tinychart} & \multicolumn{4}{c}{TinyLLaVA 3.1B} & \href{https://huggingface.co/mPLUG/TinyChart-3B-768}{mPLUG\slash TinyChart-3B-768} \\
UniChart~\cite{masry-etal-2023-unichart} &  \multicolumn{4}{c}{Donut-base 201M} & \href{https://github.com/vis-nlp/UniChart.git}{vis-nlp\slash UniChart}\\
\bottomrule
\end{tabular}
\end{adjustbox}
\vspace{0.01cm}
\caption{Open-weight Multimodal Large Language Models}
\label{appendix_tab:baseline_modelarchs_openweight}
\end{subtable}
\end{table*}


\section{Taxonomy of Tools in \chartagent{}}\label{appendix:vision-tool-description}
Table~\ref{appendix_tab:vision_tool_description} provides a summary and description of the key vision and analytical tools used in \chartagent{}. 
\begin{table*}[h!]
\caption{\textbf{Taxonomy of Tools in \chartagent{}.} Summary of key vision and analytical tools used in \chartagent{}.}
\centering
\begin{adjustbox}{max width=\linewidth}
\begin{tabular}{p{2.2cm}|l|p{13cm}}
\toprule
 \textbf{Chart Type} & \textbf{Chart Tool} & \textbf{Description} \\
 \midrule
\multicolumn{3}{l}{\cellcolor{pale4}\textit{Universal Tools}}\\
\midrule
\multirow{21}{2.2cm}{\colorbox{gray!20}{All}} & \colorbox{gray!10}{\texttt{annotate\_legend}} & Detects legend coordinates, crops the legend, and annotates it with numeric labels. \textbf{\textit{Returns the cropped and annotated legend image}} along with label mappings. \\
\cmidrule{2-3}
& \colorbox{gray!10}{\texttt{get\_marker\_rgb}} &  \textbf{\textit{Retrieves the dominant RGB color of a legend marker}}, either by label (from an annotated legend image) or by associated text. \\
\cmidrule{2-3}
& \colorbox{gray!10}{\texttt{clean\_chart\_image}} & \textit{\textbf{Detects and removes the title and legend}} (if present) from the chart image to avoid interference with downstream visual analysis such as OCR, segmentation, or edge detection. \\
\cmidrule{2-3}
& \colorbox{gray!10}{\texttt{segment\_and\_mark}} & \textbf{\textit{Segments an input image}} using the specified model and applies post-processing to clean the masks. This includes a multi-step filtering pipeline that removes small, duplicate, composite, and background-dominated masks. \textbf{\textit{Returns a labeled image with drawn contours and optional numbered labels}}, along with a cleaned list of segmentation masks. Uses Segment Anything (ViT-H) as the default segmentation model \cite{kirillov2023segment}. \\
\cmidrule{2-3}
& \colorbox{gray!10}{\texttt{axis\_localizer}} & \textbf{\textit{Localizes the specified axis}} (x-axis, left y-axis, or right y-axis) by detecting its numeric tick values and mapping them to corresponding pixel positions in the chart image. Uses Tesseract OCR \cite{TessOverview} and EasyOCR \cite{jaided2020easyocr}. \\
\cmidrule{2-3}
& \colorbox{gray!10}{\texttt{interpolate\_pixel\_to\_value}} & \textbf{\textit{Maps a pixel coordinate to its corresponding axis value}} using linear interpolation between known axis ticks and their pixel positions.\\
\cmidrule{2-3}
& \colorbox{gray!10}{\texttt{arithmetic}} & \textbf{\textit{Performs a specified arithmetic operation}} between two numeric inputs. Supports operations such as addition, subtraction, multiplication, division, percentage, and ratio. \\
\midrule
\multicolumn{3}{l}{\cellcolor{pale4}\textit{Chart-specific Tools}}\\
\midrule
\multirow{3}{2.2cm}{\colorbox{red!20}{Pie,} \colorbox{red!20}{Treemap}} & \colorbox{red!10}{\texttt{compute\_segment\_area}} & \textbf{\textit{Computes the area of a chart segment}} by: (1) counting discrete visual elements of a specified color, (2) counting pixels of a specified color, or (3) counting pixels within a segment identified by a specific label ID.\\
\midrule
\multirow{6}{2.2cm}{\colorbox{green!20}{Bar,} \colorbox{green!20}{Combination}} 
& \colorbox{green!10}{\texttt{get\_bar}} & \textbf{\textit{Detects and returns the bounding box of a bar}} in a chart image that matches a specified color and/or axis label. It segments bar regions using a model, filters by color if provided, locates the target axis label using OCR if specified, and selects the closest matching bar accordingly. \\
\cmidrule{2-3}
& \colorbox{green!10}{\texttt{compute\_bar\_height}} & \textbf{\textit{Computes the height or length of a bar}} in value space by mapping its pixel coordinates to axis values using OCR-based axis detection and localization. \\
 \midrule
\multirow{6}{2.2cm}{\colorbox{cyan!20}{Box}} & \colorbox{cyan!10}{\texttt{get\_boxplot}} & \textbf{\textit{Detects and returns boxplot segments}} filtered by color, axis label, or segmentation indices. Handles both horizontal and vertical boxplot orientations and supports fuzzy matching for axis-aligned labels and approximate color filtering.\\
\cmidrule{2-3}
& \colorbox{cyan!10}{\texttt{compute\_boxplot\_entity}} & \textbf{\textit{Computes a statistical entity (e.g., max, min, median, Q1, Q3, range, or interquartile range) of a boxplot}} by mapping its pixel coordinates to value space using axis localization. \\
\midrule
\multirow{4}{2.2cm}{\colorbox{magenta!20}{Line,} \colorbox{magenta!20}{Area,} \colorbox{magenta!20}{Scatter,} \colorbox{magenta!20}{Combination}} & \colorbox{magenta!10}{\texttt{get\_edgepoints}} & \textbf{\textit{Computes edge points of a chart segment}} filtered by color, axis label, or segmentation indices. The edge is determined by scanning perpendicular to the center of the matched label. Supports both vertical and horizontal chart orientations and optionally \textit{handles lineplot dots}. Useful for identifying segment bounds for downstream value extraction.\\
\midrule
\multirow{8}{2.2cm}{\colorbox{yellow!20}{Radial Bar}} 
& \colorbox{yellow!10}{\texttt{get\_radial}} & \textbf{\textit{Computes the coordinates for the radial bar segment}} of interest using either color-based filtering or segmentation mask labels. \\
\cmidrule{2-3}
& \colorbox{yellow!10}{\texttt{analyze\_radial\_geometry}} & Estimates the radial geometry of a radial bar chart for the segment of interest. \textbf{\textit{Identifies the chart center, detects the outer circle representing the maximum value, and computes the maximum radial extent (i.e., radius) of the contour of interest.}} \\
\cmidrule{2-3}
& \colorbox{yellow!10}{\texttt{estimate\_radial\_value}} &\textbf{\textit{Estimates the value of a radial segment in a radial bar chart}} by scaling its radial length relative to the outermost circle. The reference value for the outer circle is provided externally (e.g., by an LLM), with a default of 100. \\
\bottomrule
\end{tabular}
\end{adjustbox}
\label{appendix_tab:vision_tool_description}
\end{table*}

\twocolumn
These tools are organized into two broad categories:
\begin{itemize}
\item[(1)] \textbf{Universal tools}, which operate on fundamental chart components and are applicable across all chart types. These include legend detection and annotation (\texttt{annotate\_legend}), axis localization (\texttt{axis\_localizer}), legend marker color extraction (\texttt{get\_marker\_rgb}), chart cleaning to remove extraneous elements (e.g., titles and legends) that may interfere with downstream perception tasks (\texttt{clean\_chart\_image}), visual segmentation with post-processing (\texttt{segment\_and\_mark}), pixel-to-value interpolation (\texttt{interpolate\_pixel\_to\_value}), and basic arithmetic operations (\texttt{arithmetic}). Together, these tools provide the core perception and numeric reasoning primitives required for chart understanding.

\item[(2)] \textbf{Chart-specific tools}, which are specialized for particular chart types (e.g., pie, bar, line, box) and target subtasks unique to their underlying visual structures. For example, pie and treemap charts use \texttt{compute\_segment\_area}; bar charts use \texttt{get\_bar} and \texttt{compute\_bar\_height}; box plots use \texttt{get\_boxplot} and \texttt{compute\_boxplot\_entity}; line, area, and scatter charts use \texttt{get\_edgepoints}; and radial bar charts use \texttt{get\_radial}, \texttt{analyze\_radial\_geometry}, and \texttt{estimate\_radial\_value}. For combination charts (e.g., bar+line or bar+pie), the agent composes the relevant chart-specific tools corresponding to each constituent chart type.
\end{itemize}

The tool suite is intentionally designed to be simple, modular, and component-centric. Rather than introducing highly specialized tools for each chart subtype, we focus on a small set of reusable primitives that operate on universal chart elements such as legends, axes, segments, and geometric extents. While more complex, chart-specific tools could be engineered, doing so would sacrifice generality and make extension to new or unseen chart types more brittle. By grounding all chart-specific tools in shared visual components, the framework naturally scales to a wide range of chart types (currently covering 40+ types) and enables straightforward extension: supporting a new chart type typically requires only composing or lightly adapting existing primitives.

\subsection{Underlying Models Powering \chartagent{} Tools} 
\chartagent{} relies on a set of custom-designed, chart-aware tools, some of which are built upon a small number of off-the-shelf vision and OCR models. These underlying models provide basic perception and text extraction, while the tools introduce task-specific structure and reasoning tailored to chart understanding.

\begin{itemize}
\item[(1)] \textbf{Semantic segmentation.} 
\texttt{Segment Anything Model v1 (SAM)}~\cite{kirillov2023segment} is used by the \texttt{segment\_and\_mark} tool to extract chart foreground content and generate candidate segmentation masks corresponding to chart elements (e.g., pie slices in pie charts, bar regions in bar charts, or areas in area charts). SAM produces a dense set of object-agnostic masks, which our tool then post-processes using a multi-stage filtering pipeline to remove extraneous, duplicate, composite, or background-dominated regions, yielding a clean set of chart-relevant segments. \texttt{Segment Anything} employs a ViT-H image encoder (641M parameters) trained on large-scale, diverse segmentation data, together with a prompt encoder and a lightweight mask decoder, enabling strong generalization to previously unseen visual structures such as diverse chart layouts and styles.

\item[(2)] \textbf{Optical character recognition (OCR).} 
\texttt{Tesseract}~\cite{TessOverview} is used for fast OCR and text localization, including extracting x- and y-axis tick values in \texttt{axis\_localizer} and legend text in \texttt{annotate\_legend}. Owing to its lightweight design and computational efficiency, \texttt{Tesseract} serves as the default OCR engine. For visually complex or noisy charts where \texttt{Tesseract} may fail, \texttt{EasyOCR}~\cite{jaided2020easyocr} is used as a fallback. \texttt{EasyOCR} employs a VGG16-based \texttt{CRAFT} text detector (138M parameters), followed by a CRNN (83M parameters) for text recognition.
\end{itemize}

\subsection{Tool Outputs and Intermediate Visualizations for Self-Verification in \chartagent{}}\label{appendix:self-verification}
\begin{figure*}[!t]
    \centering
    \includegraphics[scale = 0.445]{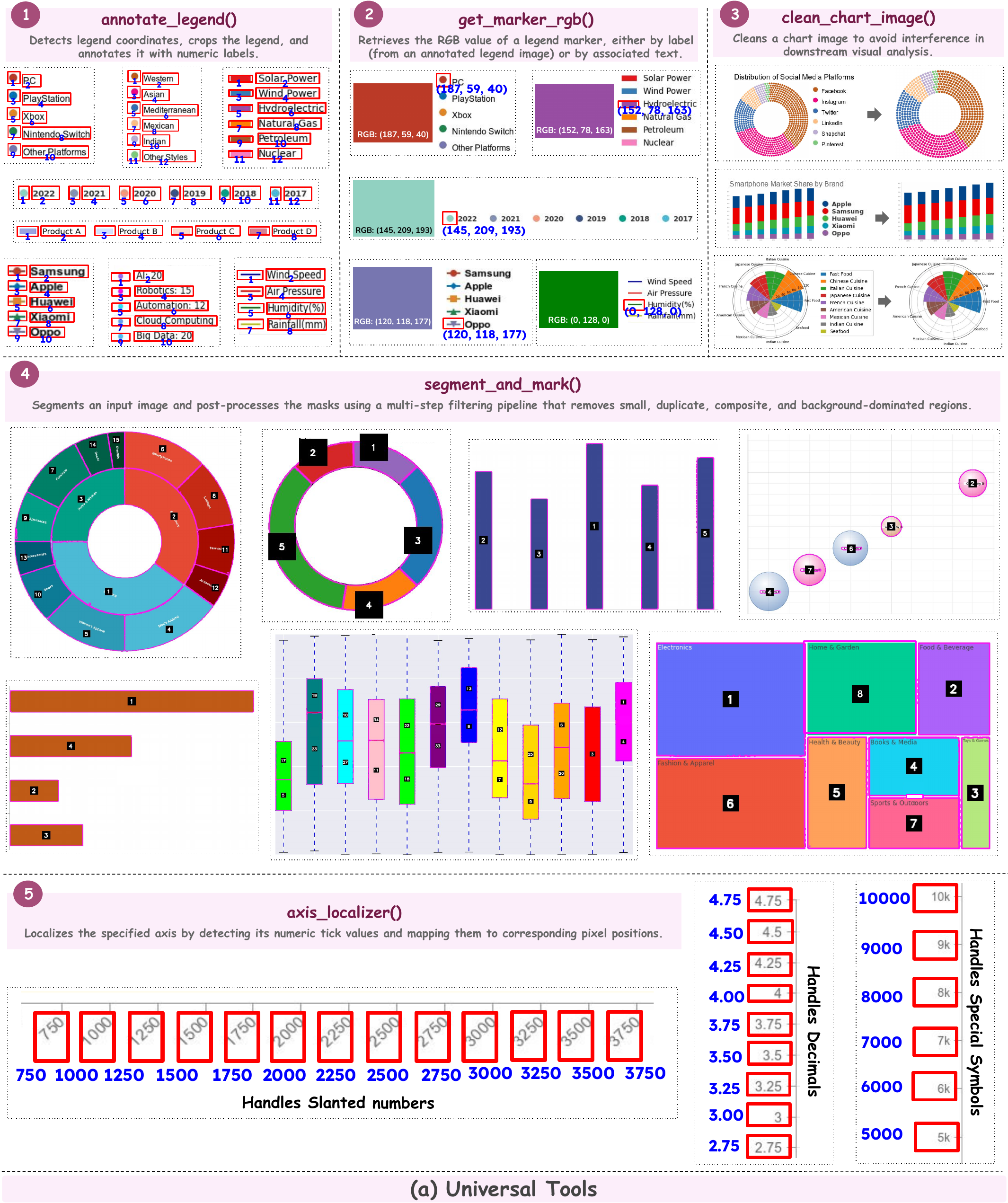}
    \caption{\textbf{Illustrative examples of key intermediate and final output visualizations for \colorbox{pink!50}{universal tools} in \chartagent{}.} These visualizations are critical to facilitating visual self-verification in \chartagent{}. Such tool observations enable \chartagent{} to perceptually assess the outputs and refine its tool usage in the next iteration—either by adjusting tool parameters or invoking a different tool if the intermediate results indicate incorrect or unexpected behavior. Note the diverse variations that our tools are capable of handling robustly.}
    \label{fig:universal-tools-viz-self-verify}
\end{figure*}

\begin{figure*}[htbp]
    \centering
    \vspace{-0.5cm}
    \includegraphics[scale = 0.37]{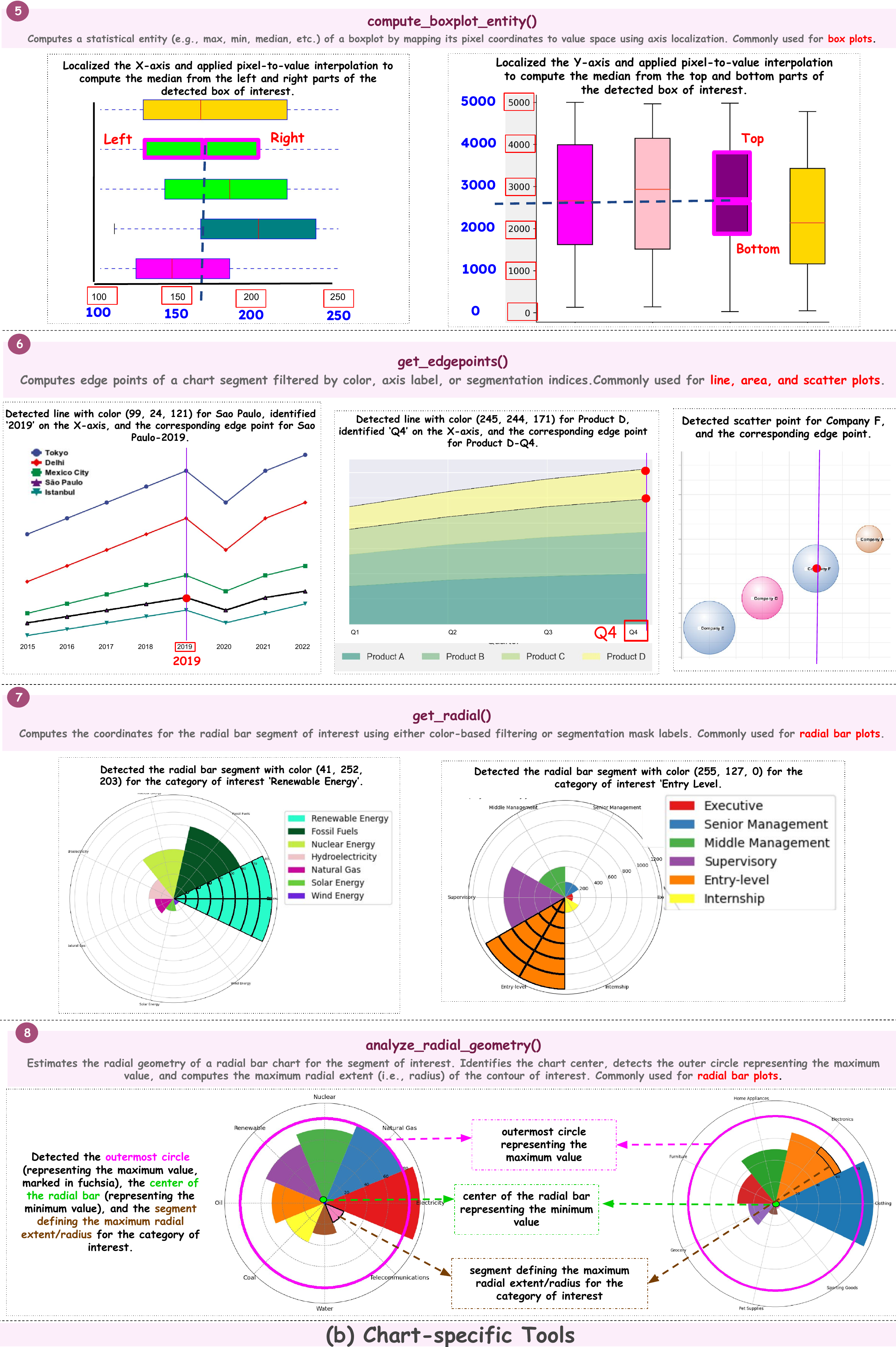}
    \caption{\textbf{Illustrative examples of key intermediate and final output visualizations for \colorbox{pink!50}{chart-specific tools} in \chartagent{}.} These visualizations enable visual self-verification in \chartagent{}, allowing it to refine tool usage through perceptual assessment and iterative correction. We intentionally present some easier examples here for illustration, to help readers quickly follow the process. However, \chartagent{} tools are capable of handling a wide range of cases, including more difficult and complex ones, as demonstrated by the overall results. }
    \label{fig:chart-specific-tools-viz-self-verify}
\end{figure*}

Our chart-specialized tools are carefully designed to produce clear, perception-friendly visualizations and outputs that \chartagent{} can interpret for self-verification. Figures~\ref{fig:universal-tools-viz-self-verify} and~\ref{fig:chart-specific-tools-viz-self-verify} show illustrative intermediate visualizations and final outputs from our universal and chart-specific tools, respectively, and also highlight the variations that these tools are able to robustly handle.
To support explicit visual inspection, tool outputs include overlays, highlights, or annotations that are optimized to be easily interpretable by the base MLLM (e.g., colored segment overlays in pie charts, bar height markers, annotated legends). These custom-designed artifacts allow \chartagent{} to reason over visual evidence grounded in the charts. When outputs appear semantically inconsistent or visually incorrect (e.g., pie segments too small, mismatched colors, negative bar heights, or responses contradicting axis values), \chartagent{} engages in a recover-and-retry process—tweaking tool parameters or invoking alternative tools. This iterative correction loop mimics human-like debugging, ensuring robust reasoning and accurate interpretation in the chart domain. These visualizations are therefore critical for enabling \chartagent{} to assess intermediate results and adapt its behavior in subsequent steps. A quantitative evaluation of the effectiveness of this visual self-verification is provided in Section~\ref{results: additional_analysis}.

Note that some tools generate additional outputs not displayed here—for example, the \texttt{annotate\_legend} tool also produces a cropped legend image, an annotated cropped legend image, and a bounding-box mapping between detected markers/text and their $(x, y, w, h)$ coordinates. In this figure, however, we highlight only the key output (the annotated cropped legend image) to focus on the most relevant artifacts for visual self-verification. In contrast, some tools produce only numeric outputs, such as \texttt{arithmetic} and \texttt{interpolate\_pixel\_to\_value}, which are not included here. Complete input–output specifications for each chart-specialized tool are provided in Table~\ref{appendix_tab:vision_tool_description} and Section~\ref{appendix-chartagent-tool-definition-prompts}.

\begin{figure*}[!t]
    \centering
    \includegraphics[scale = 0.38]{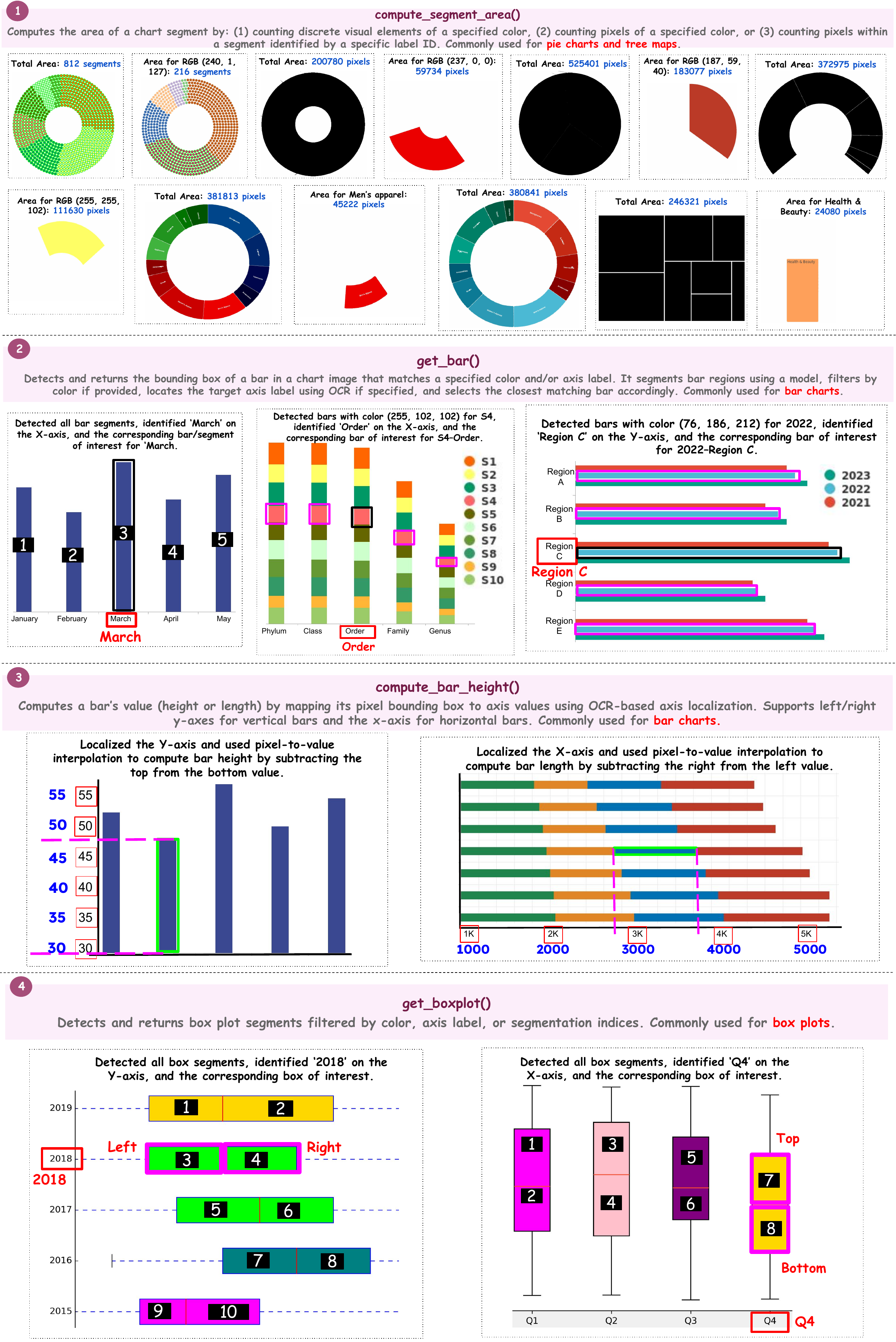}
\end{figure*}

\subsection{Adaptive, Heuristic-Free Visual Self-Verification}\label{appendix:adaptive-heuristic-free-self-verify}
In \chartagent{}, verification is not based on fixed heuristic rules (e.g., pixel-overlap thresholds or axis-consistency formulas). Instead, we adopt a flexible, agent-driven strategy in which the agent interprets tool outputs—such as segmentation masks, axis overlays, and annotated legends—and determines whether they are sufficient for the current reasoning step. This forms the core of our visual self-verification loop. We deliberately avoid hard-coded verification logic because such rules tend to be brittle and fail to generalize across the 40+ chart types and diverse layout structures supported in our framework. By contrast, learned, context-aware visual reasoning enables more robust and scalable behavior.

It is also important to note that, as with many recent agentic systems built around external tool calls (e.g., Visual Sketchpad~\cite{huvisualsketchpad}, ViperGPT~\cite{suris2023vipergpt}, VideoAgent~\cite{wang2024videoagent}, VideoAgent2~\cite{zhi2025videoagent2}), there is no standard methodology for evaluating tool-level accuracy. Ground truth for intermediate steps—such as which segment mask, axis tick, or bounding box should be considered ``correct"—typically does not exist. Consequently, these systems, like ours, focus on end-task performance while allowing the agent to interpret and adaptively incorporate visual tool outputs into its reasoning process.

\section {Implementation Details}\label{appendix:implementation_details}
\subsection{\chartagent{}} \label{appendix:implementation_details_chartagent}
\chartagent{} is implemented using the AutoGen 0.2.26 framework, running on Python 3.9 and configured to perform a maximum of 15 reasoning iterations per task. 
In practice, significantly fewer iterations are required: across all evaluated samples, trajectories use an average of 5--7 model calls, with the 15-iteration limit serving only as a safeguard for rare cases requiring extended reasoning or self-correction.

The GPT-4o model (\texttt{gpt-4o-2024-08-06}) is used as the primary multimodal LLM for reasoning in \chartagent{}, with the temperature set to 0.0 for deterministic outputs. 
Importantly, GPT-4o (\texttt{gpt-4o-2024-08-06}) has a knowledge cutoff of October~1,~2023. Since ChartBench and ChartX were released in December~2023 and February~2024, respectively, they were definitively not part of GPT-4o’s training data. 
For the variants of \chartagent{} evaluated in Section~\ref{results: effectiveness_chartagent}, we additionally use GPT-4o-mini (\texttt{gpt-4o-mini-2024-07-18}), Claude~3~Haiku (\texttt{claude-3-haiku-20240307}), and Pixtral-12B-2409 as alternative base MLLMs. 

For reproducibility, all experiments use a fixed random seed of 42.
All experiments are conducted on a Linux machine using an \texttt{AWS g4dn.xlarge} instance equipped with a single NVIDIA T4 GPU (16~GB memory). For segmentation tasks, we employ the Segment Anything (SAM, ViT-H) \cite{kirillov2023segment}, which has 641M parameters and a model size of 2.56~GB. For OCR, we use Tesseract OCR \cite{TessOverview} and EasyOCR \cite{jaided2020easyocr}. 
All \chartagent{} prompts are provided in Appendix~\ref{appendix:chart-agent-prompts}.

\subsection{Baselines} \label{appendix:implementation_details_baselines}
Similar to the \chartagent{} setup, all applicable baselines were run with a temperature setting of 0.0 to ensure deterministic outputs, with the random seed fixed at 42 for reproducibility. All proprietary baseline models, as well as open-weight general-purpose baseline models, were evaluated using both zero-shot and Chain-of-Thought (CoT) prompting styles. All baseline prompts are provided in Appendix~\ref{appendix:baseline_prompts}.
For chart-based baseline models such as DePlot~\cite{liu-etal-2023-deplot} and OneChart~\cite{chen2024onechart}, which output structured tables rather than direct answers, we apply a zero-shot GPT-4o call to extract the final answer (see Appendix~\ref{appendix:baseline-table-to-vqa-conversion} for the corresponding prompt).

\section{Examples of Response Standardization for Accuracy Evaluation} \label{appendix:standardization_examples}
As part of our two-step accuracy evaluation (Section~\ref{experiments:evaluation}), we use GPT-4o to standardize both the model’s response and the ground truth answer, before applying an arithmetic or string-matching correctness check. Below are representative examples of the standardization operations applied:

\vspace{0.5em}
\noindent
\textbf{(1) Converting Scales} e.g., \texttt{K} for thousand, \texttt{M} for million, \texttt{B} for billion
\begin{itemize}
    \item ground truth: \underline{\texttt{3000}} $\vert$ response: \texttt{4K} $\rightarrow$ \underline{\texttt{4000}}
    \item ground truth: \texttt{15\%} $\rightarrow$ \underline{\texttt{15}} $\vert$ response: \texttt{0.15 times} $\rightarrow$ \texttt{15\%} $\rightarrow$ \underline{\texttt{15}}
    \item ground truth: \texttt{2000m} $\rightarrow$ \underline{\texttt{2000}} $\vert$ response: \texttt{2.5km} $\rightarrow$ \texttt{2500m} $\rightarrow$ \underline{\texttt{2500}}
    \item ground truth: \texttt{48 hours} $\rightarrow$ \underline{\texttt{48}} $\vert$ response: \texttt{2 days} $\rightarrow$ \texttt{48 hours} $\rightarrow$ \underline{\texttt{48}}
\end{itemize}

\vspace{0.5em}
\noindent
\textbf{(2) Stripping Units} e.g., \texttt{\$}, \texttt{\%}, \texttt{K}, \texttt{M}, \texttt{B}, etc.
\begin{itemize}
    \item ground truth: \underline{\texttt{5}} $\vert$ response: \texttt{5K} $\rightarrow$ \underline{\texttt{5}}
    \item ground truth: \underline{\texttt{15}} $\vert$ response: \texttt{10\%} $\rightarrow$ \underline{\texttt{10}}
\end{itemize}

\vspace{0.5em}
\noindent
\textbf{(3) Removing Symbols}
\begin{itemize}
    \item response: \texttt{1,000} $\rightarrow$ \underline{\texttt{1000}}
\end{itemize}

\vspace{0.5em}
\noindent
\textbf{(4) Standardizing Number Formats}
\begin{itemize}
    \item ground truth: \underline{\texttt{7}} $\vert$ response: \texttt{seven} $\rightarrow$ \underline{\texttt{7}}
\end{itemize}
These standardizations of the ground truth and response ensure that formatting differences do not lead to incorrect evaluations during the subsequent arithmetic correctness check or string-matching step. Prompts for both evaluation strategies—namely, our standardization-based accuracy computation and the LLM-as-a-Judge baseline evaluation—are provided in Appendix~\ref{appendix:prompts_evaluation_metrics}.

Further, to assess the correctness of these standardization operations, we manually annotated and verified the process. We sampled 100 examples per dataset and reviewed both the model responses and the ground-truth normalizations, finding the standardized outputs to be accurate in over 97\% of cases. The few remaining errors arose in highly convoluted answers involving multiple entangled numeric values or ambiguous final quantities, edge cases that understandably challenge automatic extraction.

\section{Analysis of Numerical Tolerance Choices in the Evaluation Metric}
\label{appendix:evaluation_tolerance_analysis}

The 5\% relative error threshold used in our evaluation follows the standard protocol established across the Chart VQA literature. Widely used benchmarks such as ChartQA~\citep{masry-etal-2022-chartqa}, PlotQA~\citep{methani2020plotqa}, UniChart~\citep{masry-etal-2023-unichart}, MATCHA~\citep{liu-etal-2023-matcha}, ChartX and ChartVLM~\citep{xia2024chartx}, ChartBench~\citep{xu2023chartbench}, TinyChart~\citep{zhang2024tinychart}, ChartLLaMA~\citep{han2023chartllamamultimodalllmchart}, and ChartGemma~\citep{masry-etal-2025-chartgemma} all apply a 5\% tolerance when judging numerical correctness. This convention balances strictness with the inherent visual ambiguity in reading values from charts and enables consistent comparison across benchmarks. Our work follows this same standard.

That said, different application contexts (e.g., financial forecasting vs.\ everyday QA) may warrant different numerical tolerances. To explore this, we conducted a stratified evaluation across six thresholds: 0.1\%, 1\%, 3\%, 5\%, 10\%, and 15\%. This analysis simulates varying levels of risk sensitivity and precision requirements.
Table~\ref{tab:tolerance_stratified} reports the overall accuracy results for the top-10 performing models on ChartBench.
\begin{table}[h!]
\centering
\caption{\textbf{Accuracy under varying relative error tolerances.} Best performance in each threshold is highlighted in \textbf{bold}.}
\label{tab:tolerance_stratified}
\resizebox{0.49\textwidth}{!}{
\begin{tabular}{lcccccc}
\toprule
\textbf{Model} & \textbf{0.1\%} & \textbf{1\%} & \textbf{3\%} & \textbf{5\%} & \textbf{10\%} & \textbf{15\%} \\
\midrule
ChartAgent        & \textbf{40.16} & \textbf{59.84} & \textbf{67.84} & \textbf{71.39} & \textbf{76.63} & \textbf{79.53} \\
GPT-4o            & 39.19 & 42.14 & 46.48 & 54.53 & 57.76 & 63.48 \\
GPT-4o mini       & 30.43 & 33.38 & 35.67 & 44.03 & 45.43 & 51.10 \\
Claude 3 Haiku    & 27.43 & 31.29 & 34.90 & 44.53 & 47.00 & 51.14 \\
Phi-3 Vision      & 35.38 & 38.57 & 43.95 & 55.32 & 56.19 & 58.38 \\
Qwen2-VL          & 35.38 & 37.76 & 44.81 & 54.53 & 55.95 & 56.81 \\
Llama-3.2         & 34.86 & 37.00 & 42.81 & 52.11 & 54.52 & 58.00 \\
Pixtral           & 29.62 & 32.62 & 36.90 & 44.11 & 48.52 & 52.95 \\
DeepSeek-VL2      & 34.00 & 37.29 & 41.48 & 49.39 & 54.62 & 59.29 \\
DePlot            & 25.95 & 31.19 & 34.90 & 41.39 & 40.33 & 43.19 \\
TinyChart         & 24.81 & 29.57 & 36.81 & 46.84 & 47.90 & 52.57 \\
\bottomrule
\end{tabular}
}
\end{table}

As expected, accuracy improves as the tolerance widens (e.g., at the 10--15\% settings). However, across all thresholds, ChartAgent consistently maintains the highest accuracy, demonstrating that its advantages are robust and not overly dependent on the standard 5\% threshold. This analysis validates our evaluation choices while enabling more nuanced, scenario-specific interpretations.

\section{Complexity Analysis}\label{appendix:complexity_analysis}
To examine \chartagent{} performance under varying levels of difficulty, we divide all chart–QA samples across our evaluation datasets into difficulty levels based on (a) the \textit{visual complexity} of charts and (b) the \textit{reasoning complexity} of chart–QA pairs. This stratification enables us to analyze performance trends across distinct categories of challenge. Each dimension is categorized into three levels: Easy, Medium, and Hard.  
\begin{table*}[htb]
\caption{\textbf{Complexity Label Statistics.} Distribution of difficulty levels stratified by (a) \textit{visual complexity} of charts and (b) \textit{reasoning complexity} of chart–QA pairs in the evaluation datasets. Rows correspond to reasoning complexity; columns correspond to visual complexity. Each dimension has three levels: \colorbox{green!15}{Easy}, \colorbox{yellow!15}{Medium}, \colorbox{red!15}{Hard}.}
\centering
\begin{subtable}[t]{0.48\textwidth}
\centering
\resizebox{\textwidth}{!}{
\begin{tabular}{l|ccc|c}
\toprule
\multirow{2}{*}{\textbf{Reasoning Complexity}} & \multicolumn{3}{c|}{\textbf{Visual Complexity}} & \multirow{2}{*}{\textbf{Total}} \\ 
\cmidrule{2-4}
 & \cellcolor{green!15}\textbf{Easy} & \cellcolor{yellow!15}\textbf{Medium} & \cellcolor{red!15}\textbf{Hard} & \\ 
\midrule
\cellcolor{green!15}\textbf{Easy}   & 37.38\% & 35.88\% & 1.43\% & 74.69\% \\
\cellcolor{yellow!15}\textbf{Medium} & 0.76\%  & 8.86\%  & 6.40\% & 16.02\% \\
\cellcolor{red!15}\textbf{Hard}   & 0.98\%  & 7.07\%  & 1.24\% & 9.29\%  \\ 
\midrule
\textbf{Total}  & 39.12\% & 51.81\% & 9.07\% & 100\% \\
\bottomrule
\end{tabular}}
\vspace{0.2mm}
\caption{ChartBench Dataset}
\label{tab:chartbench_complexity}
\end{subtable}
\hfill
\begin{subtable}[t]{0.48\textwidth}
\centering
\resizebox{\textwidth}{!}{
\begin{tabular}{l|ccc|c}
\toprule
\multirow{2}{*}{\textbf{Reasoning Complexity}} & \multicolumn{3}{c|}{\textbf{Visual Complexity}} & \multirow{2}{*}{\textbf{Total}} \\ 
\cmidrule{2-4}
 & \cellcolor{green!15}\textbf{Easy} & \cellcolor{yellow!15}\textbf{Medium} & \cellcolor{red!15}\textbf{Hard} & \\ 
\midrule
\cellcolor{green!15}\textbf{Easy}   & 44.27\% & 20.83\% & 2.60\% & 67.71\% \\
\cellcolor{yellow!15}\textbf{Medium} & 9.38\%  & 7.55\%  & 5.90\% & 22.74\% \\
\cellcolor{red!15}\textbf{Hard}   & 0.52\%  & 3.12\%  & 5.82\% & 9.55\%  \\ 
\midrule
\textbf{Total}  & 54.17\% & 31.51\% & 14.32\% & 100\% \\
\bottomrule
\end{tabular}}
\vspace{0.2mm}
\caption{ChartX Dataset}
\label{tab:chartx_complexity}
\end{subtable}
\label{tab:complexity_statistics}
\end{table*}

\begin{figure*}[htb]
    \centering
    \includegraphics[scale = 0.35]{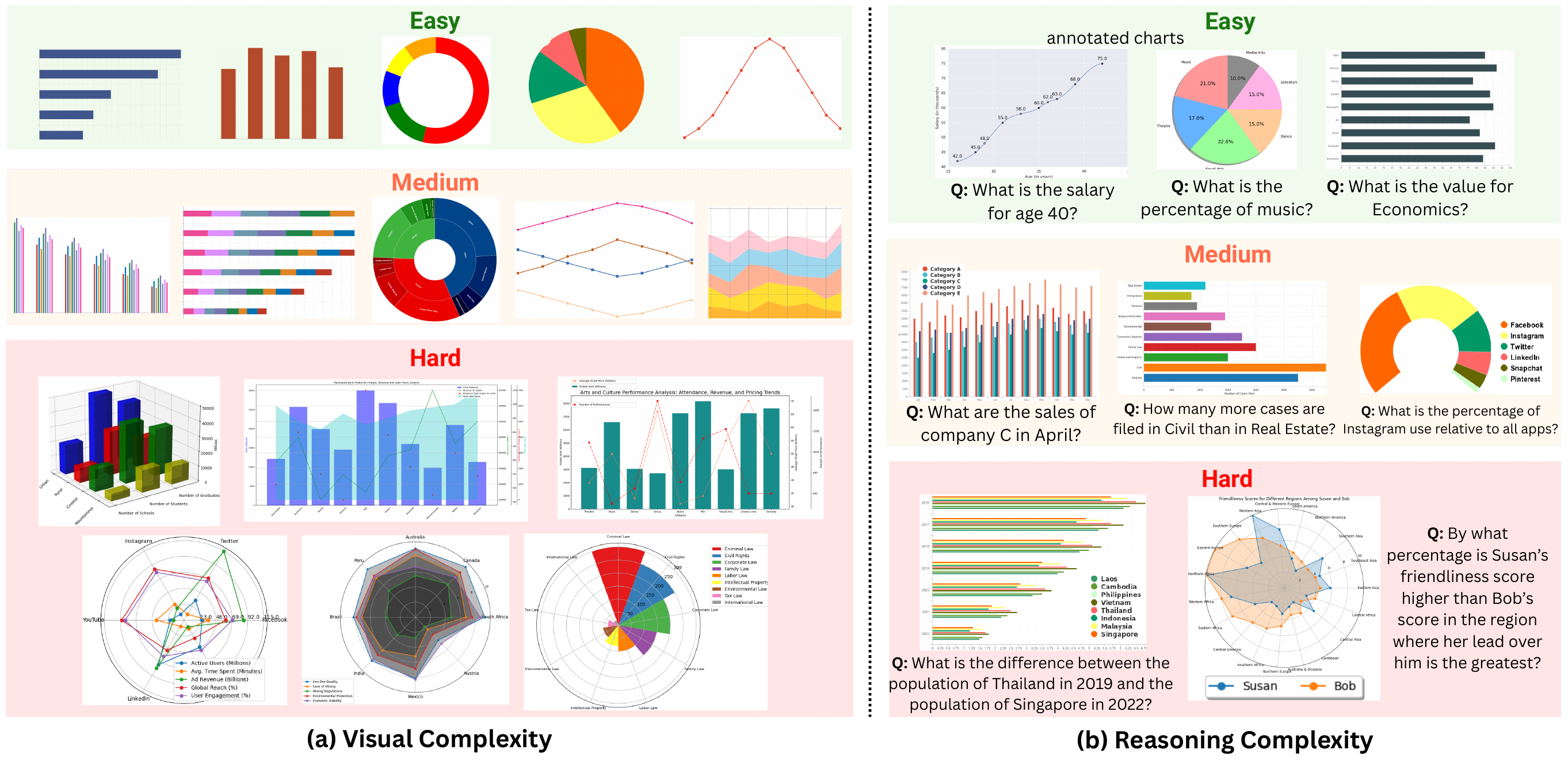}
    \caption{\textbf{Complexity dimensions in chart–QA pairs.} 
    Representative examples are shown for (a) \textit{visual complexity} of charts and (b) \textit{reasoning complexity} of chart–QA pairs, each categorized into \textcolor{green!50!black}{Easy}, \textcolor{orange}{Medium}, and \textcolor{red}{Hard} levels. 
    \textbf{(a)} For \textit{visual complexity}: Easy charts (e.g., single bar or line plots) have few elements and clean layouts; Medium charts (e.g., multi-series line or stacked bar plots) add moderate overlap; Hard charts (e.g., radar charts, 3D plots, or heavily layered visuals) are highly cluttered. 
    \textbf{(b)} For \textit{reasoning complexity}: Easy chart–QA pairs involve direct lookup; Medium pairs require comparisons or proportions; Hard pairs need complex multi-step reasoning.}
    \label{fig:visual-reasoning-complexity-examples}
\end{figure*}
\begin{itemize}
    \item \textit{Visual complexity} reflects the effort needed to interpret the chart image. Easy charts (e.g., single bar or line plots) contain few elements and clean layouts. Medium charts (e.g., multi-series line plots, grouped/stacked bar charts) introduce moderate clutter and overlapping elements. Hard charts (e.g., radar charts, 3D plots, or heavily layered visuals) are highly cluttered and visually demanding.  

    \item \textit{Reasoning complexity} captures the cognitive effort required to answer a question using the chart. Easy chart–QA pairs involve direct value lookup. Medium pairs require comparisons, ratios, or proportions. Hard pairs demand multi-step reasoning, arithmetic aggregation, or complex logical inference. 
\end{itemize}

Table~\ref{tab:complexity_statistics} reports the distribution of visual and reasoning complexity across our evaluation datasets, ChartBench and ChartX. Both datasets provide coverage across all three categories. The majority of charts fall under visually Easy or Medium categories, with fewer than 15\% classified as visually Hard. ChartX contains a larger fraction of visually Hard charts, making it slightly more challenging overall in terms of clutter and layout. A similar trend is observed for reasoning complexity: although Easy dominates, both datasets include substantial portions of Medium and Hard reasoning tasks, ensuring coverage of non-trivial scenarios.  

Further, Figure~\ref{fig:visual-reasoning-complexity-examples} illustrates representative examples spanning different chart types and subtypes across the Easy, Medium, and Hard levels for both visual and reasoning complexity. The prompts used to label chart images and chart–QA pairs into these stratified levels are provided in Appendix~\ref{appendix:prompts_complexity_analysis}.

Further, to assess human agreement with the complexity labels, we conducted a small-scale validation study with two annotators, each reviewing 10 examples per category (Easy, Medium, Hard) for both visual and reasoning complexity. We observed an average disagreement rate of 8\% between the human annotators and our automatic labeling pipeline, with most discrepancies occurring between Medium and Hard visual complexity.

\onecolumn
\section{Qualitative Analysis}\label{appendix:qualitative_examples}
This section provides qualitative insights into \chartagent{}’s behavior, illustrating how the agent integrates visual perception, tool usage, and reasoning across a diverse set of chart types and question settings. We complement the quantitative results in Section~\ref{sec:results} by analyzing representative reasoning trajectories (Section~\ref{appendix:qual_agent_trajectories}) and representative qualitative comparison examples (Section~\ref{appendix:qual_examples}).

\subsection{Illustrative Agent Trajectories}\label{appendix:qual_agent_trajectories}
We present illustrative \chartagent{} trajectories organized into three categories: unannotated charts and numeric QA (Section~\ref{appendix:qual_agent_trajectories-unannotated}), visual self-verification in action (Section~\ref{appendix:qual_agent_trajectories-visual-self-verify}), and annotated charts (Section~\ref{appendix:qual_agent_trajectories-annotated}). We additionally discuss a set of interesting and edge-case trajectories in Section~\ref{appendix:qual_agent_trajectories-interesting-extra-trajs}.

\subsubsection{Agent Trajectories on Unannotated Charts and Numeric QA}\label{appendix:qual_agent_trajectories-unannotated}
\begin{figure*}[htbp]
  \centering
  \includegraphics[scale = 0.46]{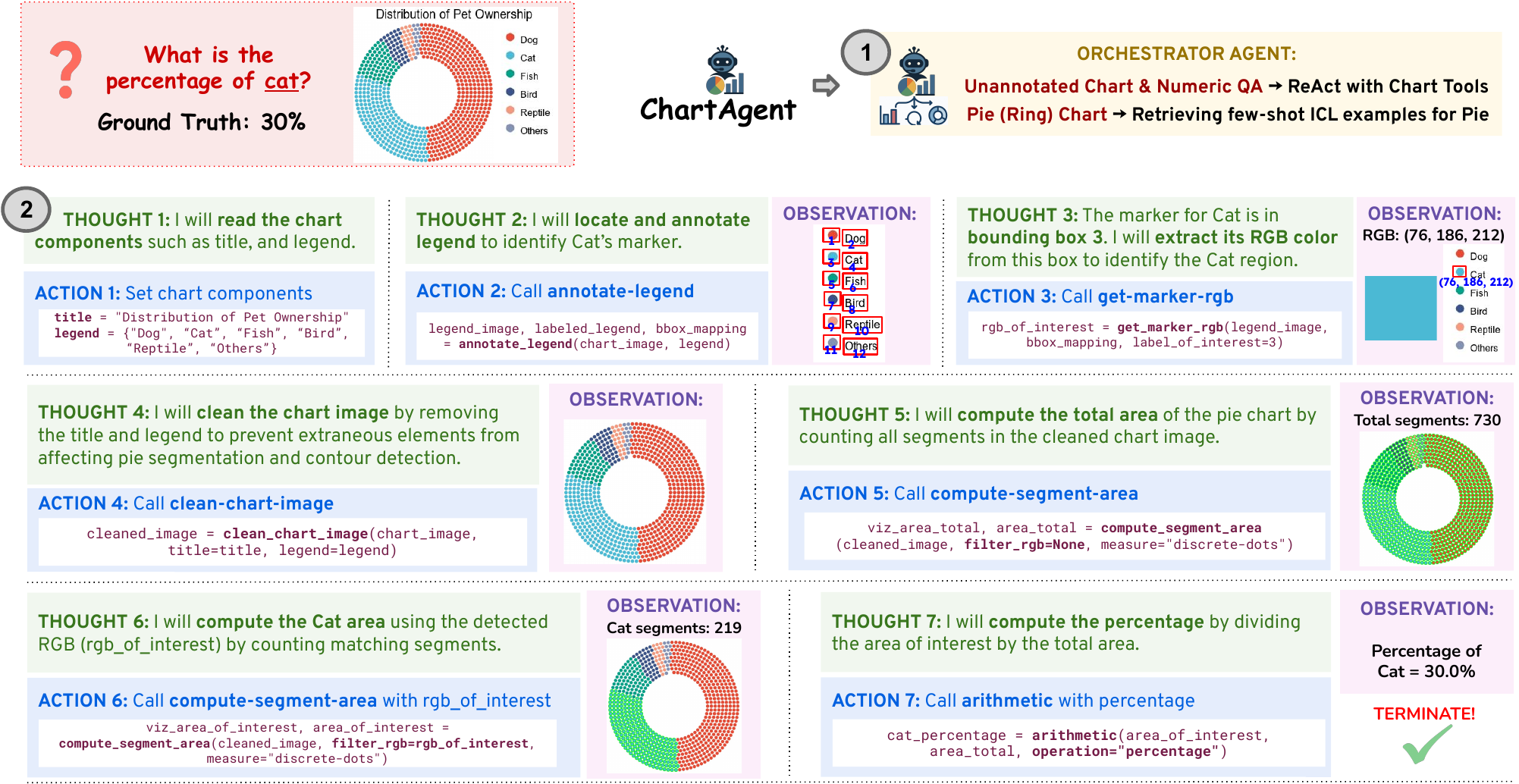}
  \caption{\textbf{Qualitative Trajectory on a Pie (Ring) Chart.} }
  \label{fig:qual_traj_pie_ring}
\end{figure*}

\begin{figure*}[htbp]
  \centering
  \includegraphics[scale = 0.46]{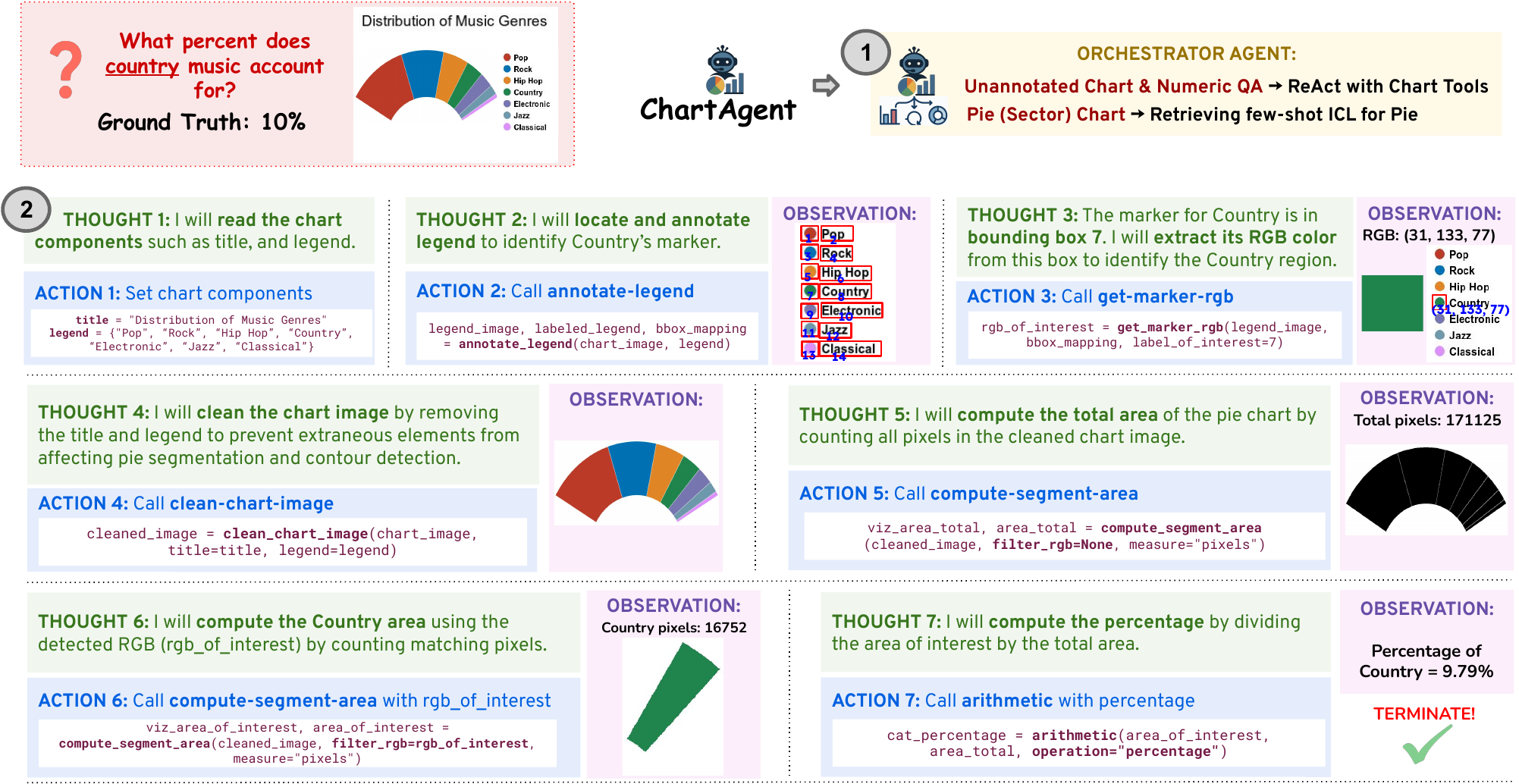}
    \caption{\textbf{Qualitative Trajectory on a Pie (Sector) Chart.}}
  \label{fig:qual_traj_pie_sector}
\end{figure*}

\begin{figure*}[htbp]
  \centering
  \includegraphics[scale = 0.485]{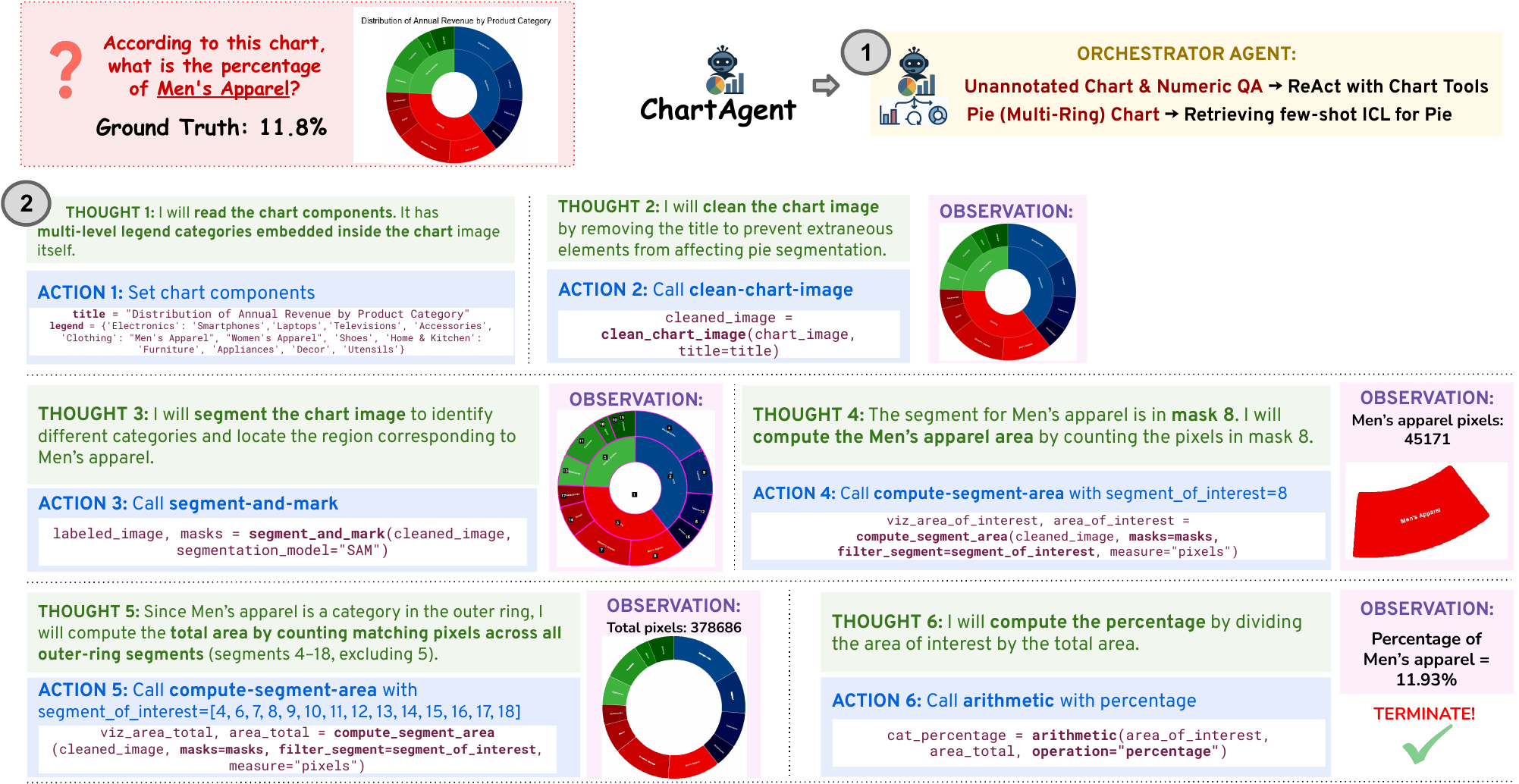}
    \caption{\textbf{Qualitative Trajectory on a Pie (Multi-Ring) Chart.}}
  \label{fig:qual_traj_pie_multi_ring}
\end{figure*}

\begin{figure*}[htbp]
  \centering
  \includegraphics[scale = 0.58]{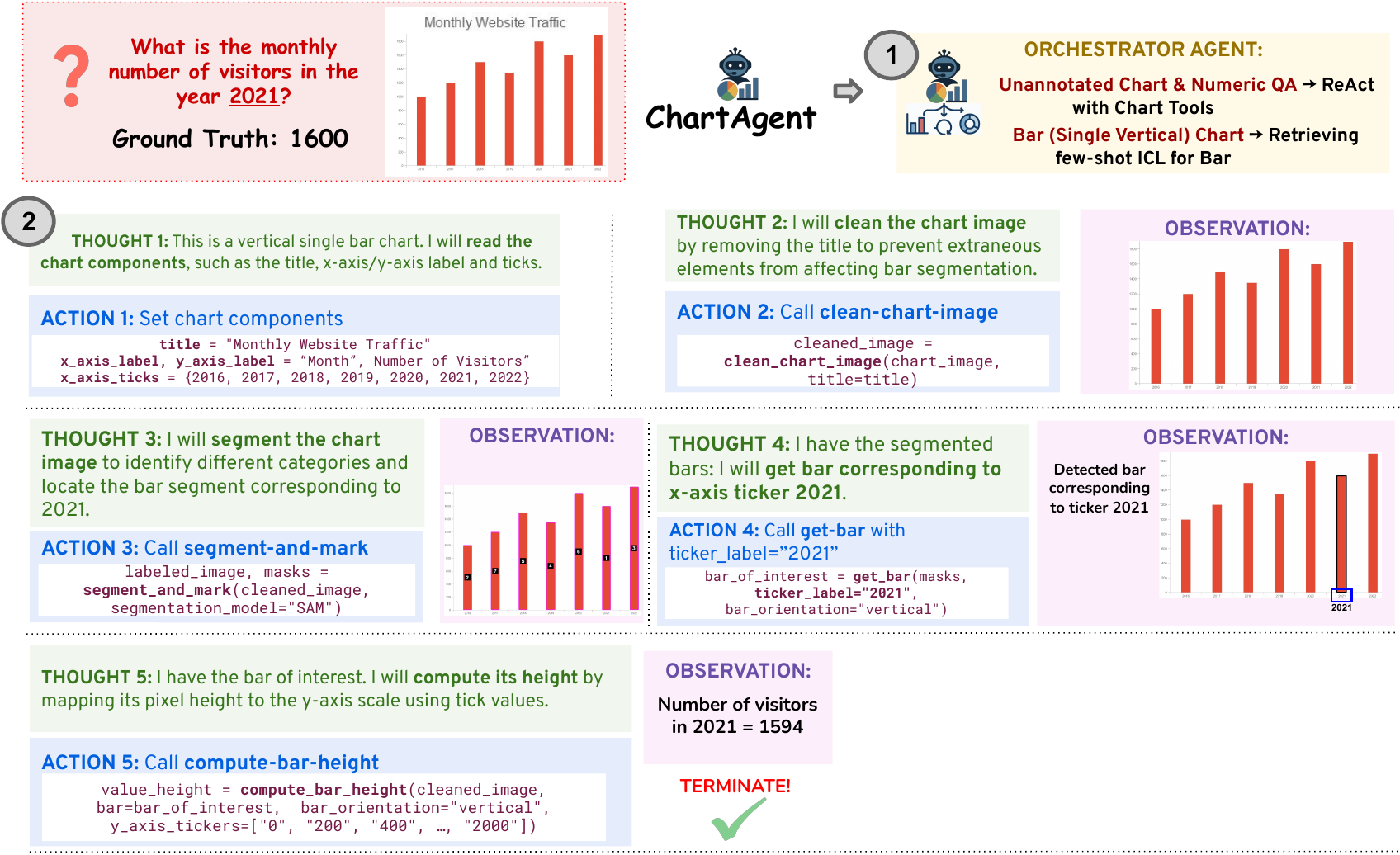}
    \caption{\textbf{Qualitative Trajectory on a Bar (Single Vertical) Chart.}}
  \label{fig:qual_traj_bar_single_vertical}
\end{figure*}

\begin{figure*}[htbp]
  \centering
  \includegraphics[scale = 0.52]{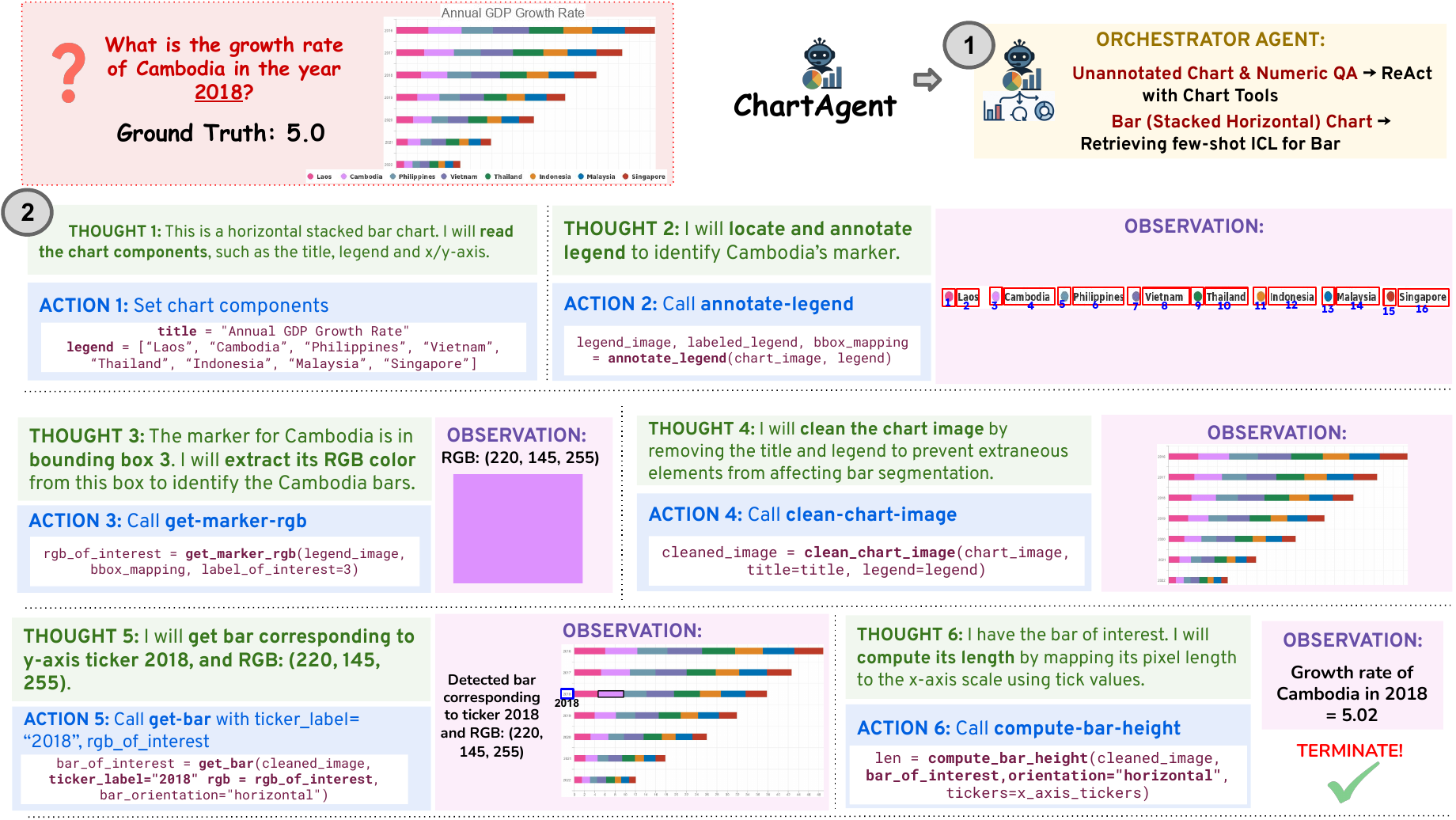}
    \caption{\textbf{Qualitative Trajectory on a Bar (Stacked Horizontal) Chart.}}
  \label{fig:qual_traj_bar_stacked_horizontal}
\end{figure*}

\begin{figure*}[htbp]
  \centering
  \includegraphics[scale = 0.53]{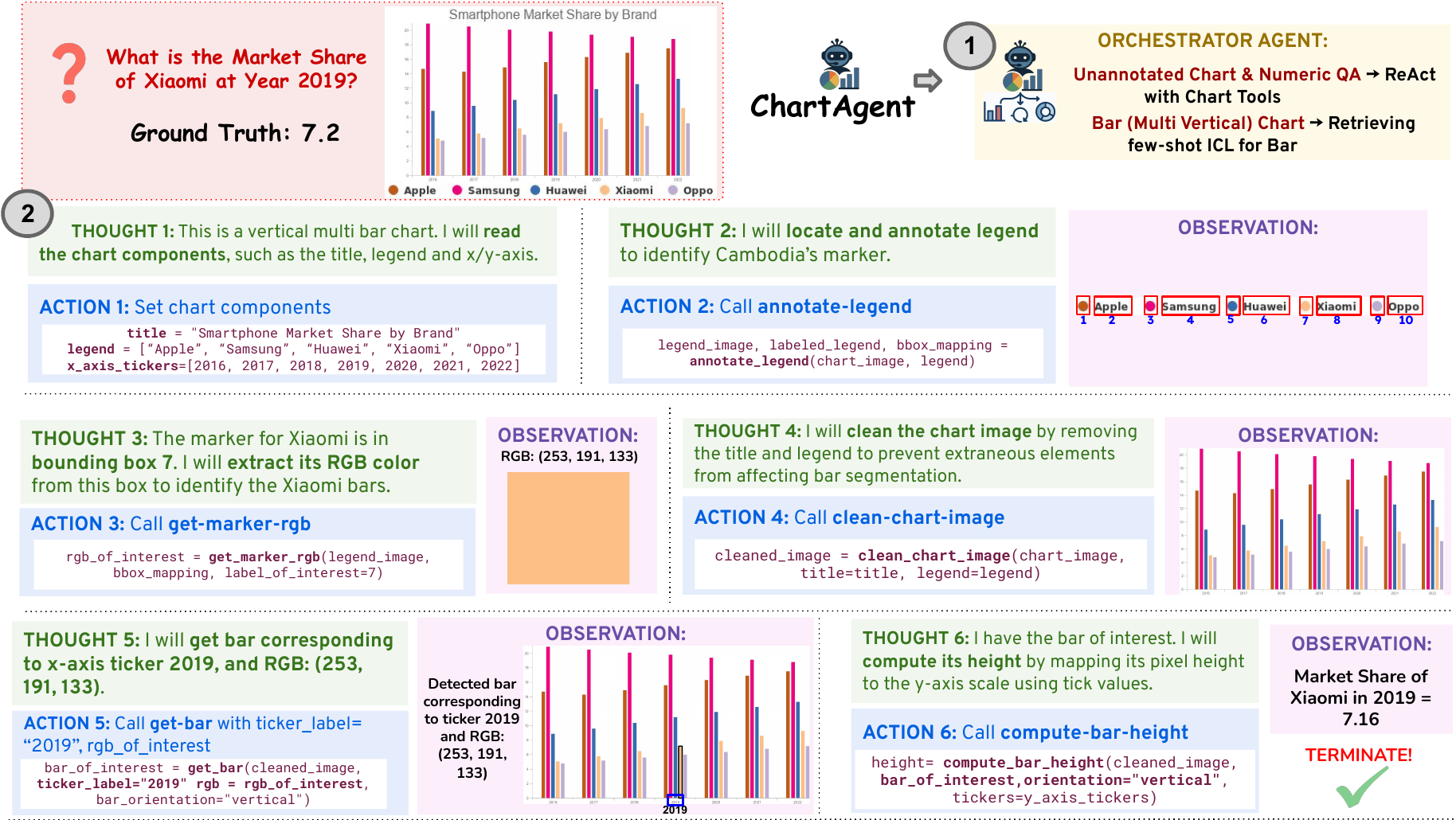}
    \caption{\textbf{Qualitative Trajectory on a Bar (Multi-grouped Vertical) Chart.}}
  \label{fig:qual_traj_bar_multi_vertical}
\end{figure*}

\begin{figure*}[htbp]
  \centering
  \includegraphics[scale = 0.465]{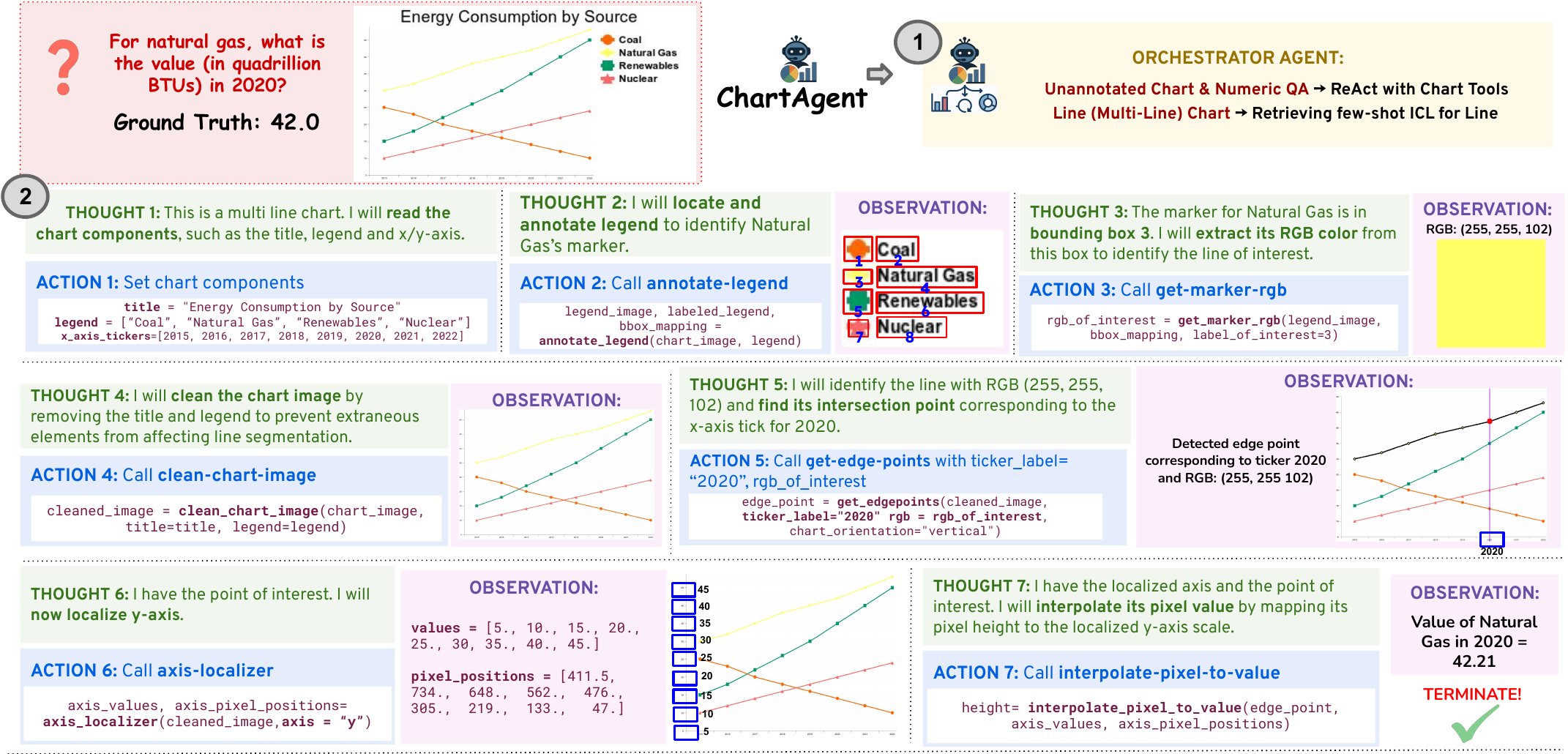}
    \caption{\textbf{Qualitative Trajectory on a Line (Multi-line) Chart.}}
  \label{fig:qual_traj_line_multi}
\end{figure*}

\begin{figure*}[htbp]
  \centering
  \includegraphics[scale = 0.465]{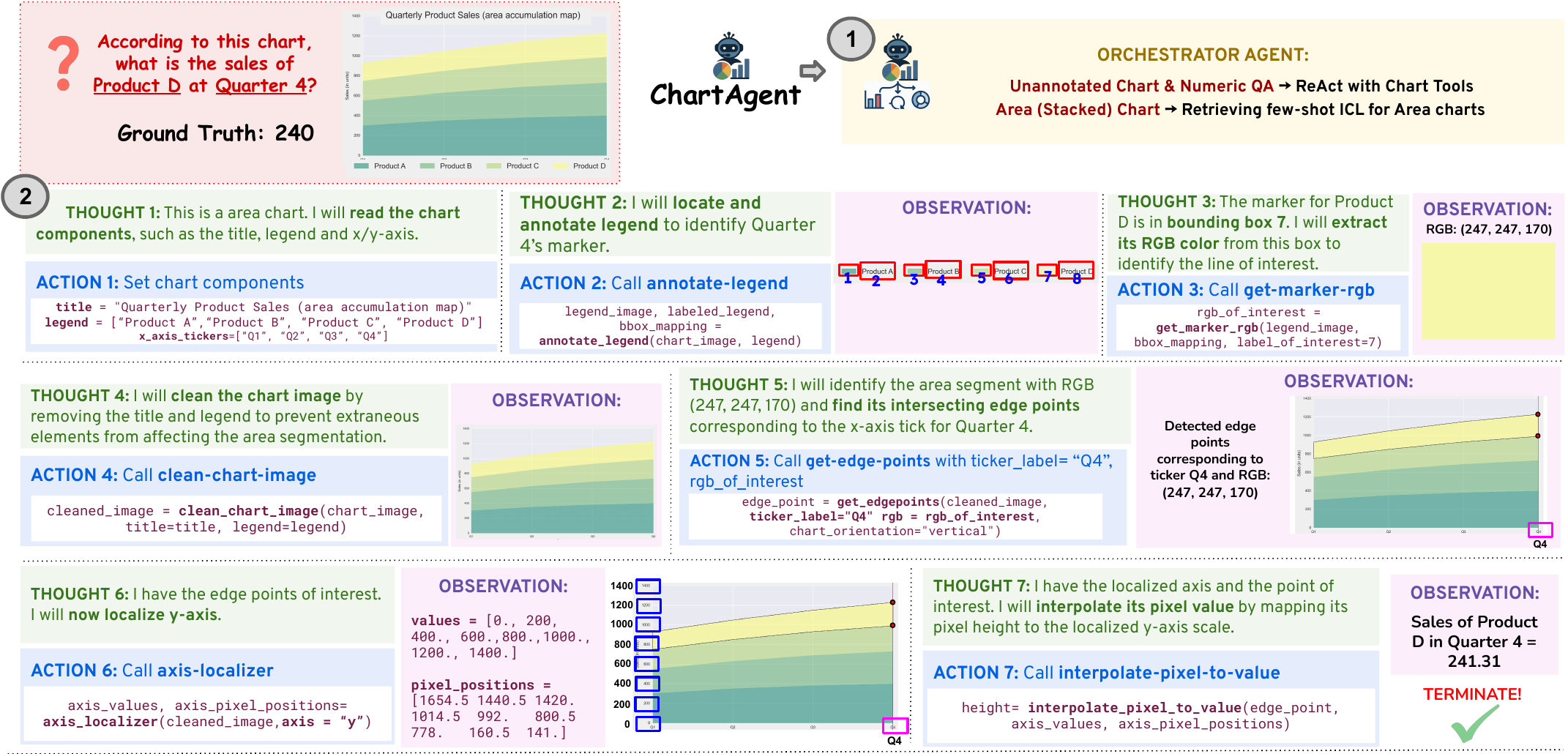}
    \caption{\textbf{Qualitative Trajectory on an Area (Stacked Area) Chart.}}
  \label{fig:qual_traj_area_stacked}
\end{figure*}

\begin{figure*}[htbp]
  \centering
  \includegraphics[scale = 0.58]{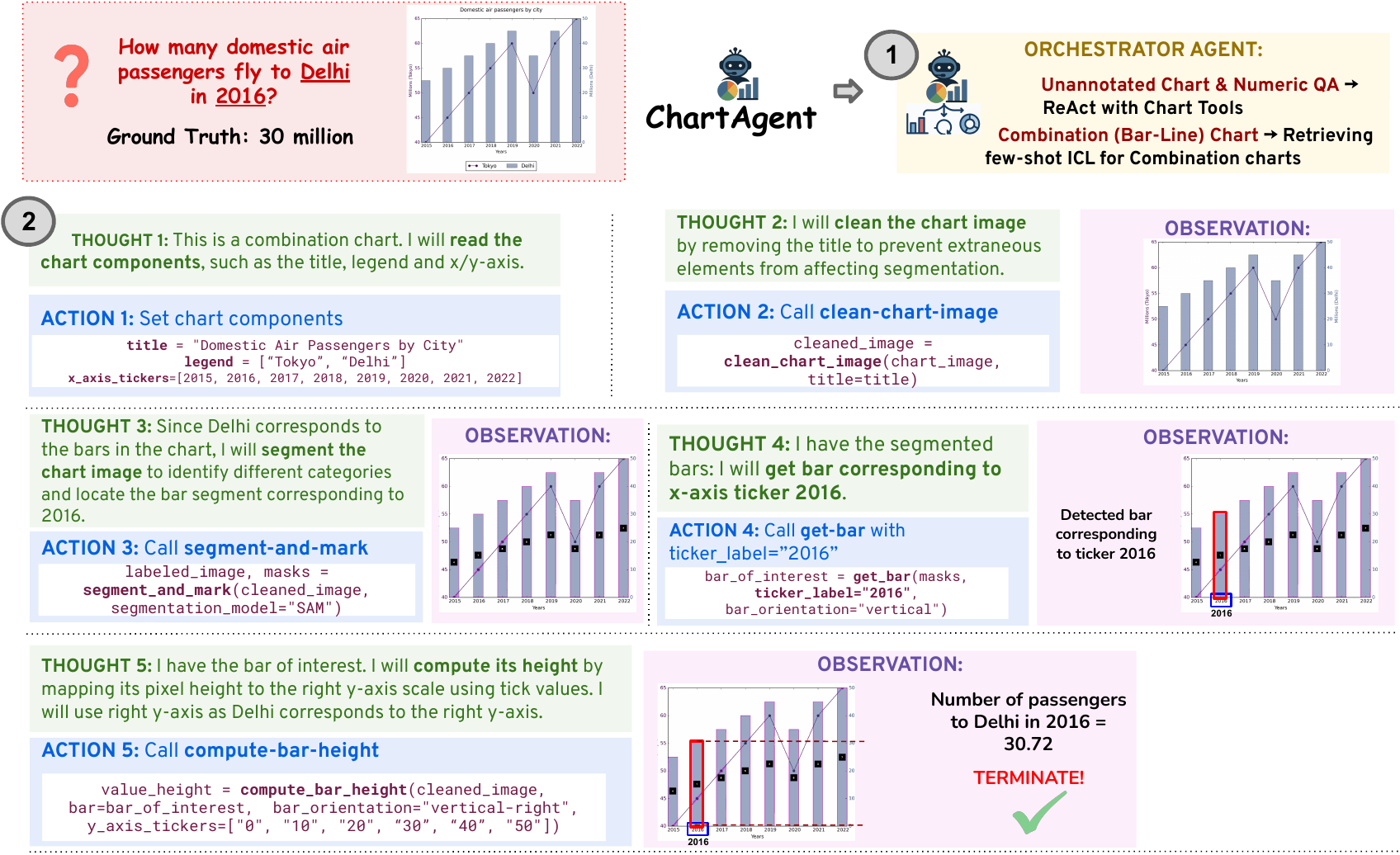}
    \caption{\textbf{Qualitative Trajectory on a Combination (Bar-Line) Chart.}}
  \label{fig:qual_traj_combo_bar_line_bar}
\end{figure*}

\begin{figure*}[htbp]
  \centering
  \includegraphics[scale = 0.465]{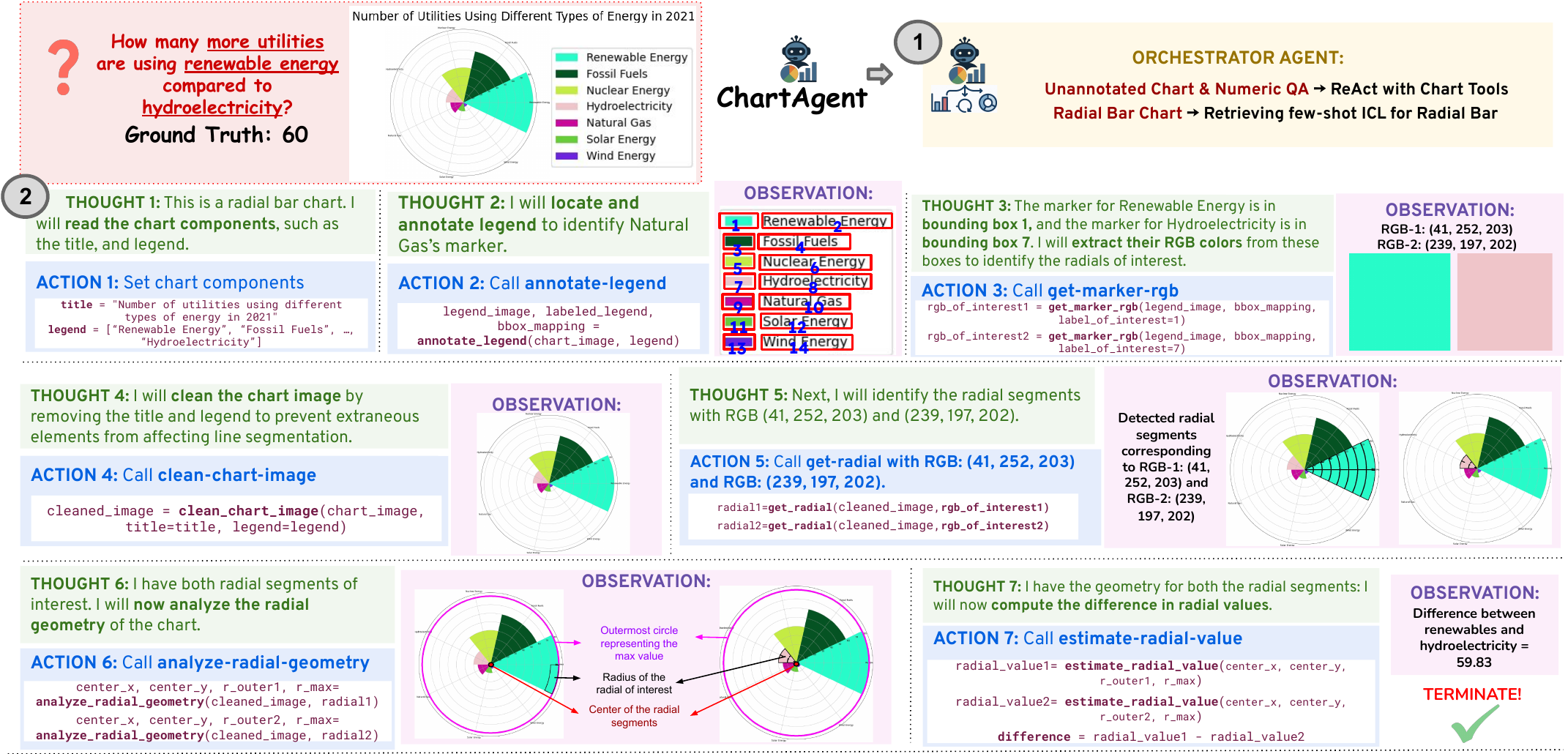}
    \caption{\textbf{Qualitative Trajectory on a Radial Bar Chart.}}
  \label{fig:qual_traj_radial_bar}
\end{figure*}

\begin{figure*}[htbp]
  \centering
  \includegraphics[scale = 0.465]{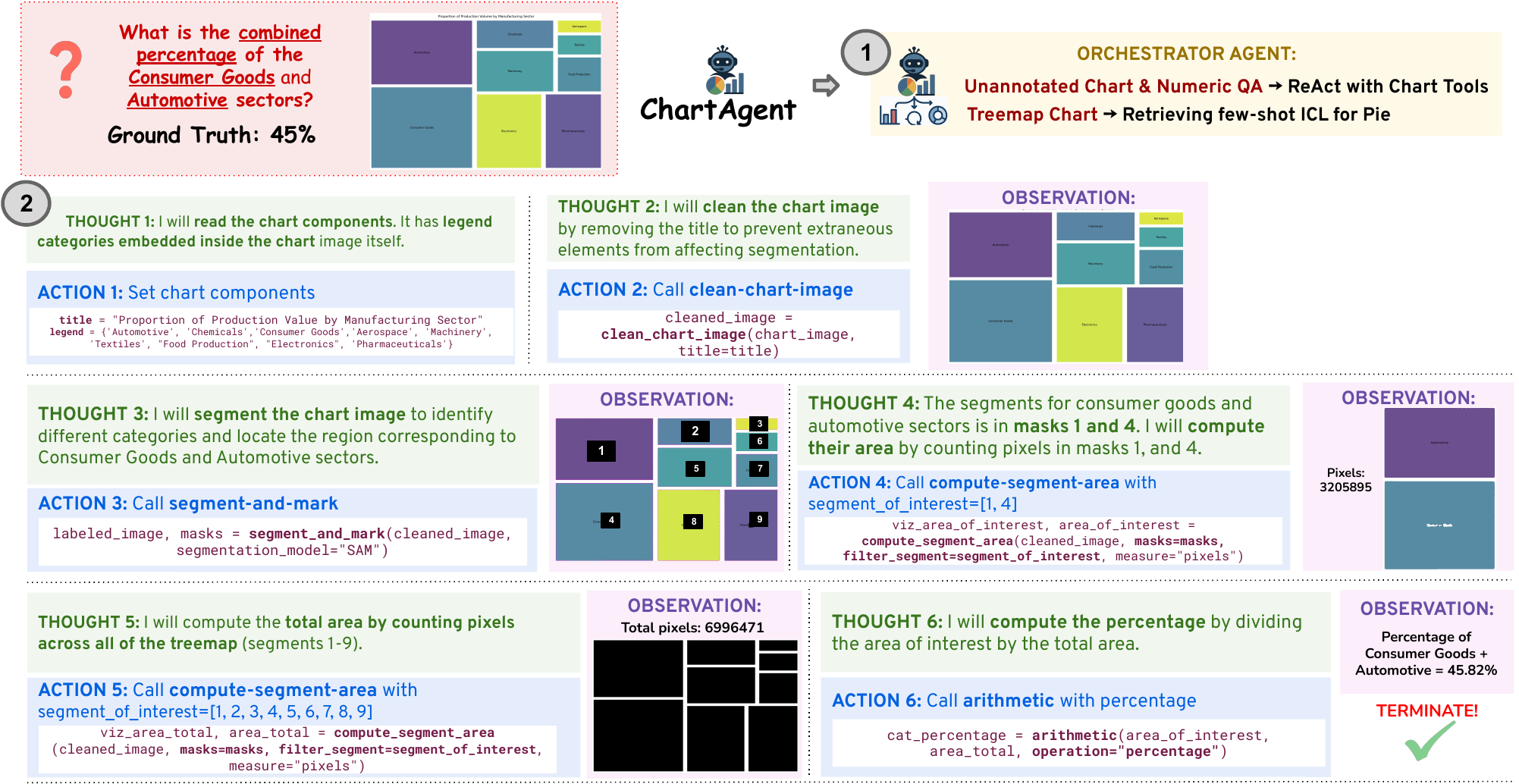}
    \caption{\textbf{Qualitative Trajectory on a Tree map Chart.}}
  \label{fig:qual_traj_treemap}
\end{figure*}

Figures~\ref{fig:qual_traj_pie_ring}-\ref{fig:qual_traj_treemap} show sample reasoning trajectories for \chartagent{} on questions involving diverse unannotated chart types.
The LLM-based orchestrator agent classifies the chart as unannotated, triggering the ReAct routine with chart tools. It also retrieves few-shot ICL examples specific to the corresponding chart type, after which the multi-turn interaction loop produces the accurate final answer. 


\subsubsection{Agent Trajectories Demonstrating Visual Self-Verification in Action}\label{appendix:qual_agent_trajectories-visual-self-verify}
\begin{figure*}[h!]
  \centering
  \includegraphics[scale = 0.49]{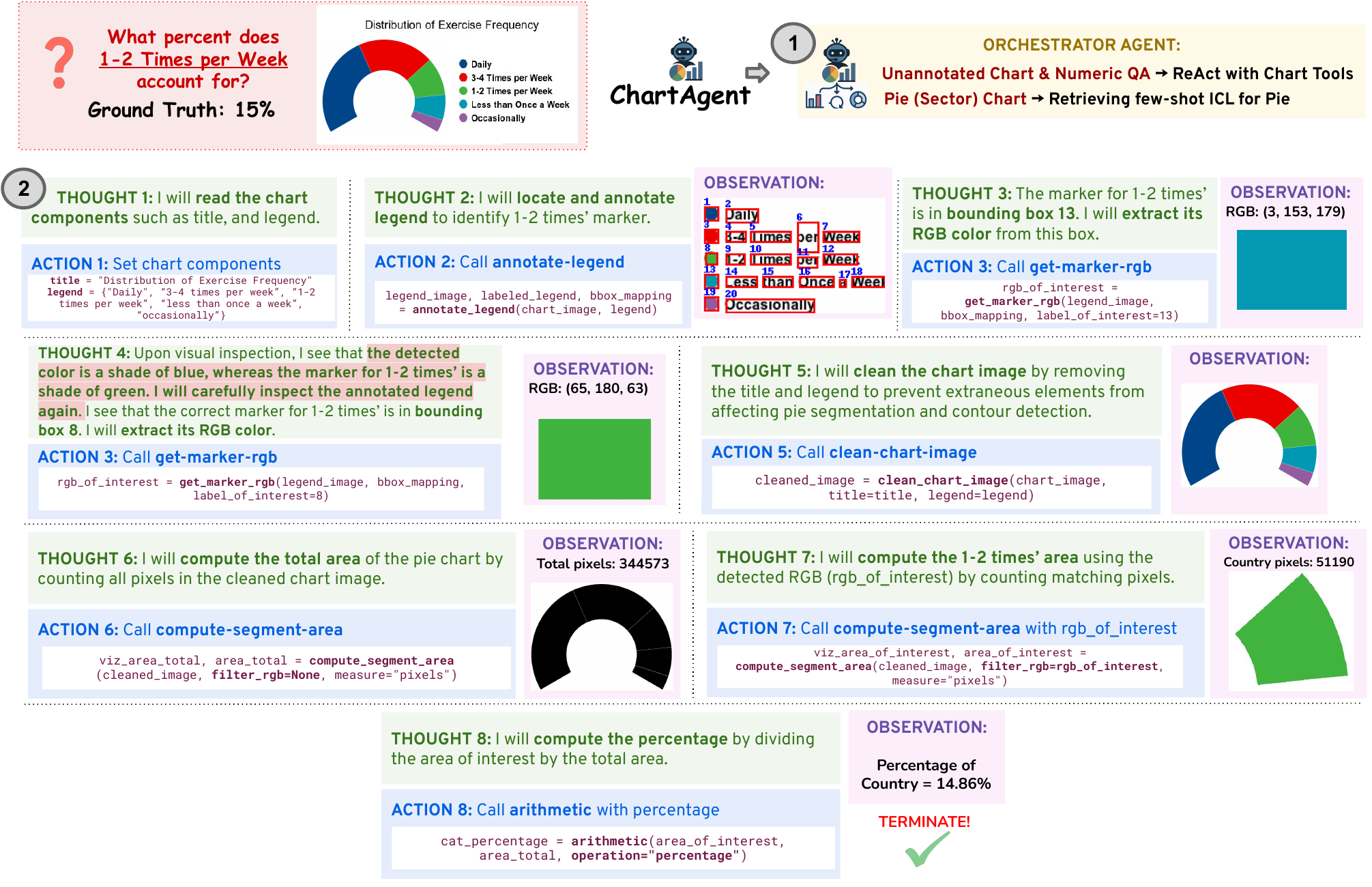}
    \caption{\textbf{Qualitative trajectory where visual self-verification is invoked (highlighted in \colorbox{red!20}{red}) during Thought~4.}}
  \label{fig:qual_traj_selfverify1}
\end{figure*}

\begin{figure*}[h!]
  \centering
  \includegraphics[scale = 0.49]{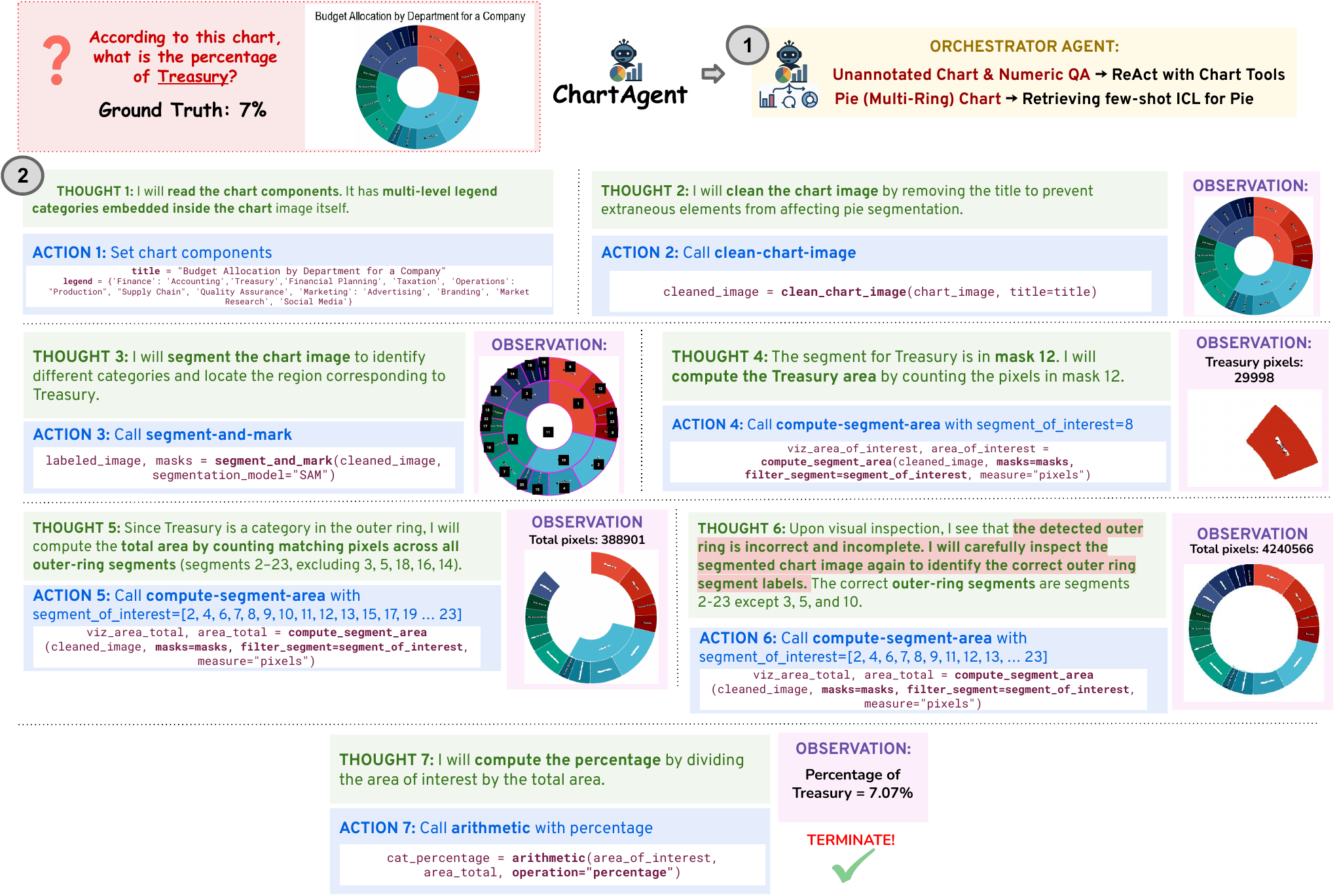}
    \caption{\textbf{Qualitative trajectory where visual self-verification is invoked (highlighted in \colorbox{red!20}{red}) during Thought~6.}}
  \label{fig:qual_traj_selfverify2}
\end{figure*}

\begin{figure*}[h!]
  \centering
  \includegraphics[scale = 0.45]{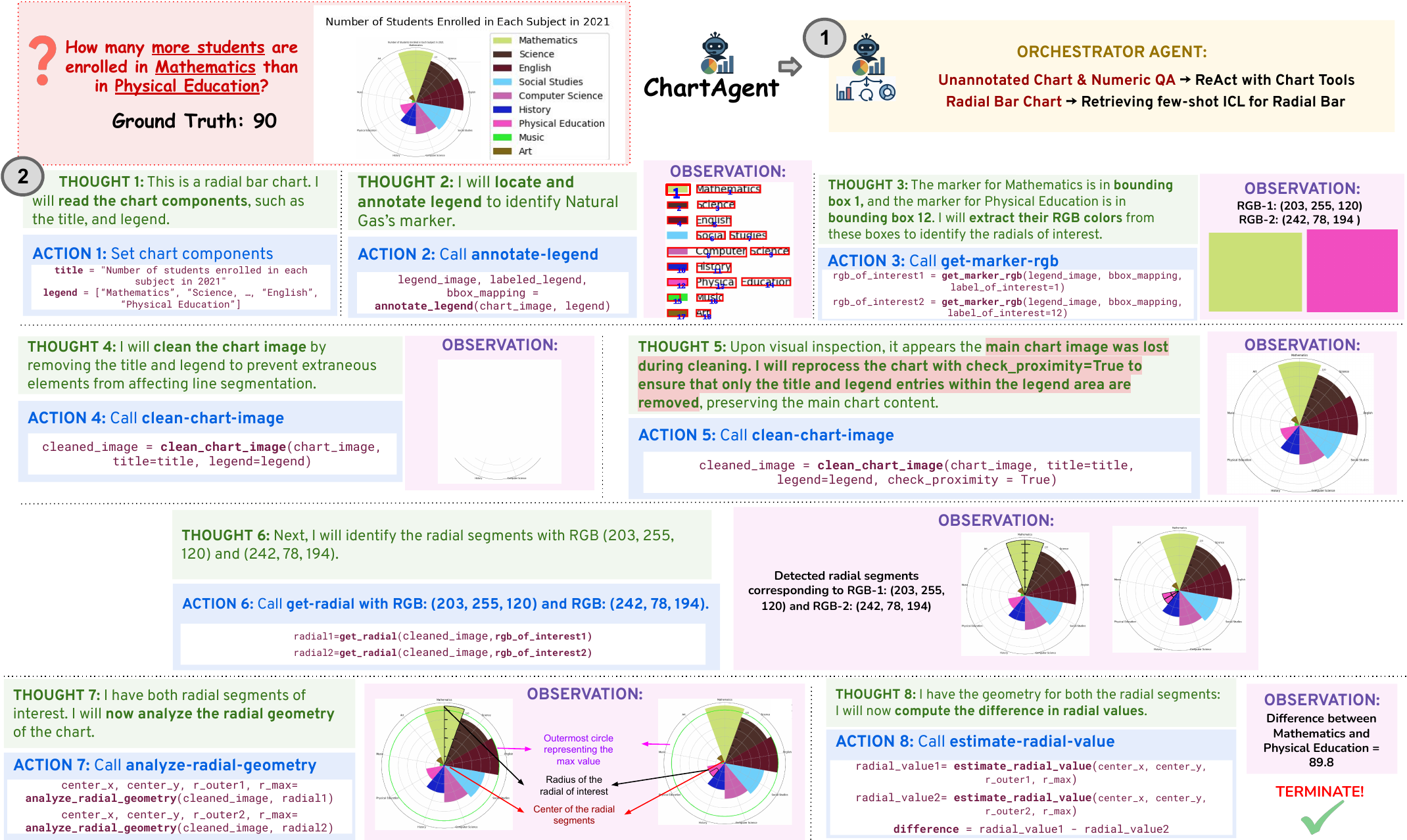}
    \caption{\textbf{Qualitative trajectory where visual self-verification is invoked (highlighted in \colorbox{red!20}{red}) during Thought~5.}}
  \label{fig:qual_traj_selfverify3}
\end{figure*}

Figures~\ref{fig:qual_traj_selfverify1}-\ref{fig:qual_traj_selfverify3} show sample reasoning trajectories for \chartagent{} where visual self-verification was invoked and the response was subsequently corrected.

\subsubsection{Agent Trajectories on Annotated Charts}\label{appendix:qual_agent_trajectories-annotated}
\begin{figure*}[htbp]
  \centering
  \includegraphics[scale = 0.59]{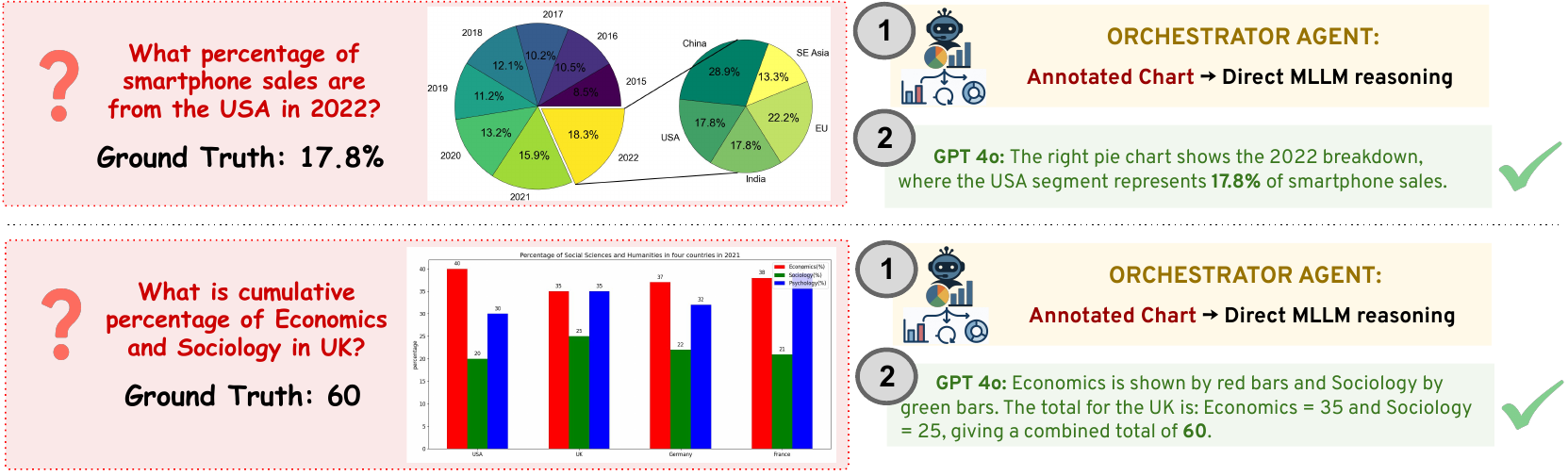}
    \caption{\textbf{Qualitative Trajectories on Annotated Chart Examples.}}
  \label{fig:qual_traj_annotated}
\end{figure*}

Figure~\ref{fig:qual_traj_annotated} shows sample reasoning trajectories for \chartagent{} on questions involving annotated charts. The LLM-based orchestrator classifies the chart as annotated and routes it to direct MLLM reasoning (GPT-4o), which produces the correct answer.



\subsubsection{Some Interesting Agent Trajectories}\label{appendix:qual_agent_trajectories-interesting-extra-trajs}
\chartagent{} exhibits adaptive decision-making during reasoning. For instance, in scatter plots with variable-sized points, it correctly identifies when certain points are too small to be captured through segmentation and instead relies on its own visual judgment to infer the answer—yielding accurate results without tool assistance. Similarly, when tool-based methods fail, the agent provides transparent and reasonable justifications for reverting to direct reasoning. For example: “\textit{THOUGHT 6: The interpolation failed because there is only one y-axis value available. I will directly estimate the Click-through Rate from the chart image using the visual position of the Campaign F bubble. ANSWER: The Click-through Rate for Campaign F when the Impressions is 700 is approximately 5.5\%. TERMINATE.}” Such cases highlight \chartagent{}’s ability to recognize tool limitations and intelligently switch to self-guided reasoning when appropriate.

\subsection{Representative Examples}\label{appendix:qual_examples}
\begin{figure*}[htbp]
  \centering
  \includegraphics[scale = 0.66]{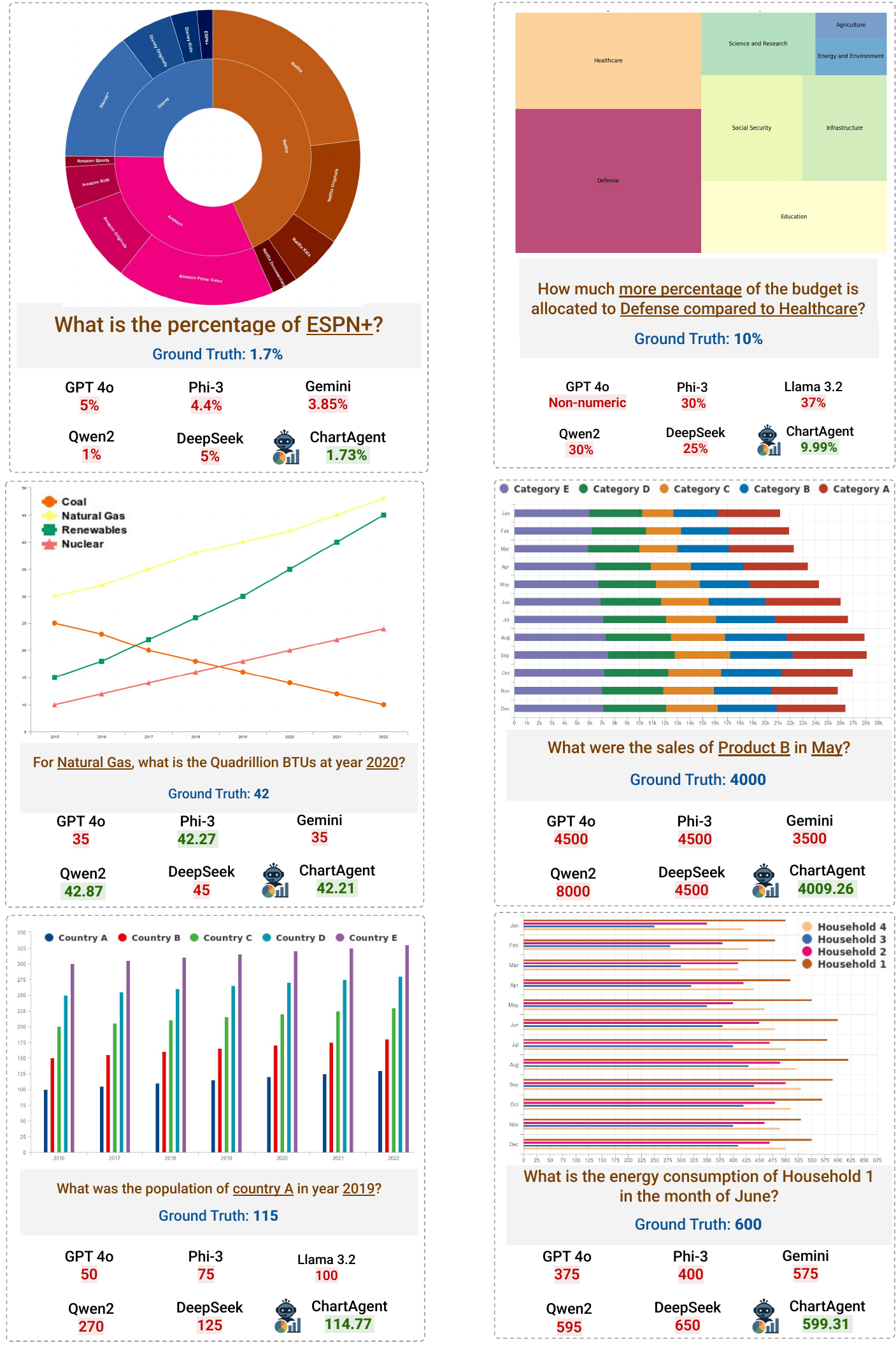}
  \caption{\textbf{Qualitative Examples.} Correct responses (within a 5\% error margin) are highlighted in \colorbox{green!15}{green}, while incorrect responses are highlighted in \colorbox{red!15}{red}.}
  \label{fig:qual_examples}
\end{figure*}

Figure~\ref{fig:qual_examples} presents qualitative comparison examples across the diverse chart types that \chartagent{} can handle, alongside several state-of-the-art baseline models (e.g., GPT, Phi, LLaMA, Qwen, Gemini, and DeepSeek). We observe improved performance across the variety of chart types in both the ChartBench and ChartX datasets.

\begin{figure*}[htbp]
  \centering
  \includegraphics[scale=0.595]{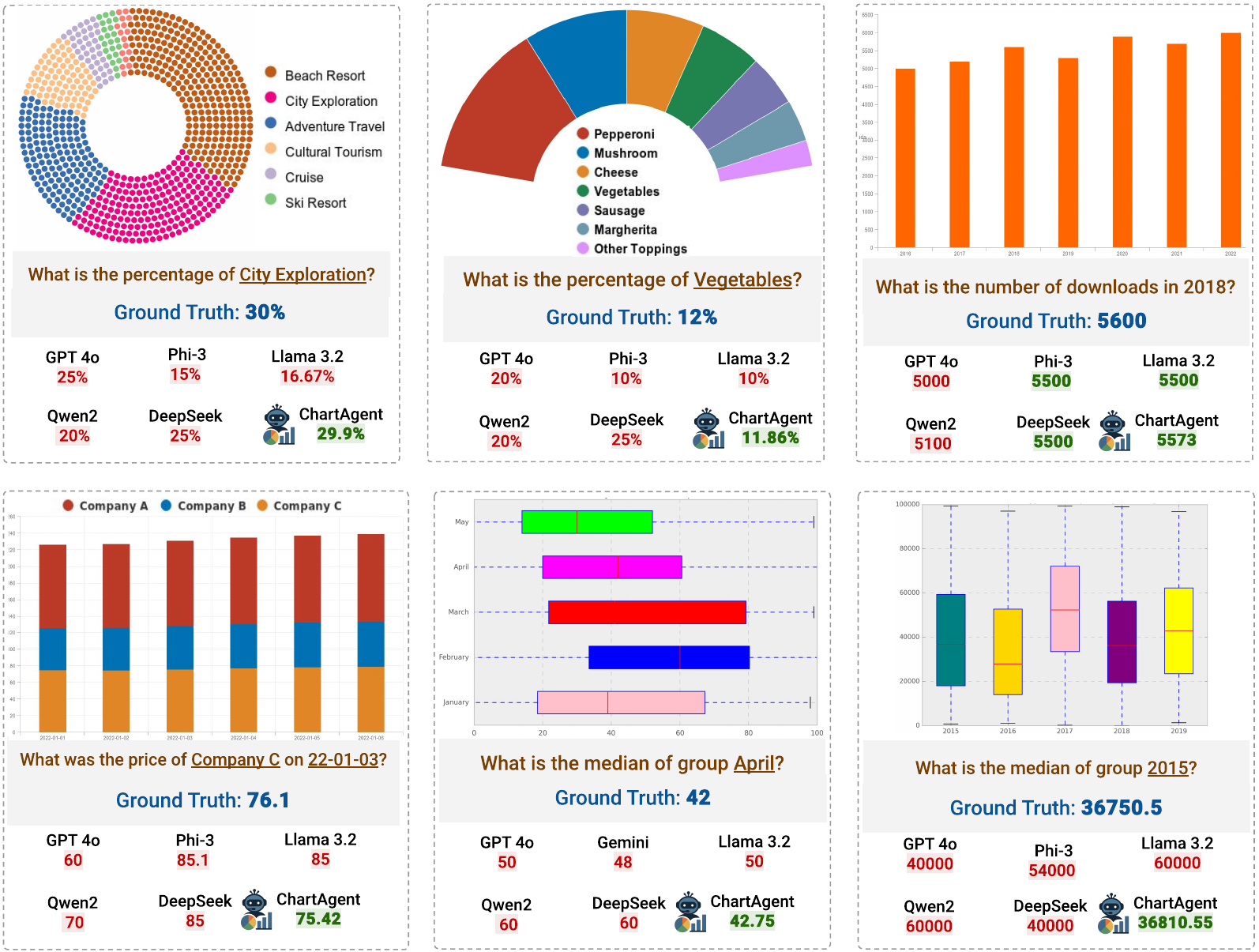}
\end{figure*}
\begin{figure*}[htbp]
  \centering
  \includegraphics[scale=0.595]{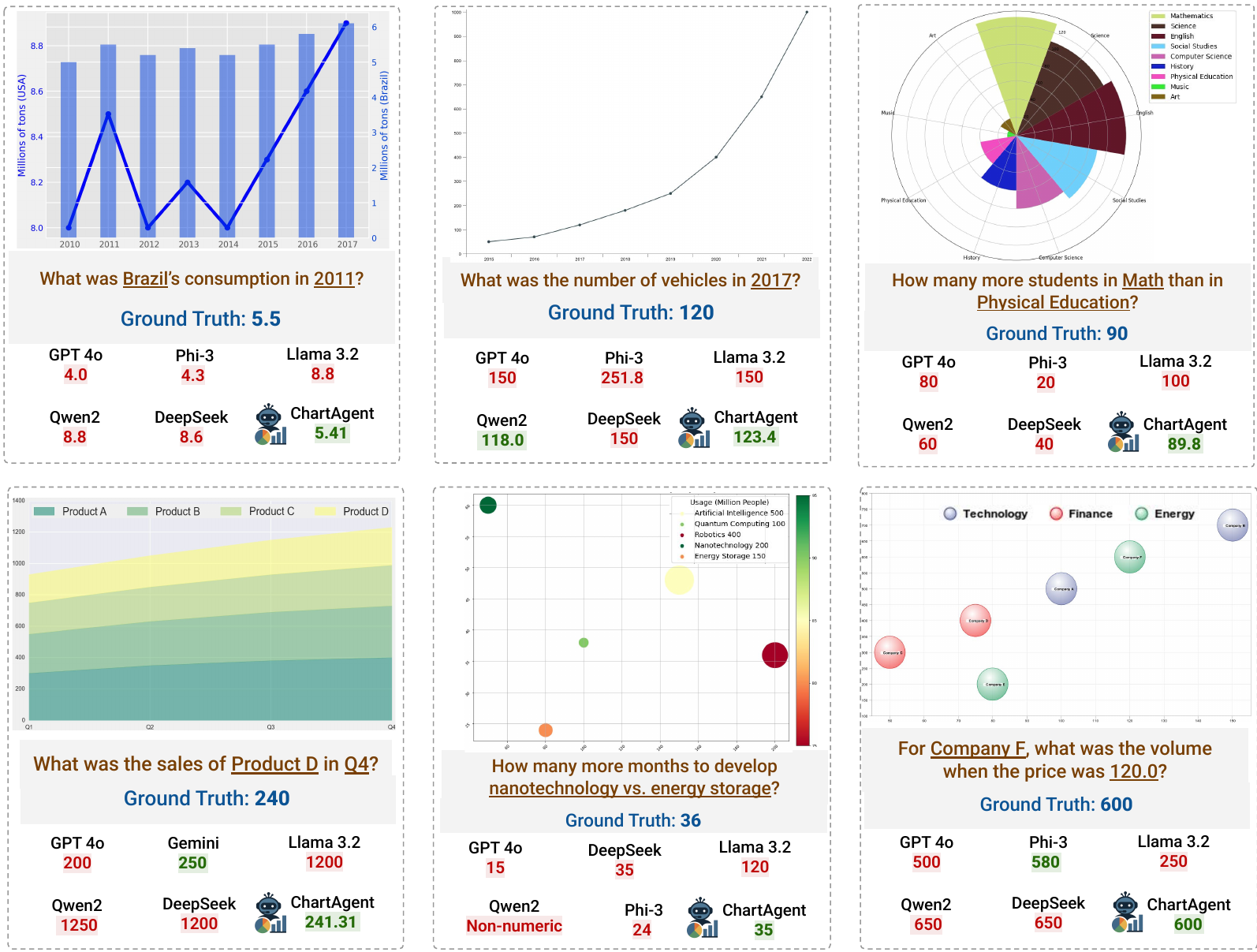}
\end{figure*}

\onecolumn
\section{Expanded Discussion on Results}\label{appendix:extended_results}

\subsection{Performance by Chart Type}\label{appendix:comparision_chart_type_full_table}
Table~\ref{tab:unannotated-results-combined} compares \chartagent{} with baseline methods on unannotated charts, stratified by chart type.
\begin{table*}[h!]
\vspace{-0.1cm}
\caption{\textbf{Accuracy on unannotated charts (\%) by chart type.}
\textcolor{red}{\textbf{Red:}} Best,
\textcolor{blue}{\underline{Blue:}} Second best. \textbf{Abbreviations:} Over: Overlay $\vert$ Stack: Stacked $\vert$ Mul: Multi $\vert$ Sing: Single $\vert$ Hor: Horizontal $\vert$ Vert: Vertical $\vert$ B-L: Bar-Line $\vert$ L-L: Line-Line $\vert$ Dir: Directed $\vert$ Undir: Undirected $\vert$ Combo: Combination. See App.~\ref{appendix:chart_types_figures} for examples of each chart type.}
\begin{subtable}{1.0\textwidth}
\centering
\small                                  
\setlength{\tabcolsep}{4pt}
\renewcommand{\arraystretch}{1.1}
\begin{adjustbox}{max width=\linewidth}
\begin{tabular}{l|cc|cccccccc|ccc|cc|cc|cc|ccc|ccc|c|c}
\toprule
\multirow[c]{2}{*}{\textbf{Model}} & \multicolumn{2}{c|}{\textbf{Area}} & \multicolumn{3}{c}{\textbf{Horizontal Bar}} & \multicolumn{2}{c}{\textbf{3D Bar}} & \multicolumn{3}{c|}{\textbf{Vertical Bar}} & \multicolumn{3}{c|}{\textbf{Box}} & \multicolumn{2}{c|}{\textbf{Combo}} & \multicolumn{2}{c|}{\textbf{Line}} & \multicolumn{2}{c|}{\textbf{Node}} & \multicolumn{3}{c|}{\textbf{Pie}} & \multicolumn{3}{c|}{\textbf{Radar}} & \textbf{Scatter} & \cellcolor{gray!15} \multirow[c]{2}{*}{\textbf{Avg.} $ \mathbf{\uparrow}$} \\
\cmidrule(lr{1em}){2-3}
\cmidrule(lr{1em}){4-6}
\cmidrule(lr{1em}){7-8}
\cmidrule(lr{1em}){9-11}
\cmidrule(lr{1em}){12-14}
\cmidrule(lr{1em}){15-16}
\cmidrule(lr{1em}){17-18}
\cmidrule(lr{1em}){19-20}
\cmidrule(lr{1em}){21-23}
\cmidrule(lr{1em}){24-26}
\cmidrule(lr{1em}){27-27}
& Over & Stack & Mul & Sing & Stack & Mul & Stack & Mul & Sing & Stack & Hor & Vert & Stock & B-L & L-L & Mul & Sing & Dir & Undir & Mul & Ring & Sector & Mul & Fill & Sing & 3D & \cellcolor{gray!15} \\
\midrule
\multicolumn{28}{l}{\cellcolor{orange!15}\textit{Proprietary Multimodal Large Language Models}}  \\
\midrule
GPT 4o & 21.0 & 18.0 &  24.0 &   59.0 &10.0 &  20.0 & 6.0 &38.0 & 73.0 &  12.0 &   \textcolor{blue}{\underline{20.0}} &   26.0 &   \textcolor{blue}{\underline{63.0}} &  35.0 &   41.0 & 37.0 &  75.0 & \textcolor{red}{\textbf{91.0}} &   \textcolor{red}{\textbf{91.0}} &  3.0 &  \textcolor{blue}{\underline{32.0}} & \textcolor{blue}{\underline{34.0}} &   \textcolor{blue}{\underline{22.0}} & 20.0 & 6.0 & 63.0 & \cellcolor{gray!15} 36.15 \\
GPT 4o-mini &   23.0 &  7.0 &  13.0 &   27.0 & 7.0 &  20.0 & 7.0 &19.0 & 56.0 &   2.0 &   13.0 &   12.0 &   57.0 &  29.0 &   36.0 & 19.0 &  50.0 & 88.0 &   \textcolor{red}{\textbf{91.0}} &  1.0 &   7.0 &16.0 &3.0 & 8.0 & 1.0 &43.0 & \cellcolor{gray!15}   25.19 \\
Claude 3 &   15.0 &  5.0 &  12.0 &   32.0 & 7.0 &  25.0 & 5.0 &51.0 & 67.0 &   6.0 &8.0 &5.0 &   62.0 &  24.0 &   23.0 & 28.0 &  50.0 & 75.0 &   71.0 &  \textcolor{blue}{\underline{7.0}} &   9.0 &12.0 &3.0 &13.0 &11.0 &51.0 & \cellcolor{gray!15}   26.04 \\
Gemini 1.5 &   5.0 &  4.0 &  28.0 &   52.0 & 7.0 &  14.0 & 4.0 & 39.05 & 49.0 & 5.0 &   13.0 & 18.0 & 24.0 &   28.0 &  5.0 &   7.0 & \textcolor{red}{\textbf{91.0}} &  48.0 & 59.26 &   1.0 &  14.0 &   29.52 & 1.0 & 7.0 & 0.0 & 45.0 & \cellcolor{gray!15} 27.27 \\
\midrule
\multicolumn{28}{l}{\cellcolor{orange!15}\textit{Open-weights Multimodal Large Language Models}}  \\
\midrule
BLIP-2 & 0.0 & 0.0 & 3.0 & 1.0 & 4.0 & 5.0 & 4.0 & 2.0 & 4.0 & 3.0 & 3.0 & 1.0 & 3.0 & 0.0 & 0.0 & 4.0 & 4.0 & 3.0 & 5.0 & 3.0 & 2.0 & 2.0 & 9.0 & 2.0 & 6.0 & 3.0& \cellcolor{gray!15} 2.92 \\
CogAgent &   14.0 &  2.0 &   3.0 &   15.0 & 6.0 &  15.0 & 4.0 &11.0 &  9.0 &   4.0 &8.0 &6.0 &   22.0 &  21.0 &   16.0 &  6.0 &  20.0 & 20.0 &   31.0 &  3.0 &  18.0 & 9.0 &2.0 & 4.0 & \textcolor{blue}{\underline{13.0}} &20.0 & \cellcolor{gray!15}   11.62 \\
CogVLM &   21.0 &  3.0 &   4.0 &   17.0 & 3.0 &  18.0 & 3.0 &11.0 & 16.0 &   4.0 &7.0 &7.0 &2.0 &  24.0 &   20.0 &  9.0 &  10.0 & 19.0 &   24.0 &  1.0 &   7.0 &25.0 &   13.0 &15.0 & 6.0 &16.0 & \cellcolor{gray!15}  11.62\\
DeepSeek-VL2 & 29.0 & 11.0 & 25.0 & 57.0 & 8.0 & 36.0 & 8.0 & 58.0 & 82.0 & 13.0 & 11.0 & 3.0 & 51.0 & 46.0 & 48.0 & 51.0 & 8.0 & 31.0 & 36.0 & 0.0 & 6.0 & 15.0 & 13.0 & 21.0 & 5.0 & 44.0 &\cellcolor{gray!15} 30.31 \\
DocOwl1.5 &   19.0 &  8.0 &  21.0 &   69.0 & 3.0 &  20.0 & 0.0 &39.0 & 78.0 &   6.0 &7.0 &   17.0 &   32.0 &  15.0 &   23.0 & 23.0 &  74.0 & 42.0 &   47.0 &  2.0 &  14.0 & 8.0 &2.0 &14.0 &10.0 &20.0 & \cellcolor{gray!15}  23.58 \\
InstructBLIP & 5.0 & 7.0 & 3.0 & 11.0 & 1.0 & 5.0 & 4.0 & 3.0 & 11.0 & 4.0 & 4.0 & 1.0 & 1.0 & 3.0 & 5.0 & 2.0 & 9.0 & 23.0 & 26.0 & 2.0 & 1.0 & 3.0 & 2.0 & 7.0 & 0.0 & 11.0 &\cellcolor{gray!15} 5.92 \\
InternVL3 &   25.0 & 16.0 &  45.0 &   80.0 & \textcolor{blue}{\underline{19.0}} &  \textcolor{blue}{\underline{38.0}} & 1.0 &44.0 & 80.0 &  16.0 &   16.0 &   23.0 &   60.0 &  27.0 &   24.0 & 30.0 &  56.0 & 62.0 &   52.0 &  0.0 &   2.0 & 9.0 &   \textcolor{red}{\textbf{24.0}} & \textcolor{blue}{\underline{24.0}} & 6.0 &25.0 & \cellcolor{gray!15}  30.92 \\
LLama3.2 & \textcolor{blue}{\underline{46.0}} & 21.0 & 58.0 & \textcolor{blue}{\underline{91.0}} & 11.0 & 31.0 & 4.0 & 71.0 & 89.0 & 10.0 & 6.0 & 6.0 & 49.0 & 42.0 & 46.0 & 63.0 & \textcolor{blue}{\underline{87.0}} & 42.0 & 58.0 & 5.0 & 4.0 & 25.0 & 8.0 & 17.0 & 10.0 & 46.0 &\cellcolor{gray!15} 36.38 \\
Llava1.6 & 7.0 & 7.0 & 11.0 & 12.0 & 8.0 & 18.0 & 1.0 & 7.0 & 19.0 & 1.0 & 5.0 & 3.0 & 0.0 & 16.0 & 15.0 & 7.0 & 5.0 & 39.0 & 45.0 & 1.0 & 4.0 & 5.0 & 3.0 & 1.0 & 2.0 & 16.0 &\cellcolor{gray!15} 9.92 \\
Llava1.5 & 1.0 & 5.0 & 8.0 & 12.0 & 7.0 & 6.0 & 3.0 & 5.0 & 9.0 & 4.0 & 4.0 & 1.0 & 2.0 & 7.0 & 1.0 & 3.0 & 5.0 & 11.0 & 22.0 & 0.0 & 8.0 & 11.0 & 9.0 & 13.0 & 11.0 & 14.0 & \cellcolor{gray!15} 7.00 \\
LlaVA-OneVision & 9.0 & 2.0 & 9.0 & 7.0 & 12.0 & 12.0 & 10.0 & 11.0 & 7.0 & 7.0 & 12.0 & 8.0 & 14.0 & 7.0 & 10.0 & 2.0 & 5.0 & 38.0 & 36.0 & 0.0 & 1.0 & 1.0 & 24.0 & 12.0 & 1.0 & 16.0 & \cellcolor{gray!15} 10.50 \\
mPLUG-Owl3 & 11.0 & 2.0 & 9.0 & 20.0 & 1.0 & 15.0 & 2.0 & 11.0 & 15.0 & 2.0 & 7.0 & 6.0 & 16.0 & 14.0 & 15.0 & 14.0 & 10.0 & 52.0 & 41.0 & 0.0 & 10.0 & 23.0 & 7.0 & 17.0 & 3.0 & 6.0 &\cellcolor{gray!15} 12.65 \\
Phi3-vision & 27.0 & \textcolor{blue}{\underline{37.0}} & 43.0 & 78.0 & 8.0 & \textcolor{red}{\textbf{40.0}} & 7.0 & 86.0 & \textcolor{blue}{\underline{92.0}}  & 30.0 & 9.0 & 15.0 & 48.0 & 31.0 & 55.0 & \textcolor{blue}{\underline{66.0}} & 84.0 & 39.0 & 51.0 & 2.0 & 14.0 & 21.0 & 11.0 & \textcolor{red}{\textbf{26.0}} & \textcolor{red}{\textbf{66.0}} & \textcolor{red}{\textbf{73.0}} &\cellcolor{gray!15} 40.77 \\
Pixtral & 26.0 & 10.0 & 25.0 & 51.0 & 6.0 & 30.0 & 5.0 & 39.0 & 89.0 & 10.0 & 16.0 & 29.0 & 39.0 & 19.0 & 24.0 & 17.0 & 32.0 & 68.0 & 59.0 & 2.0 & 21.0 & 28.0 & 13.0 & 9.0 & 8.0 & 72.0 & \cellcolor{gray!15} 28.73 \\
Qwen2VL &   \textcolor{red}{\textbf{57.0}} & 18.0 &  \textcolor{red}{\textbf{87.0}} &   \textcolor{red}{\textbf{97.0}} & 17.0 &  \textcolor{red}{\textbf{40.0}} & 7.0 & \textcolor{red}{\textbf{94.0}} & \textcolor{red}{\textbf{97.0}} &  \textcolor{blue}{\underline{24.0}} &   13.0 & 4.0 &  \textcolor{red}{\textbf{64.0}} &  37.0 &   46.0 & \textcolor{red}{\textbf{80.0}} &  85.0 & 80.0 &   \textcolor{blue}{\underline{86.0}} &  1.0 &  12.0 & 9.0 &9.0 &11.0 & 9.0 &47.0 & \cellcolor{gray!15}   \textcolor{blue}{\underline{43.50}} \\
QwenVLChat &6.0 &  8.0 &   4.0 &8.0 & 2.0 &   6.0 & 3.0 & 5.0 & 17.0 &   5.0 &0.0 &1.0 &2.0 &   9.0 &7.0 &  6.0 &   6.0 & 20.0 &   22.0 &  2.0 &   2.0 & 3.0 &8.0 & 3.0 &10.0 & 5.0 & \cellcolor{gray!15}   6.54 \\
SmolVLM &7.0 &  3.0 &  12.0 &   17.0 & 3.0 &  12.0 & 1.0 &14.0 & 26.0 &   0.0 &7.0 &7.0 &   28.0 &  15.0 &   13.0 &  5.0 &  23.0 & 62.0 &   54.0 &  0.0 &   2.0 &12.0 &   14.0 &16.0 & 9.0 &14.0 & \cellcolor{gray!15}   14.46\\
SPHINX-V & 7.0 & 2.0 & 3.0 & 17.0 & 4.0 & 16.0 & 10.0 & 9.0 & 26.0 & 4.0 & 4.0 & 7.0 & 2.0 & 16.0 & 22.0 & 7.0 & 10.0 & 46.0 & 54.0 & 2.0 & 3.0 & 16.0 & 4.0 & 8.0 & 14.0 & 7.0 & \cellcolor{gray!15} 12.30 \\
VisualGLM &6.0 &  3.0 &   1.0 &2.0 & 4.0 &   2.0 & 1.0 & 4.0 &  6.0 &   5.0 &1.0 &6.0 &0.0 &   0.0 &2.0 &  6.0 &   3.0 & 63.0 &   53.0 &  1.0 &   5.0 & 4.0 &7.0 & 4.0 & 2.0 & 8.0 & \cellcolor{gray!15}  7.65\\
\midrule
\multicolumn{28}{l}{\cellcolor{orange!15}\textit{Chart-related Models}}  \\
\midrule
ChartGemma &   25.0 &  8.0 &  21.0 &   54.0 & 9.0 &  21.0 & 3.0 &36.0 & 86.0 &   6.0 &5.0 &5.0 &   22.0 &  31.0 &   36.0 & 24.0 &  68.0 & 32.0 &   38.0 &  0.0 &   2.0 & 8.0 &3.0 & 8.0 & 3.0 &29.0 & \cellcolor{gray!15}  22.42  \\
ChartInstruct &   20.0 &  6.0 &  23.0 &   72.0 & 1.0 &  17.0 & 7.0 &36.0 & 85.0 &   6.0 &9.0 &   \textcolor{blue}{\underline{27.0}} &5.0 &  27.0 &   24.0 & 13.0 &  68.0 & 18.0 &   26.0 &  2.0 &   8.0 & 3.0 &8.0 & 6.0 & 4.0 & 4.0 & \cellcolor{gray!15}  20.19 \\
ChartLlama &   20.0 &  2.0 &   2.0 &   15.0 & 7.0 &  12.0 & 7.0 &14.0 & 20.0 &   7.0 &5.0 &9.0 &1.0 &  16.0 &   18.0 &  3.0 &  10.0 & 41.0 &   38.0 &  2.0 &   8.0 &15.0 &0.0 & 0.0 & 11.0 &14.0 & \cellcolor{gray!15}  11.42 \\
ChartVLM &   16.0 &  8.0 &  24.0 &   78.0 &10.0 &  29.0 & 7.0 &60.0 & 85.0 &   8.0 &3.0 &   23.0 &7.0 &  37.0 &   40.0 & 30.0 &  95.0 & 13.0 &   10.0 &  1.0 &   7.0 & 5.0 &2.0 & 4.0 & 6.0 &14.0 & \cellcolor{gray!15}  23.92 \\
DePlot &   18.0 &  2.0 &  43.0 &   74.0 &13.0 &  34.0 & \textcolor{blue}{\underline{9.0}} &66.0 & 78.0 &   7.0 &   \textcolor{blue}{\underline{20.0}} &   20.0 &0.0 &  \textcolor{blue}{\underline{48.0}} &   45.0 & 14.0 &  63.0 & \textcolor{blue}{\underline{84.0}}&   73.0 &  4.0 &   3.0 & 5.0 &2.0 & 2.0 & 3.0 & 2.0 & \cellcolor{gray!15}  28.15 \\
MatCha & 3.0 & 1.0 & 8.0 & 29.0 & 0.0 & 8.0 & 1.0 & 18.0 & 40.0 & 11.0 & 3.0 & 17.0 & 1.0 & 16.0 & 14.0 & 13.0 & 18.0 & 16.0 & 19.0 & 0.0 & 1.0 & 1.0 & 2.0 & 0.0 & 2.0 & 10.0 &\cellcolor{gray!15} 9.69 \\
OneChart & 0.0 &  6.0 &  27.0 &   67.0 & 2.0 &  16.0 & 2.0 &69.0 & 80.0 &  11.0 &0.0 &   17.0 &0.0 &  12.0 &   \textcolor{blue}{\underline{62.0}} & 38.0 &  90.0 & 65.0 &   60.0 &  0.0 &   0.0 & 7.0 &0.0 & 0.0 & 0.0 & 2.0 & \cellcolor{gray!15}  26.81\\
TinyChart &   32.0 & 22.0 &  71.0 &   88.0 &13.0 &  37.0 & \textcolor{red}{\textbf{15.0}} &76.0 & 82.0 &  21.0 &2.0 &3.0 &4.0 &  46.0 &   50.0 & 51.0 &  \textcolor{red}{\textbf{91.0}} & 22.0 &   35.0 &  1.0 &  20.0 &21.0 &   10.0 & 8.0 & 4.0 &27.0 & \cellcolor{gray!15}  32.77 \\
UniChart &   15.0 &  5.0 &  24.0 &   59.0 & 7.0 &  11.0 & 0.0 &32.0 & 60.0 &   1.0 &3.0 &8.0 &6.0 &  16.0 &   25.0 & 13.0 &  37.0 & 36.0 &   33.0 &  3.0 &   0.0 & 1.0 &4.0 & 4.0 & 1.0 & 11.0 & \cellcolor{gray!15}  15.96 \\
\midrule
\multicolumn{28}{l}{\cellcolor{orange!15}\textit{Multimodal Agentic Framework (Ours)}} \\
\midrule
\textbf{\chartagent{}} &  30.0 &  \textcolor{red}{\textbf{38.0}} &  \textcolor{blue}{\underline{79.0}} &  76.0 &  \textcolor{red}{\textbf{82.0}} &  20.0 &  6.0 &  \textcolor{blue}{\underline{88.0}} &  88.0 &  \textcolor{red}{\textbf{76.0}} &  \textcolor{red}{\textbf{89.0}} &  \textcolor{red}{\textbf{83.0}} & \textcolor{red}{\textbf{64.0}} & \textcolor{red}{\textbf{67.0}} & \textcolor{red}{\textbf{65.0}} &  63.0 &  81.0 &  \textcolor{red}{\textbf{91.0}} & \textcolor{red}{\textbf{91.0}} &  \textcolor{red}{\textbf{18.0}} &  \textcolor{red}{\textbf{94.0}} &  \textcolor{red}{\textbf{80.0}} & \textcolor{blue}{\underline{22.0}} &  20.0 &  6.0 &  \textcolor{blue}{\underline{64.0}} & \cellcolor{gray!15}  \textcolor{red}{\textbf{60.81}} \\

\bottomrule
\end{tabular}%
\end{adjustbox}
\vspace{0.2mm}
\caption{\textbf{ChartBench Dataset} (9 major chart types, 42 subtypes; 26 unannotated)}
\label{chartbench_unannotated_results}
\end{subtable}
\vspace{0.05in}
\begin{subtable}{1.0\textwidth}
\centering
\begin{adjustbox}{max width=0.85\linewidth}
\begin{tabular}{l|c|c|c|c|c|c|c|c|c|c|c|c|c|c|c}
\toprule
\textbf{Model} & \textbf{Area} & \textbf{Bar} & \textbf{3D Bar} & \textbf{Box} & \textbf{Bubble} & \textbf{Candlestick} & \textbf{Heatmap} & \textbf{Histogram} & \textbf{Line} & \textbf{Multi-Axes} & \textbf{Radar} & \textbf{Ring} & \textbf{Rose} & \textbf{Treemap} & \cellcolor{gray!15}  \textbf{Average $\uparrow$}\\ 
\midrule
\multicolumn{16}{l}{\cellcolor{orange!15}\textit{Proprietary Multimodal Large Language Models}}  \\
\midrule
GPT 4o &      26.0 &    35.19 & 22.0 & 40.0 & 44.0 &    \textcolor{red}{\textbf{  78.0}} &  \textcolor{blue}{\underline{50.0}} &    42.55 &     53.92 &      \textcolor{blue}{\underline{18.0}} & \textcolor{blue}{\underline{30.0}} & 30.0 &  \textcolor{red}{\textbf{34.0}} &  \textcolor{blue}{\underline{44.83}} & \cellcolor{gray!15}  39.44\\
GPT 4o-mini &      16.0 &    32.41 & \textcolor{blue}{\underline{34.0}} & 42.0 & 48.0 &       \textcolor{blue}{\underline{66.0}} &   \textcolor{blue}{\underline{50.0}} &    34.04 &     39.22 &       8.0 & 28.0 & \textcolor{blue}{\underline{35.0}} &  26.0 &  24.14 & \cellcolor{gray!15} 33.94\\
Claude 3 Haiku &      26.0 &    25.0 & 20.0 & 22.0 & 38.0 &      48.0 &  \textcolor{blue}{\underline{50.0}} &    27.66 &     33.33 &       6.0 & 22.0 & 15.0 &  20.0 &  10.34 & \cellcolor{gray!15}  25.77\\
Gemini 1.5 &  26.0 & 40.74 & 22.0 & \textcolor{red}{\textbf{  48.0}} & \textcolor{blue}{\underline{50.0}} & 8.0 & 25.0 & 44.68 & 33.33 & \textcolor{blue}{\underline{18.0}} & 20.0 & 30.0 & 30.0 & 20.69 & \cellcolor{gray!15} 31.41\\
\midrule
\multicolumn{16}{l}{\cellcolor{orange!15}\textit{Open-weights Multimodal Large Language Models}}  \\
\midrule
BLIP-2  & 0.0 & 0.9 & 2.0 & 0.0 & 2.0 & 2.0 & 0.0 & 2.1 & 2.0 & 0.0 & 6.0 & 0.0 & 4.0 & 0.0 & \cellcolor{gray!15} 1.69 \\
CogAgent &      16.0 &    23.15 & 30.0 & 30.0 & 20.0 &      48.0 &  \textcolor{blue}{\underline{50.0}} &    19.15 &     30.39 &      10.0 & 26.0 & 15.0 &  24.0 &  17.24 & \cellcolor{gray!15} 24.93\\
CogVLM &      20.0 &    31.48 & 30.0 & 28.0 & 16.0 &      34.0 &  \textcolor{blue}{\underline{50.0}} &    17.02 &     25.49 &      12.0 & 26.0 & 15.0 &  16.0 &  27.59  & \cellcolor{gray!15} 24.23\\
DeepSeek-VL2 & 24.0 & 41.7 & 24.0 & 36.0 & 34.0 & 62.0 & 50.0 & 38.3 & 54.9 & 14.0 & 26.0 & 20.0 & 26.0 & 17.2
 & \cellcolor{gray!15} 35.63  \\
DocOwl1.5 &      14.0 &    24.07 & 20.0 & 32.0 & 18.0 &      44.0 & \textcolor{blue}{\underline{50.0}} &    42.55 &     35.29 &      12.0 & 24.0 &  5.0 &  10.0 &   3.45 & \cellcolor{gray!15} 24.37\\
InstructBLIP & 6.0 & 3.7 & 20.0 & 14.0 & 10.0 & 0.0 & 25.0 & 2.1 & 17.6 & 8.0 & 8.0 & 0.0 & 6.0 & 10.3 & \cellcolor{gray!15} 8.87  \\
InternVL3 &      24.0 &    36.11 & 30.0 & \textcolor{blue}{\underline{44.0}} & 38.0 &      66.0 &  \textcolor{blue}{\underline{50.0}} &    53.19 &     49.02 &      16.0 & 24.0 & 30.0 &  \textcolor{blue}{\underline{32.0}} &   3.45 & \cellcolor{gray!15} 36.62\\
LLama3.2 & 40.0 & 37.0 & 30.0 & 30.0 & 26.0 & 58.0 & 25.0 & 70.2 & 69.6 & 16.0 & 26.0 & 25.0 & 28.0 & 20.7 & \cellcolor{gray!15} 39.86  \\
Llava1.6 & 16.0 & 19.4 & 24.0 & 26.0 & 12.0 & 30.0 & 50.0 & 14.9 & 25.5 & 4.0 & 18.0 & 10.0 & 10.0 & 3.4 & \cellcolor{gray!15} 18.17  \\
Llava1.5 & 12.0 & 11.1 & 18.0 & 36.0 & 16.0 & 6.0 & 0.0 & 8.5 & 20.6 & 8.0 & 20.0 & 5.0 & 10.0 & 6.9 & \cellcolor{gray!15} 14.51  \\
LlaVA-OneVision & 8.0 & 12.0 & 12.0 & 16.0 & 10.0 & 36.0 & 0.0 & 6.4 & 20.6 & 6.0 & 8.0 & 10.0 & 8.0 & 0.0 & \cellcolor{gray!15} 12.82  \\
mPLUG-Owl3 & 14.0 & 30.6 & 24.0 & 24.0 & 12.0 & 18.0 & 25.0 & 19.1 & 22.5 & 4.0 & 16.0 & 5.0 & 8.0 & 10.3 & \cellcolor{gray!15} 18.31 \\
Phi3-vision & 38.0 & 41.7 & 38.0 & 54.0 & 40.0 & 58.0 & 50.0 & 46.8 & 52.0 & 22.0 & 40.0 & 35.0 & 36.0 & 13.8 & \cellcolor{gray!15} 41.69   \\
Pixtral & 34.0 & 45.4 & 22.0 & 54.0 & 42.0 & 62.0 & 50.0 & 44.7 & 43.1 & 14.0 & 32.0 & 20.0 & 24.0 & 31.0 & \cellcolor{gray!15} 38.17  \\
Qwen2VL &      \textcolor{blue}{\underline{28.0}} &  \textcolor{red}{\textbf{53.70}} & \textcolor{red}{\textbf{38.0}} & 42.0 & 42.0 &      60.0 &  \textcolor{blue}{\underline{50.0}} &  \textcolor{red}{\textbf{65.96}} &     61.76 &      18.0 & 26.0 & 15.0 &  \textcolor{red}{\textbf{34.0}} &  13.79  & \cellcolor{gray!15} \textcolor{blue}{\underline{42.96}}\\
QwenVLChat &      24.0 &    17.59 & 18.0 & 20.0 & 20.0 &      28.0 &  \textcolor{blue}{\underline{50.0}} &    21.28 &     28.43 &       6.0 & \textcolor{red}{\textbf{36.0}} & 10.0 &6.0 &  13.79 & \cellcolor{gray!15} 20.42\\
SmolVLM &      26.0 &    23.15 & 20.0 & 28.0 & 14.0 &      50.0 &    0.0 &    17.02 &     31.37 &       8.0 & 20.0 &  5.0 &  16.0 &   0.0 & \cellcolor{gray!15} 22.11 \\
SPHINX-V & 18.0 & 20.4 & 20.0 & 20.0 & 16.0 & 30.0 & 0.0 & 21.3 & 28.4 & 10.0 & 30.0 & 5.0 & 18.0 & 13.8 & \cellcolor{gray!15} 20.70  \\
VisualGLM &      16.0 &     8.33 & 24.0 & 10.0 & 22.0 &       8.0 &   \textcolor{red}{\textbf{75.0}} &     8.51 &     18.63 &       8.0 & 16.0 & 10.0 &4.0 &   6.90 & \cellcolor{gray!15} 13.10\\
\midrule
\multicolumn{16}{l}{\cellcolor{orange!15}\textit{Chart-related Models}}  \\
\midrule
ChartGemma &      \textcolor{red}{\textbf{32.0}} &    36.11 & 26.0 & 30.0 & 28.0 &      42.0 &   25.0 &    31.91 &     42.16 &       8.0 & 22.0 & 10.0 &  18.0 &   6.90  & \cellcolor{gray!15} 28.87\\
ChartInstruct &       8.0 &    16.67 & 12.0 & 26.0 &  6.0 &      56.0 &    0.0 &    21.28 &     28.43 &       4.0 &  8.0 &  5.0 &  10.0 &  10.34 & \cellcolor{gray!15} 17.75 \\
ChartLlama &      12.0 &    18.52 & \textcolor{red}{\textbf{38.0}} & 28.0 & 16.0 &      44.0 &   25.0 &     8.51 &     24.51 &      10.0 & 28.0 & 15.0 &  16.0 &  13.79  & \cellcolor{gray!15} 21.55\\
ChartVLM &      12.0 &    26.85 & 28.0 & 34.0 & 26.0 &      42.0 &   \textcolor{blue}{\underline{50.0}} &    42.55 &     44.12 &      16.0 & 24.0 & 30.0 &  18.0 &  13.79 & \cellcolor{gray!15} 29.01 \\
DePlot &      16.0 &   \textcolor{blue}{\underline{52.78}} & 14.0 & 22.0 & 32.0 &      32.0 &   25.0  & \textcolor{blue}{\underline{63.83}} &   \textcolor{blue}{\underline{70.59}} &      16.0 & 22.0 &  5.0 &6.0 &  13.79 & \cellcolor{gray!15} 34.51\\
MatCha & 12.0 & 18.5 & 18.0 & 12.0 & 16.0 & 32.0 & 50.0 & 8.5 & 29.4 & 6.0 & 14.0 & 10.0 & 10.0 & 10.3 & \cellcolor{gray!15} 17.04  \\
OneChart &       9.3 &    69.52 &  5.26 & 20.41 & 10.87 &      39.58 &    0.0 &    63.04 &     \textcolor{red}{\textbf{77.0}} &      \textcolor{red}{\textbf{24.0}} &  9.3 & 30.0 &  11.11 &   3.57 & \cellcolor{gray!15} 37.14\\
TinyChart &      22.0 &    47.22 & 28.0 & 28.0 & 24.0 &      62.0 &   25.0 &    51.06 &     46.08 &      16.0 & 24.0 & 10.0 &  16.0 &   6.90 & \cellcolor{gray!15} 33.38\\
UniChart &      16.0 &    23.15 & 14.0 & 12.0 &  4.0 &      26.0 &   \textcolor{red}{\textbf{75.0}} &    42.55 &     29.41 &      12.0 & 12.0 & 10.0 &8.0 &   6.90 & \cellcolor{gray!15} 18.87 \\
\midrule
\multicolumn{16}{l}{\cellcolor{orange!15}\textit{Multimodal Agentic Framework (Ours)}} \\
\midrule
\textbf{\chartagent{}} &  \textcolor{red}{\textbf{32.0}} &  50.0 &  30.0 &  33.33 &  \textcolor{red}{\textbf{70.0}}  &  50.0 & \textcolor{blue}{\underline{50.0}} & 36.17 &  64.71 & 16.0 &  \textcolor{blue}{\underline{30.0}} &  \textcolor{red}{\textbf{50.0}} &  28.0 &  \textcolor{red}{\textbf{65.52}} & \cellcolor{gray!15} \textcolor{red}{\textbf{44.16}}\\
\bottomrule
\end{tabular}%
\end{adjustbox}
\vspace{0.2cm}
\caption{\textbf{ChartX Dataset} (18 chart types in total; 14 unannotated)}
\label{chartx_unannotated_results}
\end{subtable}
\vspace{-0.6cm}
\label{tab:unannotated-results-combined}
\end{table*}


\clearpage
\twocolumn
\subsection{Analysis of Tool Usage in \chartagent{}} \label{sec:tool_selection}
To gain deeper insight into the internal decision-making process of \chartagent{}, we examine how it selects visual tools across different chart types. Table~\ref{tab:tool_usage_by_chart_type} summarizes the most frequently used tools for each chart type, reflecting tool-usage patterns observed in agent trajectories (see Appendix Table~\ref{appendix_tab:vision_tool_description} for detailed descriptions of each tool’s functionality).
\begin{table*}[htb]
\centering
\caption{\textbf{Most frequently used tools across chart types.} Tool-usage patterns observed in agent trajectories (see Appendix Table~\ref{appendix_tab:vision_tool_description} for tool descriptions).}
\begin{adjustbox}{max width=\linewidth}
\begin{tabular}{p{6cm}|l}
\toprule
\textbf{Chart Type (Chart Subtypes)} & \textbf{Chart Tools Used} \\
\midrule
\multirow{2}{6cm}{Pie (Ring, Sector, Multi-Ring), Treemap} & \texttt{annotate\_legend} $\vert$ \texttt{get\_marker\_rgb} $\vert$ \texttt{clean\_chart\_image} $\vert$ \texttt{segment\_and\_mark} $\vert$ \\ & \texttt{compute\_segment\_area} $\vert$ \texttt{arithmetic} \\
\midrule
\multirow{2}{6cm}{Bar (Horizontal/Vertical Single/Multi/Stacked, Histogram, 3D)} & \texttt{annotate\_legend} $\vert$ \texttt{get\_marker\_rgb} $\vert$ \texttt{clean\_chart\_image} $\vert$ \texttt{segment\_and\_mark}  $\vert$ \texttt{get\_bar} $\vert$ \\ &  \texttt{compute\_bar\_height} $\vert$ \texttt{axis\_localizer} $\vert$ \texttt{interpolate\_pixel\_to\_value} \\
\midrule
Box (Horizontal/Vertical) & \texttt{clean\_chart\_image} $\vert$ \texttt{segment\_and\_mark} $\vert$ \texttt{get\_boxplot} $\vert$ \texttt{compute\_boxplot\_entity} $\vert$ \\ & \texttt{axis\_localizer} $\vert$ \texttt{interpolate\_pixel\_to\_value} \\
\midrule
\multirow{2}{6cm}{Area (Overlay, Stacked)} & \texttt{annotate\_legend} $\vert$ \texttt{get\_marker\_rgb} $\vert$ \texttt{clean\_chart\_image} $\vert$ \texttt{segment\_and\_mark} $\vert$ \\ &  \texttt{get\_edgepoints} $\vert$  \texttt{axis\_localizer} $\vert$ \texttt{interpolate\_pixel\_to\_value} $\vert$ \texttt{arithmetic} \\
\midrule
\multirow{2}{6cm}{Line (Single/Multi)} & \texttt{annotate\_legend} $\vert$ \texttt{get\_marker\_rgb} $\vert$ \texttt{clean\_chart\_image} $\vert$ \texttt{get\_edgepoints} $\vert$ \\ &  \texttt{axis\_localizer} $\vert$  \texttt{interpolate\_pixel\_to\_value} \\
\midrule
\multirow{2}{6cm}{Scatter (Bubble, 3D)} & \texttt{annotate\_legend} $\vert$ \texttt{get\_marker\_rgb} $\vert$ \texttt{clean\_chart\_image} $\vert$ \texttt{segment\_and\_mark} $\vert$ \\ & \texttt{get\_edgepoints} $\vert$ \texttt{axis\_localizer} $\vert$ \texttt{interpolate\_pixel\_to\_value} \\
\midrule
\multirow{2}{6cm}{Radial Bar, Rose} & \texttt{annotate\_legend} $\vert$ \texttt{get\_marker\_rgb} $\vert$ \texttt{clean\_chart\_image} $\vert$ \texttt{segment\_and\_mark} $\vert$ \\ & \texttt{get\_radial} $\vert$ \texttt{analyse\_radial\_geometry} $\vert$ \texttt{estimate\_radial\_value} \\
\midrule
\multirow{2}{6cm}{Combination (Bar-Line, Line-Line), Multi-Axes} & \texttt{annotate\_legend} $\vert$ \texttt{get\_marker\_rgb} $\vert$ \texttt{clean\_chart\_image} $\vert$ \texttt{segment\_and\_mark}  $\vert$ \texttt{get\_bar} $\vert$ \\ & \texttt{compute\_bar\_height} $\vert$ \texttt{get\_edgepoints} $\vert$ \texttt{axis\_localizer} $\vert$ \texttt{interpolate\_pixel\_to\_value} \\
\bottomrule
\end{tabular}
\end{adjustbox}
\label{tab:tool_usage_by_chart_type}
\end{table*}
This analysis demonstrates that \chartagent{} strategically adapts its tool usage to the structural and semantic properties of different chart types. 

\begin{figure*}[htbp]
  \centering
  \includegraphics[width = \textwidth]{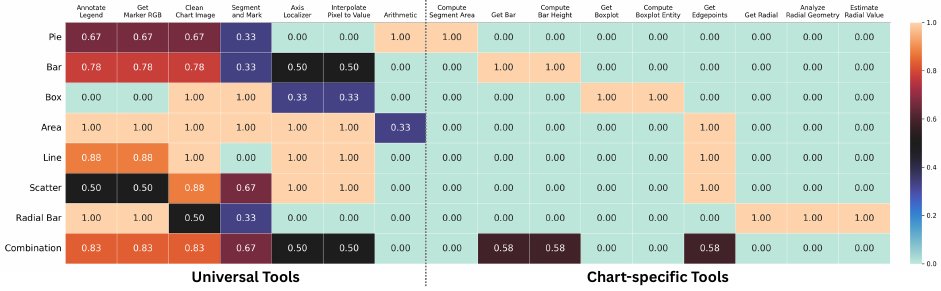}
    \caption{\textbf{Tool-use statistics across benchmark datasets.} Percentage of times \chartagent{} employs a given tool when solving queries for each chart type. As expected, universal tools are used broadly across all chart types, whereas chart-specific tools are invoked selectively depending on the chart type detected by the \chartagent{} orchestrator.}
  \label{fig:tool_use_statistics}
\end{figure*}

Further, Figure~\ref{fig:tool_use_statistics} illustrates the percentage of times \chartagent{} employs each tool across chart types. Overall, \textbf{tool usage is strongly chart-type dependent}. Universal tools (e.g., \texttt{annotate\_legend}, \texttt{get\_marker\_rgb}, \texttt{clean\_chart\_image}) are employed consistently across nearly all chart types, whereas chart-specific tools (e.g., \texttt{get\_boxplot} for boxplots or \texttt{analyze\_radial\_geometry} for radial bars) are invoked only when structurally required. Combination charts exhibit the highest diversity of tool usage, reflecting the need to simultaneously process multiple chart modalities (e.g., bar and line elements).  

Interestingly, several tools show nearly identical usage percentages, suggesting they are frequently used together in agent trajectories. For example, \texttt{annotate\_legend} and \texttt{get\_marker\_rgb} exhibit very similar distributions across chart types: once the legend is localized, the agent almost always proceeds to extract the corresponding marker color. Such patterns indicate that \textbf{certain tools are implicitly coupled in the decision-making process}, with \chartagent{} invoking them in conjunction to complete semantically linked subtasks.

\subsection{Ablation Study} \label{sec:appendix-ablation}
Prior agentic frameworks in natural image VQA rely heavily on generic tools like cropping and zooming. While effective for object localization or text spotting in natural images, these tools lack the capabilities required for structured, quantitative reasoning over charts. Chart-based QA tasks often demand operations such as axis parsing, color-based segmentation, pixel-to-value interpolation, and arithmetic reasoning, which cannot be supported by coarse manipulations like cropping or zooming. This motivates the design of chart-specialized tools tightly integrated into the reasoning loop.

Generic tools such as crop/zoom are insufficient because:
\begin{itemize}
    \item They cannot extract or match RGB values to identify legend categories.
    \item They cannot segment visual elements (e.g., pie slices, bars) based on color or structure.
    \item They cannot compute pixel areas or interpolate numerical values from axes.
\end{itemize}
As a result, agents using only natural image tools often produce reasoning traces filled with irrelevant observations, ultimately lowering accuracy. In contrast, chart-specialized tools (e.g., axis parsing, bar/pie segmentation, legend detection, numeric estimation) allow precise grounding of reasoning steps and enable recovery via visual self-verification.

To understand the contribution of chart-specialized visual tools in our framework, we conduct an ablation study comparing three variants of the ReAct agent, all implemented with GPT-4o as the underlying reasoning model and equipped with visual self-verification: (i) \textbf{ReAct (No Tools)}: reasoning without any visual tools; (ii) \textbf{ReAct + Natural Image Tools}: reasoning augmented with generic natural-image tools such as crop and zoom; and (iii) \textbf{\chartagent{} (Ours)}: reasoning supported by chart-specialized tools designed for fine-grained chart understanding.

Table~\ref{tab:ablation} presents the comparison across the three variants. 
Note that the same ReAct iteration limit (15 maximum steps) is used across all settings in the ablation study. 
We report both overall average accuracy and performance on the more challenging subset of unannotated numeric chart questions.
\begin{table}[h!]
\centering
\caption{\textbf{Ablation study} on the role of tools in chart VQA. Chart-specialized tools enable strong gains, especially for unannotated charts \& numeric QA. \textcolor{red}{\textbf{Red:}} Best.}
\label{tab:ablation}
\begin{adjustbox}{max width=\linewidth}
\begin{tabular}{@{}m{2.5cm}|M{2cm}|M{2.3cm}|M{2.3cm}@{}}
\toprule
\multirow{3}{*}{\textbf{Method}} & \multirow{3}{*}{\textbf{Tool Type}} & \textbf{Overall Acc. (\%) $\uparrow$} & \textbf{Unannotated \& Numeric Acc. (\%) $\uparrow$} \\
\midrule
ReAct & \multirow{2}{*}{None} & \multirow{2}{*}{38.84} & \multirow{2}{*}{19.46} \\
+ No Tools & & & \\
\midrule
ReAct & \multirow{3}{*}{Generic} & \multirow{3}{*}{41.35} & \multirow{3}{*}{20.50} \\
+ Natural Image Tools & & & \\
\midrule
\cellcolor{orange!20} \chartagent{} (Ours) & \cellcolor{orange!20} Chart-specialized & \cellcolor{orange!20} \multirow{2}{*}{\textcolor{red}{\textbf{71.39}}} & \cellcolor{orange!20} \multirow{2}{*}{\textcolor{red}{\textbf{58.29}}} \\
\bottomrule
\end{tabular}
\end{adjustbox}
\end{table}

The results highlight several key observations:
\begin{itemize}
    \item \textbf{ReAct without tools underperforms even GPT-4o + CoT.} While ReAct provides reasoning structure, without visual grounding it accumulates errors, producing misleading traces.
    \item \textbf{Generic tools provide marginal gains.} Crop/zoom adds limited context but cannot handle structured quantitative reasoning, resulting in only minor improvements over no tools.
    \item \textbf{Chart-specialized tools are critical.} The large performance jump with \chartagent{} demonstrates the necessity of type-specific visual grounding and self-verification mechanisms for robust chart QA.
\end{itemize}

This ablation study confirms that generic natural-image tools are fundamentally inadequate for chart reasoning. By equipping the agent with a comprehensive taxonomy of chart-specialized tools, integrated into an iterative ReAct loop with visual self-verification, \chartagent{} achieves state-of-the-art performance—particularly excelling on unannotated charts and numeric QA where prior methods fail.

\subsection{Visual and Reasoning Complexity Analysis}\label{appendix:complexity_results}
Table \ref{tab:complexity_analysis_results} presents the accuracy on unannotated charts by visual complexity of the charts and reasoning complexity of the chart–QA pairs. 
 \begin{table*}[h!]
\caption{\textbf{Accuracy by Complexity Levels.} Accuracy (\%) on unannotated charts stratified by visual complexity of the charts and reasoning complexity of the chart–QA pairs. \textcolor{red}{\textbf{Red:}} Best,
\textcolor{blue}{\underline{Blue:}} Second best. }
\begin{subtable}[t]{0.5\textwidth}
\centering
\begin{adjustbox}{max width=\linewidth}
\begin{tabular}{l|ccc|ccc|c}
\toprule
\multirow{2}{*}{\textbf{Model}} & \multicolumn{3}{c|}{\textbf{Visual Complexity}} & \multicolumn{3}{c|}{\textbf{Reasoning Complexity}} & \cellcolor{gray!15} \textbf{Overall}\\ 
\cmidrule(lr{1em}){2-4}
\cmidrule(lr{1em}){5-7}
\cmidrule(lr{1em}){8-8}
& \cellcolor{green!15} Easy & \cellcolor{yellow!15} Medium & \cellcolor{red!15} Hard & \cellcolor{green!15} Easy & \cellcolor{yellow!15}  Medium & \cellcolor{red!15} 
 Hard &  \cellcolor{gray!15} Average  $\mathbf{\uparrow}$ \\
\midrule
\multicolumn{8}{l}{\cellcolor{orange!15}\textit{Proprietary Multimodal Large Language Models}}  \\
\midrule
GPT 4o & 57.16 & 28.25 & 17.59 & 44.06 & \textcolor{blue}{\underline{20.84}} & 13.72  & \cellcolor{gray!15} 36.15  \\
GPT 4o-mini & 39.93 & 20.22 & 9.45 & 32.06 & 9.94 & 9.39  & \cellcolor{gray!15} 25.19  \\
Claude 3 Haiku & 40.53 & 21.17 & 10.42 & 33.17 & 10.33 & 9.39  & \cellcolor{gray!15} 26.04  \\
Gemini 1.5 & 46.36 & 20.83 & 6.19 & 36.43 & 9.35 & 1.08 & \cellcolor{gray!15} 27.27\\
\midrule
\multicolumn{8}{l}{\cellcolor{orange!15}\textit{Open-weights Multimodal Large Language Models}}  \\
\midrule
BLIP-2 & 3.16 & 2.45 & 4.56 & 3.06 & 3.44 & 1.08  & \cellcolor{gray!15} 2.92 \\
CogAgent & 13.23 & 11.78 & 6.51 & 13.44 & 8.41 & 5.78 & \cellcolor{gray!15} 11.62 \\
CogVLM & 15.17 & 9.94 & 11.07 & 12.94 & 9.18 & 8.66 & \cellcolor{gray!15} 11.73 \\
DeepSeek-VL2 & 43.08 & 25.39 & 19.54 & 37.00 & 16.63 & 12.64 & \cellcolor{gray!15} 30.31 \\
DocOwl1.5-Chat & 43.08 & 15.45 & 10.10 & 29.72 & 10.33 & 8.66 & \cellcolor{gray!15} 23.58 \\
InstructBLIP & 9.83 & 4.02 & 4.56 & 6.67 & 3.82 & 5.05 & \cellcolor{gray!15} 5.92 \\
InternVL3 & 49.27 & 22.67 & \underline{\textcolor{blue}{21.17}} & 37.89 & 16.83 & 12.27 & \cellcolor{gray!15} 30.92 \\
LLama3.2 & 58.01 & 28.86 & 14.33 & 45.28 & 14.15 & 20.58 & \cellcolor{gray!15} 36.38 \\
Llava1.6 & 15.66 & 7.69 & 5.21 & 12.78 & 2.68 & 5.05 & \cellcolor{gray!15} 9.92 \\
Llava1.5 & 8.50 & 6.19 & 6.84 & 7.83 & 6.31 & 2.89 & \cellcolor{gray!15} 7.00 \\
LlaVA-OneVision & 11.17 & 9.39 & 14.01 & 11.39 & 10.71 & 4.33 & \cellcolor{gray!15} 10.50 \\
mPLUG-Owl3 & 18.81 & 9.67 & 10.42 & 14.89 & 8.99 & 5.05 & \cellcolor{gray!15} 12.65 \\
Phi3-vision & 55.83 & 36.08 & \textcolor{red}{\textbf{22.48}} & \textcolor{blue}{\underline{50.11}} & 19.69 & 19.49 & \cellcolor{gray!15} 40.73 \\
Pixtral & 45.39 & 22.94 & 11.73 & 35.39 & 14.53 & 12.27 & \cellcolor{gray!15} 28.73 \\
Qwen2VL & \textcolor{blue}{\underline{66.02}} & \textcolor{blue}{\underline{36.69}} & 15.64 & 54.44 & 17.40 & \textcolor{blue}{\underline{21.66}} & \cellcolor{gray!15} \textcolor{blue}{\underline{43.50}} \\
QwenVLChat & 8.98 & 5.65 & 4.23 & 7.61 & 3.25 & 5.78 & \cellcolor{gray!15} 6.54 \\
SmolVLM & 23.06 & 10.42 & 10.75 & 17.83 & 9.37 & 2.17 & \cellcolor{gray!15} 14.46 \\
SPHINX-V & 20.26 & 8.44 &   9.44 & 15.22 & 7.07 & 3.24 & \cellcolor{gray!15} 12.30 \\
VisualGLM & 12.74 & 5.51 & 4.23 & 9.56 & 3.44 & 3.25 & \cellcolor{gray!15} 7.65 \\
\midrule
\multicolumn{8}{l}{\cellcolor{orange!15}\textit{Chart-related Models}}  \\
\midrule
ChartGemma & 39.68 & 15.66 & 8.47 & 28.72 & 7.46 & 9.75 & \cellcolor{gray!15} 22.42 \\
ChartInstruct & 38.96 & 12.05 & 8.79 & 25.67 & 7.27 & 9.03 & \cellcolor{gray!15} 20.19 \\
ChartLlama & 17.84 & 9.26 & 4.56 & 13.61 & 5.93 & 7.58 & \cellcolor{gray!15} 11.42 \\
ChartVLM & 44.90 & 15.11 & 9.77 & 31.56 & 6.31 & 7.58 & \cellcolor{gray!15} 23.92 \\
DePlot & 50.36 & 19.54 & 9.77 & 37.78 & 5.16 & 9.03 & \cellcolor{gray!15} 28.15 \\
MatCha & 17.48 & 6.81 & 2.61 & 13.22 & 1.72 & 1.81 & \cellcolor{gray!15} 9.69 \\
OneChart & 52.21 & 15.61 & 5.22 & 34.17 & 4.29 & 2.82 & \cellcolor{gray!15} 26.81 \\
TinyChart & 53.03 & 24.71 & 16.94 & 40.00 & 16.83 & 15.88 & \cellcolor{gray!15} 32.77 \\
UniChart & 30.83 & 10.28 & 3.26 & 21.06 & 3.06 & 7.22 & \cellcolor{gray!15} 15.96 \\
\midrule
\multicolumn{8}{l}{\cellcolor{orange!15}\textit{Multimodal Agentic Framework}}  \\
\midrule
 ChartAgent \textbf{(Ours)} &  \textcolor{red}{\textbf{83.98}} &  \textcolor{red}{\textbf{56.77}} &  17.92 &  \textcolor{red}{\textbf{71.33}} &  \textcolor{red}{\textbf{41.68}} &  \textcolor{red}{\textbf{28.52}} &  \textcolor{red}{\textbf{60.81}} \\
\bottomrule
\end{tabular}
\end{adjustbox}
\vspace{0.2mm}
\caption{ChartBench Dataset}
\label{tab:chartbench}
\end{subtable}
\hfill
\begin{subtable}[t]{0.5\textwidth}
\centering
\begin{adjustbox}{max width=\linewidth}
\begin{tabular}{l|ccc|ccc|c}
\toprule
\multirow{2}{*}{\textbf{Model}} & \multicolumn{3}{c|}{\textbf{Visual Complexity}} & \multicolumn{3}{c|}{\textbf{Reasoning Complexity}} & \cellcolor{gray!15} \textbf{Overall}\\ 
\cmidrule(lr{1em}){2-4}
\cmidrule(lr{1em}){5-7}
\cmidrule(lr{1em}){8-8}
& \cellcolor{green!15} Easy & \cellcolor{yellow!15} Medium & \cellcolor{red!15} Hard & \cellcolor{green!15} Easy & \cellcolor{yellow!15}  Medium & \cellcolor{red!15} 
 Hard &  \cellcolor{gray!15} Average  $\mathbf{\uparrow}$ \\
\midrule
\multicolumn{8}{l}{\cellcolor{orange!15}\textit{Proprietary Multimodal Large Language Models}}  \\
\midrule
GPT 4o & 42.11 & \textcolor{blue}{\underline{47.77}} & 22.70 & 49.86 & 31.54 & 22.22 & \cellcolor{gray!15} 39.44 \\
GPT 4o-mini & 36.22 & 39.28 & 22.09 & 43.21 & 24.48 & 24.07 & \cellcolor{gray!15} 33.94 \\
Claude 3 Haiku & 25.69 & 32.59 & 16.56 & 34.07 & 17.01 & 17.59 & \cellcolor{gray!15}  25.77 \\
Gemini 1.5 & 36.84 & 31.25 & 20.86 & 44.60 & 18.26 & 16.67 & \cellcolor{gray!15} 31.41 \\
\midrule
\multicolumn{8}{l}{\cellcolor{orange!15}\textit{Open-weights Multimodal Large Language Models}}  \\
\midrule
BLIP-2 & 0.93 & 1.78 & 3.07 & 1.38 & 2.48 & 0.93 & \cellcolor{gray!15} 1.69 \\
CogAgent & 26.06 & 25.89 & 21.47 & 30.47 & 18.67 & 20.37 & \cellcolor{gray!15} 24.93 \\
CogVLM & 26.62 & 23.21 & 20.85 & 27.98 & 19.92 & 21.30 & \cellcolor{gray!15} 24.23 \\
DeepSeek-VL2 & 42.41 & 36.61 & 20.86 & 47.37 & 24.07 & 22.22 & \cellcolor{gray!15} 35.63 \\
DocOwl1.5-Chat & 28.79 & 23.66 & 16.56 & 32.41 & 17.43 & 12.96 & \cellcolor{gray!15} 24.37 \\
InstructBLIP & 8.05 & 8.04 & 11.66 & 8.03 & 8.71 & 12.04 & \cellcolor{gray!15} 8.87 \\
InternVL3 & 40.25 & 41.52 & 22.70 & 46.26 & 28.21 & 23.15 & \cellcolor{gray!15} 36.62 \\
LLama3.2 & 49.23 & 37.95 & 23.93 & 49.31 & 31.95 & \textcolor{blue}{\underline{25.93}} & \cellcolor{gray!15} 39.86 \\
Llava1.6 & 19.20 & 18.75 & 15.33 & 21.32 & 13.27 & 18.52 & \cellcolor{gray!15} 18.17 \\
Llava1.5 & 14.55 & 14.29 & 14.72 & 16.34 & 12.45 & 12.96 & \cellcolor{gray!15} 14.51 \\
LlaVA-OneVision & 12.69 & 15.63 & 9.20 & 16.89 & 7.88 & 10.19 & \cellcolor{gray!15} 12.82 \\

mPLUG-Owl3 & 21.67 & 16.96 & 13.49 & 21.33 & 15.35 & 14.81 & \cellcolor{gray!15}  18.31 \\
Phi3-vision & 46.74 & 41.07 & 32.52 & 53.74 & 26.97 & 34.26 & \cellcolor{gray!15} 41.69 \\
Pixtral & 45.82 & 39.73 & 20.86 & 49.58 & 27.39 & 24.07 & \cellcolor{gray!15} 38.17 \\
Qwen2VL & \textcolor{blue}{\underline{51.39}} & 40.18 & \textcolor{red}{\textbf{28.83}} & \textcolor{red}{\textbf{55.13}} & 32.37 & 24.07 & \cellcolor{gray!15} \textcolor{blue}{\underline{42.96}} \\
QwenVLChat & 19.19 & 24.10 & 17.79 & 23.82 & 15.76 & 19.44 & \cellcolor{gray!15} 20.42 \\
SmolVLM & 23.22 & 25.44 & 15.33 & 25.76 & 18.25 & 18.52 & \cellcolor{gray!15}  22.11 \\
SPHINX-V & 21.67 & 20.08 &   19.63   & 25.20 & 16.59 & 14.81 & \cellcolor{gray!15} 20.70 \\
VisualGLM & 10.52 & 15.18 & 15.34 & 14.68 & 8.71 & 17.59 & \cellcolor{gray!15} 13.10 \\
\midrule
\multicolumn{8}{l}{\cellcolor{orange!15}\textit{Chart-related Models}}  \\
\midrule
ChartGemma & 31.89 & 32.58 & 17.79 & 37.67 & 18.25 & 23.14 & \cellcolor{gray!15} 28.87 \\
ChartInstruct & 22.60 & 17.85 & 7.97 & 24.37 & 11.61 & 9.25 & \cellcolor{gray!15}  17.75 \\
ChartLlama & 20.12 & 22.76 & 22.69 & 22.43 & 19.08 & 24.07 & \cellcolor{gray!15}  21.55 \\
ChartVLM & 33.13 & 28.57 & 21.47 & 35.45 & 23.65 & 19.44 & \cellcolor{gray!15} 29.01 \\
DePlot & 49.22 & 26.78 & 15.95 & 45.70 & 28.21 & 11.11 & \cellcolor{gray!15} 34.51 \\
MatCha & 17.95 & 19.64 & 11.65 & 20.77 & 13.69 & 12.03 & \cellcolor{gray!15} 17.04 \\
OneChart & \textcolor{red}{\textbf{55.73}} & 25.35 & 13.38 & 45.55 & \textcolor{blue}{\underline{36.77}} & 6.45 & \cellcolor{gray!15} 37.14 \\
TinyChart & 39.93 & 32.14 & 22.08 & 44.04 & 22.82 & 21.29 & \cellcolor{gray!15} 33.38 \\
UniChart & 25.69 & 13.83 & 12.26 & 26.31 & 11.61 & 10.18 & \cellcolor{gray!15} 18.87 \\
\midrule
\multicolumn{8}{l}{\cellcolor{orange!15}\textit{Multimodal Agentic Framework}}  \\
\midrule
 ChartAgent \textbf{(Ours)} &  50.93 &  \textcolor{red}{\textbf{49.91}} &  \textcolor{blue}{\underline{24.54}} &  \textcolor{blue}{\underline{54.14}} &  \textcolor{red}{\textbf{38.17}} &  \textcolor{red}{\textbf{27.78}} &  \textcolor{red}{\textbf{44.16}}\\
\bottomrule
\end{tabular}
\end{adjustbox}
\vspace{0.2mm}
\caption{ChartX Dataset}
\label{tab:chartx}
\end{subtable}
\label{tab:complexity_analysis_results}
\end{table*}

\subsection{Accuracy vs. LLM-as-a-Judge}\label{appendix:acc_vs_llm_as_judge}
We found that  LLM-as-a-Judge often \textit{relaxes} the 5\% margin condition, leading to inflated performance compared to arithmetic accuracy, which strictly enforces this threshold. This observation is important to share with the community, as most recent Chart VQA papers \cite{xu2023chartbench, xia2024chartx, masry-etal-2022-chartqa} rely directly on GPT-based accuracy for evaluation. 
Table~\ref{tab:accuracy_vs_llm_as_a_judge} reports the comparison between our standardized accuracy evaluation and the corresponding LLM-as-a-Judge results on the ChartBench dataset.
\begin{table}[htb]
\centering
\caption{\textbf{Accuracy vs. LLM-as-a-Judge.} Results on the ChartBench dataset. All values represent accuracy in percentage.}
\resizebox{0.48\textwidth}{!}{
\begin{tabular}{lccc}
\toprule
\textbf{Model} & \textbf{Accuracy} & \textbf{LLM-as-a-Judge} & \textbf{Gap (\%)}\\ 
\midrule
Gemini 2.0 flash & 69.90 & 76.45 & -6.55 \\
GPT 4o-mini & 42.24 & 48.47 & -6.24 \\
DeepSeek-VL2 & 49.39 & 55.16 & -5.76 \\
ChartLlama & 19.89 & 24.42 & -4.53 \\
ChartInstruct & 31.24 & 35.68 & -4.45 \\
GPT 4o & 51.47 & 55.63 & -4.16 \\
SPHINX-V & 19.76 & 23.79 & -4.03 \\
TinyChart & 46.84 & 50.82 & -3.97 \\
CogVLM & 28.11 & 31.68 & -3.58 \\
ChartGemma & 39.32 & 42.76 & -3.45 \\
\bottomrule
\end{tabular}}
\label{tab:accuracy_vs_llm_as_a_judge}
\end{table}

\subsection{Concurrent Works} \label{appendix:concurrent_works}
The ChartBench dataset was released on December 26, 2023, and ChartX on February 19, 2024. Table~\ref{appendix:concurrent_dates} shows the split of models with knowledge cutoff dates before versus after each dataset release. Since datasets may have leaked into the training data of models with knowledge cutoff dates after release, we report these concurrent model results separately.  
Notably, we use GPT-4o (\texttt{gpt-4o-2024-08-06}, with a knowledge cutoff of October 1, 2023) as the base multimodal LLM for reasoning in \chartagent{}. Since ChartBench and ChartX were released in December 2023 and February 2024, respectively, they were definitively not part of GPT-4o’s training data.

\begin{table}[htb]
\centering
\caption{\textbf{Knowledge Cutoffs and Concurrent Works.} Comparison of model and dataset release dates relative to ChartBench and ChartX, showing whether models were trained before or after these benchmarks.}
\label{appendix:concurrent_dates}
\resizebox{0.48\textwidth}{!}{
\begin{tabular}{@{}lcM{3cm}@{}}
\toprule
\textbf{Model / Dataset} & \textbf{Knowledge Cutoff} & \textbf{Relative to ChartBench / ChartX} \\
\midrule
Claude 3 Haiku  & Aug 1, 2023 & \cellcolor{red!20} Before both \\
Claude 3 Sonnet & Aug 1, 2023 & \cellcolor{red!20} Before both \\
GPT-4o & Oct 1, 2023 & \cellcolor{red!20} Before both \\
GPT-4o-mini & Oct 1, 2023 & \cellcolor{red!20} Before both \\
GPT-o1 & Oct 1, 2023 & \cellcolor{red!20} Before both \\
\rowcolor{gray!20}
ChartBench Dataset  & Dec 26, 2023  & ---\\
\rowcolor{gray!20}
ChartX Dataset  & Feb 19, 2024  & ---\\
Claude 3.5 Sonnet   & Apr 1, 2024  & \cellcolor{green!20} After both \\
GPT-o3 & May 31, 2024   & \cellcolor{green!20} After both \\
GPT-o4-mini & May 31, 2024   & \cellcolor{green!20} After both \\
GPT-4.1 & May 31, 2024   & \cellcolor{green!20} After both \\
GPT-5 mini & May 31, 2024   & \cellcolor{green!20} After both \\
Claude 3.5 Haiku & Jul 1, 2024  & \cellcolor{green!20} After both \\
Gemini 2.0 & Aug 1, 2024   & \cellcolor{green!20} After both \\
GPT-5  & Oct 1, 2024  & \cellcolor{green!20} After both \\
Claude 3.7 Sonnet   & Nov 1, 2024  & \cellcolor{green!20} After both \\
Mistral-Small  & Mar 17, 2025 & \cellcolor{green!20} After both \\
\bottomrule
\end{tabular}
}
\end{table}

\subsubsection{Performance of Concurrent Works on Public Benchmarks}
Table~\ref{tab:overall_results_concurrent} presents the accuracy comparison for concurrent works with knowledge cutoff dates after the dataset releases.  
\begin{table*}[htb]
\caption{\textbf{Accuracy on Concurrent Works (Public Benchmarks).} Comparison of accuracy (\%) on concurrent works with knowledge cut-off dates after the release of the datasets. All values correspond to the highest performance achieved across zero-shot and CoT prompting styles for each MLLM. Ann./Unann. denote Annotated and Unannotated charts. RL QA: Relationship QA; VC/GC QA: Value Comparison \& Global Conception QA. }
\begin{subtable}[t]{0.48\textwidth}
\centering
\resizebox{\textwidth}{!}{
\begin{tabular}{lcc|cc|c}
\toprule
\multirow{2}{*}{\textbf{Model}} & \multicolumn{2}{c|}{\textbf{Chart Types}} & \multicolumn{2}{c|}{\textbf{Question Types}} &  \cellcolor{gray!15} \textbf{Overall}\\ 
\cmidrule(lr{1em}){2-3}
\cmidrule(lr{1em}){4-5}
\cmidrule(lr{1em}){6-6}
& Ann. & Unann. & Numeric QA & RL QA & \cellcolor{gray!15} Avg.  $ \mathbf{\uparrow}$ \\
\midrule
\multicolumn{6}{l}{\cellcolor{orange!15}\textit{Proprietary Multimodal Large Language Models}}  \\
\midrule
GPT o3 & 98.18 & 76.56 & 82.55 & 98.44 & \cellcolor{gray!15} 83.39\\
GPT o4-mini & 98.50 & 71.73 & 79.14 & 99.00 & \cellcolor{gray!15} 80.18 \\
GPT 4.1 & 97.33 & 67.00 & 75.61 & 94.00 & \cellcolor{gray!15} 76.58  \\
Gemini 2.0 flash & 97.79 & 58.31 & 71.81 & 41.00 & \cellcolor{gray!15} 69.90 \\
Claude 3.7 Sonnet & 97.75 & 60.38 & 71.64 & 82.00 & \cellcolor{gray!15} 72.18 \\
Claude 3.5 Sonnet & 96.50 & 56.23 & 68.14 & 83.50 & \cellcolor{gray!15} 68.95 \\
Claude 3.5 Haiku & 90.67 & 38.58 & 53.89 & 75.50 & \cellcolor{gray!15} 55.03 \\
\midrule
\multicolumn{6}{l}{\cellcolor{orange!15}\textit{Open-weights Multimodal Large Language Models}}  \\
\midrule
Mistral & 91.75 & 43.23 & 57.08 & 90.00 & \cellcolor{gray!15} 58.55 \\
\midrule
\multicolumn{6}{l}{\cellcolor{orange!15}\textit{Multimodal Agentic Framework}}  \\
\midrule
\cellcolor{blue!15} ChartAgent \textbf{(Ours)} & \cellcolor{blue!15} \textbf{94.33} & \cellcolor{blue!15} \textbf{60.81} & \cellcolor{blue!15} \textbf{70.91} & \cellcolor{blue!15} \textbf{91.00} & \cellcolor{blue!15} \textbf{71.39}\\
\bottomrule
\end{tabular}}
\vspace{0.2mm}
\caption{ChartBench Dataset}
\label{tab:chartbench}
\end{subtable}
\hfill
\begin{subtable}[t]{0.515\textwidth}
\centering
\resizebox{\textwidth}{!}{
\begin{tabular}{lcc|cc|c}
\toprule
\multirow{2}{*}{\textbf{Model}} & \multicolumn{2}{c|}{\textbf{Chart Types}} & \multicolumn{2}{c|}{\textbf{Question Types}} & \cellcolor{gray!15} \textbf{Overall}\\ 
\cmidrule(lr{1em}){2-3}
\cmidrule(lr{1em}){4-5}
\cmidrule(lr{1em}){6-6}
& Ann. & Unann. & Numeric QA & VC/GC QA & \cellcolor{gray!15} Avg.  $ \mathbf{\uparrow}$  \\
\midrule
\multicolumn{6}{l}{\cellcolor{orange!15}\textit{Proprietary Multimodal Large Language Models}}  \\
\midrule
GPT o3 & 91.18 & 71.13 & 79.59 & 76.85 & \cellcolor{gray!15} 78.82 \\
GPT o4-mini & 91.18 & 72.68 & 80.92 & 76.85 & \cellcolor{gray!15} 79.77 \\
GPT 4.1 & 92.99 & 69.58 & 77.90 & 80.25 & \cellcolor{gray!15} 78.56\\
Gemini 2.0 flash & 89.37 & 58.31 & 68.72 & 74.07 & \cellcolor{gray!15} 70.23\\
Claude 3.7 Sonnet & 89.37 & 60.28 & 69.81 & 75.62 & \cellcolor{gray!15} 71.44\\
Claude 3.5 Sonnet & 87.78 & 57.32 & 67.39 & 73.15 & \cellcolor{gray!15} 69.01\\
Claude 3.5 Haiku & 80.32 & 40.70 & 50.97 & 68.52 & \cellcolor{gray!15} 55.90 \\
\midrule
\multicolumn{6}{l}{\cellcolor{orange!15}\textit{Open-weights Multimodal Large Language Models}}  \\
\midrule
Mistral & 84.84 & 48.59 & 59.06 & 71.30 & \cellcolor{gray!15} 62.50\\
\midrule
\multicolumn{6}{l}{\cellcolor{orange!15}\textit{Multimodal Agentic Framework}}  \\
\midrule
\cellcolor{blue!15} ChartAgent \textbf{(Ours)} & \cellcolor{blue!15} \textbf{84.84} & \cellcolor{blue!15} \textbf{44.16} & \cellcolor{blue!15} \textbf{55.93} & \cellcolor{blue!15} \textbf{69.14} & \cellcolor{blue!15} \textbf{59.69}\\
\bottomrule
\end{tabular}}
\vspace{0.2mm}
\caption{ChartX Dataset}
\label{tab:chartx}
\end{subtable}
\label{tab:overall_results_concurrent}
\end{table*}

We suspect that benchmark data (ChartBench and ChartX, released in December 2023 and February 2024, respectively) may have been included in the training data of GPT-o3 and GPT-o4-mini (knowledge cutoff: May 2024). In several cases, particularly with GPT-o3, we observed that the model produced correct answers despite incorrect reasoning steps or tool outputs. For example, even when the agent misidentified key visual elements or generated invalid intermediate outputs, the final answer was still correct. We also noted this behavior in instances where it was humanly very difficult to provide the exact answer, yet GPT-o3 and GPT-o4-mini produced outputs with decimal-level precision. Such patterns suggest possible memorization or exposure to similar instances during training.

While preliminary, these observations provide strong evidence of potential data leakage from public benchmarks into newer models. To strengthen this analysis, we curated a new held-out internal dataset that mirrors the complexity of ChartBench and ChartX, enabling a more rigorous evaluation.

\subsubsection{Performance of Concurrent Works on the Internal Dataset}
We created a new dataset with 125 chart–QA pairs that we are confident were not included in the training data of newer models, and conducted evaluations for a fairer comparison of these models against \chartagent{}. Specifically, we collected unannotated charts such as bar, line, pie, and bar–line combinations requiring numeric QA from the open web, selecting only those whose ground-truth answers are unavailable online, thereby increasing confidence that they were not included in the training data of newer models.

Table~\ref{tab:appendix-concurrent-internal-dataset} reports the overall accuracy (within a 5\% margin) and average numeric error on this curated dataset.  
\begin{table}[ht]
\centering
\caption{\textbf{Accuracy on Concurrent Works (Internal Benchmarks).} Overall average accuracy (within 5\% margin) and average error across models on the curated internal dataset. \textcolor{red}{\textbf{Red:}} Best, \textcolor{blue}{\underline{Blue:}} Second best.}
\label{tab:appendix-concurrent-internal-dataset}
\resizebox{0.48\textwidth}{!}{
\begin{tabular}{lcc}
\toprule
\textbf{Model} & \textbf{Accuracy (\%) $\uparrow$} & \textbf{Avg. Error (\%) $\downarrow$} \\
\midrule
ChartAgent & \textcolor{red}{\textbf{85.19}} & \textcolor{red}{\textbf{3.42}} \\
GPT 5 & \textcolor{blue}{\underline{74.71}} & 24.09 \\
GPT 5-mini & 73.18 & 11.24 \\
Claude 3.7 Sonnet & 69.71 & 15.52 \\
GPT o4-mini & 69.68 & 21.88 \\
Gemini 2.0 & 67.24 & 21.07 \\
GPT-4.1 & 66.61 & 24.32 \\
GPT-o3 & 62.93 & \textcolor{blue}{\underline{9.14}} \\
Claude 3.5 Haiku & 42.11 & 37.31 \\
Mistral & 38.54 & 38.74 \\
o1  & 33.07 & 44.31 \\
GPT-4o  & 22.02 & 64.34 \\
\bottomrule
\end{tabular}
}
\end{table}
Clearly, \chartagent{} outperforms all newer models by a significant margin in both accuracy and average error, achieving a +10.48\% absolute accuracy gain over the second-best model (GPT-5) and a 5.72-point reduction in average absolute error relative to GPT-o3. 
Notably, the baselines include both recent closed-source models (e.g., GPT-5) and agentic variants (e.g., o3 and o4-mini). These results further reinforce \chartagent{}'s effectiveness as a chart-focused visually-grounded reasoning framework.


\subsection{Visual Self-Verification and Recovery Behavior}\label{appendix:visual_self_verification_results} 
In addition to analyzing difficulty-based trends, we studied whether \chartagent{} could detect unsatisfactory tool outputs and recover using its visual self-verification mechanism. We manually evaluated 30 randomly selected agent trajectories from the ChartBench dataset to assess this behavior. The results are summarized in Table~\ref{tab:self_verification_stats}.
\begin{table}[htb]
\centering
\caption{Visual self-verification and recovery outcomes in \chartagent{} trajectories.}
\resizebox{0.46\textwidth}{!}{
\begin{tabular}{p{6.5cm}c}
\toprule
\textbf{Metric} & \textbf{Value} \\
\midrule
Cases where recovery was needed (i.e., tool output deemed unsatisfactory) & 50\% \\
\midrule
Successful recoveries among needed cases & \textbf{70\%} \\
\midrule
Correct final answers following recovery & \textbf{70\%} \\
\midrule
Cases where tool error propagated to final answer (i.e., remained incorrect) & 15\% \\
\bottomrule
\end{tabular}
}
\label{tab:self_verification_stats}
\end{table}
In 50\% of the sampled cases, the tool outputs were correct, and no recovery was needed. In the remaining 50\%, the agent correctly identified the tool outputs as unsatisfactory and triggered its self-verification mechanism. Among these, 70\% resulted in successful recovery, leading to correct final answers. The remaining 30\% failed to recover, contributing to a 15\% overall error rate attributable to unresolved tool-level failures.
These findings demonstrate that \chartagent{}'s visual self-verification mechanism is both frequently invoked and often effective, enhancing robustness in the presence of imperfect tool outputs—especially critical for unannotated chart understanding.

\subsection{Fallback Analysis: When \chartagent{} Reverts to the Base Model and Common Trigger Conditions}\label{appendix:fallback_analysis} 
We conducted a manual analysis of 30 randomly selected agent trajectories from ChartBench, focusing on unannotated charts and numeric QA, to better understand when and why the agent reverts to the base model (GPT-4o). We found that the fallback rate was relatively low—less than 10\% across the sample. The most common reasons for fallback included the following:

\noindent \textbullet\hspace{1pt} Bar charts: When the computed bar height was negative or highly inconsistent with the axis values, indicating a failure in visual estimation, the agent abandoned tool-based reasoning and allowed GPT-4o to attempt a direct response.

\noindent \textbullet\hspace{1pt} OCR-based tools returning \texttt{None}: For example, if legend or axis label detection failed to locate any relevant entities, the agent deemed the output unsatisfactory and reverted to GPT-4o.

\noindent \textbullet\hspace{1pt} Line charts: When edge-point detection or interpolation tools produced empty outputs or values that were highly inconsistent with the axis, the agent once again defaulted to GPT-4o.

In all such cases, the agent judged tool-based reasoning to be unreliable and defaulted to the base model. While rare, this fallback mechanism serves as a valuable fail-safe.

\subsection{Runtime and Inference Efficiency Analysis}\label{appendix:inference_time_analysis}
We conducted a preliminary timing analysis on a representative subset of chart types to evaluate the inference efficiency of \chartagent{} in comparison to baseline models.
In practice, \chartagent{} required an average of 5--7 ReAct iterations per sample. On average:
\begin{itemize}
    \item A single GPT-4o call with chain-of-thought reasoning required approximately 6--10 seconds per query.
    \item A full \chartagent{} trajectory, including multi-step tool usage and self-verification, required roughly \(90\,\mathrm{s}\) per query in the non-parallelized setting, and about \(30\,\mathrm{s}\) when parallelizable steps were executed concurrently. For reference, OpenAI’s agentic model \texttt{o3} required \(25\text{--}40\,\mathrm{s}\) on the same tasks, even when predictions were inaccurate.
\end{itemize}

This increase in inference time is expected due to the agentic design, which involves iterative reasoning, multiple visual-perception tool calls, and self-verification steps. We note that runtime can be substantially reduced in practice by optimizing tool efficiency—several intermediate outputs currently computed for visualization and debugging can be streamlined or skipped entirely in deployment scenarios. Despite the additional overhead, we believe the significant accuracy gains, particularly on unannotated charts for numeric QA, justify the increased computational cost in applications where precision is critical.

Beyond parallelization, we identify two additional directions for reducing latency:
\begin{itemize}
    \item \textbf{Smart routing.} As shown in Section~\ref{results:main_performance} (Performance by Chart Type) and Table~\ref{tab:unannotated-results-mainpaper}, the benefits of agentic reasoning vary notably across chart subtypes, visual and reasoning complexity levels (Section~\ref{results: effectiveness_chartagent}, Figure~\ref{fig:additional_results_figure}), and question types. A lightweight classifier could exploit these patterns to determine when full \chartagent{} reasoning is necessary versus when a faster baseline model would suffice.
    \item \textbf{Caching.} Intermediate visual artifacts, such as axis maps, segmentation masks, and legend annotations, are often reusable across related queries for the same chart. Incorporating caching would avoid redundant tool calls and substantially reduce latency in multi-query or conversational settings.
\end{itemize} 

\subsection{Monetary Cost Analysis}\label{appendix_sec:monetary_costs}
Our approach incurs monetary costs due to the use of OpenAI's GPT-4o~\cite{hurst2024gpt} as the base reasoning model. We spent approximately~\$2000 to run \chartagent{} on both datasets, covering 4952 chart image and QA pairs across diverse chart types—resulting in an average cost of approximately~\$0.40 per sample. This cost can be substantially reduced by using smaller models such as GPT-4o-mini, or eliminated entirely with open-source models like Pixtral, Llama, or Qwen, since our framework is designed to be plug-and-play. For example, switching from GPT-4o to GPT-4o-mini would reduce the average cost per sample by more than $15\times$ (to roughly \$0.025), making large-scale evaluation far more economical. Thus, monetary cost should not be considered a serious limitation, as our approach can seamlessly adapt to free or low-cost models as well.

\section{Details on Failure Mode Analysis}\label{appendix:failure_modes}
\chartagent{} encounters two main categories of failure: visual perception challenges and reasoning ambiguities.

\begin{itemize}
\item [1)] \textbf{Perception-based failures.}  

\noindent (1.1)~\textit{OCR obstruction by visual overlays:} Black overlays or dense chart elements often cover axis or legend text, preventing accurate OCR extraction.  

\noindent (1.2)~\textit{Poor color contrast:} Labels in white placed over fluorescent yellow or similarly bright backgrounds are difficult for vision tools to detect.  

\noindent (1.3)~\textit{Legend occlusion:} In some charts, the legend overlaps with key visual elements—such as bars of interest—hindering accurate region detection.  

\noindent (1.4)~\textit{Chart element invisibility:} Median lines in box plots that share the same color as the box become indistinguishable, making it hard to extract correct values.  

\noindent (1.5)~\textit{Segmentation failure due to axis overlap:} Axis lines overlapping with chart elements confuse the segmentation tool and result in incorrect extraction.  

\noindent (1.6)~\textit{Overlap-induced indistinguishability:} When multiple data series substantially overlap in charts (e.g., radar plots, line charts, scatterplots with dense clusters, or filled regions), subtle differences between categories become imperceptible. This occurs due to coincident paths, stacked fills, or saturation effects, preventing reliable detection of fine-grained deviations. 

\noindent (1.7)~\textit{Axis interpretation failures:} When unusual or complex axes (e.g., 3D distorted axes, multiple Y-axes with different scales) make it visually hard to map chart elements to the correct reference values.

\item [2)] \textbf{Reasoning-based failures.}  

\noindent (2.1)~\textit{Unit mismatches:} The agent sometimes multiplies values based on axis labels (e.g., reading 160 as 160,000 due to “in thousands”), which may not match the ground truth.  

\noindent (2.2)~\textit{Incorrect tool selection:} Occasionally, the agent chooses the wrong measurement tool—for instance, computing area instead of height—leading to incorrect results despite correct region localization.  

\noindent (2.3)~\textit{Question ambiguity:} Some questions, such as those from multi-ring pie charts in ChartBench, lack clear context (e.g., undefined denominators), resulting in ambiguous interpretation. We plan to address such cases in future work by enabling the agent to detect ambiguity and proactively request user clarification when necessary.

\noindent (2.4)~\textit{Label duplication:} Charts with the same label used at multiple hierarchy levels (e.g., parent and child segments both labeled “Netflix”) confuse the model during segment selection and reasoning. See Appendix \ref{appendix:failure_modes} for examples.

\noindent (2.5)~\textit{Subtype misclassification in area charts:} Overlay and stacked area charts can appear visually similar, and misclassifying them leads to incorrect answer logic (e.g., value subtraction errors), even if all other steps are executed correctly
\end{itemize}

\begin{figure*}[htbp]
\vspace{-0.6cm}
 \begin{subfigure}[htbp]{\linewidth}
  \centering
  \includegraphics[scale = 0.35]{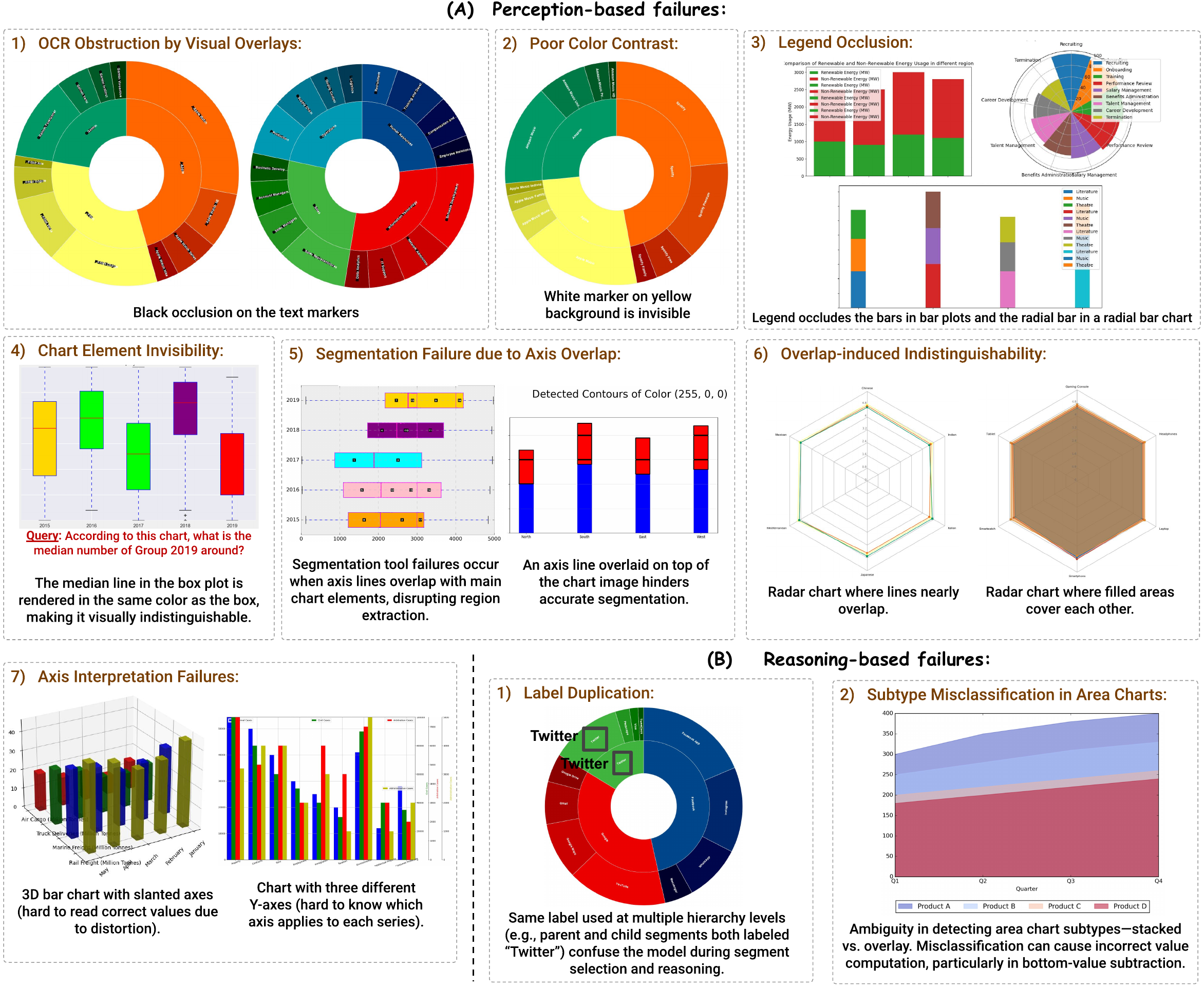}
  \caption{\textbf{Illustrations of common failure modes in \chartagent{}.}}
   \label{fig:failure_mode_illustrations}
\end{subfigure}

\vspace{0.4cm}
\begin{subfigure}[htbp]{\linewidth}
  \centering
  \includegraphics[scale = 0.39]{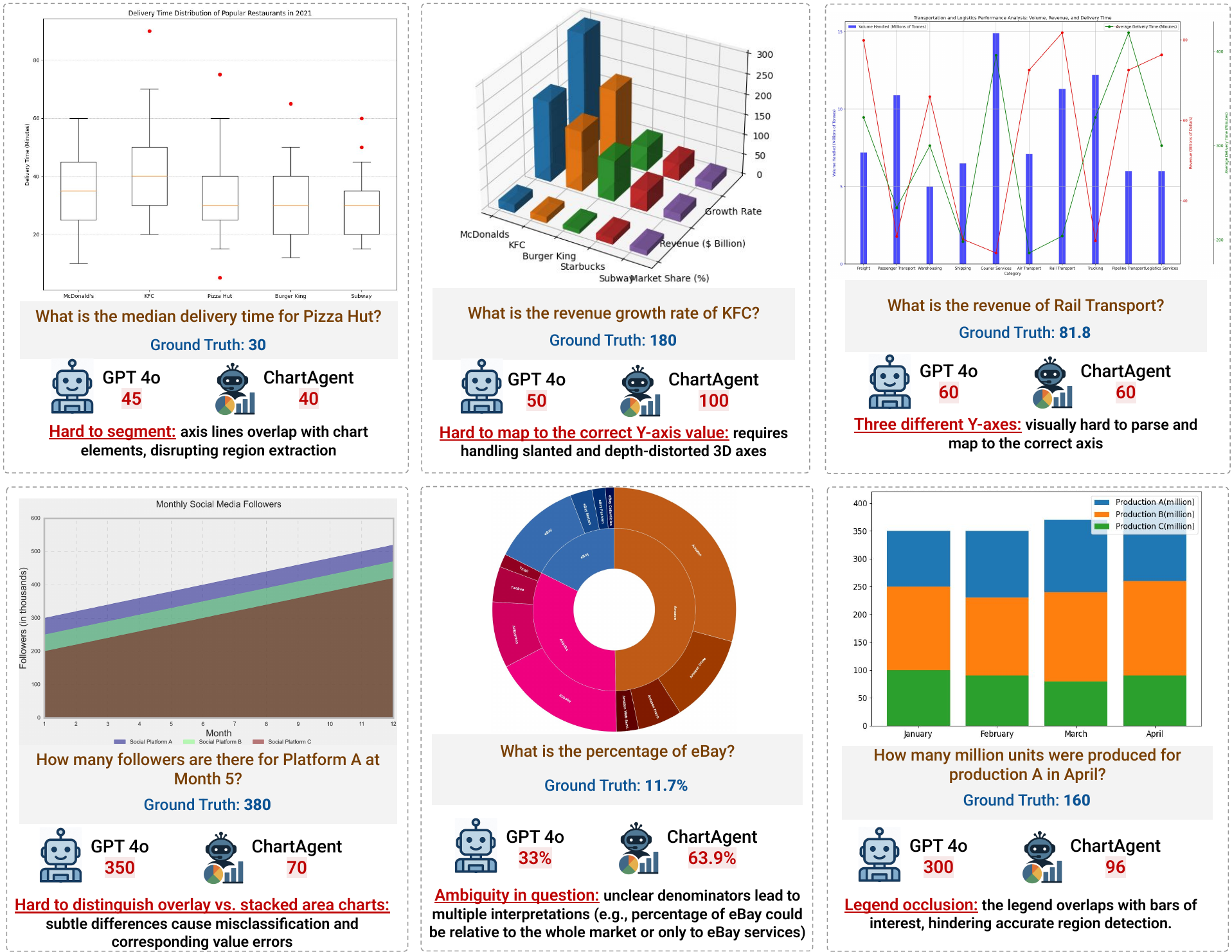}
   \caption{\textbf{Qualitative failure cases where \chartagent{} produces incorrect responses.}} 
    \label{fig:failure_mode_qual_examples}
   \end{subfigure}
   \caption{\textbf{Failure Mode Analysis.} Examples where \chartagent{} fails to produce the correct response due to visual perception challenges or reasoning ambiguities. \textit{(A) Perception-based failures} include OCR obstruction by overlays, poor color contrast, key chart element occlusions (e.g., legends blocking bars), chart element invisibility, difficult segmentation (e.g., overlapping axes or cluttered regions), overlap confusion, 3D depth distortion, and multiple Y-axis mapping errors. \textit{(B) Reasoning-based failures} include label duplication, ambiguous questions (e.g., undefined denominators) and misclassification of visually similar chart subtypes (e.g., stacked vs.\ overlay area).}
   \label{fig:appendix-failure-mode-figure}
\end{figure*}

See Figure~\ref{fig:appendix-failure-mode-figure} for illustrations of common failure modes (\ref{fig:failure_mode_illustrations}) and qualitative failure cases where \chartagent{} produces incorrect responses (\ref{fig:failure_mode_qual_examples}).
Overall, most failures are perception-driven, originating from chart tool errors rather than complex reasoning or planning.

\clearpage 
\onecolumn
\section{Prompts}\label{appendix:Prompts}
We present the prompts used for \chartagent{}~\ref{appendix:chart-agent-prompts}, baselines~\ref{appendix:baseline_prompts}, evaluation~\ref{appendix:prompts_evaluation_metrics}, and complexity analysis~\ref{appendix:prompts_complexity_analysis}. Note that some low-level prompt details are omitted below for space constraints. 

\subsection{ChartAgent Prompts}\label{appendix:chart-agent-prompts}
\chartagent{} comprises a structured set of prompts that specify reasoning, tool usage, metadata extraction, and in-context learning (ICL).
For clarity, we first present the overall concatenated prompt, followed by its individual components: the System Prompt~(\ref{appendix:chart-agent-system-prompt}), Chart Tool Definitions~(\ref{appendix-chartagent-tool-definition-prompts}), Chart Metadata Extraction Prompt~(\ref{appendix:chart-metadata-prompt}), and ICL Examples~(\ref{appendix:chartagent-ICL-prompt}).

\begin{tcolorbox}[
  title=ChartAgent Prompt,
  colback=white,
  colframe=black,
  fonttitle=\bfseries,
  sharp corners=south,
  boxrule=0.6pt,
  breakable,
  enhanced,
  fontupper=\small
]

\begin{tcolorbox}[colback=red!5, colframe=red!50, boxrule=0pt, sharp corners=south, left=2mm, right=2mm]
\textcolor{red!70!black}{\textbf{SYSTEM PROMPT}} [\ref{appendix:chart-agent-system-prompt}]
\end{tcolorbox}

\begin{tcolorbox}[
  enhanced,
  colback=blue!5,
  colframe=blue!50,
  boxrule=0pt,
  sharp corners=south,
 left=2mm, right=2mm,
  top=1mm,
  bottom=1mm
]
\textbf{Instruction:} \\[0.3em]
To support your analysis, several Python-based tools are available in \texttt{tools.py} and will be pre-imported for you.  

\begin{itemize}[leftmargin=1.2em, itemsep=0.1em, topsep=0.5em]
\item Bounding boxes follow the format \([x, y, w, h]\), where \(x\) and \(y\) denote the horizontal and vertical coordinates of the upper-left corner, and \(w\) and \(h\) represent the width and height of the box.  

\item Use the provided tools for precise numeric analysis by extracting properties such as area, height, and other quantitative attributes of chart components.

\item Execute one tool at a time, and wait for its output before proceeding. If the output seems uncertain, you may re-run the tool with adjusted parameters or switch to a different tool.  
\end{itemize}

Below are the tools defined in \texttt{tools.py}:
\par 
\vspace{0.5em}
\textcolor{magenta!70!black}{\textbf{TOOL DEFINITIONS PROMPT}} [\ref{appendix-chartagent-tool-definition-prompts}]

---

\vspace{1em}
Below are examples demonstrating how to use the tools to address user requests. You may refer to them for guidance.
\par 
\vspace{0.5em}
\textcolor{magenta!70!black}{\textbf{IN CONTEXT LEARNING EXAMPLES PROMPT}} [\ref{appendix:chartagent-ICL-prompt}]

\textcolor{gray}{Note for readers: Only ICL examples corresponding to the chart type detected in the chart metadata extraction stage are retrieved and used. For example, if the chart is detected as a pie chart, only pie chart ICL examples are included.}

---

\par 
\vspace{1em}
\textbf{GOAL:} 
\par
Using the tools above, reason about how to solve \texttt{\# USER REQUEST \#} and generate step-by-step actions (each action is a Python Jupyter notebook code block) to solve the request.

\begin{itemize}[leftmargin=1.2em, itemsep=0.1em, topsep=0.5em]
\item You may need to use the tools above to process chart images and numerical values, and to make decisions based on the visual and numerical outputs of previous code blocks.

\item The Jupyter notebook has already executed the following code to import the necessary packages:
\begin{tcblisting}{
  listing only,
  breakable,
  enhanced jigsaw,
  boxrule=0pt,
  colback=blue!5,
  colframe=blue!5,
  top=0mm, bottom=0mm, left=1mm, right=0.2mm,
  boxsep=0pt,
  before skip=0pt, after skip=0pt,
  listing options={style=pydoc}
}
```python
from PIL import Image
from IPython.display import display
from tools import *
\end{tcblisting}

\item The generated actions should fully resolve the user request \texttt{\# USER REQUEST \#}. Assume the request is reasonable and solvable; do your best to solve it.

\item If you believe you have the answer, output \texttt{ANSWER: <your answer>} and end with \texttt{TERMINATE}.
\end{itemize}
---
\par 
\vspace{0.5em}
Here is the chart metadata:
\par 
\vspace{0.5em}
\textcolor{magenta!70!black}{\textbf{CHART METADATA}} [\ref{appendix:chart-metadata-prompt}]

\textcolor{gray}{Note for readers: Extracted during the chart metadata extraction stage using the Chart Metadata Extraction Prompt~(\ref{appendix:chart-metadata-prompt}).}

\par 
\vspace{0.5em}
\end{tcolorbox}

\begin{tcolorbox}[
  enhanced,
  colback=green!5,
  colframe=green!50,
  boxrule=0pt,
  sharp corners=south,
  top=1mm, bottom=1mm, left=2mm, right=2mm
]
\textbf{Input:}
\begin{verbatim}
<chart image> # USER IMAGE #: {entry["image"]}
<question> # USER REQUEST #: {entry["query"]}
\end{verbatim}
\end{tcolorbox}

\begin{tcolorbox}[
  enhanced,
  colback=blue!5,
  colframe=blue!50,
  boxrule=0pt,
  sharp corners=south,
  top=1mm, bottom=1mm, left=2mm, right=2mm
]
\textbf{Instruction (continued):}\\[0.3em]
Now please generate only \texttt{THOUGHT 0}, and \texttt{ACTION 0} in \texttt{RESULT}. If no action needed, also reply with \texttt{ANSWER: <your answer>} and ends with \texttt{TERMINATE} in the \texttt{RESULT}.
\begin{verbatim}
# RESULT #:
\end{verbatim}
\end{tcolorbox}

\begin{tcolorbox}[
  enhanced,
  colback=yellow!5,
  colframe=yellow!50,
  boxrule=0pt,
  sharp corners=south,
  top=1mm, bottom=1mm, left=2mm, right=2mm
]
\textbf{Agent:}\\[0.3em]
\thought{0}{I should annotate the legend in the chart image to correctly identify the marker of interest.} 

\action{0}{}
\begin{tcblisting}{ 
listing only, breakable, enhanced jigsaw, boxrule=0pt, colback=yellow!5, colframe=yellow!5, top=0mm, bottom=0mm, left=1mm, right=0.2mm, boxsep=0pt, before skip=0pt, after skip=0pt, listing options={style=pydoc} 
}
```python
legend_image, labeled_legend, bbox_map = annotate_legend(chart_image, legend_texts)
display(labeled_legend)
print (bbox_map)
```
\end{tcblisting}
\end{tcolorbox}

\begin{tcolorbox}[
  enhanced,
  colback=blue!5,
  colframe=blue!50,
  boxrule=0pt,
  sharp corners=south,
  top=1mm, bottom=1mm, left=2mm, right=2mm
]
\textbf{Instruction (continued):}\\[0.3em]
\texttt{OBSERVATION:} Execution success. The output is as follows: \texttt{<output>}
\par 
\vspace{0.5em}
Please generate the next \texttt{THOUGHT} and \texttt{ACTION}. 
If you can get the answer, please also reply with \texttt{ANSWER: <your answer>} and end with \texttt{TERMINATE}.

\end{tcolorbox}
\textit{Trajectory continues as the agent generates \texttt{THOUGHT 1}, \texttt{ACTION 1}, … until termination.}
\end{tcolorbox}

\noindent For visualization purposes, prompts containing code are formatted differently here; in practice, however, all prompts are provided as plain text inputs to the MLLM.
The same prompt template is used across all chart VQA samples and datasets in this work.

\subsubsection{System Prompt}\label{appendix:chart-agent-system-prompt}
The system prompt establishes the agent’s role and high-level objectives. It instructs the model to follow structured reasoning, invoke tools where appropriate, and return answers in a well-defined format.
\begin{tcolorbox}[
  title=System Prompt,
  colback=white,
  colframe=black,
  fonttitle=\bfseries,
  sharp corners=south,
  boxrule=0.6pt,
  breakable,
  enhanced,
  fontupper=\small
]

\begin{tcolorbox}[
  enhanced,
  colback=red!5,
  colframe=red!50,
  boxrule=0pt,
  sharp corners=south,
  top=1mm, bottom=1mm, left=2mm, right=2mm
]
You are a helpful multimodal AI assistant for answering questions about chart images, including numeric QA, free-form QA, and multiple-choice QA.
\\
\\
You operate in a Python Jupyter notebook environment and can:
\begin{itemize}[leftmargin=1.2em, itemsep=0.1em, topsep=0.5em]
    \item Suggest Python code (in executable code blocks) to process images, text, or data.
    \item Use variables and states from previous cells.
    \item Provide complete code, not partial snippets.
    \item The notebook imports \texttt{Image} from the PIL package and \texttt{display} from the IPython.display package. Display all image outputs using \texttt{display()}.

\end{itemize}
\vspace{0.2cm}
\textbf{[MORE INSTRUCTIONS ...]}  
(The actual system prompt contains additional detailed guidelines.)
\\
\\
Execution Instructions:
\begin{itemize}[leftmargin=1.2em, itemsep=0.1em, topsep=0.5em]
    \item Execute one tool at a time and wait for results.
    \item If an error occurs, fix it and re-run.
    \item If unsure about output, try different parameters or tools.
\end{itemize}

\vspace{0.5em}
For each turn, you should first do a \textbf{\texttt{"THOUGHT"}}, based on the chart images and textual question you see.  
If you think you get the answer to the initial user request, you can reply with:  
\textbf{\texttt{ANSWER: <your answer>}} and end with \textbf{\texttt{TERMINATE}}.
\end{tcolorbox}

\end{tcolorbox}

\subsubsection{Chart Tool Definitions}\label{appendix-chartagent-tool-definition-prompts}
The following are the Python-based tools available to \chartagent{}, along with their inputs, outputs, and expected behaviors. An abridged parameter set is shown for some tools to save space and aid readability.
\begin{tcolorbox}[
  title=Tool Definitions Prompt,
  colback=white,
  colframe=black,
  fonttitle=\bfseries,
  sharp corners=south,
  boxrule=0.6pt,
  breakable,
  enhanced jigsaw,
  fontupper=\small,
]


\definecolor{py-keyword}{RGB}{0,0,0}       
\definecolor{py-func}{RGB}{163,21,21}        
\definecolor{py-arg}{RGB}{0,16,128}          
\definecolor{py-type}{RGB}{38,127,153}       
\definecolor{py-num}{RGB}{9,134,88}          
\definecolor{py-doc}{RGB}{0,128,0}       

\lstdefinestyle{pydoc}{
  basicstyle=\ttfamily\small\color{py-doc},
  breaklines=true,
  breakatwhitespace=false,
  columns=fullflexible,
  keepspaces=false,
  showstringspaces=false,
  xleftmargin=2em,
}

\verb|```python|
\par
\begin{center}
\noindent\colorbox{black!5}{\parbox{\dimexpr\linewidth-2\fboxsep}{\centering\fontsize{11pt}{12pt}\selectfont\textbf{Universal Tools}}}
\end{center}
\vspace{0.5em} 

\textcolor{py-keyword}{\textbf{def}} \textcolor{py-func}{\textbf{annotate\_legend }}\textcolor{black}{(}%
\textcolor{py-arg}{image}\textcolor{black}{:} \textcolor{py-type}{PIL.Image.Image},\,
\textcolor{py-arg}{legend}\textcolor{black}{:} \textcolor{py-type}{dict[str]}%
\textcolor{black}{) $\rightarrow$ \textcolor{py-type}{tuple[}%
\textcolor{py-type}{PIL.Image.Image},\,
\textcolor{py-type}{PIL.Image.Image},\,
\textcolor{py-type}{dict[int, tuple[str, tuple[int,int,int,int]]]}%
\textcolor{py-type}{]}\textcolor{black}{:}}

\medskip
\hspace*{1.5em}\textcolor{py-doc}{""" Detects legend coordinates, crops the legend, and annotates it with numeric labels.}
\par
\vspace{0.2cm}
\hspace*{1.5em}\textcolor{py-doc}{Args:}
{\color{py-doc}
\hspace*{1.5em}\begin{itemize}[label={},nosep]
  \item \texttt{image} (\texttt{PIL.Image.Image}): Input chart image
  \item \texttt{legend} (\texttt{dict[str]}): Legend strings
\end{itemize}
}
\vspace{0.5cm}
\hspace*{1.5em}\textcolor{py-doc}{Returns:}
{\color{py-doc}
\hspace*{1.5em}\begin{itemize}[label={},nosep]
  \item \textcolor{py-doc}{\texttt{legend\_image} (\texttt{PIL.Image.Image}): Cropped legend image
  \item \texttt{labeled\_legend} (\texttt{PIL.Image.Image}): Annotated legend image with numeric label mappings
  \item \texttt{bbox\_mapping} (\texttt{dict[int, (str, (int, int, int, int))]}): Maps numeric labels to (text,  bounding box coordinates [x\_min,y\_min,x\_max,y\_max]).}
\end{itemize}
}
\vspace{0.5cm}
\hspace*{1.5em}\textcolor{py-doc}{Example:}
\begin{tcblisting}{
  listing only,
  breakable,
  enhanced jigsaw,
  boxrule=0pt,
  colback=white,
  colframe=white,
  top=0mm, bottom=0mm, left=1mm, right=0.2mm,
  boxsep=0pt,
  before skip=0pt, after skip=0pt,
  listing options={style=pydoc}
}
image = PIL.Image.open("chart_image.png")
legend_image, labeled_legend, bbox_mapping = annotate_legend(image, 
legend={"Legend1", "Legend2"})
display(labeled_legend)
\end{tcblisting}
\hspace*{1.5em}\textcolor{py-doc}{"""}
\par
\vspace{0.2cm}

\textcolor{py-keyword}{\textbf{def}} \textcolor{py-func}{\textbf{get\_marker\_rgb }}%
\textcolor{black}{ (}%
\textcolor{py-arg}{image}\textcolor{black}{:} \textcolor{py-type}{PIL.Image.Image},\,
\textcolor{py-arg}{bbox\_mapping}\textcolor{black}{:} \textcolor{py-type}{dict[int, tuple[str, tuple[int, int, int, int]]]},\,
\textcolor{py-arg}{text\_of\_interest}\textcolor{black}{:} \textcolor{py-type}{str},\,
\textcolor{py-arg}{label\_of\_interest}\textcolor{black}{:} \textcolor{py-type}{int},\,
\textcolor{py-arg}{distance\_between\_text\_and\_marker}\textcolor{black}{:} \textcolor{py-type}{int}%
\textcolor{black}{) $\rightarrow$ \textcolor{py-type}{tuple[int, int, int]}\textcolor{black}{:}}

\medskip
\hspace*{1.5em}\textcolor{py-doc}{"""
Retrieves the dominant RGB color of a legend marker, either by label (from an annotated legend image) or by associated text.}
\par
\vspace{0.2cm}

\hspace*{1.5em}\textcolor{py-doc}{Args:}
{\color{py-doc}
\hspace*{1.5em}\begin{itemize}[label={},nosep]
  \item \texttt{image} (\texttt{PIL.Image.Image}): Input legend image 
  \item \texttt{bbox\_mapping} (\texttt{dict}): Mapping of label numbers to (text, bbox) tuples. The bounding box is (x\_min, y\_min, x\_max, y\_max).
  \item \texttt{text\_of\_interest} (\texttt{str}, optional): The legend text whose marker color should be retrieved. If provided, fuzzy matching is applied.
  \item \texttt{label\_of\_interest} (\texttt{int}, optional): The label number in \texttt{bbox\_mapping} whose marker color should be retrieved.
  \item \texttt{distance\_between\_text\_and\_marker} (\texttt{int}): Approximate distance in pixels between the legend text and its marker (default: 5).
\end{itemize}
}
\vspace{0.5cm}

\hspace*{1.5em}\textcolor{py-doc}{Returns:}
{\color{py-doc}
\hspace*{1.5em}\begin{itemize}[label={},nosep]
  \item \color{py-doc} \texttt{tuple[int, int, int]}: The (R, G, B) color of the detected marker.
\end{itemize}
}
\vspace{0.5cm}

\hspace*{1.5em}\textcolor{py-doc}{Examples:}
\begin{tcblisting}{
  listing only,
  breakable,
  enhanced jigsaw,
  boxrule=0pt,
  colback=white,
  colframe=white,
  top=0mm, bottom=0mm, left=1mm, right=0.2mm,
  boxsep=0pt,
  before skip=0pt, after skip=0pt,
  listing options={style=pydoc}
}
# Example 1: Using text label
image = Image.open("chart_image.png")
legend_image, labeled_legend, bbox_mapping = annotate_legend(image)
rgb_color = get_marker_rgb(legend_image, bbox_mapping, text_of_interest="Rock")
print("Detected RGB color:", rgb_color)  # Output: (0, 0, 255)

# Example 2: Using label number
rgb_color = get_marker_rgb(legend_image, bbox_mapping, label_of_interest=5)
print("Detected RGB color:", rgb_color)   # Output: (255, 0, 0)
\end{tcblisting}
\hspace*{1.5em}\textcolor{py-doc}{"""}
\par
\vspace{0.2cm}

\textcolor{py-keyword}{\textbf{def}} \textcolor{py-func}{\textbf{clean\_chart\_image }}\textcolor{black}{(}%
\textcolor{py-arg}{image}\textcolor{black}{:} \textcolor{py-type}{PIL.Image.Image},  
\textcolor{py-arg}{title}\textcolor{black}{:} \textcolor{py-type}{str},  
\textcolor{py-arg}{legend}\textcolor{black}{:} \textcolor{py-type}%
{dict[str]}{) $\rightarrow$ \textcolor{py-type}{PIL.Image.Image}:}

\medskip

\hspace*{1.5em}\textcolor{py-doc}{""" Cleans a chart image by removing title and legend if provided.}
\par
\vspace{0.2cm}
\hspace*{1.5em}\textcolor{py-doc}{Args:}
\hspace*{1.5em}\textcolor{py-doc}{\begin{itemize}[label={},nosep]
  \item \color{py-doc} image (\texttt{PIL.Image.Image}): Input PIL image of the chart
  \item title (\texttt{str}): Title to remove (\texttt{None} to skip title removal)
  \item legend (\texttt{dict[str]}): Legend strings to remove (\texttt{None} to skip legend removal)
  \item Thresholds and expand values control removal bounding boxes.
\end{itemize}
}
\vspace{0.5cm}
\hspace*{1.5em}\textcolor{py-doc}{Returns:}
\textcolor{py-doc}{\begin{itemize}[label={},nosep]
  \item cleaned\_image (\texttt{PIL.Image.Image}): Cleaned chart image
\end{itemize}
}
\vspace{0.5cm}
\hspace*{1.5em}\textcolor{py-doc}{Example:}
\begin{tcblisting}{
  listing only,
  breakable,
  enhanced jigsaw,
  boxrule=0pt,
  colback=white,
  colframe=white,
  top=0mm, bottom=0mm, left=1mm, right=0.2mm,
  boxsep=0pt,
  before skip=0pt, after skip=0pt,
  listing options={style=pydoc}
}
image = PIL.Image.open("chart_image.png")
cleaned_image = clean_chart_image(image, title="Title", legend={"Legend1", "Legend2"})
display(cleaned_image)
\end{tcblisting}
\hspace*{1.5em}\textcolor{py-doc}{"""}
\par
\vspace{0.2cm}

\textcolor{py-keyword}{\textbf{def}} \textcolor{py-func}{\textbf{segment\_and\_mark }}\textcolor{black}{(}%
\textcolor{py-arg}{image}\textcolor{black}{:} \textcolor{py-type}{PIL.Image.Image},\,
\textcolor{py-arg}{segmentation\_model}\textcolor{black}{:} \textcolor{py-type}{str},\,
\textcolor{py-arg}{min\_area}\textcolor{black}{:} \textcolor{py-type}{int},\,
\textcolor{py-arg}{iou\_thresh\_unique}\textcolor{black}{:} \textcolor{py-type}{float},\,
\textcolor{py-arg}{iou\_thresh\_composite}\textcolor{black}{:} \textcolor{py-type}{float},\,
\textcolor{py-arg}{white\_ratio\_thresh}\textcolor{black}{:} \textcolor{py-type}{float},\,
\textcolor{py-arg}{remove\_background\_color}\textcolor{black}{:} \textcolor{py-type}{bool}%
\textcolor{black}{) $\rightarrow$ \textcolor{py-type}{tuple[}%
\textcolor{py-type}{PIL.Image.Image},\,
\textcolor{py-type}{list[dict]}%
\textcolor{py-type}{]}\textcolor{black}{:}}

\medskip
\hspace*{1.5em}\textcolor{py-doc}{""" 
Segments an input image using the specified model and applies post-processing to clean the masks through a multi-step filtering pipeline that removes small, duplicate, composite, and background-dominated masks. Returns a labeled image with drawn contours and optional numbered labels, along with a cleaned list of segmentation masks.}
\par
\vspace{0.2cm}
\hspace*{1.5em}\textcolor{py-doc}{Args:}
{\color{py-doc}
\hspace*{1.5em}\begin{itemize}[label={},nosep]
  \item \texttt{image} (\texttt{PIL.Image.Image}): The input chart image to be segmented
  \item \texttt{segmentation\_model} (\texttt{str}): Segmentation model (Segment Anything (\texttt{"SAM"}) by default)
  \item \texttt{min\_area} (\texttt{int}): Minimum pixel area to keep a mask (\texttt{5000} default)
  \item \texttt{iou\_thresh\_unique} (\texttt{float}): IoU threshold for duplicate removal (\texttt{0.9} default)
  \item \texttt{iou\_thresh\_composite} (\texttt{float}): IoU threshold for composite mask removal (\texttt{0.98} default)
  \item \texttt{white\_ratio\_thresh} (\texttt{float}): White pixel ratio to discard mask (\texttt{0.95} default)
  \item \texttt{remove\_background\_color} (\texttt{bool}): If \texttt{True}, remove background color pixels
\end{itemize}
}
\vspace{0.5cm}
\hspace*{1.5em}\textcolor{py-doc}{Returns:}
{\color{py-doc}
\hspace*{1.5em}\begin{itemize}[label={},nosep]
  \item \texttt{labeled\_image} (\texttt{PIL.Image.Image}): Segmented and labeled image with drawn contours and numeric labels
  \item \texttt{cleaned\_masks} (\texttt{list[dict]}): Cleaned list of segmentation masks
\end{itemize}
}
\vspace{0.5cm}
\hspace*{1.5em}\textcolor{py-doc}{Example:}
\begin{tcblisting}{
  listing only,
  breakable,
  enhanced jigsaw,
  boxrule=0pt,
  colback=white,
  colframe=white,
  top=0mm, bottom=0mm, left=1mm, right=0.2mm,
  boxsep=0pt,
  before skip=0pt, after skip=0pt,
  listing options={style=pydoc}
}
image = Image.open("chart_image.png")
labeled_image, cleaned_masks = segment_and_mark(image)
display(labeled_image)
print(f"Total masks: {len(cleaned_masks)}")
\end{tcblisting}
\hspace*{1.5em}\textcolor{py-doc}{"""}
\par
\vspace{0.2cm}

\textcolor{py-keyword}{\textbf{def}} \textcolor{py-func}{\textbf{axis\_localizer }}\textcolor{black}{(}%
\textcolor{py-arg}{image}\textcolor{black}{:} \textcolor{py-type}{PIL.Image.Image},\,
\textcolor{py-arg}{axis}\textcolor{black}{:} \textcolor{py-type}{str},\,
\textcolor{py-arg}{axis\_threshold}\textcolor{black}{:} \textcolor{py-type}{float},\,
\textcolor{py-arg}{axis\_tickers}\textcolor{black}{:} \textcolor{py-type}{list}%
\textcolor{black}{) $\rightarrow$ \textcolor{py-type}{tuple[}\textcolor{py-type}{list[float]},\,\textcolor{py-type}{list[int]}\textcolor{py-type}{]}\textcolor{black}{:}}

\medskip
\hspace*{1.5em}\textcolor{py-doc}{""" 
Localizes the specified axis (x-axis, left y-axis, or right y-axis) by detecting its numeric tick values and mapping them to corresponding pixel positions in the chart image. Uses Tesseract OCR and EasyOCR.
}
\par
\vspace{0.2cm}
\hspace*{1.5em}\textcolor{py-doc}{Args:}
{\color{py-doc}
\hspace*{1.5em}\begin{itemize}[label={},nosep]
  \item \texttt{pil\_image} (\texttt{PIL.Image.Image}): Input chart image
  \item \texttt{axis} (\texttt{str}): Axis to localize; \texttt{'x'} (x-axis), \texttt{'y'} (left y-axis), or \texttt{'right\_y'} (right y-axis)
  \item \texttt{axis\_threshold} (\texttt{float}): Fraction of the image to scan for tick labels along the axis direction (\texttt{0.2} default)
  \item \texttt{axis\_tickers} (\texttt{list} or \texttt{None}): Optional pre-supplied axis tick strings to improve matching
\end{itemize}
}
\vspace{0.5cm}
\hspace*{1.5em}\textcolor{py-doc}{Returns:}
{\color{py-doc}
\hspace*{1.5em}\begin{itemize}[label={},nosep]
  \item \texttt{axis\_values} (\texttt{list[float]}): Detected numeric tick values (e.g., \texttt{[0, 200, 400, 600]})
  \item \texttt{axis\_pixel\_positions} (\texttt{list[int]}): Corresponding pixel positions (e.g., \texttt{[950, 850, 750, 650]})
\end{itemize}
}
\vspace{0.5cm}
\hspace*{1.5em}\textcolor{py-doc}{Example:}
\begin{tcblisting}{
  listing only,
  breakable,
  enhanced jigsaw,
  boxrule=0pt,
  colback=white,
  colframe=white,
  top=0mm, bottom=0mm, left=1mm, right=0.2mm,
  boxsep=0pt,
  before skip=0pt, after skip=0pt,
  listing options={style=pydoc}
}
axis_values, axis_pixel_positions = axis_localizer(image, axis='y', axis_threshold=0.2,
    axis_tickers=["200", "400", "600", "800", "1000", "1200", "1400"])
print(axis_values, axis_pixel_positions)
\end{tcblisting}
\hspace*{1.5em}\textcolor{py-doc}{"""}
\par
\vspace{0.2cm}

\textcolor{py-keyword}{\textbf{def}} \textcolor{py-func}{\textbf{interpolate\_pixel\_to\_value }}\textcolor{black}{(}%
\textcolor{py-arg}{pixel}\textcolor{black}{:} \textcolor{py-type}{float},\,
\textcolor{py-arg}{axis\_values}\textcolor{black}{:} \textcolor{py-type}{list[float]},\,
\textcolor{py-arg}{axis\_pixel\_positions}\textcolor{black}{:} \textcolor{py-type}{list[int]}%
\textcolor{black}{) $\rightarrow$ \textcolor{py-type}{float}\textcolor{black}{:}}

\medskip
\hspace*{1.5em}\textcolor{py-doc}{""" 
Maps a pixel coordinate to its corresponding axis value using linear interpolation between known axis ticks and \color{py-doc} their pixel positions.}
\par
\vspace{0.2cm}
\hspace*{1.5em}\textcolor{py-doc}{Args:}
{\color{py-doc}
\hspace*{1.5em}\begin{itemize}[label={},nosep]
\item \textcolor{py-doc}{\texttt{pixel} (\texttt{float} or \texttt{int}): Pixel coordinate to map
  \item \texttt{axis\_values} (\texttt{list[float]}): Numeric axis values (e.g., \texttt{[0, 200, 400, 600]})
  \item \texttt{axis\_pixel\_positions} (\texttt{list[int]}): Pixel positions corresponding to \texttt{axis\_values} (e.g., \texttt{[950, 850, 750, 650]})}
\end{itemize}
}
\vspace{0.5cm}
\hspace*{1.5em}\textcolor{py-doc}{Returns:}
{\color{py-doc}
\hspace*{1.5em}\begin{itemize}[label={},nosep]
  \item \texttt{float}: Interpolated axis value corresponding to the given pixel
\end{itemize}
}
\vspace{0.5cm}
\hspace*{1.5em}\textcolor{py-doc}{Example:}
\begin{tcblisting}{
  listing only,
  breakable,
  enhanced jigsaw,
  boxrule=0pt,
  colback=white,
  colframe=white,
  top=0mm, bottom=0mm, left=1mm, right=0.2mm,
  boxsep=0pt,
  before skip=0pt, after skip=0pt,
  listing options={style=pydoc}
}
axis_values = [0, 200, 400, 600]
axis_pixel_positions = [950, 850, 750, 650]
val = interpolate_pixel_to_value(800, axis_values, axis_pixel_positions)
print(val)  # Expected interpolation between 200 and 400
\end{tcblisting}
\hspace*{1.5em}\textcolor{py-doc}{"""}
\par
\vspace{0.2cm}

\textcolor{py-keyword}{\textbf{def}} \textcolor{py-func}{\textbf{arithmetic }}\textcolor{black}{(}%
\textcolor{py-arg}{a}\textcolor{black}{:} \textcolor{py-type}{float},\,
\textcolor{py-arg}{b}\textcolor{black}{:} \textcolor{py-type}{float},\,
\textcolor{py-arg}{operation}\textcolor{black}{:} \textcolor{py-type}{str}%
\textcolor{black}{) $\rightarrow$ \textcolor{py-type}{float}\textcolor{black}{:}}

\medskip
\hspace*{1.5em}\textcolor{py-doc}{""" 
Performs a specified arithmetic operation between two numeric inputs. Supports operations such as addition, subtraction, multiplication, division, percentage, and ratio. 
}
\par
\vspace{0.2cm}
\hspace*{1.5em}\textcolor{py-doc}{Args:}
{\color{py-doc}
\hspace*{1.5em}\begin{itemize}[label={},nosep]
  \item \texttt{a} (\texttt{float}): First operand
  \item \texttt{b} (\texttt{float}): Second operand
  \item \texttt{operation} (\texttt{str}): Arithmetic operation to perform. Supported: \texttt{"add"}, \texttt{"subtract"}, \texttt{"multiply"}, \texttt{"divide"}, \texttt{"percentage"}, \texttt{"ratio"} (\texttt{"percentage"} by default)
\end{itemize}
}
\vspace{0.5cm}
\hspace*{1.5em}\textcolor{py-doc}{Returns / Raises:}
{\color{py-doc}
\hspace*{1.5em}\begin{itemize}[label={},nosep]
  \item \texttt{float}: Result of the arithmetic operation
  \item \texttt{ValueError}: If an unsupported operation is provided or division by zero occurs
\end{itemize}
}
\vspace{0.5cm}
\hspace*{1.5em}\textcolor{py-doc}{Example:}
\begin{tcblisting}{
  listing only,
  breakable,
  enhanced jigsaw,
  boxrule=0pt,
  colback=white,
  colframe=white,
  top=0mm, bottom=0mm, left=1mm, right=0.2mm,
  boxsep=0pt,
  before skip=0pt, after skip=0pt,
  listing options={style=pydoc}
}
total = 1200
part = 300
result = arithmetic(part, total, operation="percentage")
print("Percentage:", result)  # Output: 25.0
\end{tcblisting}
\hspace*{1.5em}\textcolor{py-doc}{"""}
\par
\vspace{0.2cm}

\begin{center}
\noindent\colorbox{black!5}{\parbox{\dimexpr\linewidth-2\fboxsep}{\centering\fontsize{11pt}{12pt}\selectfont\textbf{Chart-specific Tools}}}
\end{center}
\vspace{0.5em} 

\noindent{\parbox{\dimexpr\linewidth-2\fboxsep}{\fontsize{11pt}{12pt}\selectfont\textbf{\underline{Pie Chart $\vert$ Treemap}}}}
\vspace{1em} 

\textcolor{py-keyword}{\textbf{def}} \textcolor{py-func}{\textbf{compute\_segment\_area }}\textcolor{black}{(}%
\textcolor{py-arg}{image}\textcolor{black}{:} \textcolor{py-type}{PIL.Image.Image},\,
\textcolor{py-arg}{filter\_rgb}\textcolor{black}{:} \textcolor{py-type}{tuple[int,int,int]},\,
\textcolor{py-arg}{measure}\textcolor{black}{:} \textcolor{py-type}{str},\,
\textcolor{py-arg}{masks}\textcolor{black}{:} \textcolor{py-type}{list},\,
\textcolor{py-arg}{filter\_segment}\textcolor{black}{:} \textcolor{py-type}{list}%
\textcolor{black}{) $\rightarrow$ \textcolor{py-type}{tuple[}%
\textcolor{py-type}{PIL.Image.Image},\,
\textcolor{py-type}{int}%
\textcolor{py-type}{]}\textcolor{black}{:}}

\medskip
\hspace*{1.5em}\textcolor{py-doc}{""" 
Computes the area of a chart segment by: (1) counting discrete visual elements of a specified color, (2) counting pixels of a specified color, or (3) counting pixels within a segment identified by a specific label ID. Commonly used for pie charts and tree maps.
}
\par
\vspace{0.2cm}
\hspace*{1.5em}\textcolor{py-doc}{Args:}
{\color{py-doc}
\hspace*{1.5em}\begin{itemize}[label={},nosep]
  \item \texttt{image} (\texttt{PIL.Image.Image}): Input chart image (cleaned if necessary)
  \item \texttt{filter\_rgb} (\texttt{tuple[int,int,int]}, optional): RGB values to filter by; if \texttt{None}, uses full chart
  \item \texttt{measure} (\texttt{str}): Method to measure area —  \texttt{"pixels"} or \texttt{"discrete-dots"}
  \item \texttt{masks} (\texttt{list}, optional): Segmentation masks (SAM-style)
  \item \texttt{filter\_segment} (\texttt{list}, optional): Segment label numbers to include in pixel counting
\end{itemize}
}
\vspace{0.5cm}
\hspace*{1.5em}\textcolor{py-doc}{Returns / Raises:}
{\color{py-doc}
\hspace*{1.5em}\begin{itemize}[label={},nosep]
  \item \texttt{visualization} (\texttt{PIL.Image.Image}): Image with detected/filtered areas highlighted
  \item \texttt{int}: Computed area (discrete-dots or pixels)
  \item \texttt{ValueError}: If \texttt{measure} is unsupported
\end{itemize}
}
\vspace{0.5cm}
\hspace*{1.5em}\textcolor{py-doc}{Examples:}
\begin{tcblisting}{
  listing only,
  breakable,
  enhanced jigsaw,
  boxrule=0pt,
  colback=white,
  colframe=white,
  top=0mm, bottom=0mm, left=1mm, right=0.2mm,
  boxsep=0pt,
  before skip=0pt, after skip=0pt,
  listing options={style=pydoc}
}
# Example 1: Full pie chart area (discrete-dots)
image = Image.open("pie_chart.png")
vis, area = compute_segment_area(image, measure="discrete-dots")
print(area)  # e.g., 500

# Example 2: Area of RGB-colored section (pixels)
rgb_interest = (255, 0, 0)  # red
vis, area = compute_segment_area(image, filter_rgb=rgb_interest, measure="pixels")
print(area)  # e.g., 5000

# Example 3: Area of specific segments via masks
labeled_img, masks = segment_and_mark(image)
vis, area = compute_segment_area(image, measure="pixels", masks=masks, filter_segment=[3,5,7])
print(area)  # e.g., 8453
\end{tcblisting}
\hspace*{1.5em}\textcolor{py-doc}{"""}
\par
\vspace{0.2cm}

\vspace{0.5em}
\noindent{\parbox{\dimexpr\linewidth-2\fboxsep}{\fontsize{11pt}{12pt}\selectfont\textbf{\underline{Bar Chart}}}}
\vspace{1em} 

\textcolor{py-keyword}{\textbf{def}} \textcolor{py-func}{\textbf{get\_bar }}\textcolor{black}{(}%
\textcolor{py-arg}{image}\textcolor{black}{:} \textcolor{py-type}{PIL.Image.Image},\,
\textcolor{py-arg}{rgb\_of\_interest}\textcolor{black}{:} \textcolor{py-type}{tuple[int,int,int]},\,
\textcolor{py-arg}{ticker\_label}\textcolor{black}{:} \textcolor{py-type}{str},\,
\textcolor{py-arg}{segmentation\_model}\textcolor{black}{:} \textcolor{py-type}{str},\,
\textcolor{py-arg}{bar\_orientation}\textcolor{black}{:} \textcolor{py-type}{str}
\textcolor{black}{) $\rightarrow$ \textcolor{py-type}{tuple[}%
\textcolor{py-type}{int},\,
\textcolor{py-type}{int},\,
\textcolor{py-type}{int},\,
\textcolor{py-type}{int}%
\textcolor{py-type}{]}\textcolor{black}{:}}

\medskip
\hspace*{1.5em}\textcolor{py-doc}{"""
Detects and returns the bounding box of a bar in a chart image that matches a specified color and/or axis label. It segments bar regions using a model, filters by color if provided, locates the target axis label using OCR if specified, and selects the closest matching bar accordingly. Commonly used for bar charts.
}
\par
\vspace{0.2cm}
\hspace*{1.5em}\textcolor{py-doc}{Args:}
{\color{py-doc}
\hspace*{1.5em}\begin{itemize}[label={},nosep]
  \item \texttt{image} (\texttt{PIL.Image.Image}): Input chart image
  \item \texttt{rgb\_of\_interest} (\texttt{tuple[int,int,int]}, optional): RGB color of target bar
  \item \texttt{ticker\_label} (\texttt{str}, optional): Axis label text of interest
  \item \texttt{segmentation\_model} (\texttt{str}): Segmentation model for detection (\texttt{"SAM"} default)
  \item \texttt{bar\_orientation} (\texttt{str}): \texttt{"vertical"}, \texttt{"horizontal"}, or \texttt{"vertical-right"} (\texttt{"vertical"} default)
\end{itemize}
}
\vspace{0.5cm}
\hspace*{1.5em}\textcolor{py-doc}{Returns:}
{\color{py-doc}
\hspace*{1.5em}\begin{itemize}[label={},nosep]
  \item \texttt{tuple[int,int,int,int]}: Bounding box \texttt{(x, y, w, h)} if bar is found, else \texttt{None}
\end{itemize}
}
\vspace{0.5cm}
\hspace*{1.5em}\textcolor{py-doc}{Examples:}
\begin{tcblisting}{
  listing only,
  breakable,
  enhanced jigsaw,
  boxrule=0pt,
  colback=white,
  colframe=white,
  top=0mm, bottom=0mm, left=1mm, right=0.2mm,
  boxsep=0pt,
  before skip=0pt, after skip=0pt,
  listing options={style=pydoc}
}
# Example 1: Vertical bar plot
image = Image.open("bar_chart.png")
bbox = get_bar(image, rgb_of_interest=(100,128,45), ticker_label="2016")
print(bbox)  # e.g., (50, 100, 30, 200)

# Example 2: Combination bar-line plot
bbox = get_bar(image, rgb_of_interest=(100,128,45), ticker_label="2016", bar_orientation = "vertical-right")
print(bbox)
\end{tcblisting}
\hspace*{1.5em}\textcolor{py-doc}{"""}
\par
\vspace{0.2cm}

\textcolor{py-keyword}{\textbf{def}} \textcolor{py-func}{\textbf{compute\_bar\_height }}\textcolor{black}{(}%
\textcolor{py-arg}{image}\textcolor{black}{:} \textcolor{py-type}{PIL.Image.Image},\,
\textcolor{py-arg}{bar\_of\_interest}\textcolor{black}{:} \textcolor{py-type}{tuple[int,int,int,int]},\,
\textcolor{py-arg}{bar\_orientation}\textcolor{black}{:} \textcolor{py-type}{str},\,
\textcolor{py-arg}{axis\_threshold}\textcolor{black}{:} \textcolor{py-type}{float},\,
\textcolor{py-arg}{x\_axis\_tickers}\textcolor{black}{:} \textcolor{py-type}{list},\,
\textcolor{py-arg}{y\_axis\_tickers}\textcolor{black}{:} \textcolor{py-type}{list},\,
\textcolor{py-arg}{x\_axis\_title}\textcolor{black}{:} \textcolor{py-type}{str},\,
\textcolor{py-arg}{y\_axis\_title}\textcolor{black}{:} \textcolor{py-type}{str}%
\textcolor{black}{) $\rightarrow$ \textcolor{py-type}{float}\textcolor{black}{:}}

\medskip
\hspace*{1.5em}\textcolor{py-doc}{"""
Computes a bar’s value (height or length) by mapping its pixel bounding box to axis values using OCR-based axis localization. Supports left/right y-axes for vertical bars and the x-axis for horizontal bars. Commonly used for bar charts.}
\par
\vspace{0.2cm}
\hspace*{1.5em}\textcolor{py-doc}{Args:}
{\color{py-doc}
\hspace*{1.5em}\begin{itemize}[label={},nosep]
  \item \texttt{image} (\texttt{PIL.Image.Image}): Input chart image
  \item \texttt{bar\_of\_interest} (\texttt{tuple[int,int,int,int]}): Bounding box \texttt{(x, y, w, h)} of the bar
  \item \texttt{bar\_orientation} (\texttt{str}): \texttt{"vertical"}, \texttt{"vertical-right"}, or \texttt{"horizontal"} (\texttt{"vertical"} default)
  \item \texttt{axis\_threshold} (\texttt{float}): Fraction of the image scanned for tick labels during axis localization (\texttt{0.15} default)
  \item \texttt{x\_axis\_tickers} (\texttt{list} or \texttt{None}): Optional pre-read x-axis tick labels
  \item \texttt{y\_axis\_tickers} (\texttt{list} or \texttt{None}): Optional pre-read y-axis tick labels
  \item \texttt{x\_axis\_title} (\texttt{str} or \texttt{None}): X-axis title, if available
  \item \texttt{y\_axis\_title} (\texttt{str} or \texttt{None}): Y-axis title, if available
\end{itemize}
}
\vspace{0.5cm}
\hspace*{1.5em}\textcolor{py-doc}{Returns:}
{\color{py-doc}
\hspace*{1.5em}\begin{itemize}[label={},nosep]
  \item \texttt{float}: Estimated bar value (height for vertical; length for horizontal)
\end{itemize}
}
\vspace{0.5cm}
\hspace*{1.5em}\textcolor{py-doc}{Examples:}
\begin{tcblisting}{
  listing only,
  breakable,
  enhanced jigsaw,
  boxrule=0pt,
  colback=white,
  colframe=white,
  top=0mm, bottom=0mm, left=1mm, right=0.2mm,
  boxsep=0pt,
  before skip=0pt, after skip=0pt,
  listing options={style=pydoc}
}
# Example 1: Vertical bar on left y-axis
image = Image.open("bar_chart.png")
bar = (120, 210, 35, 180)  # (x, y, w, h) from get_bar()
bar_height = compute_bar_height(image, bar, bar_orientation="vertical")
print(bar_height)

# Example 2: Horizontal bar (value from x-axis)
bar = (100, 70, 150, 25)
bar_length = compute_bar_height(image, bar, bar_orientation="horizontal", x_axis_tickers=["200", "400", "600", "800", "1000", "1200", "1400"])
print(bar_length)
\end{tcblisting}
\hspace*{1.5em}\textcolor{py-doc}{"""}
\par
\vspace{0.2cm}

\vspace{0.5em}
\noindent{\parbox{\dimexpr\linewidth-2\fboxsep}{\fontsize{11pt}{12pt}\selectfont\textbf{\underline{Box Plot}}}}
\vspace{1em} 

\textcolor{py-keyword}{\textbf{def}} \textcolor{py-func}{\textbf{get\_boxplot }}\textcolor{black}{(}%
\textcolor{py-arg}{image}\textcolor{black}{:} \textcolor{py-type}{PIL.Image.Image},\,
\textcolor{py-arg}{masks}\textcolor{black}{:} \textcolor{py-type}{list},\,
\textcolor{py-arg}{rgb\_of\_interest}\textcolor{black}{:} \textcolor{py-type}{tuple[int,int,int]},\,
\textcolor{py-arg}{ticker\_label}\textcolor{black}{:} \textcolor{py-type}{str},\,
\textcolor{py-arg}{box\_labels\_of\_interest}\textcolor{black}{:} \textcolor{py-type}{list},\,
\textcolor{py-arg}{boxplot\_orientation}\textcolor{black}{:} \textcolor{py-type}{str},\,
\textcolor{py-arg}{axis\_threshold}\textcolor{black}{:} \textcolor{py-type}{float}%
\textcolor{black}{) $\rightarrow$ \textcolor{py-type}{list}\textcolor{black}{:}}

\medskip
\hspace*{1.5em}\textcolor{py-doc}{"""
Detects and returns boxplot segments filtered by color, axis label, or segmentation indices. Handles both horizontal and vertical boxplot orientations and supports fuzzy matching for axis-aligned labels and approximate color filtering. Commonly used for box plots.}
\par
\vspace{0.2cm}
\hspace*{1.5em}\textcolor{py-doc}{Args:}
{\color{py-doc}
\hspace*{1.5em}\begin{itemize}[label={},nosep]
  \item \texttt{image} (\texttt{PIL.Image.Image}): Input chart image
  \item \texttt{masks} (\texttt{list}): List of segmentation masks
  \item \texttt{rgb\_of\_interest} (\texttt{tuple[int,int,int]} or \texttt{None}): RGB color to filter segments
  \item \texttt{ticker\_label} (\texttt{str} or \texttt{None}): Axis label (e.g., "Tuesday") to filter segments
  \item \texttt{box\_labels\_of\_interest} (\texttt{list} or \texttt{None}): Segmentation mask indices to select
  \item \texttt{boxplot\_orientation} (\texttt{str}): \texttt{"vertical"} or \texttt{"horizontal"} (\texttt{"vertical"} default)
  \item \texttt{axis\_threshold} (\texttt{float}): Fraction of image scanned for axis values (\texttt{0.15} default)
\end{itemize}
}
\vspace{0.5cm}
\hspace*{1.5em}\textcolor{py-doc}{Returns:}
{\color{py-doc}
\hspace*{1.5em}\begin{itemize}[label={},nosep]
  \item \texttt{list[tuple[int,int,int,int]]}: Final filtered \texttt{(x, y, w, h)} segments
\end{itemize}
}
\vspace{0.5cm}
\hspace*{1.5em}\textcolor{py-doc}{Examples:}
\begin{tcblisting}{
  listing only,
  breakable,
  enhanced jigsaw,
  boxrule=0pt,
  colback=white,
  colframe=white,
  top=0mm, bottom=0mm, left=1mm, right=0.2mm,
  boxsep=0pt,
  before skip=0pt, after skip=0pt,
  listing options={style=pydoc}
}
# Example 1: Filter by RGB color (vertical boxplot)
image = Image.open("box_plot.png")
boxplot_of_interest = get_boxplot(image, masks=masks, rgb_of_interest=(106, 184, 209))
print(boxplot_of_interest)

# Example 2: Filter by ticker label (vertical boxplot)
boxplot_of_interest = get_boxplot(image, masks=masks, ticker_label="Tuesday")
print(boxplot_of_interest)

# Example 3: Filter by segmentation indices (horizontal boxplot)
boxplot_of_interest = get_boxplot(image, masks=masks, box_labels_of_interest=[3, 7], boxplot_orientation="horizontal")
print(boxplot_of_interest)
\end{tcblisting}
\hspace*{1.5em}\textcolor{py-doc}{"""}
\par
\vspace{0.2cm}

\textcolor{py-keyword}{\textbf{def}} \textcolor{py-func}{\textbf{compute\_boxplot\_entity }}\textcolor{black}{(}%
\textcolor{py-arg}{image}\textcolor{black}{:} \textcolor{py-type}{PIL.Image.Image},\,
\textcolor{py-arg}{boxplot\_of\_interest}\textcolor{black}{:} \textcolor{py-type}{list[tuple[int,int,int,int]]},\,
\textcolor{py-arg}{boxplot\_orientation}\textcolor{black}{:} \textcolor{py-type}{str},\,
\textcolor{py-arg}{entity\_of\_interest}\textcolor{black}{:} \textcolor{py-type}{str},\,
\textcolor{py-arg}{axis\_threshold}\textcolor{black}{:} \textcolor{py-type}{float},\,
\textcolor{py-arg}{x\_axis\_tickers}\textcolor{black}{:} \textcolor{py-type}{list},\,
\textcolor{py-arg}{y\_axis\_tickers}\textcolor{black}{:} \textcolor{py-type}{list}%
\textcolor{black}{) $\rightarrow$ \textcolor{py-type}{float}\textcolor{black}{:}}

\medskip
\hspace*{1.5em}\textcolor{py-doc}{"""
Computes a statistical entity (e.g., max, min, median, Q1, Q3, range, or interquartile range) of a boxplot by mapping its pixel coordinates to value space using axis localization. Commonly used for box plots.}
\par
\vspace{0.2cm}
\hspace*{1.5em}\textcolor{py-doc}{Args:}
{\color{py-doc}
\hspace*{1.5em}\begin{itemize}[label={},nosep]
  \item \texttt{image} (\texttt{PIL.Image.Image}): Input chart image
  \item \texttt{boxplot\_of\_interest} (\texttt{list[tuple[int,int,int,int]]}): Bounding boxes of the boxplot segments
  \item \texttt{boxplot\_orientation} (\texttt{str}): \texttt{"vertical"} or \texttt{"horizontal"} (\texttt{"vertical"} default)
  \item \texttt{entity\_of\_interest} (\texttt{str}): One of \texttt{"median"}, \texttt{"max"}, \texttt{"min"}, \texttt{"range"}, \texttt{"iqr"}, \texttt{"q1"}, \texttt{"q3"}, \texttt{"q2"} (\texttt{"median"} default)
  \item \texttt{axis\_threshold} (\texttt{float}): Fraction of image scanned for tick labels during axis localization (\texttt{0.15} default)
  \item \texttt{x\_axis\_tickers} (\texttt{list} or \texttt{None}): Optional pre-read x-axis tick labels
  \item \texttt{y\_axis\_tickers} (\texttt{list} or \texttt{None}): Optional pre-read y-axis tick labels
\end{itemize}
}
\vspace{0.5cm}
\hspace*{1.5em}\textcolor{py-doc}{Returns:}
{\color{py-doc}
\hspace*{1.5em}\begin{itemize}[label={},nosep]
  \item \texttt{float}: Computed value of the requested boxplot entity
\end{itemize}
}
\vspace{0.5cm}
\hspace*{1.5em}\textcolor{py-doc}{Examples:}
\begin{tcblisting}{
  listing only,
  breakable,
  enhanced jigsaw,
  boxrule=0pt,
  colback=white,
  colframe=white,
  top=0mm, bottom=0mm, left=1mm, right=0.2mm,
  boxsep=0pt,
  before skip=0pt, after skip=0pt,
  listing options={style=pydoc}
}
# Example 1: Median value of vertical boxplot
image = Image.open("box_plot.png")
boxplot_segments = [(120, 150, 40, 80), (120, 250, 40, 70)] # from get_boxplot()
median_val = compute_boxplot_entity(image, boxplot_segments, entity_of_interest="median")
print(median_val)

# Example 2: Maximum value (Q1) of horizontal boxplot
boxplot_segments = [(100, 70, 120, 30), (250, 70, 140, 30)]
max_val = compute_boxplot_entity(image, boxplot_segments, "horizontal", "max")
print(max_val)

# Example 3: Interquartile range (IQR) of vertical boxplot
boxplot_segments = [(130, 160, 30, 70), (130, 260, 30, 90)]
iqr_val = compute_boxplot_entity(image, boxplot_segments, entity_of_interest="iqr")
print(iqr_val)
\end{tcblisting}
\hspace*{1.5em}\textcolor{py-doc}{"""}
\par
\vspace{0.2cm}

\vspace{0.5em}
\noindent{\parbox{\dimexpr\linewidth-2\fboxsep}{\fontsize{11pt}{12pt}\selectfont\textbf{\underline{Line $\vert$ Area $\vert$ Scatter Plots}}}}
\vspace{1em} 

\textcolor{py-keyword}{\textbf{def}} \textcolor{py-func}{\textbf{get\_edgepoints }}\textcolor{black}{(}%
\textcolor{py-arg}{image}\textcolor{black}{:} \textcolor{py-type}{PIL.Image.Image},\,
\textcolor{py-arg}{masks}\textcolor{black}{:}\textcolor{py-type}{list},\,
\textcolor{py-arg}{rgb\_of\_interest}\textcolor{black}{:} \textcolor{py-type}{tuple[int,int,int]},\,
\textcolor{py-arg}{ticker\_label}\textcolor{black}{:} \textcolor{py-type}{str},\,
\textcolor{py-arg}{mask\_labels\_of\_interest}\textcolor{black}{:} \textcolor{py-type}{list},\,
\textcolor{py-arg}{chart\_orientation}\textcolor{black}{:} \textcolor{py-type}{str},\,
\textcolor{py-arg}{lineplot\_get\_dot}\textcolor{black}{:} \textcolor{py-type}{bool},\,
\textcolor{py-arg}{axis\_threshold}\textcolor{black}{:} \textcolor{py-type}{float}%
\textcolor{black}{) $\rightarrow$ \textcolor{py-type}{list[tuple[int,int]]}\textcolor{black}{:}}

\medskip
\hspace*{1.5em}\textcolor{py-doc}{"""
Computes edge points of a chart segment filtered by color, axis label, or segmentation indices. The edge is determined by scanning perpendicular to the center of the matched label. Supports both vertical and horizontal chart orientations and handles lineplot dots. Useful for identifying segment bounds for downstream value extraction. Commonly used for line, area, and scatter plots.}
\par
\vspace{0.2cm}
\hspace*{1.5em}\textcolor{py-doc}{Args:}
{\color{py-doc}
\hspace*{1.5em}\begin{itemize}[label={},nosep]
  \item \texttt{image} (\texttt{PIL.Image.Image}): Input chart image
  \item \texttt{masks} (\texttt{list} or \texttt{None}): SAM masks with \texttt{bbox} and \texttt{segmentation} fields
  \item \texttt{rgb\_of\_interest} (\texttt{tuple[int,int,int]} or \texttt{None}): Target RGB color for filtering
  \item \texttt{ticker\_label} (\texttt{str} or \texttt{None}): Axis label (e.g., "Q3") for filtering
  \item \texttt{mask\_labels\_of\_interest} (\texttt{list} or \texttt{None}): SAM mask indices to select
  \item \texttt{chart\_orientation} (\texttt{str}): \texttt{"vertical"} or \texttt{"horizontal"} (\texttt{"vertical"} default)
  \item \texttt{lineplot\_get\_dot} (\texttt{bool}): Whether to get edge points for lineplot dots (\texttt{True}) or area chart segments (\texttt{False}) (\texttt{False} default)
  \item \texttt{axis\_threshold} (\texttt{float}): Portion of the image scanned for axis localization (\texttt{0.15} default)
\end{itemize}
}
\vspace{0.5cm}
\hspace*{1.5em}\textcolor{py-doc}{Returns:}
{\color{py-doc}
\hspace*{1.5em}\begin{itemize}[label={},nosep]
  \item \texttt{list[tuple[int,int]]}: Edge points perpendicular to the label center:
    \begin{itemize}
      \item Vertical: \texttt{[(x, top\_y), (x, bottom\_y)]}
      \item Horizontal: \texttt{[(left\_x, y), (right\_x, y)]}
    \end{itemize}
\end{itemize}
}
\vspace{0.5cm}
\hspace*{1.5em}\textcolor{py-doc}{Examples:}
\begin{tcblisting}{
  listing only,
  breakable,
  enhanced jigsaw,
  boxrule=0pt,
  colback=white,
  colframe=white,
  top=0mm, bottom=0mm, left=1mm, right=0.2mm,
  boxsep=0pt,
  before skip=0pt, after skip=0pt,
  listing options={style=pydoc}
}
# Example 1: RGB color + ticker label (vertical area chart)
image = Image.open("area_chart.png")
edge_points = get_edgepoints(image, rgb_of_interest=(106, 184, 209), ticker_label="Q2")
print(edge_points)

# Example 2: SAM mask index + ticker label (vertical area chart)
edge_points = get_edgepoints(image, masks, ticker_label="A", mask_labels_of_interest=[3])
print(edge_points)

# Example 3: RGB color + ticker label (line plot)
image = Image.open("line_plot.png")
line_dots = get_edgepoints(image, rgb_of_interest=(237, 0, 209), ticker_label="Q1", lineplot_get_dot = True)
print(line_dots)
\end{tcblisting}
\hspace*{1.5em}\textcolor{py-doc}{"""}
\par
\vspace{0.2cm}

\vspace{0.5em}
\noindent{\parbox{\dimexpr\linewidth-2\fboxsep}{\fontsize{11pt}{12pt}\selectfont\textbf{\underline{Radial Bar Plot}}}}
\vspace{1em} 

\textcolor{py-keyword}{\textbf{def}} \textcolor{py-func}{\textbf{get\_radial }}\textcolor{black}{(}%
\textcolor{py-arg}{image}\textcolor{black}{:} \textcolor{py-type}{PIL.Image.Image},\,
\textcolor{py-arg}{rgb\_of\_interest}\textcolor{black}{:} \textcolor{py-type}{tuple[int,int,int]},\,
\textcolor{py-arg}{ticker\_label}\textcolor{black}{:} \textcolor{py-type}{str},\,
\textcolor{py-arg}{segmentation\_model}\textcolor{black}{:} \textcolor{py-type}{str}%
\textcolor{black}{) $\rightarrow$ \textcolor{py-type}{tuple[int,int,int,int]}\textcolor{black}{:}}

\medskip
\hspace*{1.5em}\textcolor{py-doc}{"""
Computes the coordinates for the radial bar segment of interest using either color-based filtering or segmentation mask labels. Commonly used for radial bar plots.}
\par
\vspace{0.2cm}
\hspace*{1.5em}\textcolor{py-doc}{Args:}
{\color{py-doc}
\hspace*{1.5em}\begin{itemize}[label={},nosep]
  \item \texttt{image} (\texttt{PIL.Image.Image}): Input chart image
  \item \texttt{rgb\_of\_interest} (\texttt{tuple[int,int,int]} or \texttt{None}): Target RGB color of segment
  \item \textcolor{py-doc}{\texttt{ticker\_label} (\texttt{str} or \texttt{None}): Axis label (e.g., "Q3") for filtering
  \item \texttt{segmentation\_model} (\texttt{str}): Segmentation model to use; \texttt{"color"} for color-based filtering or \texttt{"SAM"} for Segment Anything (\texttt{"color"} default)}
\end{itemize}
}
\vspace{0.5cm}
\hspace*{1.5em}\textcolor{py-doc}{Returns:}
{\color{py-doc}
\hspace*{1.5em}\begin{itemize}[label={},nosep]
  \item \texttt{tuple[int,int,int,int]}: Bounding box \texttt{(x, y, w, h)} representing the segment's radial coordinates
\end{itemize}
}
\vspace{0.5cm}
\hspace*{1.5em}\textcolor{py-doc}{Example:}
\begin{tcblisting}{
  listing only,
  breakable,
  enhanced jigsaw,
  boxrule=0pt,
  colback=white,
  colframe=white,
  top=0mm, bottom=0mm, left=1mm, right=0.2mm,
  boxsep=0pt,
  before skip=0pt, after skip=0pt,
  listing options={style=pydoc}
}
image = Image.open("radial_bar_plot.png")
radial_coords = get_radial(image, rgb_of_interest=(106, 184, 209), ticker_label="Q2")
print(radial_coords)
\end{tcblisting}
\hspace*{1.5em}\textcolor{py-doc}{"""}
\par
\vspace{0.2cm}

\textcolor{py-keyword}{\textbf{def}} \textcolor{py-func}{\textbf{analyze\_radial\_geometry }}\textcolor{black}{(}%
\textcolor{py-arg}{image}\textcolor{black}{:} \textcolor{py-type}{PIL.Image.Image},\,
\textcolor{py-arg}{contour\_of\_interest}\textcolor{black}{:} \textcolor{py-type}{np.ndarray}%
\textcolor{black}{) $\rightarrow$ \textcolor{py-type}{tuple[PIL.Image.Image, int, int, float, float]}\textcolor{black}{:}}

\medskip
\hspace*{1.5em}\textcolor{py-doc}{"""
Estimates the radial geometry of a radial bar chart for the segment of interest. Identifies the chart center, detects the outer circle representing the maximum value, and computes the maximum radial extent (i.e., radius) of the contour of interest. Commonly used for radial bar plots.
}
\par
\vspace{0.2cm}
\hspace*{1.5em}\textcolor{py-doc}{Args:}
{\color{py-doc}
\hspace*{1.5em}\begin{itemize}[label={},nosep]
  \item \texttt{image} (\texttt{PIL.Image.Image}): Input chart image
  \item \texttt{contour\_of\_interest} (\texttt{np.ndarray}): Contour representing the segment of interest
\end{itemize}
}
\vspace{0.5cm}
\hspace*{1.5em}\textcolor{py-doc}{Returns:}
{\color{py-doc}
\hspace*{1.5em}\begin{itemize}[label={},nosep]
  \item \texttt{PIL.Image.Image}: Image with detected outer circle and chart center marked
  \item \texttt{int}: X-coordinate of the circle's center
  \item \texttt{int}: Y-coordinate of the circle's center
  \item \texttt{float}: Outer circle radius (\texttt{r\_outer})
  \item \texttt{float}: Maximum radius from center to the contour (\texttt{r\_max})
\end{itemize}
}
\vspace{0.5cm}
\hspace*{1.5em}\textcolor{py-doc}{Example:}
\begin{tcblisting}{
  listing only,
  breakable,
  enhanced jigsaw,
  boxrule=0pt,
  colback=white,
  colframe=white,
  top=0mm, bottom=0mm, left=1mm, right=0.2mm,
  boxsep=0pt,
  before skip=0pt, after skip=0pt,
  listing options={style=pydoc}
}
image_radial_geometry, center_x, center_y, r_outer, r_max = analyze_radial_geometry(image, contour_of_interest=contour) #contour from get_radial()
display(image_radial_geometry)
print("Center coordinates:", center_x, center_y, "Outer circle radius:", r_outer, "Max radius:", r_max)
\end{tcblisting}
\hspace*{1.5em}\textcolor{py-doc}{"""}
\par
\vspace{0.2cm}

\textcolor{py-keyword}{\textbf{def}} \textcolor{py-func}{\textbf{estimate\_radial\_value }}\textcolor{black}{(}%
\textcolor{py-arg}{image}\textcolor{black}{:} \textcolor{py-type}{PIL.Image.Image},\,
\textcolor{py-arg}{center\_x}\textcolor{black}{:} \textcolor{py-type}{int},\,
\textcolor{py-arg}{center\_y}\textcolor{black}{:} \textcolor{py-type}{int},\,
\textcolor{py-arg}{r\_outer}\textcolor{black}{:} \textcolor{py-type}{int},\,
\textcolor{py-arg}{r\_max}\textcolor{black}{:} \textcolor{py-type}{int},\,
\textcolor{py-arg}{reference\_circle\_value}\textcolor{black}{:} \textcolor{py-type}{float}%
\textcolor{black}{) $\rightarrow$ \textcolor{py-type}{float}\textcolor{black}{:}}

\medskip
\hspace*{1.5em}\textcolor{py-doc}{"""
Estimates the value of a radial segment in a radial bar chart by scaling its radial length relative to the outermost circle. The reference value for the outer circle is provided externally (e.g., by an LLM), with a default of 100.
}
\par
\vspace{0.2cm}
\hspace*{1.5em}\textcolor{py-doc}{Args:}
{\color{py-doc}
\hspace*{1.5em}\begin{itemize}[label={},nosep]
  \item \texttt{image} (\texttt{PIL.Image.Image}): Input chart image
  \item \texttt{center\_x} (\texttt{int}): X-coordinate of the circle center
  \item \texttt{center\_y} (\texttt{int}): Y-coordinate of the circle center
  \item \texttt{r\_outer} (\texttt{int}): Radius of the outer circle
  \item \texttt{r\_max} (\texttt{int}): Maximum radius from the center to the contour
  \item \texttt{reference\_circle\_value} (\texttt{float}): Value corresponding to the outer reference circle (default: \texttt{100})
\end{itemize}
}
\vspace{0.5cm}
\hspace*{1.5em}\textcolor{py-doc}{Returns:}
{\color{py-doc}
\hspace*{1.5em}\begin{itemize}[label={},nosep]
  \item \texttt{float}: Estimated value of the radial segment
\end{itemize}
}
\vspace{0.5cm}
\hspace*{1.5em}\textcolor{py-doc}{Example:}
\begin{tcblisting}{
  listing only,
  breakable,
  enhanced jigsaw,
  boxrule=0pt,
  colback=white,
  colframe=white,
  top=0mm, bottom=0mm, left=1mm, right=0.2mm,
  boxsep=0pt,
  before skip=0pt, after skip=0pt,
  listing options={style=pydoc}
}
radial_value = estimate_radial_value(image, center_x=250, center_y=250, r_outer=200, r_max=150, reference_circle_value=100) #center_x, center_y, r_outer, r_max from analyze_radial_geometry()
print("Estimated value:", radial_value)
\end{tcblisting}
\hspace*{1.5em}\textcolor{py-doc}{"""}
\par
\vspace{0.2cm}

\verb|```|
\end{tcolorbox}

\subsubsection{Chart Metadata Extraction}\label{appendix:chart-metadata-prompt}
The metadata extraction prompt guides the agent to identify essential chart components, such as chart type, axis ranges, and legend entries. This metadata is then used to retrieve and condition the appropriate ICL examples, and to parameterize subsequent tool calls.
\begin{tcolorbox}[
  title= Chart Metadata Extraction Prompt,
  colback=white,
  colframe=black,
  fonttitle=\bfseries,
  sharp corners=south,
  boxrule=0.6pt,
  breakable,
  enhanced,
  fontupper=\small
]

\begin{tcolorbox}[
  enhanced,
  colback=blue!5,
  colframe=blue!50,
  boxrule=0pt,
  sharp corners=south,
  top=1mm, bottom=1mm, left=2mm, right=2mm
]
\textbf{Instruction:} \\
You are a vision-language model tasked with analyzing a data visualization chart image.

Extract and return the following information as a JSON dictionary using the exact keys specified below.
\begin{itemize}[leftmargin=1.5em, noitemsep, topsep=0.5em]
    \item \texttt{chart\_type}: e.g., pie chart, multi-ring pie chart, bar chart, line chart, box plot, etc.
    \item \texttt{title}: Exact chart title as shown.
    \item \texttt{legend}: List or dictionary of all legend entries.
    \item \texttt{highlevel\_legend\_categories} and \texttt{finegrained\_legend\_subcategories}: If the chart shows category hierarchy, list both, even if names overlap.
    \item \texttt{legend\_embedded}: \texttt{true} if legend is within the chart; \texttt{false} if outside.
    \item \texttt{x axis}/ \texttt{y axis}/ \texttt{right-y axis}/ \texttt{color-bar} labels: Axis labels (strings). May be empty.
    \item \texttt{x axis}/ \texttt{y axis}/ \texttt{right-y axis}/ \texttt{color-bar}/ \texttt{radial axis} ticker values: Tick values (List). May be empty.
    \item \texttt{annotation\_type}: Either \texttt{"annotated"} or \texttt{"unannotated"}.
    \begin{itemize}[leftmargin=1.5em]
        \item \texttt{"annotated"} – if numeric values are written directly in the chart.
        \item \texttt{"unannotated"} – if such values are not shown in the chart.
    \end{itemize}
    \item \texttt{visual\_description}: Concise summary of the chart’s visual structure.
\end{itemize}

Only output the JSON object.
\end{tcolorbox}


\begin{tcolorbox}[
  enhanced,
  colback=green!5,
  colframe=green!50,
  boxrule=0pt,
  sharp corners=south,
  top=1mm, bottom=1mm, left=2mm, right=2mm
]
\textbf{Input:} \\
\begin{center}
\includegraphics[scale = 0.35]{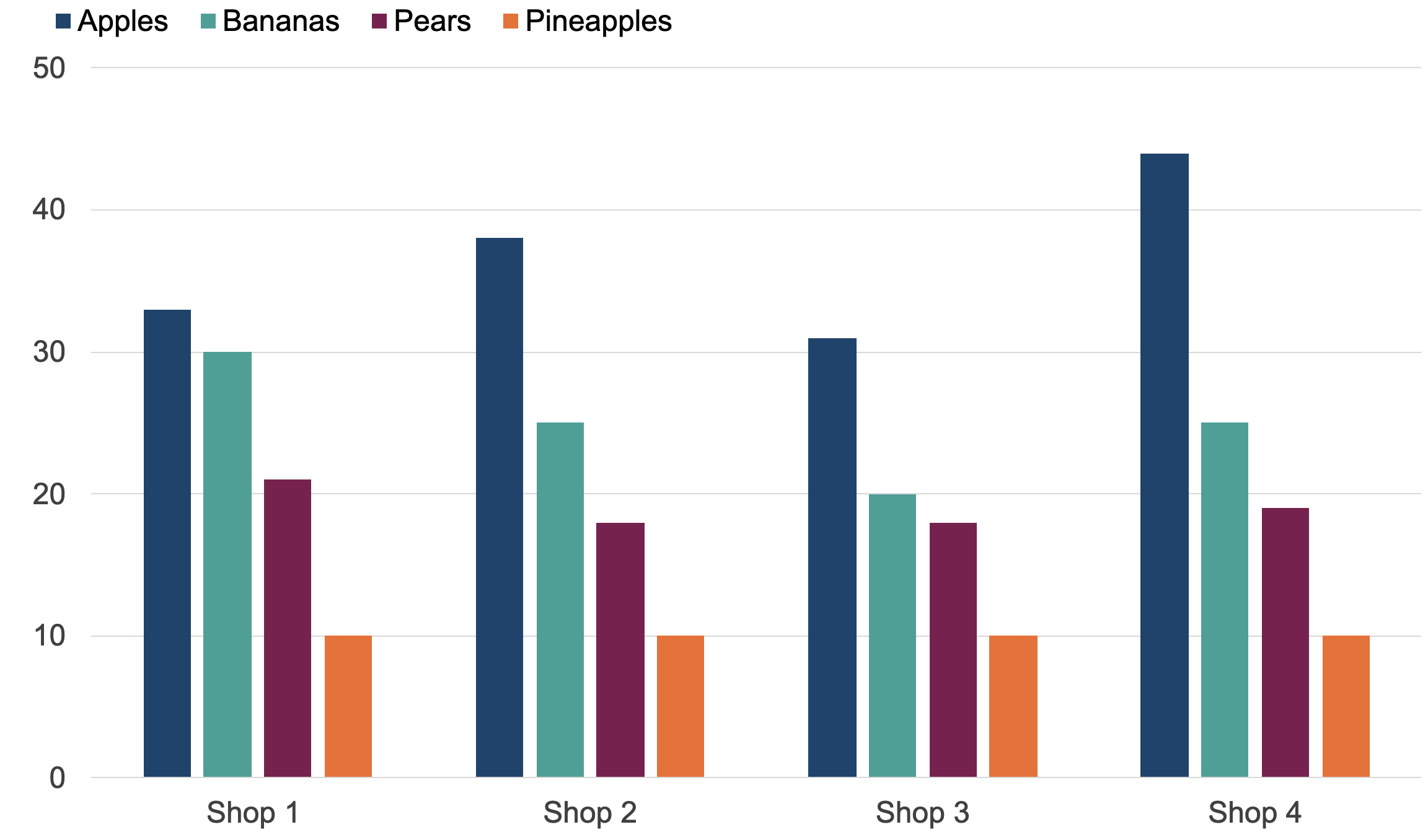}
\end{center}
\begin{center}
\textit{Chart image corresponding to the current sample.}
\end{center}
\end{tcolorbox}
\end{tcolorbox}
\vspace{0.5em}

\subsubsection{In Context Learning}\label{appendix:chartagent-ICL-prompt}
We provide ICL examples corresponding to each major chart type. At inference time, only the examples matching the detected chart type are retrieved and used. For instance, if a chart is classified as a pie chart during the metadata extraction stage, only pie chart ICL examples are appended to the prompt. If no ICL examples exist for the detected chart type, then no ICL is added.

\begin{tcolorbox}[
  title=ICL Example Prompt,
  colback=white,
  colframe=black,
  fonttitle=\bfseries,
  sharp corners=south,
  boxrule=0.6pt,
  breakable,
  enhanced,
  fontupper=\small
]

Below is an example of how to use the tools to solve a user query with a vertical grouped bar chart.

\begin{center}
\noindent\colorbox{black!5}{\parbox{\dimexpr\linewidth-2\fboxsep}{
\textbf{[MORE INSTRUCTIONS ...]}  
The actual ICL prompt contains additional detailed guidelines, including self-verification, recovery strategies, changing tool parameters, using alternative tool options in case of incorrect trajectories, and fallback options to the base MLLM.
}}
\end{center}
\vspace{0.6em}
\par

{\color{red!70!black} 
\texttt{\# USER REQUEST \#:} \texttt{<An image here>}  
According to this chart, what is the GDP growth rate for India in 2020?  

\texttt{\# USER IMAGE} stored in \texttt{chart\_image}, as a PIL image.
}



\vspace{0.8em}


\thought{0}{First, I will display and read the image.}  

\action{0}
\begin{tcblisting}{
  listing only,
  breakable,
  enhanced jigsaw,
  boxrule=0pt,
  colback=white,
  colframe=white,
  top=0mm, bottom=0mm, left=1mm, right=0.2mm,
  boxsep=0pt,
  before skip=0pt, after skip=0pt,
  listing options={style=pydoc}
}
```python
display(image=chart_image)
```
\end{tcblisting}

\observation{Execution success. The chart image is displayed: \texttt{<image displayed here>}}

---

\thought{1}{Next, I will read the chart components such as title, legend, and axes.}

\action{1}{}
\begin{tcblisting}{
  listing only,
  breakable,
  enhanced jigsaw,
  boxrule=0pt,
  colback=white,
  colframe=white,
  top=0mm, bottom=0mm, left=1mm, right=0.2mm,
  boxsep=0pt,
  before skip=0pt, after skip=0pt,
  listing options={style=pydoc}
}
```python
title_text = "Annual GDP Growth Rate"
legend_texts = {"Singapore", "India", "United States", "United Kingdom", "China"}
x_axis_tickers = ["2016", "2017", "2018", "2019", "2020"]
y_axis_tickers = ["0", "1.50", "3.0", "4.50", "6.0", "7.50", "9.0", "10.50"]
x_axis_title = "Year"
y_axis_title = "GDP Growth Rate (%)"
print(title_text, legend_texts)
```
\end{tcblisting}

\observation{Execution success. Title, legend, and axes are recorded.}  

---

\thought{2}{Now, I should annotate the legend in the chart image so that I can correctly identify the marker corresponding to the category of interest "India".} 

\action{2}{}
\begin{tcblisting}{
  listing only,
  breakable,
  enhanced jigsaw,
  boxrule=0pt,
  colback=white,
  colframe=white,
  top=0mm, bottom=0mm, left=1mm, right=0.2mm,
  boxsep=0pt,
  before skip=0pt, after skip=0pt,
  listing options={style=pydoc}
}
```python
legend_image, labeled_legend, bbox_mapping = annotate_legend(chart_image, legend_texts)
display(labeled_legend)
print (bbox_mapping)
```
\end{tcblisting}

\observation{Execution success. The output is a labeled image of legend elements and their bounding boxes. Labeled legend image is displayed: \texttt{<image displayed here>}}  

---

\thought{3}{I will now use the \texttt{labeled\_legend} to determine the bounding box label corresponding to the legend marker of interest --- \textit{India}. 
Markers are small colored circles or squares placed to the left of their respective labels.  
The \textit{India} marker is red, with the number 5 above it, so I assign 5 as the \texttt{label\_of\_interest}.}

\action{3}{}

\begin{tcblisting}{
  listing only,
  breakable,
  enhanced jigsaw,
  boxrule=0pt,
  colback=white,
  colframe=white,
  top=0mm, bottom=0mm, left=1mm, right=0.2mm,
  boxsep=0pt,
  before skip=0pt, after skip=0pt,
  listing options={style=pydoc}
}
```python
label_of_interest = 5
print(label_of_interest)
```
\end{tcblisting}

\observation{Execution success. Output: \texttt{label\_of\_interest = 5}}

---

\thought{4}{I will now extract the RGB color of the \textit{India} legend marker using its bounding box label.}

\action{4}{}

\begin{tcblisting}{
  listing only,
  breakable,
  enhanced jigsaw,
  boxrule=0pt,
  colback=white,
  colframe=white,
  top=0mm, bottom=0mm, left=1mm, right=0.2mm,
  boxsep=0pt,
  before skip=0pt, after skip=0pt,
  listing options={style=pydoc}
}
```python
rgb_of_interest = get_marker_rgb(legend_image, bbox_mapping, label_of_interest=5)
print ("Detected RGB color:", rgb_of_interest)
```
\end{tcblisting}

\observation{Execution success. Output: \texttt{Detected RGB color: (145, 20, 45)}} 

---

\thought{5}{I will now clean the chart image by removing both the title and the legend to avoid interference with bar identification using segmentation and color matching. 
This ensures that extraneous elements do not affect contour detection or segmentation. 
If the legend is too close to the x- or y-axis, I will not crop the legend to avoid removing parts of the axis; in that case, only the title will be removed.}

\action{5}{}

\begin{tcblisting}{
  listing only,
  breakable,
  enhanced jigsaw,
  boxrule=0pt,
  colback=white,
  colframe=white,
  top=0mm, bottom=0mm, left=1mm, right=0.2mm,
  boxsep=0pt,
  before skip=0pt, after skip=0pt,
  listing options={style=pydoc}
}
```python
cleaned_image = clean_chart_image(chart_image, title = title_text, legend=legend_texts)
display(cleaned_image)
```
\end{tcblisting}

\observation{Execution success. Cleaned image displayed without title and legend: \texttt{<image displayed here>}}

---

\thought{6}{I will now detect the bar of interest in the \texttt{cleaned\_image} corresponding to \textit{India} (marker color = \texttt{rgb\_of\_interest}) and the year \textit{2020} (\texttt{ticker\_label}).  
The bar orientation is vertical.}

\action{6}{}

\begin{tcblisting}{
  listing only,
  breakable,
  enhanced jigsaw,
  boxrule=0pt,
  colback=white,
  colframe=white,
  top=0mm, bottom=0mm, left=1mm, right=0.2mm,
  boxsep=0pt,
  before skip=0pt, after skip=0pt,
  listing options={style=pydoc}
}
```python
bar_of_interest = get_bar(cleaned_image, rgb_of_interest=rgb_of_interest, ticker_label="2020", bar_orientation="vertical")
print(bar_of_interest)
```
\end{tcblisting}

\observation{Execution success. Output: \texttt{bar\_of\_interest = (120, 210, 35, 180)}} 

---

\thought{7}{Now, I will compute the value height of the detected bar (\texttt{bar\_of\_interest}) by mapping its pixel height to the axis scale.  
Since this is a vertical bar chart, I will apply a threshold of 0.2 to automatically detect the axis region in the left 20\% of the image.  
I will then use the \texttt{x\_axis\_tickers} and \texttt{y\_axis\_tickers} (already extracted in previous steps) to perform the mapping accurately.}

\action{7}{}

\begin{tcblisting}{
  listing only,
  breakable,
  enhanced jigsaw,
  boxrule=0pt,
  colback=white,
  colframe=white,
  top=0mm, bottom=0mm, left=1mm, right=0.2mm,
  boxsep=0pt,
  before skip=0pt, after skip=0pt,
  listing options={style=pydoc}
}
value_height = compute_bar_height(cleaned_image, bar_of_interest, bar_orientation="vertical", axis_threshold=0.2, x_axis_tickers=x_axis_tickers, y_axis_tickers=y_axis_tickers, x_axis_title=x_axis_title, y_axis_title=y_axis_title)
print("Bar value height:", value_height)
\end{tcblisting}

\observation{Execution success. Output: \texttt{Bar value height: 4.2}}  

---

{\color{green!50!black} 
\answer{The GDP growth rate for India in 2020 is 4.2. \terminate{}}
}

---

\end{tcolorbox}

\subsection{Baseline Prompts}\label{appendix:baseline_prompts}
To benchmark \chartagent{}, we compare against several baseline prompting strategies. 
We apply zero-shot~(\ref{appendix:baseline-zeroshot-prompt}) and chain-of-thought (CoT)~(\ref{appendix:baseline-cot-prompt}) prompts across all proprietary and open-weight MLLM baselines. 
In addition, we include a ReAct prompt~(\ref{appendix:baseline-react-ablation-prompt}) for ablation studies, comparing \chartagent{} with a ReAct-style agent to isolate the effect of chart-specialized visual tools.
Finally, we use a tabular question-answering prompt~(\ref{appendix:baseline-table-to-vqa-conversion}) for a few chart-based baselines that output structured tables instead of direct answers.

\subsubsection{Zero-shot}\label{appendix:baseline-zeroshot-prompt}
The zero-shot prompt provides only minimal task instructions, requiring the model to answer directly from the chart without intermediate reasoning or tool use.
\begin{tcolorbox}[
  title=Zero-shot Prompt,
  colback=white,
  colframe=black,
  fonttitle=\bfseries,
  sharp corners=south,
  boxrule=0.6pt,
  breakable,
  enhanced,
  fontupper=\small
]

\begin{tcolorbox}[
  enhanced,
  colback=blue!5,
  colframe=blue!50,
  boxrule=0pt,
  sharp corners=south,
  top=1mm, bottom=1mm, left=2mm, right=2mm
]
\textbf{Instruction:} \\[0.3em]
You are a data analyst skilled at analyzing chart data. Carefully examine the chart and answer the User’s question with a single word or short phrase.
\end{tcolorbox}

\begin{tcolorbox}[
  enhanced,
  colback=green!5,
  colframe=green!50,
  boxrule=0pt,
  sharp corners=south,
  top=1mm, bottom=1mm, left=2mm, right=2mm
]
\textbf{Input Format:}
\begin{verbatim}
<chart image> {entry["image"]}
<question> {entry["query"]}
\end{verbatim}
\end{tcolorbox}

\end{tcolorbox}

\subsubsection{Chain-of-Thought}\label{appendix:baseline-cot-prompt}
The chain-of-thought (CoT) prompt encourages the model to reason step by step before providing its final answer, resulting in more structured and coherent reasoning compared to zero-shot prompting.
\begin{tcolorbox}[
  title=Chain-of-Thought (CoT) Prompt,
  colback=white,
  colframe=black,
  fonttitle=\bfseries,
  sharp corners=south,
  boxrule=0.6pt,
  breakable,
  enhanced,
  fontupper=\small
]

\begin{tcolorbox}[
  enhanced,
  colback=blue!5,
  colframe=blue!50,
  boxrule=0pt,
  sharp corners=south,
  top=1mm, bottom=1mm, left=2mm, right=2mm
]
\textbf{Instruction:} \\[0.3em]
You are a data analyst skilled at analyzing chart data. Analyze the user’s chart, carefully examine it, think step by step, and answer the user’s question. Provide your final answer in the exact format: `My final answer is \{answer here\}'.
\end{tcolorbox}

\begin{tcolorbox}[
  enhanced,
  colback=green!5,
  colframe=green!50,
  boxrule=0pt,
  sharp corners=south,
  top=1mm, bottom=1mm, left=2mm, right=2mm
]
\textbf{Input Format:}
\begin{verbatim}
<chart image> {entry["image"]}
<question> {entry["query"]}
\end{verbatim}
\end{tcolorbox}

\end{tcolorbox}

\subsubsection{ReAct}\label{appendix:baseline-react-ablation-prompt}
The ReAct prompt~\cite{yao2023react} combines reasoning traces with action steps, allowing the model to interleave thought, tool/code invocation, and observations until a final answer is reached. We use this prompt in our ablation studies to isolate the contribution of chart-specialized visual tools in our framework.
\begin{tcolorbox}[
  title=ReAct Prompt,
  colback=white,
  colframe=black,
  fonttitle=\bfseries,
  sharp corners=south,
  boxrule=0.6pt,
  breakable,
  enhanced,
  fontupper=\small
]

\begin{tcolorbox}[
  enhanced,
  colback=blue!5,
  colframe=blue!50,
  boxrule=0pt,
  sharp corners=south,
  top=1mm, bottom=1mm, left=2mm, right=2mm
]
\textbf{Instruction:} \\[0.3em]
You are a data analyst skilled at analyzing chart data. Your task is to analyze the provided chart and answer the user’s question. Carefully examine the chart, reason step by step, and invoke actions (e.g., tool calls or code) when helpful. 
Follow this exact format:
\begin{verbatim}
Thought: ...
Action: ...
Observation: ...
(repeat Thought/Action/Observation as needed)

Final Answer: ...
\end{verbatim}
If no action is needed, go directly to Final Answer.

\end{tcolorbox}

\begin{tcolorbox}[
  enhanced,
  colback=green!5,
  colframe=green!50,
  boxrule=0pt,
  sharp corners=south,
  top=1mm, bottom=1mm, left=2mm, right=2mm
]
\textbf{Input Format:}
\begin{verbatim}
<chart image> {entry["image"]}
<question> {entry["query"]}
\end{verbatim}
\end{tcolorbox}

\end{tcolorbox}

\subsubsection{Tabular Question-Answering}\label{appendix:baseline-table-to-vqa-conversion}
For a few chart-based baselines that output structured tables rather than direct answers, we apply a tabular question-answering prompt. This prompt instructs the GPT-4o model to use the extracted table together with the user’s question to produce a concise answer.
\begin{tcolorbox}[
  title=Tabular Question-Answering Prompt,
  colback=white,
  colframe=black,
  fonttitle=\bfseries,
  sharp corners=south,
  boxrule=0.6pt,
  breakable,
  enhanced,
  fontupper=\small
]

\begin{tcolorbox}[
  enhanced,
  colback=blue!5,
  colframe=blue!50,
  boxrule=0pt,
  sharp corners=south,
  top=1mm, bottom=1mm, left=2mm, right=2mm
]
\textbf{Instruction:} \\
Given the data table and the user’s question, use the table to determine and provide the answer in a single word or short phrase.
\end{tcolorbox}


\begin{tcolorbox}[
  enhanced,
  colback=green!5,
  colframe=green!50,
  boxrule=0pt,
  sharp corners=south,
  top=1mm, bottom=1mm, left=2mm, right=2mm
]
\textbf{Input Format:}
\begin{verbatim}
<extracted_table> {entry["extracted_table"]}
<user_question> {entry["query]}
\end{verbatim}
\end{tcolorbox}

\end{tcolorbox}

\subsection{Evaluation Prompts}\label{appendix:prompts_evaluation_metrics}
Recall that we evaluate model predictions using two strategies: (1) a standardization-based accuracy computation, and (2) a GPT-Accuracy metric based on the LLM-as-a-Judge paradigm. The first method uses GPT-4o to standardize responses before applying an arithmetic-based correctness check, with a strict 5\% relative error tolerance for numeric responses and string matching for non-numeric ones. The second method prompts an LLM to assess correctness directly, also applying a 5\% tolerance for numeric responses. The prompts used for both evaluation strategies are provided in~\ref{appendix:prompts_accuracy} and~\ref{appendix:prompts_gpt_accuracy}, respectively.

\subsubsection{Accuracy}\label{appendix:prompts_accuracy}
The following prompt is used to standardize both the ground truth and predicted responses before performing the accuracy check. GPT-4o is instructed to remove units (e.g., “K” for thousand, “M” for million, “B” for billion), convert scales, eliminate symbols, and standardize number formats. Once standardized, numeric responses are evaluated arithmetically using a strict 5\% relative error tolerance, while non-numeric responses require string match.
\begin{tcolorbox}[
  title=Prompt for Standardizing Ground Truth and Predicted Responses,
  colback=white,
  colframe=black,
  fonttitle=\bfseries,
  sharp corners=south,
  boxrule=0.6pt,
  breakable,
  enhanced,
  fontupper=\small
]

\begin{tcolorbox}[
  enhanced,
  colback=blue!5,
  colframe=blue!50,
  boxrule=0pt,
  sharp corners=south,
  top=1mm, bottom=1mm, left=2mm, right=2mm
]
\textbf{Instruction:} \\
You are given a question, a ground truth answer, and a model's predicted answer. Your task is to determine whether the prediction is correct.

\vspace{1mm}
\textbf{Follow these steps exactly:}
\begin{enumerate}[noitemsep, topsep=0.5em]
    \item If both answers are \textbf{numeric}, first extract the numeric portion of each value.
    \begin{itemize}[noitemsep, topsep=0.5em]
        \item Normalize both answers carefully by checking both the \texttt{<groundtruth answer>} and \texttt{<predicted answer>} values in context.
        \item If both values include the \textbf{same unit} (e.g., \texttt{K}, \texttt{M}, \texttt{B}, or their full-word equivalents like \texttt{thousand}, \texttt{million}, \texttt{billion}), \textbf{do not scale}. Simply strip the unit and compare the base numbers.
        \item Example: \texttt{20K} vs \texttt{21K} $\rightarrow$ compare \texttt{20} vs \texttt{21}.
        \item If one value has a unit and the other is already scaled (e.g., \texttt{21K} vs \texttt{21000}), convert the unit-based value to its full numeric form before comparison.
        \item Example: \texttt{21K} vs \texttt{21000} $\rightarrow$ compare \texttt{21000} vs \texttt{21000}.
        \item If only one value includes a unit and the other is \textbf{not clearly scaled}, strip the unit and compare only the numeric parts (without scaling either).
        \item Examples:
        \begin{itemize}
            \item \texttt{16M} vs \texttt{16} $\rightarrow$ compare \texttt{16} vs \texttt{16}
            \item \texttt{500M} vs \texttt{500} $\rightarrow$ compare \texttt{500} vs \texttt{500}
        \end{itemize}
        \item Apply scaling \textbf{only when the context requires it}, based on whether the values represent scaled vs. unscaled forms of the same quantity.
        \item Additional examples:
        \begin{itemize}
            \item \texttt{20K} vs \texttt{21K} $\rightarrow$ compare \texttt{20} vs \texttt{21}
            \item \texttt{20K} vs \texttt{21000} $\rightarrow$ compare \texttt{20000} vs \texttt{21000}
            \item \texttt{16M} vs \texttt{16} $\rightarrow$ compare \texttt{16} vs \texttt{16}
            \item \texttt{500} vs \texttt{500M} $\rightarrow$ compare \texttt{500} vs \texttt{500}
            \item \texttt{142.6} vs \texttt{1,350 million} $\rightarrow$ compare \texttt{142.6} vs \texttt{1350}
            \item \texttt{170.0} vs \texttt{160 million} $\rightarrow$ compare \texttt{170} vs \texttt{160}
            \item \texttt{20B} vs \texttt{21} $\rightarrow$ compare \texttt{20} vs \texttt{21}
        \end{itemize}
        \item Convert number words (e.g., \texttt{ten}, \texttt{eleven}, \texttt{forty-two}) to digits.
        \item Remove commas, currency symbols, and surrounding text.
        \item Ignore case when processing text.
    \end{itemize}

    \item If the answers are not numeric, return the string response.
\end{enumerate}

\vspace{1mm}
\textbf{Return a JSON object in the following format:}
\begin{verbatim}
{
  "ground_truth_filtered": <normalized ground truth value>,
  "response_filtered": <normalized predicted value>
}
\end{verbatim}
\end{tcolorbox}


\begin{tcolorbox}[
  enhanced,
  colback=green!5,
  colframe=green!50,
  boxrule=0pt,
  sharp corners=south,
  top=1mm, bottom=1mm, left=2mm, right=2mm
]
\textbf{Input Format:}
\begin{verbatim}
<question> {entry["query"]}
<groundtruth answer> {entry["ground_truth"]}
<response> {original_response}
\end{verbatim}
\end{tcolorbox}

\end{tcolorbox}

\subsubsection{LLM-as-a-Judge}\label{appendix:prompts_gpt_accuracy}
The following prompt is used to evaluate response correctness using the LLM-as-a-Judge baseline, also referred to as \textit{GPT-Accuracy} in prior literature~\cite{xu2023chartbench, masry-etal-2022-chartqa, xia2024chartx}. The LLM (GPT-4o) is shown the question, ground truth, and model prediction, and is asked to assess whether the prediction is correct, with a 5\% error tolerance applied to numeric answers. While flexible, this method may be imprecise for fine-grained numeric evaluation, as discussed in Sections~\ref{experiments:evaluation} and~\ref{appendix:acc_vs_llm_as_judge}.
\begin{tcolorbox}[
  title=LLM-as-a-Judge Prompt for Evaluating Response Correctness,
  colback=white,
  colframe=black,
  fonttitle=\bfseries,
  sharp corners=south,
  boxrule=0.6pt,
  breakable,
  enhanced,
  fontupper=\small
]

\begin{tcolorbox}[
  enhanced,
  colback=blue!5,
  colframe=blue!50,
  boxrule=0pt,
  sharp corners=south,
  top=1mm, bottom=1mm, left=2mm, right=2mm
]
\textbf{Instruction:} \\
Given multiple QA pairs and the corresponding predictions, evaluate the correctness of each prediction.  
Return only a single word: \texttt{"True"} or \texttt{"False"}.

If the ground truth answer is a numeric value (with or without units), allow a \textbf{5\% error tolerance} when comparing against the prediction.
\end{tcolorbox}


\begin{tcolorbox}[
  enhanced,
  colback=green!5,
  colframe=green!50,
  boxrule=0pt,
  sharp corners=south,
  top=1mm, bottom=1mm, left=2mm, right=2mm
]
\textbf{Input Format:}
\begin{verbatim}
<question> {entry["query"]}
<groundtruth answer> {entry["ground_truth"]}
<response> {original_response}
\end{verbatim}
\end{tcolorbox}

\end{tcolorbox}

\subsection{Complexity Analysis Prompts}\label{appendix:prompts_complexity_analysis}
Each chart–question pair in our dataset is annotated with two types of complexity labels: visual complexity and reasoning complexity. The prompts used to generate these labels are shown in~\ref{appendix:prompts_visual_complexity} and~\ref{appendix:prompts_reasoning_complexity}, respectively.

\subsubsection{Visual Complexity}\label{appendix:prompts_visual_complexity}
The following prompt categorizes charts by visual complexity—Easy, Medium, or Hard—based solely on 
the visual effort needed to interpret the information presented in the chart image.
\begin{tcolorbox}[
  title=Visual Complexity Rating Prompt,
  colback=white,
  colframe=black,
  fonttitle=\bfseries,
  sharp corners=south,
  boxrule=0.6pt,
  breakable,
  enhanced,
  fontupper=\small
]

\begin{tcolorbox}[
  enhanced,
  colback=blue!5,
  colframe=blue!50,
  boxrule=0pt,
  sharp corners=south,
  top=1mm, bottom=1mm, left=2mm, right=2mm
]
\textbf{Instruction:} \\
You are a data analyst, skilled at visually interpreting charts. You are shown only a \textbf{chart image}, with no additional context or question. Your task is to assess the \textbf{visual complexity} of the chart based solely on what you see.

\vspace{1mm}
\textbf{Think step by step:}
\begin{itemize}[noitemsep, topsep=0.5em]
    \item Count the number of visible elements (e.g., data points, lines, bars, rings, legends, labels, colors, gridlines).
    \item Assess how cluttered or clean the layout appears.
    \item Judge whether understanding the chart requires visual comparisons, dense reading, or reasoning over multiple components.
    \item Examine the clarity of legends, axis ticks, text placement, and grouping.
\end{itemize}

\vspace{1mm}
\textbf{Charts tend to be visually complex if they include:}
\begin{itemize}[noitemsep, topsep=0.5em]
    \item Multiple nested or layered elements
    \item 3D perspectives or overlapping dimensions
    \item Low contrast or overlapping visual elements
    \item Radar/polar charts with filled or intersecting shapes
    \item Multi-axis designs (e.g., dual Y-axes)
    \item Overlaid plot types (e.g., bars + lines)
    \item Dense scatter plots with tightly packed points
    \item Stacked formats requiring segment comparison
    \item Ambiguous or visually similar elements that are hard to distinguish
\end{itemize}

When in doubt — especially in the presence of distortion, depth, layering, or ambiguity — prefer labeling as \textbf{Hard}.

\vspace{1mm}
\textbf{Respond in the following JSON format:}
\begin{verbatim}
{
  "label": "Easy / Medium / Hard",
  "reasoning": "Step-by-step explanation of how you reached your conclusion."
}
\end{verbatim}
\end{tcolorbox}


\begin{tcolorbox}[
  enhanced,
  colback=green!5,
  colframe=green!50,
  boxrule=0pt,
  sharp corners=south,
  top=1mm, bottom=1mm, left=2mm, right=2mm
]
\textbf{Input Format:}
\begin{verbatim}
<chart image> {entry["image"]}
\end{verbatim}
\end{tcolorbox}

\end{tcolorbox}

\subsubsection{Reasoning Complexity}\label{appendix:prompts_reasoning_complexity}
The following prompt categorizes chart–question pairs by reasoning complexity—Easy, Medium, or Hard—based solely on the level of reasoning needed to interpret and answer the question using the chart image.
\begin{tcolorbox}[
  title=Reasoning Complexity Rating Prompt,
  colback=white,
  colframe=black,
  fonttitle=\bfseries,
  sharp corners=south,
  boxrule=0.6pt,
  breakable,
  enhanced,
  fontupper=\small
]

\begin{tcolorbox}[
  enhanced,
  colback=blue!5,
  colframe=blue!50,
  boxrule=0pt,
  sharp corners=south,
  top=1mm, bottom=1mm, left=2mm, right=2mm
]
\textbf{Instruction:} \\
You are a data analyst, skilled at solving visual questions over chart images. You are shown a chart image and a corresponding question. Your task is to assess the \textbf{reasoning complexity} required to answer the question correctly.

\vspace{3mm}
\textbf{Think step by step:}
\begin{enumerate}[noitemsep, topsep=0.5em]
    \item Identify the key visual elements referenced by the question.
    \item Determine the number of distinct reasoning steps needed to answer it.
    \item Evaluate the complexity of each step, considering:
    \begin{itemize}[noitemsep, topsep=0.5em]
        \item The need for precise perception (e.g., color or shape differentiation, relative positioning)
        \item Cross-referencing multiple regions, axes, or visual types
        \item Complex chart features (e.g., stacked vs. overlayed areas, 3D perspective)
        \item Occlusion or ambiguity in label visibility (e.g., overlapping text or hidden legends)
        \item Requirement for highly precise numerical interpretation (especially in visually challenging layouts)
    \end{itemize}
\end{enumerate}

\vspace{1mm}
\textbf{Typically \textit{Hard} cases include:}
\begin{itemize}[noitemsep, topsep=0.5em]
    \item Area charts requiring distinction between stacking vs overlay
    \item 3D charts with visual distortion or unclear projections
    \item Double-ring pie charts requiring ring disambiguation
    \item Radar charts with overlapping regions
    \item Multi-axis charts with axis disambiguation needs
    \item Perceptually ambiguous cases (e.g., boxplots with red boxes and red medians)
    \item Multi-step numerical comparison questions (e.g., "How much higher is X than Y?")
    \item Charts with occluded or obscured labels or legends
\end{itemize}

\vspace{1mm}
\textbf{Typically \textit{Medium} cases include:}
\begin{itemize}[noitemsep, topsep=0.5em]
    \item Multi-bar or multi-line charts with separated groups
    \item Node-link diagrams requiring structured inspection
    \item Two-step quantitative reasoning (e.g., compute and compare X and Y)
    \item Tasks involving comparison or arithmetic over multiple extracted values
\end{itemize}

\vspace{1mm}
\textbf{Typically \textit{Easy} cases include:}
\begin{itemize}[noitemsep, topsep=0.5em]
    \item Annotated charts where answers can be read off directly
    \item Questions solvable via clearly readable text
    \item Simple selection (e.g., identifying the maximum value)
\end{itemize}

\vspace{1mm}
\textbf{Respond in the following JSON format:}
\begin{verbatim}
{
  "label": "Easy / Medium / Hard",
  "reasoning": "Step-by-step explanation of how you reached your conclusion."
}
\end{verbatim}
\end{tcolorbox}


\begin{tcolorbox}[
  enhanced,
  colback=green!5,
  colframe=green!50,
  boxrule=0pt,
  sharp corners=south,
  top=1mm, bottom=1mm, left=2mm, right=2mm
]
\textbf{Input Format:}
\begin{verbatim}
<chart image> {entry["image"]}
<question> {entry["query"]}
\end{verbatim}
\end{tcolorbox}

\end{tcolorbox}

\end{document}